## An Open-Source Knowledge Graph Ecosystem for the Life Sciences

## Authors

Tiffany J. Callahan[1,2], Ignacio J. Tripodi[3], Adrianne L. Stefanski[1], Luca Cappelletti[4], Sanya B. Taneja[5], Jordan M. Wyrwa[6], Elena Casiraghi[4,7], Nicolas A. Matentzoglu[8], Justin Reese[7], Jonathan C. Silverstein[9], Charles Tapley Hoyt[10], Richard D. Boyce[9], Scott A. Malec[11], Deepak R. Unni[12], Marcin P. Joachimiak[7], Peter N. Robinson[13], Christopher J. Mungall[7], Emanuele Cavalleri[4], Tommaso Fontana[4], Giorgio Valentini[4,14], Marco Mesiti[4], Lucas A. Gillenwater[1,15], Brook Santangelo[1,15], Nicole A. Vasilevsky[16], Robert Hoehndorf[17], Tellen D. Bennett[15,18], Patrick B. Ryan[19], George Hripcsak[2], Michael G. Kahn[15], Michael Bada[20], William A. Baumgartner Jr[20], Lawrence E. Hunter[1,15]

**Affiliations**

1. Computational Bioscience Program, University of Colorado Anschutz Medical Campus, Aurora, CO, 80045, USA
2. Department of Biomedical Informatics, Columbia University Irving Medical Center, New York, NY 10032, USA
3. Computer Science Department, Interdisciplinary Quantitative Biology, University of Colorado Boulder, Boulder, CO, 80301, USA
4. AnacletoLab, Computer Science Department, University of Milan, 20122, Italy
5. Intelligent Systems Program, University of Pittsburgh, Pittsburgh, PA, 15260, USA
6. Department of Physical Medicine and Rehabilitation, School of Medicine, University of Colorado Anschutz Medical Campus, Aurora, CO, 80045, USA
7. Division of Environmental Genomics and Systems Biology, Lawrence Berkeley National Laboratory, Berkeley, CA 94720, USA
8. Semanticly Ltd, Athens, Greece
9. Department of Biomedical Informatics, University of Pittsburgh School of Medicine, Pittsburgh, PA, 15206, USA
10. Laboratory of Systems Pharmacology, Harvard Medical School, Boston, MA, 02115, USA
11. Division of Translational Informatics, University of New Mexico School of Medicine, Albuquerque, NM, 87131, USA
12. SIB Swiss Institute of Bioinformatics, Basel, Switzerland
13. Berlin Institute of Health at Charité-Universitatsmedizin, 10117 Berlin, Germany
14. ELLIS, European Laboratory for Learning and Intelligent Systems
15. Department of Biomedical Informatics, University of Colorado School of Medicine, Aurora, CO 80045, USA
16. Data Collaboration Center, Critical Path Institute, 1840 E River Rd. Suite 100, Tucson, AZ, 85718, USA
17. Computer, Electrical and Mathematical Sciences & Engineering Division, Computational Bioscience Research Center, King Abdullah University of Science and Technology, Thuwal 23955-6900, Kingdom of Saudi Arabia
18. Department of Pediatrics, University of Colorado School of Medicine, Aurora, CO 80045, USA
19. Janssen Research and Development, Raritan, NJ 08869, USA
20. Division of General Internal Medicine, University of Colorado School of Medicine, Aurora, CO 80045, USA

corresponding author(s): Tiffany J. Callahan (tiffany.callahan@cuanschutz.edu), William A. Baumgartner (william.baumgartner@cuanschutz.edu), and Lawrence E. Hunter (prof.larry.hunter@gmail.com)

## Abstract

Translational research requires data at multiple scales of biological organization. Advancements in sequencing and multi-omics technologies have increased the availability of these data, but researchers face significant integration challenges. Knowledge graphs (KGs) are used to model complex phenomena, and methods exist to construct them automatically. However, tackling complex biomedical integration problems requires flexibility in the way knowledge is modeled. Moreover, existing KG construction methods provide robust tooling at the cost of fixed or limited choices among knowledge representation models. PheKnowLator (Phenotype Knowledge Translator) is a semantic ecosystem for automating the FAIR (Findable, Accessible, Interoperable, and Reusable) construction of ontologically grounded KGs with fully customizable knowledge representation. The ecosystem includes KG construction resources (e.g., data preparation APIs), analysis tools (e.g., SPARQL endpoints and abstraction algorithms), and benchmarks (e.g., prebuilt KGs and embeddings). We evaluated the ecosystem by systematically comparing it to existing open-source KG construction methods and by analyzing its computational performance when used to construct 12 large-scale KGs. With flexible knowledge representation, PheKnowLator enables fully customizable KGs without compromising performance or usability.

## Introduction

The worldwide growth of biomedical data is exponential, with the volume of molecular data alone expected to surpass more than four exabytes by 2025[1]. Translational science requires integrating data and knowledge at multiple scales of biological organization. Rapid advancements in sequencing and multi-omics technologies have made tremendous amounts of diverse data available for secondary use[2–5]. Multimodal data capture different views and, when properly combined, help characterize complex systems[6]. Unfortunately, these data are highly distributed and heterogeneous, can be difficult to access due to licensing restrictions, lack interoperability, and often have inconsistent underlying models or representations, which limit most researchers from fully utilizing them[7,8].

Knowledge graphs (KGs) have frequently been used to systematically model and interrogate the biology underlying complicated systems, organisms, and diseases[9]. For example, Figure 1 provides a high-level overview of the main biomedical concepts needed to model our currently accepted knowledge of the Central Dogma[10] and has been expanded to include pathways, variants, pharmaceutical treatments, and diseases. In the life sciences, KGs are usually constructed from a wide range of data sources such as Linked Open Data[11] ontologies, the scientific literature, data derived from electronic health records, and multi-omics experiments[8,12]. In the biomedical context, nodes usually represent different kinds of biological entities such as genes, proteins or diseases, and edges (or triples) are used to specify different types of relationships that can exist between a pair of nodes (e.g., “interaction”, “substance that treats”). Multiple definitions of KGs have been proposed in the literature, all sharing the assumption that KGs are more than simple large-scale graphs[13–15]. Existing definitions are best summarized by Ehrlinger's and Wöb's (2016) definition: "A knowledge graph acquires and integrates information into an ontology and applies a reasoner to derive new knowledge"[13]. We provide an alternative definition and consider a KG to be a graph-based data structure representing a variety of heterogeneous entities and multiple types of relationships between them and serving as an abstract framework that is able to infer new knowledge (as well as reveal and resolve discrepancies or contradictions) to address a variety of applications and use cases.

KG construction is not a simple process, requiring significant data preprocessing or wrangling before edge lists can be assembled. Fortunately, several methods have been developed to tackle the primary challenges faced when constructing a KG, including: the integration or harmonization of disparate resources (e.g., SPOKE[16], RTX-KG2[17], Petagraph[18], Bio2RDF[19], and Hetionet[20]), processing and formatting of structured data and KGs (e.g., Dipper[21], the Knowledge Graph Exchange [KGX][22]), enhancement or extraction of relationships (e.g., Biomedical Knowledge Discovery Engine [BioKDE][23], KG-COVID-19[24]) and evidence (e.g., PrimeKG[25]) from the literature, and the exchange or sharing of constructed KGs (e.g., Network Data Exchange [NDEx][26], KGX[22]). Recently, several frameworks such as KG-HUB[27], the Clinical KG [CKG][28], RTX-KG2[17], BioCypher[29], and the Knowledge Base Of Biomedicine (KaBOB)[7] which provide all of the aforementioned functionalities, have been developed. While methods have been developed for each of the processes or steps required to construct KGs, robust tools and resources to evaluate constructed KGs are lacking[8] Traditionally, the evaluation of constructed KGs has been task- or domain-specific and largely limited to case studies[16,17,20,24,25,28,29]. Ideally, constructed KGs would be evaluated in the same manner as other network science (e.g., community detection and link prediction algorithms) and KG or node embedding methods using benchmarks such as Zachary’s Karate Club graph[30], DBPedia[31], and OpenBioLink[32]. KG benchmarks could be used to assess the computational performance of KG construction methods and to evaluate the implications of different knowledge representations on specific tasks. To the best of our knowledge, no existing benchmarks exist to evaluate knowledge representation systematically.

Tackling complex problems within the life sciences requires flexible knowledge representations. An important limitation of existing KG construction methods is fixed or limited flexibility in the way that knowledge is modeled. Within the biomedical domain, knowledge is typically modeled in one of three ways (Figure 2), though the nomenclature used to describe these different approaches differs widely in the literature. For simplicity's sake, we will refer to the three different approaches as simple, hybrid, and complex. The first approach results in a simple graph (Figure 2a). Simple graphs are the most common type of network used in the literature. Examples of simple graphs include Zachary's Karate Club graph[30], Hetionet[20], and SPOKE[16]. In these graphs, entities are represented as nodes, and edges are used to model relationships between them. These graphs usually lack formal semantics for the edges and nodes. Edges are often semantically overloaded, ignoring the distinction between data (e.g., a protein participating in a process) and metadata (e.g., the source of information about the protein's participation in that process). Simple graphs are usually straightforward to construct and can be stored as key-value pairs, resulting in small file sizes and using modest amounts of memory. Disadvantages of simple graphs include ad hoc semantics, which decreases interoperability, and a lack of clear specification, making machine inference difficult. The second approach results in a hybrid or property graph (Figure 2b). Example hybrid graphs include KG-COVID-19[24], DisGeNET[33], OpenBioLink[32], Petagraph[18], the Monarch KG[34], and Bio2RDF[19]. Hybrid graphs aim to model entities and their relations using a mix of standard network representations and formal semantics, usually the Resource Description Framework (RDF)[35] and RDF Schema (RDFS)[36]. Compared to simple graphs, standards-based hybrid graphs facilitate integration with other resources[37] and are more amenable to automated inference. They also provide faceted querying as nodes and edges are typed. One cost of hybrid graphs is that they require substantially more storage space than simple graphs. The third approach results in a complex graph, such as KaBOB[7], often built on the Web Ontology Language (OWL)[38] standard (Figure 2c). Complex graphs are more expressive, facilitating the generation of new knowledge via deductive inference[39]. By enforcing explicit semantics, OWL provides advantages over RDF/RDFS in the integration of large biomedical data[40]. Complex graphs are fully machine-readable, highly expressive, and, because they are built on Description Logics[39], are able to leverage reasoners to verify their logical consistency and do deductive inference. Unlike simple graphs, both hybrid and complex graphs are able to distinguish between data and metadata as demonstrated in Figure 2. Unless defining custom relations, hybrid graphs do this by primarily using RDFS and resources like the OBO Format metamodel[41], whereas complex graphs formally define these types and their attributes using RDF and OWL. Unfortunately, complex graphs are very large, can be difficult for humans to understand, and have been shown to perform poorly on some inductive inference tasks[42]. To date, none of the existing KG construction methods enable the construction of multiple or alternative versions of the same KG utilizing different underlying knowledge representations, making comparisons and benchmarking difficult.

To address the lack of relevant benchmarks and flexibility in knowledge representation, we developed PheKnowLator (Phenotype KnowledgeTransLator, referred to as "PKT" throughout the remainder of this manuscript), a semantic ecosystem for automating the FAIR (Findable, Accessible, Interoperable, and Reusable)[43] construction of ontologically grounded KGs with fully customizable knowledge representation. The ecosystem consists of three components (Figure 3): (1) **KG Construction Resources**, a set of tools to download and process heterogeneous data and algorithms to construct custom KGs; (2) **KG Benchmarks,** a collection of prebuilt KGs that can be used to systematically assess the effects of different knowledge representations on downstream analyses, workflows, and learning algorithms; and (3) **KG Tools** to analyze KGs, including Jupyter Notebook-based use cases and tutorials, archive-based data storage, application programming interfaces (APIs), and triplestores. We evaluate the PheKnowLator ecosystem by systematically comparing its components with existing open-source KG construction software using a survey developed to assess the functionality, availability, usability, maturity, and reproducibility of KG construction software. We also assess the ecosystem's computational performance when constructing 12 benchmark KGs designed to provide alternative representations for modeling the molecular mechanisms underlying human disease.

## Results

PheKnowLator is open-source and available through GitHub (https://github.com/callahantiff/PheKnowLator) and PyPI (https://pypi.org/project/pkt-kg). Important manuscript definitions are provided in Supplementary Table 1, acronyms are provided in Supplementary Table 2, and PheKnowLator ecosystem resources are listed in Supplementary Tables 3 and 4.

## Evaluation

The PheKnowLator ecosystem was evaluated in two ways. First, publicly available software to construct biomedical KGs was identified and systematically compared using a survey developed to assess each method's functionality, availability, usability, maturity, and reproducibility. Second, the computational performance of the ecosystem was assessed when used to construct 12 different types of benchmark KGs designed to provide alternative representations for modeling the molecular mechanisms underlying human disease. The resources used for each task are listed in Supplementary Table 4.

### *Systematic Comparison of Open-Source KG Construction Software*

Open-source biomedical KG construction methods available on GitHub were identified and compared to the PheKnowLator ecosystem. A survey was used to compare the methods for the task of constructing biomedical KGs and consisted of 44 questions designed to assess five criteria: KG construction functionality, maturity, availability, usability, and reproducibility (Supplementary Table 5). Of the 1,905 repositories identified on GitHub, 231 contained course, tutorial, or presentation material (i.e., manuscript reviews and slide decks), 278 were duplicate or cloned repositories, 79 were KG applications or services, 60 were websites or resource lists, and 1,253 were determined to be irrelevant (i.e., mislabeled, not biomedical, or not a KG construction method). This initial list was supplemented with 11 methods identified through a review article[8]. The final list included 15 methods (see Table 1 with additional details provided in Supplementary Table 6): Bio2Bel[44], Bio2RDF[45], Bio4J[46], BioGrakn[47], the Clinical Knowledge Graph[48], COVID-19-Community[49], Dipper[21], Hetionet[50], IASiS Open Data Graph[51], KG-COVID-19[52], KaBOB[53], KGX[22], the Knowledge Graph Toolkit[54], ProNet[55], and the SEmantic Modeling machIne[56]. The methods are visualized by date of GitHub publication in Figure 4a.

The average coverage score of the five assessment criteria was 3.93 (min=2.79, max=4.90). The coverage of each assessment criterion by method is shown in Figure 4b. Examining the results by assessment criteria revealed interesting patterns. **KG Construction Functionality** (Supplementary Table 7): The majority of the methods (81.3%; n=13) included functionality to download data, while 31.3% (n=5) were able to process free text and 37.5% (n=6) were able to process clinical data. **Availability** (Supplementary Table 8): Three-fourths of the methods (75%; n=12) were written in Python and 18.8% (n=3) were written in a Java-based language. All of the methods but one were licensed with GPL, MIT, or BSD-3. **Usability** (Supplementary Table 9): Sample data were provided by 94.4% (n=17) of the methods, and 80% (n=14) provided tutorials via R Markdown or Jupyter Notebook. **Maturity** (Supplementary Table 10): On average, the number of commits per year ranged from 17 to 1,000. Over half of the methods (68.8%, n=11) had been published, and 43.8% (n=7) provided collaboration guidelines. **Reproducibility** (Supplementary Table 11): Tools to enable reproducible workflows and aid in installing the method were provided by 75% (n=12) of the methods. Most often, these tools included Docker containers (n=6) and Jupyter or R Notebooks (n=7), and more than 37.5% (n=6) of the methods used a dependency management program like PyPI or CRAN.

While the PheKnowLator ecosystem was comparable to the other methods on the assessed criteria, we identified three important differentiating factors relative to the other methods: (i) tools to assess the quality of underlying ontologies; (ii) logging and documentation of metadata including the KG construction process, the data downloaded, the processing steps applied to each data source, and the node and edge types each source contributes to; and (iii) customizable knowledge representation making it possible to take advantage of advanced Semantic Web tools like description logic reasoners (which we have successfully applied in the construction of KGs by the PheKnowLator ecosystem). The ability to generate multiple versions of the same KGs enables the ecosystem to provide benchmark KGs, which can be used to evaluate modeling decisions and to study the impact of knowledge representation on downstream learning. PheKnowLator included all the functionalities in the five assessment criteria except for tools to process clinical data, which only 37.5% (n=6) of the methods provided.

***Human Disease Knowledge Graph Benchmark Comparison and Construction Performance***

The PheKnowLator ecosystem enables users to fully customize KG construction by providing the following parameters (described in detail in the *Construct Knowledge Graphs* section of **Component 1: Knowledge Graph Construction Resources** in the Methods): knowledge model (i.e., complex graphs using class- or instance-based knowledge models), relation strategy (i.e., standard directed relations or inverse bidirectional relations), and semantic abstraction (i.e., transformation of complex graphs into hybrid graphs) with or without knowledge model harmonization (i.e., ensuring a hybrid KG is consistent with the class- or instance-based complex graph it was abstracted from). These parameters enable 12 different versions or benchmarks of each KG to be constructed for a given build. Descriptive statistics and computational performance of the PheKnowLator ecosystem was assessed when used to build a large-scale heterogeneous KG designed to represent the molecular mechanisms underlying human disease and its 12 associated KG types or benchmarks (referred throughout the remainder of manuscript as the PKT [PheKnowLator] Human Disease benchmark KGs).

*Benchmark Comparison*

Under the advice of domain experts (ALS, IJT, LH, and CJM), the PKT Human Disease benchmark KGs were constructed from 12 OBO Foundry ontologies, 31 Linked Open Data sets, and results from two large-scale molecular experiments (all build data are listed and described in Supplementary Table 12). The knowledge representation used for the build is shown in Supplementary Figure 1. A simplified overview of this knowledge representation is provided in Figure 5, which highlights the connectivity between the 12 OBO Foundry ontologies (Figure 5a) and their relationship to the primary node types. The 18 primary node types are listed in Table 2 (visualized in Figure 5b), and 33 primary edge types are shown in Table 3. The primary node and edge types do not include all possible node and edge types made available in the core set of 12

OBO Foundry ontologies, only those that are explicitly modeled in our knowledge representation.

Descriptive statistics for the OBO Foundry ontologies, pre- and post-data quality assessment, are shown in Table 4 (and detailed statistics are provided in Supplementary Table 13). Please note that when reporting results, we will refer to edges as triples, but they both refer to node-relation-node statements. The size of the ontologies varied widely, with the Chemical Entities of Biological Interest (ChEBI)[57] containing the largest number of triples (n=5,190,458) and the Protein Ontology (PRO; modified to exclude all non-human proteins)[58] containing the most classes (n=148,243). The Relation Ontology (RO)[59] contained the fewest triples (n=34,901), and the Sequence Ontology (SO)[60] contained the fewest classes (n=2,569). The merged set of cleaned OBO Foundry ontologies (i.e., core OBO Foundry ontologies; for additional detail on the ontology cleaning process, please see the Component 1: Knowledge Graph Construction Resources section of the **Methods**) contained 545,259 classes and 13,748,009 triples. Statistics for triples added to the core OBO Foundry ontologies are listed by edge type in Table 5. The largest edge sets were protein-protein (n=618,069 triples), transcript-anatomy (n=439,917 triples), and disease-phenotype (n=408,702 triples). The smallest edge sets were biological process-pathway (n=665 triples), gene-gene (n=1,668 triples), and protein-cofactor (n=1,961 triples).

Descriptive statistics for the 12 PKT Human Disease benchmark KGs are shown in Table 6. The PKT Human Disease benchmark KGs constructed using the class-based knowledge model with inverse relations and without semantic abstraction were the largest (13,803,521 nodes; 41,116,791 triples). All of the PKT Human Disease benchmark KGs built without semantic abstraction, regardless of the knowledge model or relation strategy, contained two connected components and three self-loops. All of the PKT Human Disease benchmark KGs were highly sparse, with the average density[61] ranging from $2.16 \times 10^{-7}$ to $3.50 \times 10^{-7}$ and $3.03 \times 10^{-7}$ to $3.40 \times 10^{-7}$ for benchmark KGs constructed using class-based and instance-based knowledge models, respectively. When applying semantic abstraction, the PKT Human Disease benchmark KGs constructed using instance-based knowledge models (743,829 nodes; 4,967,391 to 9,624,232 triples) were on average larger than those constructed using the class-based knowledge models (743,829 nodes; 4,967,427 to 7,629,599 triples). All PKT Human Disease benchmark KGs constructed using the instance-based knowledge model with semantic abstraction, regardless of the relation strategy employed, were larger, had a higher average degree, and contained more self-loops when knowledge model harmonization was applied. The average density (6.68 standard relations; 10.26 inverse relations) and number of self-loops (445 standard and inverse relations) did not differ for the PKT Human Disease benchmark KGs constructed using the class-based knowledge model with semantic abstraction and when applying knowledge model harmonization. The PKT Human Disease benchmark KGs constructed with semantic abstraction, with and without knowledge model harmonization, are visualized in Figure 6.

*Construction Performance*

Performance metrics by KG construction step for each of the 12 PKT Human Disease benchmark KGs are shown in Supplementary Figure 2. On average, **Step 1** (*Data Download*) took 2.30 minutes (1.80-3.72 minutes) and used an average of 7.93 GB of memory (7.86-7.99 GB). **Step 2** (*Edge List Creation*) took an average of 4.82 minutes to complete (4.80-4.87 minutes) and used an average of 39.55 GB of memory (38.93-40.43 GB). **Step 3** (*Graph Construction*) took an average of 391.56 minutes (6.53 hours) to complete (265.98-615.92 minutes; 4.43-10.27 hours) and used an average of 118.69 GB of memory (104.30-147.10 GB). On average, the PKT Human Disease benchmark KGs constructed using class-based knowledge models took roughly the same amount of time and used roughly the same maximum amount of memory as those constructed using instance-based knowledge models. Additionally, regardless of the knowledge model, on average, the PKT Human Disease benchmark KGs built using inverse relations and semantic abstraction took longer to run and required more memory.

## Discussion

In this paper, we have presented PheKnowLator, a semantic ecosystem for automating the FAIR construction of ontologically grounded KGs with customizable knowledge representation. The ecosystem includes KG construction resources, analysis tools (i.e., SPARQL endpoints and cloud-based APIs), and benchmarks (i.e., prebuilt KGs in multiple formats and embeddings). PheKnowLator enables users to build Semantic Web-compliant complex KGs that are amenable to automatic OWL reasoning, conform to contemporary graph standards, and are importable by popular graph toolkits. By providing flexibility in the way KGs are constructed and generating multiple types of KGs, PheKnowLator also enables the use of cutting-edge graph-based learning and sophisticated inference algorithms. We demonstrated PheKnowLator's utility by comparing its features to 14 existing open-source KG construction methods and by analyzing its computational performance when constructing 12 different large-scale heterogeneous benchmark KGs. Comparing these methods to PheKnowLator revealed similarities but also highlighted important differentiating factors lacking in other systems, namely: (1) tools to assess the quality of ontologies (which identify, repair, and document syntactic and semantic errors); (2) logging and metadata documentation (which enable users to debug errors quickly and ensures builds can be rigorously reproduced); and (3) customizable data preprocessing pipelines (which enable users to use ecosystem tools to develop custom pipelines for processing a wide variety of data, leverage complex mappings, and appropriately resolve missing data), knowledge representation (class- or instance-based), and benchmarks (the ability to construct different types of KGs from the same data, which enables users to empirically evaluate modeling decisions and find the optimal knowledge model or representation for a particular task). These differences highlight PheKnowLator's ability to provide fully customizable KGs without compromising performance or usability.

One of the biggest challenges to developing novel KG construction methods is properly verifying and robustly validating the resulting KGs. Network-science-based algorithms and machine learning methods typically used within the biomedical domain such as link prediction and knowledge graph embedding are able to make use of well-established benchmarks like YAGO[62], DBPedia[63], and Wikidata[64], which are not specific to the biomedical domain. OpenBioLink[32] was developed as a benchmark for biomedical KGs, but is almost exclusively used for link prediction tasks. While it might not be possible to create a universal benchmark to verify or validate biomedical KG construction methods or biomedical KGs, development of trusted resources that are not task-specific (e.g., entity prediction or node classification) would benefit the community. The PheKnowLator ecosystem introduces a set of benchmarks to serve this purpose. These benchmarks were specifically designed to enable two types of tasks: (1) the validation of tools and algorithms designed to analyze KGs (e.g., link prediction algorithms and graph representation learning methods); and (2) the validation and comparison of KGs built using different underlying knowledge representations. The ability to empirically evaluate knowledge modeling decisions is important when designing knowledge-based systems[8] and will become more important as more performant graph representation learning methods are developed, especially with respect to explainability[65].

## PheKnowLator Applications and Use Cases

The majority of existing published KGs and KG construction software within the biomedical domain rely on case studies as a form of evaluation[16,18,20,24,25,29]. While we did not explicitly include case studies as part of our validation, the PheKnowLator ecosystem has fostered substantial collaborations and led to several publications. PheKnowLator benchmark KGs have been used in applications of toxicogenomic mechanistic inference[66], to enable the exploration of large-scale biomedical hypergraphs[67], and to facilitate deeper sub-phenotyping of pediatric rare disease patients[68]. Recently, PheKnowLator was used to create a disease-specific KG that combined ontology-grounded resources with literature-derived computable knowledge from machine reading[69]. The resulting KG was then used to identify causal features suitable for addressing confounding bias. PheKnowlator has also been used to generate hypotheses for potential pharmacokinetic natural-product/drug interactions, by facilitating the design and implementation of a KG involving biomedical ontologies, natural-product-ontology extensions, and machine reading from literature[70]. Finally, the PheKnowLator ecosystem was recently selected as the primary infrastructure to facilitate the development of a large-scale KG (denoted RNA-KG) dedicated to the study and development of RNA-based drugs by integrating more than 50 public data sources[71,72]. PheKnowLator is also the foundation for novel KG approaches in microbiome research: The microbe-relevant KG Microbe-Gene-Metabolite Link (MGMLink) was constructed by augmenting PheKnowLator with information on microbes from the integrated database gutMGene. GutMGene relationships describing observed microbe-metabolite or microbe-gene associations were introduced to the PheKnowLator knowledge base, enabling a search space for mechanistic understanding of microbial influence on disease at the molecular level[73].

In addition to the use of the PheKnowLator KG construction software and benchmark KGs, the ecosystem has also contributed to the development of novel tools and resources. Although results are not yet available, PheKnowLator is currently included in the Continuous Evaluation of Relational Learning in Biomedicine (https://biochallenge.bio2vec.net/) task. This task aims to provide a means for evaluating prediction models as new knowledge becomes available over time. Results from this task will provide insight into the usefulness of the PheKnowLator builds and will be used to identify areas where the ecosystem can be improved. Additionally, subsets of prebuilt PheKnowLator KGs have been used to help develop and evaluate novel, cutting-edge graph embedding AI tools (i.e., GRAPE[74]), including random-walk-based embedding methods for extremely large-scale heterogeneous graphs using the PheKnowLator KG builds[75]. In addition to graph representation learning, prebuilt PheKnowLator KGs were used for prototyping a novel method for knowledge-driven mechanistic enrichment of ignorome genes (i.e., differentially expressed genes which are associated with a disease experimentally but that have no known association to the disease in the literature)[76]. When applied to preeclampsia, this method was able to identify 53 novel clinically relevant and biologically actionable disease associations. The NIH Common Fund Human BioMolecular Atlas Program (HuBMAP)[77] needed to assemble a KG based on its own preferred graph schema[78–80], with one focus being to maximize the leverage of external references among ontologies for translation. The PheKnowLator ecosystem tool OWL-NETS[42] is currently being used to ingest other operational ontologies (whether in OWL or not) into HuBMAP and NIH Common Fund Cellular Senescence Network (SenNet)[81]. PheKnowLator was also applied to methods in generating pathway diagrams using biomedically relevant KGs[82]. This novel approach was able to recapitulate existing figures regarding neuroinflammation and Down Syndrome from literature with more detailed and semantically consistent molecular interactions using PheKnowLator[83].

## Limitations and Future Work

The current work has several important limitations. First, it is important to point out that the systematic comparison we performed of open-source KG construction methods on GitHub was subjective, included only three researchers actively involved in developing PheKnowLator, and was originally performed in 2020. While the results were updated in 2021 and re-reviewed in 2023, it is possible that new methods might not have been included. Further, only a qualitative comparison was carried out that considered each method's GitHub and associated publications. Ideally, a fair evaluation would be performed where each method would be downloaded and compared when used to build a KG from the same set of data. Unfortunately, this type of analysis requires significant resources and was not within the scope of our analysis. Similarly, given their success within the Semantic Web Domain, future work should formally evaluate our data integration and ontology alignment pipelines to tools like Web Karma[84], OpenRefine[85], and mapping languages like R2RML[86]. Second, computational performance metrics were only computed over a single build run due to the amount of resources required to build the KGs. While it is not expected that the results for these metrics would significantly change, small deviations related to data provider constraints with respect to accessing build data could result in different outcomes. Third, we mention that the PheKnowLator ecosystem includes two types of benchmarks: KGs and embeddings. Currently, embeddings are only available for one build (v1.0.0) because the size of the generated KGs were quite small. Subsequent builds have resulted in KGs that are so large that generating embeddings has not been feasible. Fortunately, the recent development of performant embedding tools like GRAPE will enable us to provide embeddings for future builds[74] Fourth, while the ecosystem includes robust logging to monitor metadata and builds, it does not formally integrate resources like the Bioregistry[87] and BioLink[88], which are becoming important new KG standards[17,27]. Similarly, the PheKnowLator ecosystem relies heavily on OWLTools[89] but newer and more stable tools like ROBOT[90] should be leveraged because it supports the integration of the OWL API and has improved Jena-based functionality. Fifth, as mentioned above, validating very large KGs, like the ones produced by PheKnowLator, is challenging but important. Additional validation of the PheKnowLator ecosystem, including the construction tools and benchmarks is needed, especially with respect to the different KG builds it produces. Finally, while we have worked hard to ensure that the ecosystem tools and infrastructure are user-friendly, additional work is needed to simplify the inputs and make them more machine-readable (e.g., converting input text files into configurable yaml files) and also develop Graphical User Interfaces for supporting the users in all the steps of KG construction.

# Methods

## The PheKnowLator ecosystem

The PheKnowLator ecosystem was developed to provide a more comprehensive resource to aid in the construction of KGs within the Life Sciences and consists of three components: (1) **KG Construction Resources**; (2) **Benchmark KGs**; and (3) **KG Tools**. Each component is modular; all features and elements can be replaced or extended as technology evolves or to fit a particular use case. The PheKnowLator ecosystem resources are listed by component in Supplementary Table 3.

### *Component 1: Knowledge Graph Construction Resources*

This component is represented by the largest gray box in Figure 3 and consists of two elements: (1) **Process Data.** Resources to process a variety of heterogeneous data; and (2) **Construct KGs.** An algorithm that enables the construction of different types of heterogeneous KGs. The resources that support these elements are detailed in the *ecosystem Component 1: Knowledge Graph Construction Resources* section of Supplementary Table 3.

*Process Data*

This element consists of two features and was designed to help users download and prepare a wide variety of heterogeneous data sources needed to construct KGs. The two primary features of this component are: (i) Download and (ii) Preparation.

Download

This feature has been configured to download two types of data: (i) ontologies (e.g., HPO[91], GO[92], and PRO[58]) and databases (i.e., a data source not represented as an ontology), which includes Linked Open Data (e.g., Comparative Toxicogenomics Database[93], UniProt Knowledgebase[94], STRING[95]), data from molecular experiments (e.g., the Human Protein Atlas[96] the Genotype-Tissue Expression Project[97]), and existing networks and KGs (e.g., Hetionet[20], the Monarch KG[98]). Ontologies are downloaded using OWLTools[89] (April 06, 2020 release) and databases are downloaded using a custom-built API capable of processing a variety of file formats (e.g., zip, gzip, tar) from different types of servers and APIs.

Preparation

A collection of tools were developed to help users perform a variety of tasks when preparing data that will be used to construct a KG. This feature provides services to map different types of identifiers (e.g., aligning gene identifiers from the Human Gene Nomenclature Committee [HGNC][99] to Entrez Gene[100] and Ensembl[101]), annotate concepts (e.g., convert strings of tissue names from the Human Protein Atlas[96] to Uber-Anatomy Ontology [Uberon][102] concepts), filter data (e.g., identify variant-disease relationships from Clinvar[103] with a specific type of experimental validation), and process entity metadata (e.g., obtain PubMed identifiers for exposure-outcome relationships from the Comparative Toxicogenomics Database[93] and extract synonyms and definitions for OBO Foundry ontology concepts). The Data Preparation Notebook (Data_Preparation.ipynb[104]) illustrates some of these features. There are also features to assess and repair the quality of OBO Foundry ontologies, which are known to be subject to a variety of errors[105–107]. The Ontology Cleaning Notebook (Ontology_Cleaning.ipynb[108]) includes detailed descriptions and examples of the data quality checks[109]. A report is generated after assessing the quality of each ontology, which provides statistics before and after applying each check (ontology_cleaning_report.txt).

*Construct Knowledge Graphs*

This element consists of four features designed to facilitate the construction of large-scale heterogeneous KGs. Together, these features comprise the core functionality of the PheKnowLator KG construction algorithm (referred to as PKT-KG throughout the remainder of the manuscript). The PKT-KG algorithm requires three input documents: (i) a list of one or more OBO Foundry ontologies; (ii) a list of one or more databases; and (iii) edge list assembly instructions (i.e., instructions for filtering input data sources and references to resources needed to normalize concept identifiers). Additional information on each input is available on GitHub (https://github.com/callahantiff/PheKnowLator/wiki/Dependencies). The four primary features of this component are: (i) Edge List Construction, (ii) Ontology Alignment, (iii) Customize Knowledge Representation, and (iv) Output Generation.

Edge List Construction

Using information in the edge list assembly instructions, the edge list construction procedure merges data, applies filtering and evidence criteria, and removes unneeded attributes. To automate this process, we have developed a universal file parser (and constantly update it with procedures for parsing new file types) that currently processes over 30 distinct file types. Once the edge lists are constructed, they are serialized in a JSON file.

Ontology Alignment

OBO Foundry ontologies were selected because they represent canonical knowledge and exist for nearly all scales of biological organization[110]. PKT-KG assumes that every KG is logically grounded[111] in one or more OBO Foundry ontologies. This feature leverages OWLTools[89] (April 06, 2020 release) to merge the ontologies into a single integrated core ontology.

Customize Knowledge Representation

To enable customization in the way that knowledge is represented when constructing a KG, three configurable parameters are provided:

1. **Knowledge Model.** Following Semantic Web standards[112], PKT-KG defines a KG as $K = \langle T, A \rangle$, where $T$ is the TBox and $A$ is the ABox. The TBox represents the taxonomy of a particular domain[113,114]. It describes classes, properties/relationships, and assertions that are assumed to generally hold within a domain (e.g., a gene is a heritable unit of DNA located in the nucleus of cells [Figure 7a]). The ABox describes attributes and roles of instances of classes (i.e., individuals) and assertions about their membership in classes within the TBox (e.g., A2M is a type of gene that may cause Alzheimer's Disease [Figure 7b])[113,114]. PKT KGs are logically grounded in one or more OBO Foundry ontology[111]. Database entities (i.e., entities from a data source that is not an OBO Foundry ontology) are added to the core OBO Foundry ontologies using either a TBox (i.e., class-based) or ABox (i.e., instance-based) knowledge model. For the class-based approach, each database entity is made a subclass of an existing core OBO Foundry ontology class (see the "Class-based" section of Supplementary Table 14). For the instance-based approach, each database entity is made an instance of an existing core OBO Foundry ontology class (see the "Instance-based" section of Supplementary Table 14). Both approaches require the alignment of database entities to an existing core OBO Foundry ontology class, which is managed by a dictionary that is constructed using tools in the Process Data Element of the **Knowledge Graph Construction Resources** component (subclass_construction_map.pkl).
2. **Relation Strategy.** PKT-KG provides two relation strategies. The first strategy is standard or directed relations, through a single directed edge (e.g., "gene causes phenotype"). The second strategy is inverse or bidirectional relations, through inference if the relation is from an ontology like the RO (e.g., "chemical participates in pathway" and "pathway has participant chemical") or through inferring implicitly symmetric relations for edge types that represent biological interactions (e.g., gene-gene interactions).
3. **Semantic Abstraction.** KGs built using expressive languages like OWL are structurally complex and composed of triples or edges that are logically necessary but not biologically meaningful (e.g., anonymous subclasses used to express TBox assertions with all-some quantification). PKT-KG currently uses the OWL-NETS[42] semantic abstraction algorithm to convert or transform complex KGs into hybrid KGs. OWL-NETS v2.0[115] includes additional functionality that harmonizes a semantically abstracted KG to be consistent with a class- or instance-based knowledge model. For class-based knowledge models, all triples containing *rdf:type* are updated to *rdfs:subClassOf. For* instance-based knowledge models, all triples containing *rdfs:subClassOf* are updated to *rdf:type*. For additional details, see OWL-NETS v2.0 documentation[115].

Output Generation

To ensure features of the Process Data element (**KG Construction Resources** component) are transparent and reproducible, metadata are output for all downloaded (downloaded_build_metadata.txt; Supplementary Document 1)) and processed (preprocessed_build_metadata.txt; Supplementary Document 2) data, including the details of the processing steps applied to each database (edge_source_metadata.txt; Supplementary Document 3) and OBO Foundry ontology (ontology_source_metadata.txt and ontology_cleaning_report.txt; Supplementary Documents 4-5). The PKT KG construction process is logged extensively (data download and preprocessing [pkt_builder_phases12_log.log; Supplementary Document 6] and KG construction [[pkt_build_log.log; Supplementary Document 7]). PKT KGs, including node and relation metadata, are output to a variety of standard formats. A description of all file types is available from the Zenodo Community archive (PheKnowLator_HumanDiseaseKG_Output_FileInformation.xlsx)[116].

***Component 2: Knowledge Graph Benchmarks***

This component consists of different types of prebuilt KGs or benchmarks that can be used to systematically assess the effects of different knowledge representations on downstream analyses, workflows, and learning algorithms (Figure 3). Current benchmarks and their supporting features are detailed in the *ecosystem Component 2: Knowledge Graph Benchmarks* section of Supplementary Table 3. Currently, the PheKnowLator ecosystem supports two types of benchmarks: (i) KGs and (ii) embeddings. An end-to-end example demonstrating how a single data source is transformed through each build step of Component 2 is provided in Figure 8. This figure also demonstrates how this data source would be modeled across the 12 different types of KGs that can be configured from a single build using the ecosystem.

*Knowledge Graphs*

The PKT Human Disease KG was built to model mechanisms of human disease, which includes the Central Dogma and represents multiple biological scales of organization including molecular, cellular, tissue, and organ. The knowledge representation was designed in collaboration with a PhD-level molecular biologist (Supplementary Figure 1). The PKT Human Disease KG was constructed using 12 OBO Foundry ontologies, 31 Linked Open Data sets, and results from two large-scale experiments (Supplementary Table 12). The 12 OBO Foundry ontologies were selected to represent chemicals and vaccines (i.e., ChEBI[57] and Vaccine Ontology [VO][117,118]), cells and cell lines (i.e., Cell Ontology [CL][119], Cell Line Ontology [CLO][120]), gene/gene product attributes (i.e., Gene Ontology [GO][92,121]), phenotypes and diseases (i.e., Human Phenotype Ontology [HPO][91], Mondo Disease Ontology [Mondo][122]), proteins, including complexes and isoforms (i.e., PRO[58]), pathways (i.e., Pathway Ontology [PW][123]), types and attributes of biological sequences (i.e., SO[60]), and anatomical entities (Uberon[102]). The RO[59] is used to provide relationships between the core OBO Foundry ontologies and database entities. As shown in Figure 5, the PKT Human Disease KG contained 18 node types (Table 2) and 33 edge types (listed by relation in Table 3). Note that the number of nodes and edge types reflects those that are explicitly added to the core set of OBO Foundry ontologies and does not take into account the node and edge types provided by the ontologies. These nodes and edge types were used to construct 12 different PKT Human Disease benchmark KGs by altering the Knowledge Model (i.e., class- vs. instance-based), Relation Strategy (i.e., standard vs. inverse relations), and Semantic Abstraction (i.e., OWL-NETS (yes/no) with and without Knowledge Model harmonization [OWL-NETS Only vs. OWL-NETS + Harmonization]) parameters. Benchmarks within the PheKnowLator ecosystem are different versions of a KG that can be built under alternative knowledge models, relation strategies, and with or without semantic abstraction. They provide users with the ability to evaluate different modeling decisions (based on the prior mentioned parameters) and to examine the impact of these decisions on different downstream tasks.

*Embeddings*

To provide a version of the benchmarks that can more easily be used for downstream learning tasks or to aid in the evaluation of graph-based machine learning algorithms, we have also made some of the monthly builds available with embeddings. A modified version of DeepWalk[124] was used to create node embeddings for the v1.0.0 PKT Human Disease benchmark KGs. Embeddings were trained using 128, 256, and 512 dimensions (i.e., the length of the embedding), 100 walks (i.e., the number of paths generated for each node), a walk length of 20 (i.e., the length or number of nodes included in each path), and a sliding window length of 10 (i.e., the number of nodes to the right and left of the target node, which are used as training data for the target node embedding).

Eleven monthly PKT Human Disease benchmark KG builds were created between September 2, 2019 and November 1, 2021, each containing 12 different types of KGs. Each monthly build was executed using GitHub Actions-scheduled Cron jobs and implemented using dedicated Docker containers, which output all data directly to a Google Cloud Storage (GCS) Bucket . The PKT Human Disease benchmark KGs, metadata, and logs are made available through a dedicated Zenodo Community (https://zenodo.org/communities/pheknowlator-benchmark-human-disease-kg; build archive repository116).

***Component 3: Knowledge Graph Tools***

This component consists of tools to analyze and use KGs (Figure 3), which includes Jupyter Notebook-based use cases and tutorials, cloud-based data storage, APIs, and triplestores. The features that support these elements are detailed in the *ecosystem Component 3: Knowledge Graph Tools* section of Supplementary Table 3. The Jupyter Notebooks are available on GitHub and currently include tutorials on using OWL-NETS (OWLNETS_Example_Application.ipynb[125]), querying an RDF KG (RDF_Graph_Processing_Example.ipynb[126]), and searching for paths between two entities in a PKT Human Disease KG (Entity_Search.ipynb[127]). As mentioned above, data are publicly available through the PKT Human Disease benchmark KGs Zenodo Community archive (https://zenodo.org/communities/pheknowlator-benchmark-human-disease-kg). Code is provided within the GitHub repository to build and host a SPARQL Endpoint (http://sparql.pheknowlator.com/). The Database Center for Life Science SPARQL proxy web application[128] is used as the front end, and the data is served from a Blazegraph triplestore[129].

***FAIR Data Principles***

The PheKnowLator ecosystem is built on the FAIR principles[43] (Supplementary Figure 3)[43]. **Findability.** Unique persistent identifiers are used for all data (i.e., downloaded, processed, and generated), metadata (i.e., for all downloaded and processed resources, data quality reports, and logged processes), and infrastructure (i.e., Docker containers, compute instances, and KG builds run via GitHub Actions[130] and the Google AI Platform[131]). All benchmark KGs are built using standardized and persistent node and relation identifiers. **Accessibility.** All data (i.e., downloaded, processed, and generated), constructed KGs, and metadata generated during the KG construction process, are publicly available and accessible via RESTful API access to a dedicated Zenodo Community archive. Additionally, all builds are versioned on GitHub, Google's Container Registry[132], and DockerHub[133]. Finally, PheKnowLator provides Jupyter Notebooks and automated dependency generation scripts to improve the usability of its resources. **Interoperability.** The PheKnowLator ecosystem is built on Semantic Web standards, the KGs benchmarks and construction processes are grounded in OBO Foundry ontologies, and, whenever possible, standard identifiers are assigned for all database resources. Additionally, all constructed KGs and KG metadata are output to a variety of standardized file formats like RDF/XML, N-Triples, JSON, and text files. **Reusability.** Benchmark KG builds are automated, containerized, and deployed through GitHub Actions workflows, which makes the build process and resulting KGs consistent across versions. Semantic Versioning[134] is used for all code and documentation. The ecosystem is licensed (Apache-2.0[135]), and all ingested data sources are described transparently on the ecosystem's GitHub wiki by build version (https://github.com/callahantiff/PheKnowLator/wiki).

## Evaluation

The PheKnowLator ecosystem was evaluated in two ways: (1) **Systematic Comparison of Open-Source KG Construction Software.** Publicly available software to construct biomedical KGs was identified and systematically compared using a survey developed to assess the functionality, availability, usability, maturity, and reproducibility of each method. (2) **Human Disease KG Benchmark Comparison and Construction Performance.** The computational performance of the ecosystem was assessed when used to construct 12 benchmark KGs designed to represent the molecular mechanisms underlying human disease. The resources used for each task are listed in Supplementary Table 4.

### ***Systematic Comparison of Open-Source KG Construction Software***

A systematic comparison was performed to examine how the PheKnowLator ecosystem compared to existing open-source biomedical KG construction methods available on GitHub. To provide an unbiased comparison, no assumptions were made regarding a specific set of user requirements. Instead, the goal of the comparison was to provide a detailed overview of existing methods. A survey[136] was constructed from five criteria (adapted from the evaluation methodology of Babar et al.[137]) including: KG construction functionality, maturity, availability, usability, and reproducibility. Example questions used to assess each criterion are provided in Supplementary Table 5. The full set of survey questions (n=44) are available as a Google Form from Zenodo[136]. Existing open-source biomedical KG construction methods were identified by performing a keyword search against the GitHub API. The following words were combined to form 31 distinct keyword phrases, which were queried against existing GitHub repository descriptions and README content: "biological", "bio", "medical", "biomedical", "life science", "semantic", "knowledge graph", "kg", "graph", "network", "build", "construction", "construct", "create", "creation". The GitHub scraper is publicly available from Zenodo[138] and was run in May 2020. The systematic comparison was completed in May 2020 (and updated in June 2021) by TJC with consultation and oversight from WAB and LEH. The survey was scored out of a total score of five points, which was derived as the sum of the ratio of coverage out of one point for each category (i.e., the number of answerable questions out of the number of questions for that category): KG Construction Functionality (10 questions); Availability (two questions); Usability (nine questions); Maturity (five questions); and Reproducibility (six questions). The GitHub scraper and survey results are available from Zenodo[136].

### ***Human Disease Knowledge Graph Benchmark Comparison and Construction Performance***

Performance metrics were evaluated when building the PKT Human Disease benchmark KGs (v2.1.0 April 11, 2021; testing version not officially released, logs and descriptives statistics available from Zenodo[139]), which included total runtime (minutes) and minimum, maximum, and average memory use (GB). The PKT Human Disease benchmark KGs (v2.1.0 May 1, 2021) were used to compare builds and produce descriptive statistics. Statistics were calculated to help characterize each benchmark KG, including counts of nodes, edges or triples, self-loops, average degree, the number of connected components, and the density. The semantically abstracted (with and without knowledge model harmonization) PKT Human Disease benchmark KGs were visualized and examined for patterns. The v2.1.0_01MAY2021 PKT Human Disease benchmark KGs are publicly available in several formats from Zenodo[140–147]. Additional build details, including data sources, build metadata, and logs, can be found on GitHub (https://github.com/callahantiff/PheKnowLator/wiki/May-01%2C-2021).

## Technical Specifications

The PheKnowLator ecosystem resources, including data used to construct KGs and constructed PKT Human Disease benchmark KGs, and code are listed by component in Supplementary Table 3. The PKT Human Disease KG builds were visualized using Gephi[148] (v0.9.2). The OpenOrd Force-Directed layout[149] was applied with an edge cut of 0.5, a fixed time of 0.2, and trained for 750 iterations. To help with interpretation, nodes were colored according to node type. When assessing computational performance, all PKT Human Disease KGs were constructed using Docker (v19.03.8) on a Google Cloud Platform N1 Container-Optimized OS instance configured with 24 CPUs, 500 GB of memory, and a 500 GB solid-state drive Boot Disk. PKT Human Disease KG statistics were calculated using Networkx (v2.4).

## Data Availability

***PKT Human Disease Benchmark KG Archive Resources***

Eleven monthly PKT Human Disease benchmark KG builds were created between September 2, 2019 and November 1, 2021. Each monthly build contains 12 different benchmarks or types of KGs, which were created by altering the following KG construction parameters: knowledge model (i.e., class- or instance-based), relation strategy (i.e., standard directed relations or inverse bidirectional relations), and semantic abstraction (i.e., transformation of complex graphs into OWL-NETS hybrid KGs) with or without knowledge model purification (i.e., ensuring a OWL-NETS KG is consistent with the knowledge model it was abstracted from). The 12 different KG types created by altering these parameters are:

1. Class-based knowledge + Standard Relations + OWL
2. Class-based knowledge + Standard Relations + OWL-NETS
3. Class-based knowledge + Standard Relations + OWL-NETS (Purified)
4. Class-based knowledge + Inverse Relations + OWL
5. Class-based knowledge + Inverse Relations + OWL-NETS
6. Class-based knowledge + Inverse Relations + OWL-NETS (Purified)
7. Instance-based knowledge + Standard Relations + OWL
8. Instance-based knowledge + Standard Relations + OWL-NETS
9. Instance-based knowledge + Standard Relations + OWL-NETS (Purified)
10. Instance-based knowledge + Inverse Relations + OWL
11. Instance-based knowledge + Inverse Relations + OWL-NETS
12. Instance-based knowledge + Inverse Relations + OWL-NETS (Purified)

The builds are available through a Zenodo Community archive (https://zenodo.org/communities/pheknowlator-benchmark-human-disease-kg) with all builds listed and linked on the primary archive page[116]. The monthly builds can also be accessed through the PheKnowLator GitHub wiki (https://github.com/callahantiff/PheKnowLator/wiki/Archived-Builds). The GitHub wiki build pages serve as a companion resource to each corresponding Zenodo build archive providing detailed descriptions of the output data files, links to input data sources and Jupyter notebook-based workflows, and lists of generated metadata and logs. Each wiki build page also includes direct links to each of the 12 benchmark KGs on Zenodo. A detailed description of the build KG files types, including required input documents and curated data, generated build metadata and logs, and output KG data files can be found in the PheKnowLator_HumanDiseaseKG_Output_FileInformation.xlsx[116] file available on the Zenodo Community archive and from each build's GitHub wiki page. Please note that the Zenodo Community archive and associated GitHub pages list 8 KG types rather than 12 as the non-purified and purified OWL-NETS KG type files are combined into a single repository.

*Build Data Sources*

The curated data sources required for each build are provided in the Zenodo Community archive. All other data are not included with each monthly build due to the large number of required files and their size. These files are all publicly available and can be obtained using information provided with each build, including URL and date of download. For the September 3, 2021 build, links and date of download information are provided on the GitHub wiki. For the May 01, 2020 build, see the build metadata (edge_source_metadata.txt and ontology_source_metadata.txt) and logs (*_Stats_Terminal_Output.txt), which are all available from the Zenodo Community archive. For all other builds, see downloaded_build_metadata.txt, also available from the Zenodo Community archive. A detailed description of the data sources used to build the PKT Human Disease KG is provided in Supplemental Material Table 12. This table includes the following information for each data source: data provider, filenames, download URLs, literature citations, license types, and a brief description of how each data source was used.

***Evaluation Resources***

The v2.1.0 May 2021 PKT Human Disease benchmark KGs are available through the GitHub wiki ( https://github.com/callahantiff/PheKnowLator/wiki/May-01%2C-2021), and on Zenodo by KG type (Class-based + StandardRelations + OWL[140]; Class-based + StandardRelations + OWL-NETS[141]; Class-based + InverseRelations + OWL[142]; Class-based + InverseRelations + OWL-NETS[143]; Instance-based + StandardRelations + OWL[144]; Instance-based + StandardRelations + OWL-NETS[145]; Instance-based + InverseRelations + OWL[146]; Instance-based + InverseRelations + OWL-NETS[147]). Build logs and statistics are also available for the v2.1.0 April 2012 PKT Human Disease benchmark KGs on Zenodo[139]. As mentioned in the prior section, a table describing the output file types for each build type can be found on the Zenodo Community archive (https://zenodo.org/communities/pheknowlator-benchmark-human-disease-kg)[116]. Descriptions of the data sources used to build the PKT Human Disease KG are provided in Supplemental Material Table 12.

## Code Availability

The PheKnowLator ecosystem coding resources are described in detail in Supplementary Table 3 by ecosystem component. The PKT-KG algorithm is publicly available through GitHub (https://github.com/callahantiff/PheKnowLator) and PyPI (https://pypi.org/project/pkt-kg). The SPARQL Endpoint deployment code and documentation are also available through GitHub: https://github.com/callahantiff/PheKnowLator/tree/master/builds/deploy/triple-store#readme. A list of the computational resources used to evaluate the PheKnowLator ecosystem is provided in Supplementary Table 4. The code used to scrape the GitHub API is available from Zenodo[138]. The survey of open-source KG construction tools is also available on Zenodo[136]. The v2.1.0 PheKnowLator code is available on GitHub (https://github.com/callahantiff/PheKnowLator/releases/tag/v2.1.0) and from Zenodo (https://zenodo.org/record/4685943)[150].

## Acknowledgements
This work was supported by funding from the National Library of Medicine (T15LM009451 and T15LM007079) to TJC, (4R00LM013367-03) to SAM, (R01LM013400 and 5R01LM008111-16) to LEH, and (R01LM006910) GH. This work was also supported by funding from the Director, Office of Science, Office of Basic Energy Sciences, of the U.S. Department of Energy under Contract (DE-AC02-05CH11231) to JR, the King Abdullah University of Science and Technology (KAUST) Office of Sponsored Research (OSR), Award (URF/1/4355-01-01 and URF/1/5041-01-01) to RH, the NIH Common Fund (CFDE OT2OD030545, HuBMAP OT2OD033759, and SenNet U24CA268108) to JCS, the National Center for Complementary and Integrative Health (U54 AT008909) to SBT and RDB, a National Recovery and Resilience Plan-NextGenerationEU award from the National Center for Gene Therapy and Drugs based on RNA Technology (G43C22001320007) to GV, and the Defense Advanced Research Projects Agency (DARPA) Young Faculty Award (W911NF-20-1-0255) and the DARPA Automating Scientific Knowledge Extraction and Modeling program (HR00112220036) to CTH. The authors would like to thank the OHDSI community, especially Adam Black as well as members of Dr. Hunter's lab at the University of Colorado Anschutz Medical Campus, specifically Drs. Mayla Boguslav and Harrison Pielke-Lombardo for testing different builds and helping conceive and pilot test tutorials to demonstrate different PheKnowLator use cases. The authors would also like to thank GitHub users ablack3, Bancherd-DeLong, Bsantan, GuarinoValentina, and nomisto, who identified and helped troubleshoot bugs through the PheKnowLator GitHub.

## Author Contributions
MGK, WAB, and LEH served as primary supervisors of this work. TJC, BAW, ALS, IJT, RH, and ALS conceived and helped develop the analyses performed in this work. TJC and WAB developed the PheKnowLator ecosystem. ALS, IJT, and JMW provided insight into the development of documentation for the GitHub site. ALS, BS, CJM, CTH, FM, GH, JCS, JH, JMW, JR, MB, NAM, NAV, PBR, PNR, RDB, RH, and TDB provided domain expertise and/or commented on the PheKnowLator ecosystem, data sources, or other important resources used in its development. BS, ElC, EC, GV, LC, LG, MM, RB, SAM, SBT, and TF evaluated PheKnowLator builds and provided feedback on the resulting KGs. TJC drafted the manuscript and all authors reviewed the manuscript and provided feedback. All authors read and approved the final version of the manuscript.

## Competing Interests
The authors declare no competing interests.

**Table 1. Open-Source Knowledge Graph Construction Methods.**

| Method | GitHub repository |
|---|---|
| Bio2BEL | https://github.com/bio2bel/ |
| Bio2RDF | https://github.com/bio2rdf |
| Bio4J | https://github.com/bio4j/bio4j |
| BioGrakn | https://github.com/graknlabs/biograkn |
| Clinical Knowledge Graph (CKG) | https://github.com/MannLabs/CKG |
| COVID-19-Community | https://github.com/covid-19-net/covid-19-community |
| Dipper | https://github.com/monarch-initiative/dipper |
| Hetionet | https://github.com/hetio/hetionet |
| iASiS Open Data Graph | https://github.com/tasosnent/Biomedical-Knowledge-Integration |
| KG-COVID-19 | https://github.com/Knowledge-Graph-Hub/kg-covid-19 |
| Knowledge Base Of Biomedicine (KaBOB) | https://github.com/UCDenver-ccp/kabob/tree/bg-integration |
| Knowledge Graph Exchange (KGX) | https://github.com/NCATS-Tangerine/kgx |
| Knowledge Graph Toolkit (KGTK) | https://github.com/usc-isi-i2/kgtk/ |
| ProNet | https://github.com/cran/ProNet |
| SEmantic Modeling machIne (SEMi) | https://github.com/giuseppefutia/semi |

**Table 2. PKT Human Disease Knowledge Graph Primary Node Types.**

| Node | Universal Resource Identifier |
|---|---|
| Anatomical Entities | http://purl.obolibrary.org/obo/UBERON |
| Biological Processes | http://purl.obolibrary.org/obo/GO |
| Catalysts | http://purl.obolibrary.org/obo/CHEBI |
| Cells | http://purl.obolibrary.org/obo/CL |
| Cell Lines | http://purl.obolibrary.org/obo/CLO |
| Cellular Components | http://purl.obolibrary.org/obo/GO |
| Chemicals | http://purl.obolibrary.org/obo/CHEBI |
| Cofactors | http://purl.obolibrary.org/obo/CHEBI |
| Diseases | http://purl.obolibrary.org/obo/MONDO |
| Genes | http://www.ncbi.nlm.nih.gov/gene/ |
| Molecular Functions | http://purl.obolibrary.org/obo/GO |
| Pathways[a] | http://purl.obolibrary.org/obo/PW<br>https://reactome.org/content/detail/R-HSA- |
| Phenotypes | http://purl.obolibrary.org/obo/HP |
| Proteins | http://purl.obolibrary.org/obo/PR |
| Sequences[b] | http://purl.obolibrary.org/obo/SO |
| Transcripts | https://uswest.ensembl.org/Homo_sapiens/Transcript/Summary?t=ENST |
| Vaccines[b] | http://purl.obolibrary.org/obo/VO |
| Variants | https://www.ncbi.nlm.nih.gov/snp/rs |

Note: The node types listed above apply to the PKT Human Disease KG v2.1.0. The node types listed above do not include all of the classes that exist in each Open Biological and Biomedical Ontology (OBO) Foundry ontology. The Cell Ontology is included with the extended version of Uberon.

[a]Two URIs are shown for pathways as the OBO Found ontology is the core ontology used to connect Reactome entities to the core set of OBO Foundry ontologies.

[b]OBO node type. Includes all of the classes that are contained in the ontology even though they are not all explicitly listed here.

Acronyms: CL (Cell ontology); CLO (Cell Line Ontology); CHEBI (Chemical Entities of Biological Interest); GO (Gene Ontology); HPO (Human Phenotype Ontology); MONDO (Mondo Disease Ontology); PKT (PheKnowlator); PRO (Protein Ontology); PW (Pathway Ontology); SO (Sequence Ontology); VO (Vaccine Ontology); UBERON (Uber-Anatomy Ontology).

**Table 3. PKT Human Disease Knowledge Graph Primary Edge Types by Relation.**

| Relations | Edge Types |
|---|---|
| participates in (RO_0000056)<br>has participant (RO_0000057) | chemical-pathway; gene-pathway; protein-biological process; protein-pathway |
| has function (RO_0000085)<br>function of (RO_0000079) | pathway-molecular function; protein-molecular function |
| located in (RO_0001025)<br>location of (RO_0001015) | protein-anatomy; protein-cell[a]; protein-cellular component; transcript-anatomy; transcript-cell[a] |
| has component (RO_0002180)[b] | pathway-cellular component |
| has phenotype (RO_0002200)<br>phenotype of (RO_0002201) | disease-phenotype |
| has gene product (RO_0002205)<br>gene product of (RO_0002204) | gene-protein |
| interacts with (RO_0002434)[c] | chemical-gene; chemical-protein |
| genetically interacts with (RO_0002435)[c] | gene-gene |
| molecularly interacts with (RO_0002436)[c] | chemical-biological process; chemical-cellular component; chemical-molecular function; protein-catalyst; protein-cofactor; protein-protein |
| transcribed to (RO_0002511)<br>transcribed from (RO_0002510) | gene-transcript |
| ribosomally translates to (RO_0002513)<br>ribosomal Translation of (RO_0002512) | transcript-protein |
| causally influences (RO_0002566)<br>causally influenced by (RO_0002559) | variant-gene |
| is substance that treats (RO_0002606)<br>is treated by substance (RO_0002302) | chemical-disease; chemical-phenotype |
| causes or contributes to condition (RO_0003302)[b] | gene-disease; gene-phenotype; variant-disease; variant-phenotype |
| realized in response to (RO_0009501)[b] | biological process-pathway |

Note: The primary relations and edge types listed above apply to the PKT Human Disease KG v2.1.0. These relations are added to the core set of Open Biological and Biomedical Ontology Foundry ontologies.

[a]The word “cell” above is used to represent cell lines from the Cell Line Ontology and cell types from the Cell Ontology.

[b]Relation Ontology concepts that do not have an inverse.

[c]Relations with symmetrical inverse relations.

Acronyms: PKT (PheKnowLator).

**Table 4. Ontology Statistics Pre- and Post-Data Quality Assessment.**

| Ontology | Before Cleaning | | After Cleaning | |
|---|---|---|---|---|
| | Classes | Triples | Classes | Triples |
| Cell Line Ontology | 111,712 | 1,387,096 | 111,696 | 1,422,153 |
| Chemical Entities of Biological Interest | 156,098 | 5,264,571 | 137,592 | 5,190,485 |
| Gene Ontology | 62,237 | 1,425,434 | 55,807 | 1,343,218 |
| Human Phenotype Ontology | 38,843 | 884,999 | 38,530 | 885,379 |
| Mondo Disease Ontology | 55,478 | 2,313,343 | 52,937 | 2,277,425 |
| Protein Ontology[a] | 148,243 | 2,079,356 | 148,243 | 2,079,356 |
| Pathway Ontology | 2,642 | 35,291 | 2,600 | 34,901 |
| Relation Ontology | 116 | 7,970 | 115 | 7,873 |
| Sequence Ontology | 2,910 | 44,655 | 2,569 | 41,980 |
| Uber-Anatomy Ontology[b] | 28,738 | 752,291 | 27,170 | 734,768 |
| Vaccine Ontology | 7,089 | 86,454 | 7,085 | 89,764 |
| Core OBO Foundry ontologies (merged)[c] | 548,947 | 13,746,883 | 545,259 | 13,748,009 |

Note: The numbers for the ontologies are calculated using the versions of the ontologies that include all imported ontologies referenced by the primary ontology. This means that the counts of classes include all OWL classes used for logical definitions, not only those that are explicitly part of the primary ontology's namespace.

[a]The Protein Ontology version references the human subset created for the PheKnowLator ecosystem.

[b]The extended version of the Uber-Anatomy Ontology contains the Cell Ontology.

[c]Consistency was evaluated using the ELK reasoner. The reasoner was only applied to individual ontologies.

**Table 5. PKT Human Disease Knowledge Graph Descriptive Statistics by Primary Edge Type.**

| Edge | Relation | Subjects | Objects | Standard Relations | Inverse Relations |
|---|---|---|---|---|---|
| chemical-disease | substance that treats | 4,289 | 4,494 | 167,681 | 335,362 |
| chemical-gene[a] | interacts with | 462 | 11,922 | 16,639 | 33,278 |
| chemical-biological process[a] | molecularly interacts with | 1,338 | 1,569 | 287,068 | 574,136 |
| chemical-cellular component[a] | molecularly interacts with | 1,085 | 226 | 40,992 | 81,984 |
| chemical-molecular function[a] | molecularly interacts with | 1,105 | 200 | 25,385 | 50,770 |
| chemical-pathway | participates in | 2,104 | 2,213 | 28,685 | 57,370 |
| chemical-phenotype | substance that treats | 4,053 | 1,712 | 107,962 | 215,924 |
| chemical-protein[a] | interacts with | 4,178 | 6,379 | 64,991 | 129,982 |
| disease-phenotype | has phenotype | 11,620 | 9,714 | 408,702 | 817,404 |
| gene-disease[b] | causes or contributes to | 5,031 | 4,420 | 12,717 | --- |
| gene-gene[a] | genetically interacts with | 247 | 263 | 1,668 | 3,336 |
| gene-pathway | participates in | 10,371 | 1,809 | 104,906 | 209,812 |
| gene-phenotype[b] | causes or contributes to | 6,780 | 1,528 | 23,501 | --- |
| gene-protein | has gene product | 19,327 | 19,143 | 19,534 | 39,068 |
| gene-transcript | transcribed to | 25,529 | 179,870 | 182,736 | 365,472 |
| biological process-pathway[b] | realized in response to | 471 | 665 | 665 | --- |
| pathway-cellular component[b] | has component | 11,134 | 99 | 15,846 | --- |
| pathway-molecular function | has function | 2,412 | 726 | 2,416 | 4,832 |
| protein-anatomy | located in | 10,747 | 68 | 30,682 | 61,364 |
| protein-catalyst[a] | molecularly interacts with | 3,024 | 3,730 | 23,629 | 47,258 |
| protein-cell[c] | located in | 10,045 | 125 | 73,530 | 147,060 |
| protein-cofactor[a] | molecularly interacts with | 1,584 | 44 | 1,961 | 3,922 |
| protein-biological process | participates in | 17,527 | 12,246 | 137,812 | 275,624 |
| protein-cellular component | located in | 18,427 | 1,757 | 81,602 | 163,204 |
| protein-molecular function | has function | 17,779 | 4,324 | 68,633 | 137,266 |
| protein-pathway | participates in | 10,852 | 2,468 | 117,182 | 234,364 |
| protein-protein[d] | molecularly interacts with | 14,320 | 14,230 | 618,069 | --- |
| transcript-anatomy | located in | 29,104 | 102 | 439,917 | 879,834 |
| transcript-cell[c] | located in | 14,038 | 127 | 64,427 | 128,854 |
| transcript-protein | ribosomally translates to | 44,144 | 19,200 | 44,147 | 88,294 |
| variant-disease[b] | causes or contributes to | 13,291 | 3,565 | 37,861 | --- |
| variant-gene | causally influences | 121,790 | 3,236 | 121,790 | 243,580 |
| variant-phenotype[b] | causes or contributes to | 1,822 | 371 | 2,470 | --- |

Please see Table 3 for Relation Ontology for inverse relations and identifiers.
[a]Symmetric relations were computationally inferred.
[b]The Relation Ontology does not provide an inverse relation.
[c]The word "cell" above is used to represent cell lines from the Cell Line Ontology and cell types from the Cell Ontology.
[d]The data source already included symmetrical edges.
Acronyms: PKT (PheKnowlator).

**Table 6. PheKnowLator Human Disease Knowledge Graph Descriptive Statistics.**

| Knowledge Model | Relation Strategy | Semantic Abstraction | Edges (triples) | Nodes | Relations | Self-Loops | Average Degree |
|---|---|---|---|---|---|---|---|
| [a]Core OBO Foundry ontologies | N/A | N/A | 4,044,658 | 1,399,756 | 847 | 3 | 2.89 |
| Class-based | Standard Relations | None | 25,143,729 | 8,479,167 | 847 | 3 | 2.97 |
| | | Semantic Abstraction Only | 4,967,427 | 743,829 | 294 | 445 | 6.68 |
| | | Semantic Abstraction + Harmonization | 4,967,429 | 743,829 | 293 | 445 | 6.68 |
| | Inverse Relations | None | 41,116,791 | 13,803,521 | 847 | 3 | 2.98 |
| | | Semantic Abstraction Only | 7,629,597 | 743,829 | 301 | 445 | 10.26 |
| | | Semantic Abstraction + Harmonization | 7,629,599 | 743,829 | 300 | 445 | 10.26 |
| Instance-based | Standard Relations | None | 21,770,455 | 8,479,167 | 847 | 3 | 2.57 |
| | | Semantic Abstraction Only | 4,967,391 | 743,829 | 294 | 409 | 6.68 |
| | | Semantic Abstraction + Harmonization | 7,285,496 | 743,829 | 293 | 649 | 9.79 |
| | Inverse Relations | None | 24,432,633 | 8,479,167 | 847 | 3 | 2.88 |
| | | Semantic Abstraction Only | 7,629,594 | 743,829 | 301 | 409 | 10.26 |
| | | Semantic Abstraction + Harmonization | 9,624,232 | 743,829 | 300 | 650 | 12.94 |

Note. Edges and triples are synonymous with respect to the results reported in this table.

[a]Relation Strategy and Semantic Abstraction information are not provided as this row of the table reports information on the core set of merged ontologies.

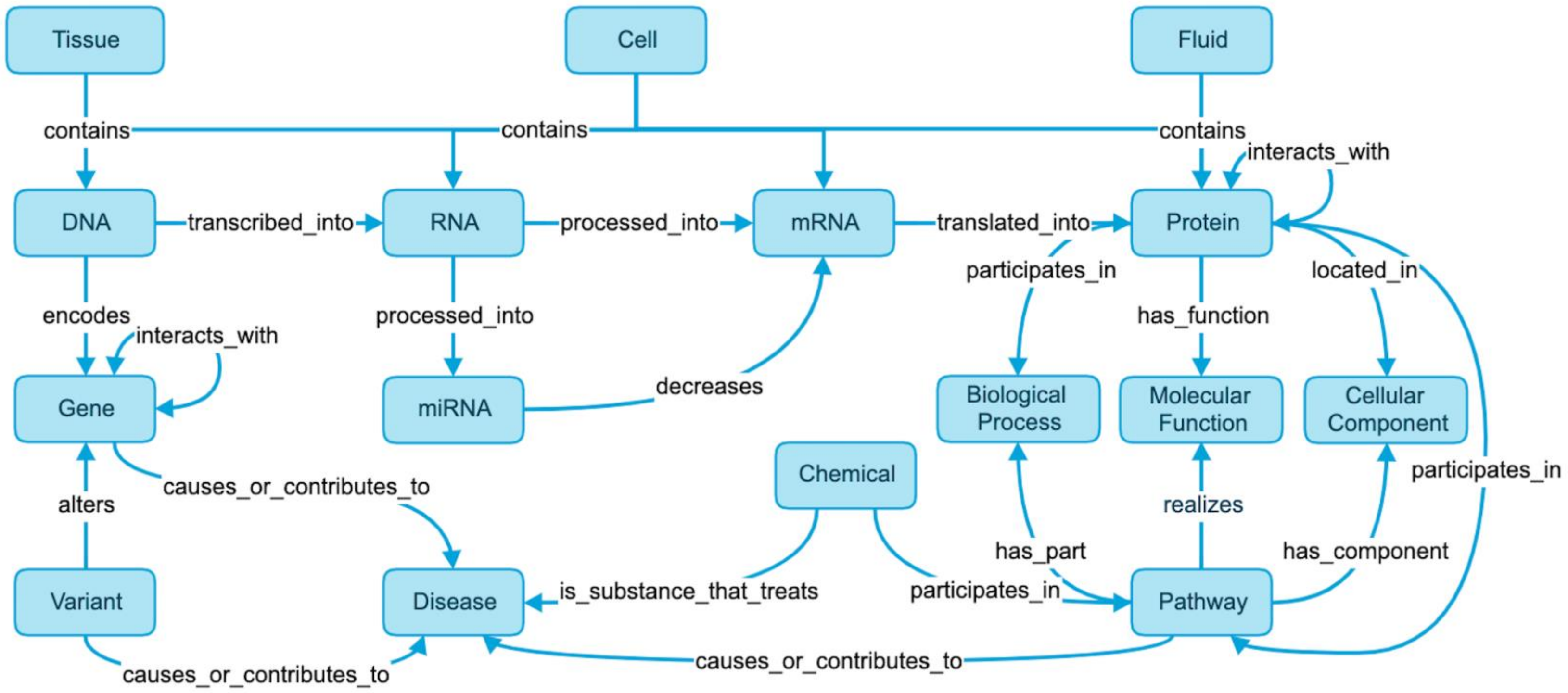

**Figure 1. A Knowledge Representation of the Levels of Biological Organization Underlying Human Disease.**

This knowledge graph provides a representation of our currently accepted knowledge of the Central Dogma expanded to include pathways, variants, pharmaceutical treatments, and diseases.[10] At a high level this knowledge graph represents anatomical entities such as tissues, cells, and bodily fluids containing genomic entities such as DNA, RNA, mRNA, and proteins. DNA encodes genes that are processed into mRNA and translated into proteins, which can interact with each other. Genes can also be altered by variants and may cause disease. Finally, proteins also have molecular functions and participate in pathways and biological processes.

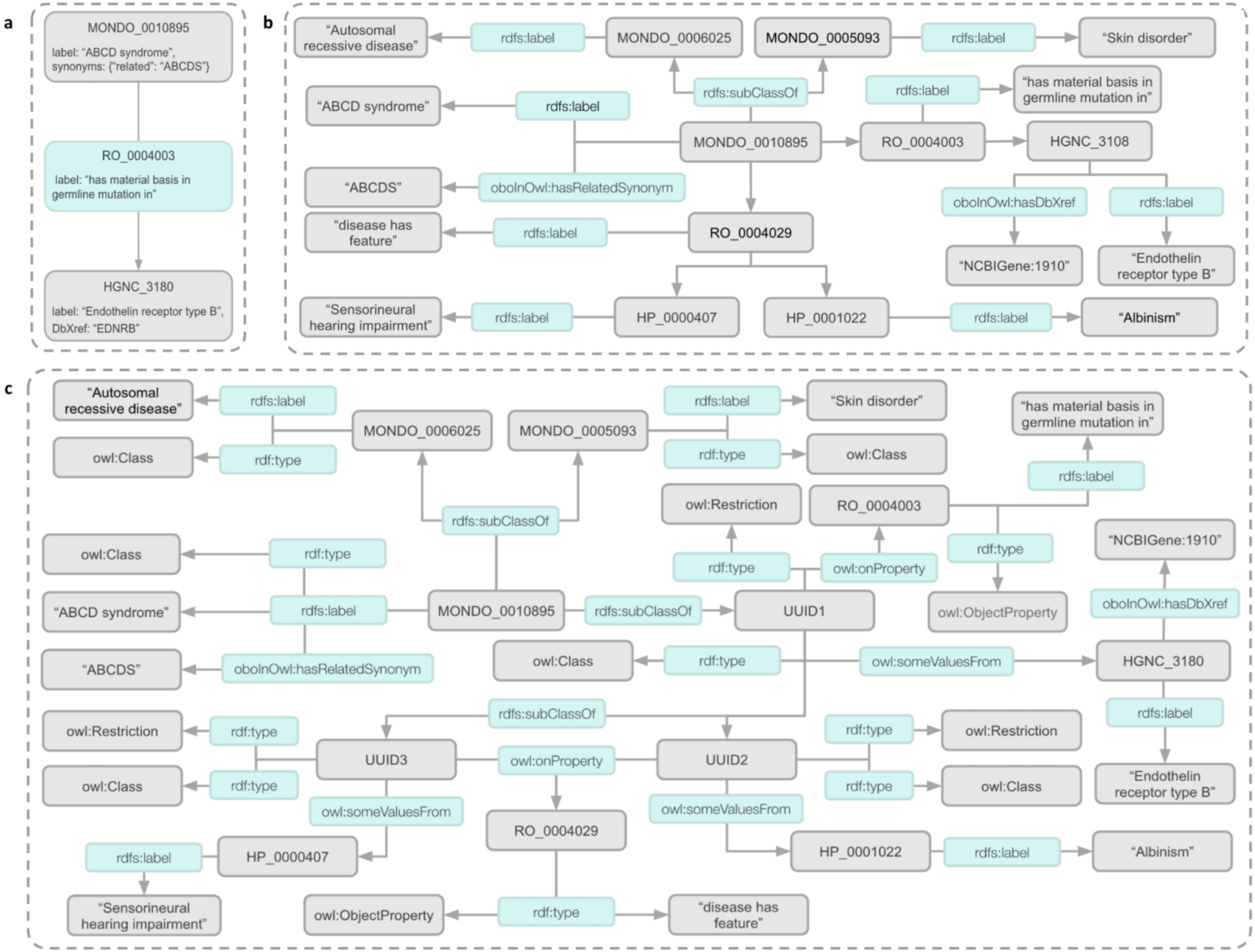
a
MONDO_0010895
label: "ABCD syndrome",
synonyms: {"related": "ABCDS"}
RO_0004003
label: "has material basis in germline mutation in"
HGNC_3180
label: "Endothelin receptor type B",
DbXref: "EDNRB"
b
"Autosomal recessive disease"
rdfs:label
MONDO_0006025
MONDO_0005093
rdfs:label
"Skin disorder"
rdfs:subClassOf
rdfs:label
"has material basis in germline mutation in"
"ABCD syndrome"
rdfs:label
MONDO_0010895
RO_0004003
HGNC_3108
"ABCDS"
oboInOwl:hasRelatedSynonym
oboInOwl:hasDbXref
rdfs:label
"disease has feature"
rdfs:label
RO_0004029
"NCBIGene:1910"
"Endothelin receptor type B"
"Sensorineural hearing impairment"
rdfs:label
HP_0000407
HP_0001022
rdfs:label
"Albinism"
c
"Autosomal recessive disease"
rdfs:label
MONDO_0006025
MONDO_0005093
rdfs:label
"Skin disorder"
owl:Class
rdf:type
rdf:type
owl:Class
"has material basis in germline mutation in"
rdfs:label
rdfs:subClassOf
owl:Restriction
RO_0004003
owl:Class
rdf:type
rdf:type
owl:onProperty
rdf:type
"NCBIGene:1910"
"ABCD syndrome"
rdfs:label
MONDO_0010895
rdfs:subClassOf
UUID1
owl:ObjectProperty
oboInOwl:hasDbXref
"ABCDS"
oboInOwl:hasRelatedSynonym
owl:Class
rdf:type
owl:someValuesFrom
HGNC_3180
rdfs:subClassOf
rdfs:label
owl:Restriction
rdf:type
rdf:type
owl:Restriction
UUID3
owl:onProperty
UUID2
owl:Class
rdf:type
rdf:type
owl:Class
"Endothelin receptor type B"
owl:someValuesFrom
owl:someValuesFrom
RO_0004029
rdfs:label
HP_0000407
HP_0001022
rdfs:label
"Albinism"
"Sensorineural hearing impairment"
owl:ObjectProperty
rdf:type
"disease has feature"

**Figure 2. Types of Knowledge Graphs used in the Life Sciences.**

This figure provides examples of three types of knowledge graphs that are typically used in the Life Sciences. All knowledge graphs are modeling the Mondo concept ABCD syndrome (*MONDO:0010895*). (**A**) illustrates a simple graph-based representation where two nodes are connected by an edge and nodes and edges are assigned attributes in the form of key-value pairs. (**B**) illustrates a hybrid or property graph-based representation where edges are represented as sets of three nodes (each composed of a subject, predicate, and object) called triples, often based on the RDF/RDFS standards. (**C**) illustrates a complex or OWL-graph-based representation where edges are represented as triples and these representations are augmented with additional OWL expressivities such as domain/range or cardinality restrictions. Acronyms: HP (Human Phenotype Ontology); MONDO (Mondo Disease Ontology); OWL (Web Ontology Language); RDF (Resource Description Framework); RDFS (Resource Description Framework Syntax); RO (Relation Ontology).

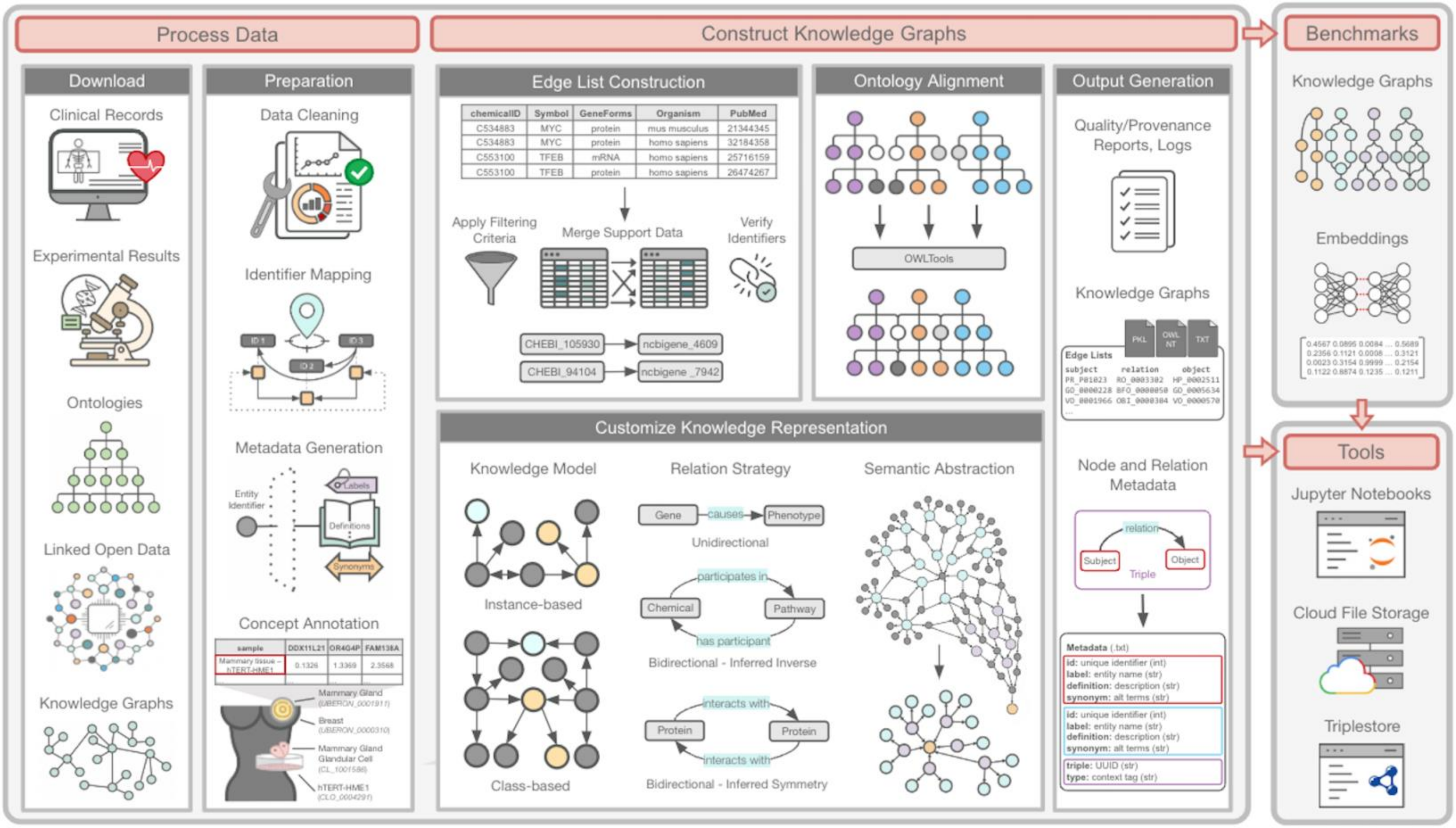

**Figure 3. The PheKnowLator Ecosystem.**

This figure provides an overview of the PheKnowLator ecosystem.[151] The ecosystem consists of three components as indicated by the gray boxes: (1) **Knowledge Graph Construction Resources**, which consist of resources to download and process data and an algorithm to customize the construction of large-scale heterogeneous biomedical knowledge graphs; (2) **Knowledge Graph Benchmarks**, which consist of prebuilt KGs that can be used to systematically assess the effects of different knowledge representations on downstream analyses, workflows, and learning algorithms; and (3) **Knowledge Graph Tools** to use knowledge graphs, cloud-based data storage, APIs, and triplestores. Acronyms: NT (N-Triples file format); OWL (Web Ontology Language); PKL (Python pickle file format); SPARQL (SPARQL Protocol and RDF Query Language).

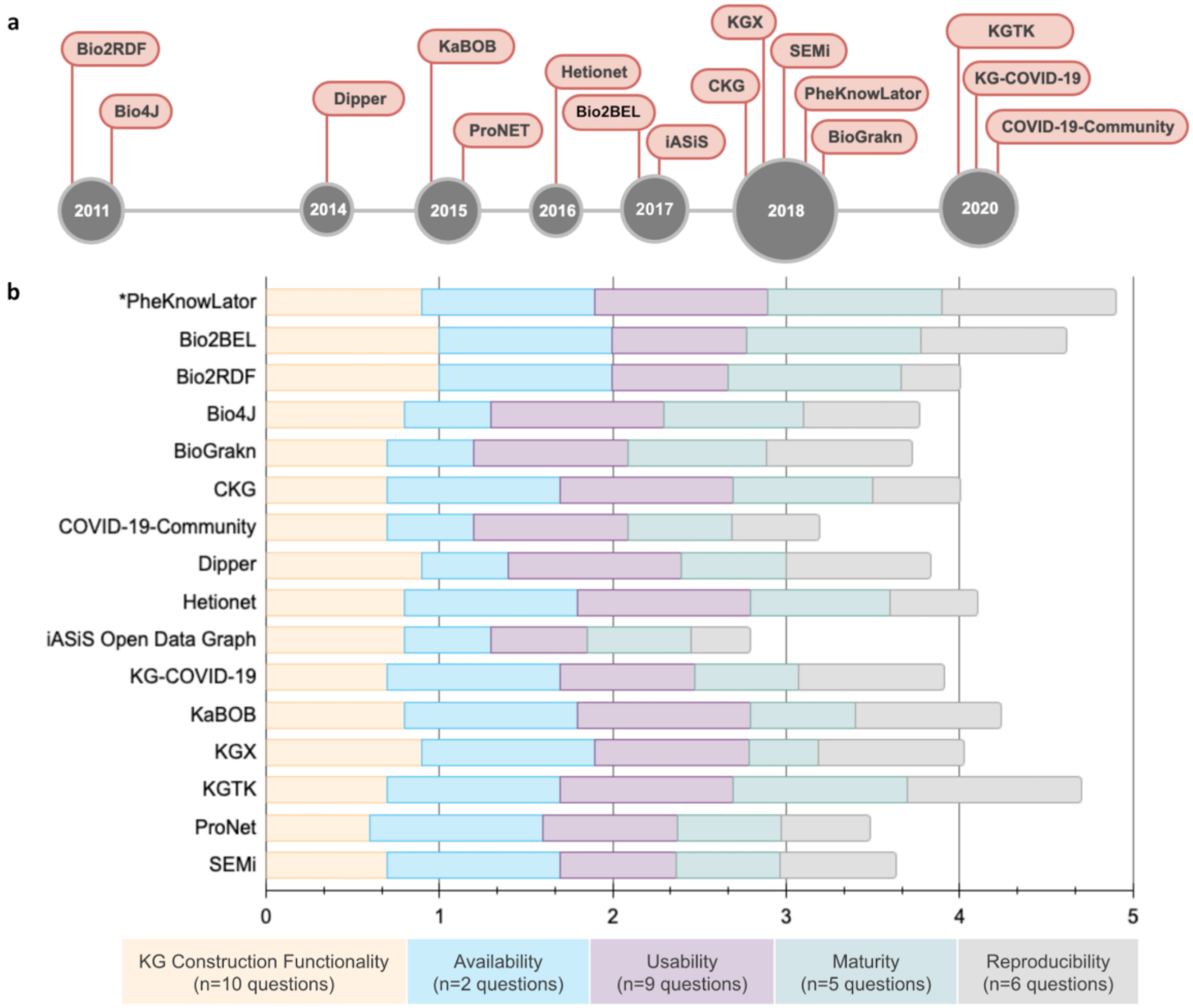

**Figure 4. Open-Source Knowledge Graph Construction Methods - Survey Results.**

This figure presents the open-source knowledge graph construction methods identified on GitHub and the results of the survey assessment. (**A**) The final set of 16 knowledge graph construction methods surveyed according to the year they were first published on GitHub. (**B**) A chart of the methods evaluated in terms of the different survey categories. The survey was scored out of a total score of five points, which was derived as the sum of the ratios of coverage, each out of one point, for the five categories: KG Construction Functionality (10 questions); Availability (two questions); Usability (nine questions); Maturity (five questions); and Reproducibility (six questions). Acronyms: iASiS, Automated Semantic Integration of Disease-Specific Knowledge; KaBOB, Knowledge Base Of Biomedicine; KG, (Knowledge Graph); KGX (Knowledge Graph Exchange); KGTK (Knowledge Graph Toolkit); SEMi (SEmantic Modeling machine).

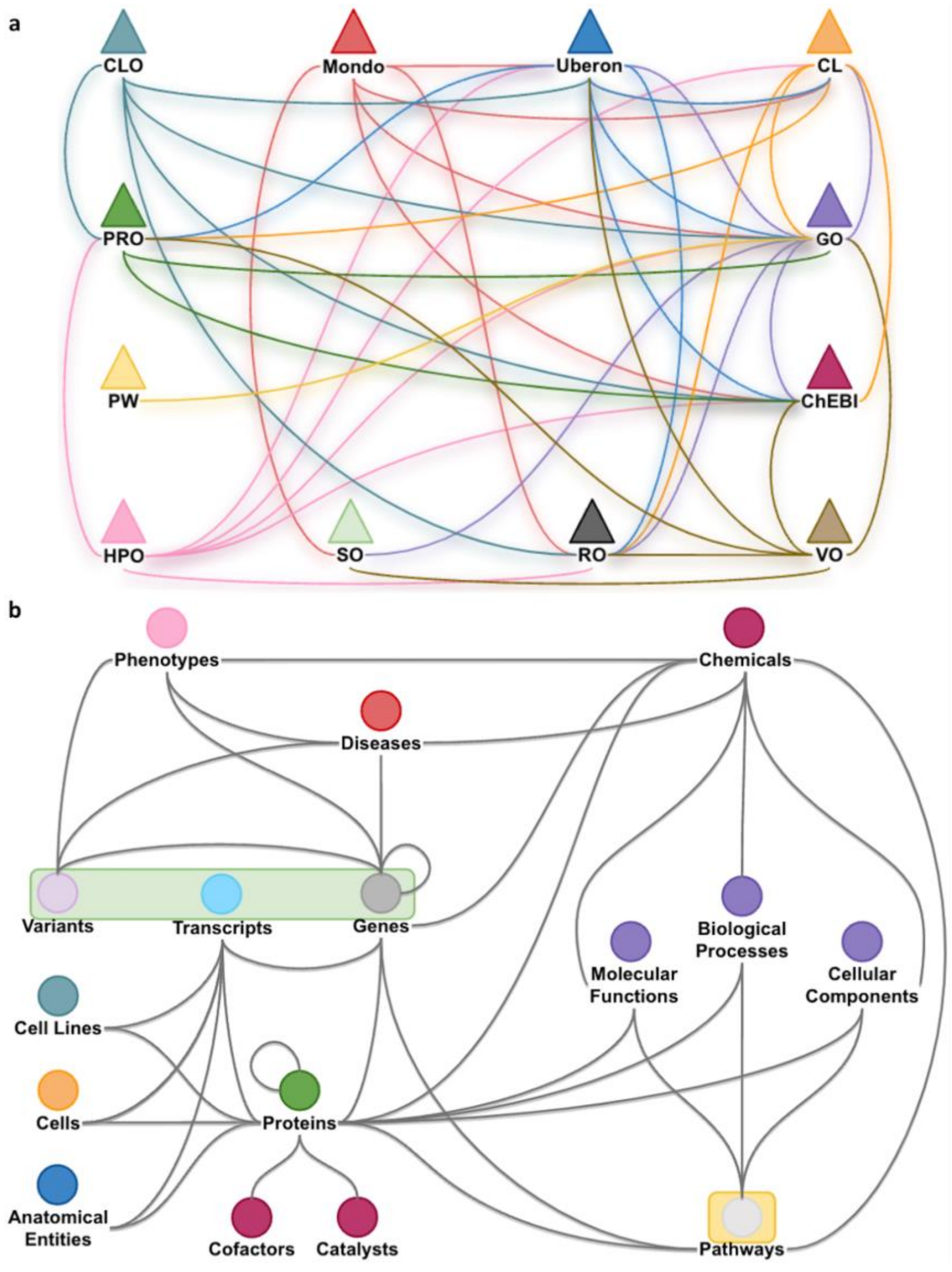

**Figure 5. An Overview of the PKT Human Disease Mechanism Knowledge Graph.**

This figure provides a high-level overview of the primary node and edge types in the PKT Human Disease Mechanism knowledge graph. (**A**) illustrates the relationships between the core set of Open Biological and Biomedical Ontology (OBO) Foundry ontologies when including their imported ontologies (as of August 2022). (**B**) illustrates the edges or triples that are added to the core set of merged ontologies in (A). Shared colors between (A) and (B) represent a single resource. For example, chemicals, cofactors, and catalysts share the same color (maroon) and are part of ChEBI. This is the same for the RO, which is represented in (B) as the black lines between nodes. The green and yellow rectangles indicate data sources that are not from an OBO Foundry ontology and the specific ontology used to integrate them with the core set of ontologies in (A). For example, variant, transcript, and gene data are connected to the core ontology set via the SO. Acronyms: CL (Cell ontology); CLO (Cell Line Ontology); ChEBI (Chemical Entities of Biological Interest); GO (Gene Ontology); HPO (Human Phenotype Ontology); Mondo (Mondo Disease Ontology); PRO (Protein Ontology); PW (Pathway Ontology); SO (Sequence Ontology); VO (Vaccine Ontology); Uberon (Uber-Anatomy Ontology).

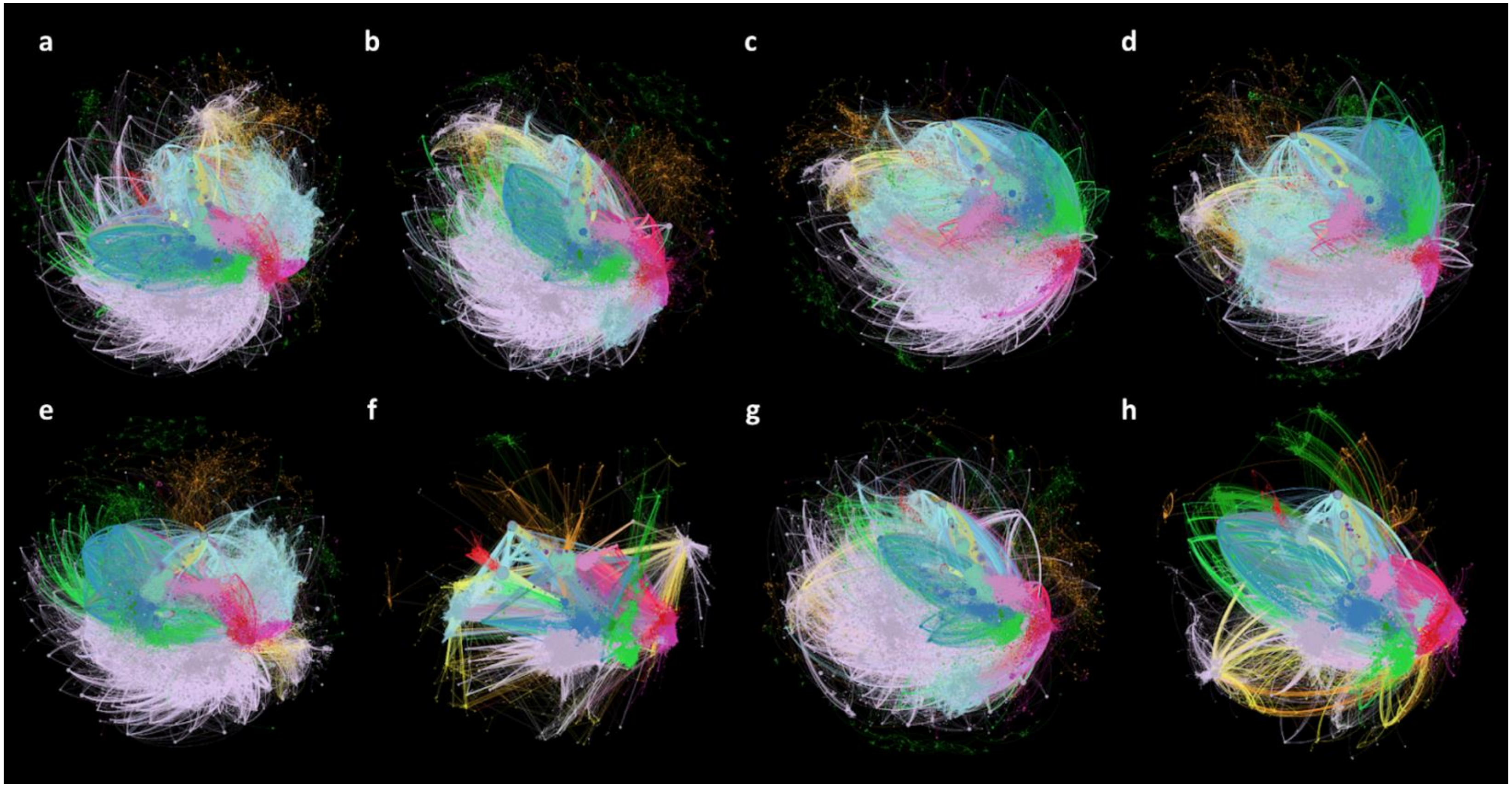

**Figure 6. The Impact of Knowledge Model Harmonization on the Semantically Abstracted PKT Human Disease Knowledge Graphs.**

The figure visualizes the impact of knowledge model harmonization on the semantically abstracted PKT Human Disease benchmark Knowledge Graphs. The top row of figures (A-D) were built using the class-based knowledge model varying: (**A**) standard relations without harmonization; (**B**) standard relations with harmonization; (**C**) inverse relations without harmonization; (**D**) inverse relations with harmonization. The bottom row of figures (E-H) were built using the instance-based knowledge model varying: (**E**) standard relations without harmonization; (**F**) standard relations with harmonization; (**G**) inverse relations without harmonization; (**H**) inverse relations with harmonization. Nodes are colored by type: anatomical entities (light blue), chemical entities (light purple), diseases (red), genes (purple), genomic features (light green), organisms (yellow), pathways (dark green), phenotypes (magenta), proteins (dark blue), molecular sequences (orange), transcripts (turquoise), and variants (light pink).

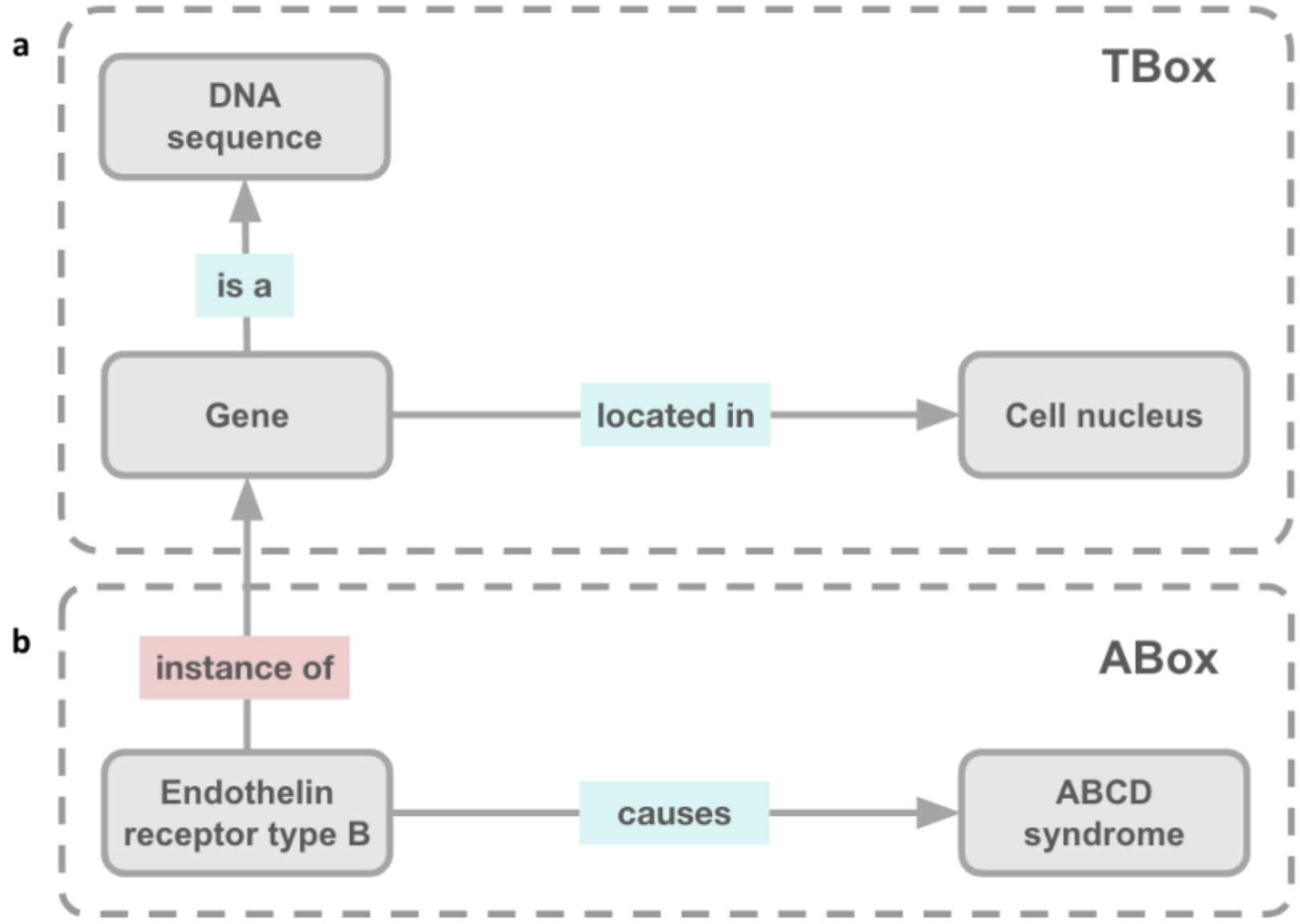

**Figure 7. Description Logics Approaches to Knowledge Modeling.**

This figure provides a simple example of two approaches for modeling knowledge within a Description Logics architecture. (**A**) The TBox includes classes (i.e., “Gene”, “DNA sequence”, and “Cell nucleus”), properties (i.e., “located in” and “is a”), and the assertions between classes (i.e., “Gene is a DNA sequence” and “Gene located in Cell nucleus”). (**B**) The ABox includes instances of classes (i.e., “Endothelin receptor type B”) represented in the TBox and assertions about those instances (i.e., “Endothelin receptor type B, instance of, Gene” and “Endothelin receptor type B, causes, ABCD syndrome”). Please note that this figure is a simplification and was inspired by Figure 2 from Thessen et al. (2020)[114].

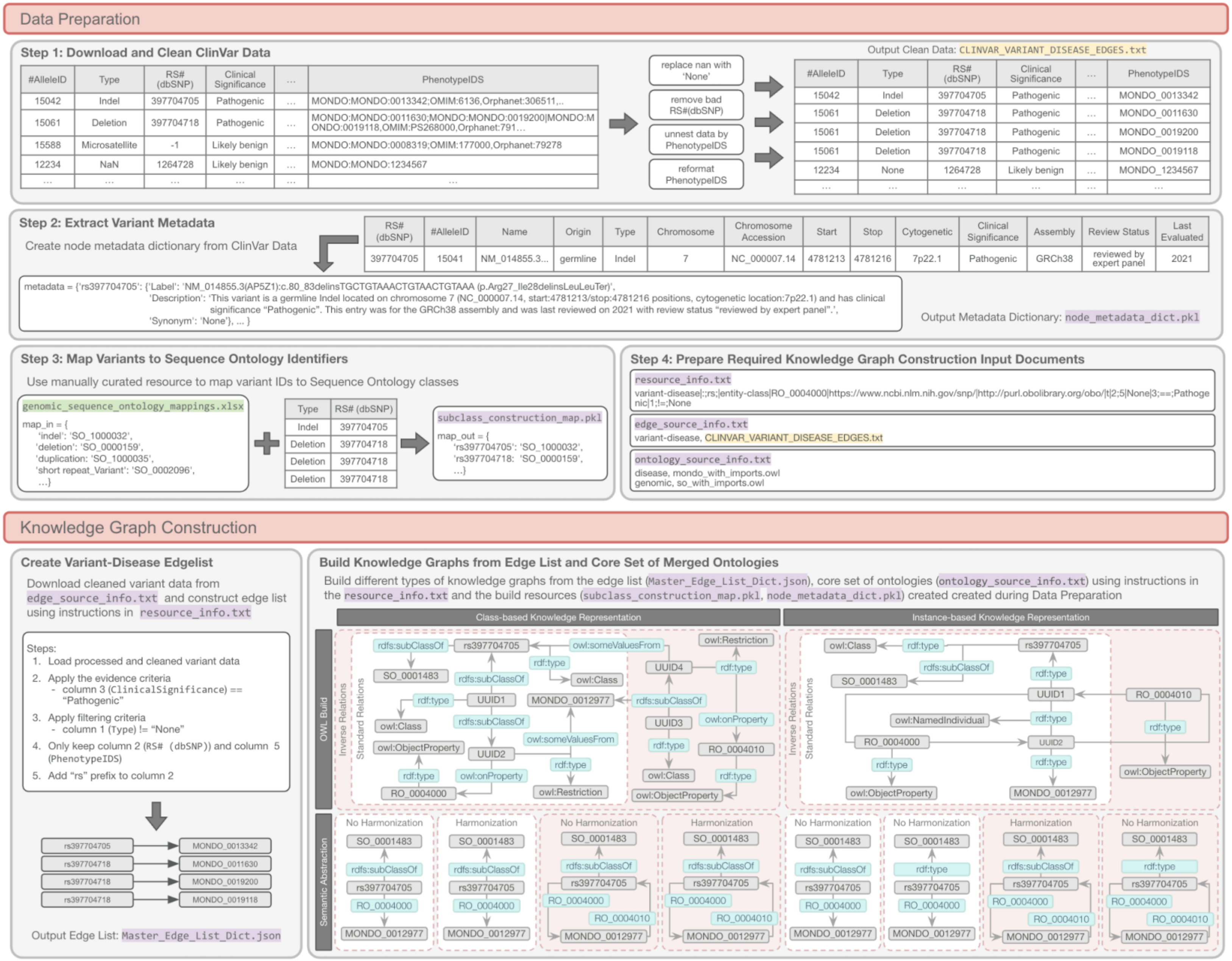
Data Preparation
Step 1: Download and Clean ClinVar Data
#AlleleID | Type | RS# (dbSNP) | Clinical Significance | ... | PhenotypeIDS
15042 | Indel | 397704705 | Pathogenic | ... | MONDO:MONDO:0013342;OMIM:6136,Orphanet:306511,..
15061 | Deletion | 397704718 | Pathogenic | ... | MONDO:MONDO:0011630;MONDO:MONDO:0019200|MONDO:MONDO:0019118,OMIM:PS268000,Orphanet:791…
15588 | Microsatellite | -1 | Likely benign | ... | MONDO:MONDO:0008319;OMIM:177000,Orphanet:79278
12234 | NaN | 1264728 | Likely benign | ... | MONDO:MONDO:1234567
replace nan with 'None'
remove bad RS#(dbSNP)
unnest data by PhenotypeIDS
reformat PhenotypeIDS
Output Clean Data: CLINVAR_VARIANT_DISEASE_EDGES.txt
#AlleleID | Type | RS# (dbSNP) | Clinical Significance | ... | PhenotypeIDS
15042 | Indel | 397704705 | Pathogenic | ... | MONDO_0013342
15061 | Deletion | 397704718 | Pathogenic | ... | MONDO_0011630
15061 | Deletion | 397704718 | Pathogenic | ... | MONDO_0019200
15061 | Deletion | 397704718 | Pathogenic | ... | MONDO_0019118
12234 | None | 1264728 | Likely benign | ... | MONDO_1234567
Step 2: Extract Variant Metadata
Create node metadata dictionary from ClinVar Data
RS# (dbSNP) | #AlleleID | Name | Origin | Type | Chromosome | Chromosome Accession | Start | Stop | Cytogenetic | Clinical Significance | Assembly | Review Status | Last Evaluated
397704705 | 15041 | NM_014855.3... | germline | Indel | 7 | NC_000007.14 | 4781213 | 4781216 | 7p22.1 | Pathogenic | GRCh38 | reviewed by expert panel | 2021
metadata = {'rs397704705': {'Label': 'NM_014855.3(AP5Z1):c.80_83delinsTGCTGTAAACTGTAACTGTAAA (p.Arg27_Ile28delinsLeuLeuTer)',
'Description': 'This variant is a germline Indel located on chromosome 7 (NC_000007.14, start:4781213/stop:4781216 positions, cytogenetic location:7p22.1) and has clinical significance "Pathogenic". This entry was for the GRCh38 assembly and was last reviewed on 2021 with review status "reviewed by expert panel".',
'Synonym': 'None'}, ... }
Output Metadata Dictionary: node_metadata_dict.pkl
Step 3: Map Variants to Sequence Ontology Identifiers
Use manually curated resource to map variant IDs to Sequence Ontology classes
genomic_sequence_ontology_mappings.xlsx
map_in = {
'indel': 'SO_1000032',
'deletion': 'SO_0000159',
'duplication: 'SO_1000035',
'short repeat_Variant': 'SO_0002096',
...}
Type | RS# (dbSNP)
Indel | 397704705
Deletion | 397704718
Deletion | 397704718
Deletion | 397704718
subclass_construction_map.pkl
map_out = {
'rs397704705': 'SO_1000032',
'rs397704718: 'SO_0000159',
...}
Step 4: Prepare Required Knowledge Graph Construction Input Documents
resource_info.txt
variant-disease|:;rs;|entity-class|RO_0004000|https://www.ncbi.nlm.nih.gov/snp/|http://purl.obolibrary.org/obo/|t|2;5|None|3;==;Pathogenic|1;!=;None
edge_source_info.txt
variant-disease, CLINVAR_VARIANT_DISEASE_EDGES.txt
ontology_source_info.txt
disease, mondo_with_imports.owl
genomic, so_with_imports.owl
Knowledge Graph Construction
Create Variant-Disease Edgelist
Download cleaned variant data from edge_source_info.txt and construct edge list using instructions in resource_info.txt
Steps:
1. Load processed and cleaned variant data
2. Apply the evidence criteria
- column 3 (ClinicalSignificance) == "Pathogenic"
3. Apply filtering criteria
- column 1 (Type) != "None"
4. Only keep column 2 (RS# (dbSNP)) and column 5 (PhenotypeIDS)
5. Add "rs" prefix to column 2
rs397704705 → MONDO_0013342
rs397704718 → MONDO_0011630
rs397704718 → MONDO_0019200
rs397704718 → MONDO_0019118
Output Edge List: Master_Edge_List_Dict.json
Build Knowledge Graphs from Edge List and Core Set of Merged Ontologies
Build different types of knowledge graphs from the edge list (Master_Edge_List_Dict.json), core set of ontologies (ontology_source_info.txt) using instructions in the resource_info.txt and the build resources (subclass_construction_map.pkl, node_metadata_dict.pkl) created created during Data Preparation
Class-based Knowledge Representation
Instance-based Knowledge Representation
OWL Build
Semantic Abstraction
Inverse Relations
Standard Relations
rdfs:subClassOf
rs397704705
owl:someValuesFrom
owl:Restriction
SO_0001483
UUID1
UUID2
UUID3
UUID4
MONDO_0012977
owl:Class
owl:ObjectProperty
rdf:type
owl:onProperty
RO_0004000
RO_0004010
owl:NamedIndividual
No Harmonization
Harmonization

**Figure 8. An Example of How Variant-Disease Edges are Created in the PKT Human Disease Mechanism Knowledge Graph**

This figure provides an end-to-end example of how variant-disease edges are created in the PKT Human Disease Mechanism knowledge graph. Beginning with the Data Preparation stage, in Step 1, the primary data source (i.e., ClinVar data) is downloaded and cleaned, which includes steps such as replacing “NaN” values with “None”, removing bad or missing identifiers, unnesting the data, and reformatting identifiers. The cleaned data (highlighted in yellow) are output for ingestion into the Knowledge Graph Construction stage. In Step 2, metadata are extracted from the primary data source to create labels, synonyms, and descriptions for each identifier. Step 3 leverages a manually curated resource (highlighted in green) to map variant identifiers to a PKT core ontology. In this case, variant identifiers are aligned to the Sequence Ontology (SO) by their type, and the final mapping is output to subclass_construction_map.pkl which is one of the required inputs for constructing a knowledge graph (highlighted in purple; cited example is from the May 2021 Class-Standard Relation-OWL build). In Step 4, the final step of this stage, the remaining required input documents for constructing a knowledge graph are updated with the resources created in the prior steps. In the Knowledge Graph Construction stage, the cleaned variant data are downloaded and an edge list is built. This edge list can then be used to construct the 12 different knowledge graphs shown in the bottom right gray box. In this example, the class-based semantically abstracted knowledge graphs are the same whether harmonization is applied or not, which is often the case for class-based builds that leverage Open Biological and Biomedical Ontology Foundry ontologies. See the Data_Preparation.ipynb Jupyter Notebook for code to process all resources used in the PKT Human Disease knowledge graph. Acronyms: PKT (PheKnowLator).

Note. A UUID is a blank or anonymous node that is created from an md5 hash of concatenated Universal Resource Identifiers (URIs). The URIs used in the hash string include the subject and object URIs (each appended with "subject" and "object," respectively) in addition to a relation. All UUIDs created during construction are explicitly defined within the PKT namespace (https://github.com/callahantiff/PheKnowLator/pkt/).

# An Open-Source Knowledge Graph Ecosystem for the Life Sciences
# SUPPLEMENTARY MATERIAL

Tiffany J. Callahan, Ignacio J. Tripodi, Adrianne L. Stefanski, Luca Cappelletti, Sanya B. Taneja, Jordan M. Wyrwa, Elena Casiraghi, Nicolas A. Matentzoglu, Justin Reese, Jonathan C. Silverstein, Charles Tapley Hoyt, Richard D. Boyce, Scott A. Malec, Deepak R. Unni, Marcin P. Joachimiak, Peter N. Robinson, Christopher J. Mungall, Emanuele Cavalleri, Tommaso Fontana, Giorgio Valentini, Marco Mesiti, Lucas A. Gillenwater, Brook Santangelo, Nicole A. Vasilevsky, Robert Hoehndorf, Tellen D. Bennett, Patrick B. Ryan, George Hripcsak, Michael G. Kahn, Michael Bada, William A. Baumgartner Jr, Lawrence E. Hunter

## Table of Contents

# Supplementary Tables and Figures

**Supplementary Table 1. Important Definitions.**

| Concept | Definition |
|---|---|
| Database | A data source not represented as an ontology, which can include Linked Open Data, data from experiments, clinical data, and existing networks and knowledge graphs. |
| Edge | Observed connections between nodes. Edges or triples can also be thought of as node-relation-node statements (e.g., geneA - interacts with - geneB). |
| Graph | An undirected, unweighted network $G(N,\ L)$, where $N$ is the set of nodes and $L$ is the set of observed edges between these nodes. |
| Knowledge Graph | A graph-based data structure representing a variety of heterogeneous entities (i.e., nodes) and multiple types of relationships between them and serving as an abstract framework that is able to infer new knowledge to address a variety of applications and use cases. |
| Knowledge Model | Within the PheKnowLator Ecosystem, there are two types of Knowledge Models that can be used when constructing a knowledge graph: (i) class-based, in which, KGs are constructed using classes, with database entities connected to the core set of merged ontologies as subclasses of existing ontology classes and (ii) instance-based, in which KGs are constructed using instances, with database entities connected to the core set of merged ontologies as instances of existing ontology classes). |
| Node | Entities or concepts, which are the subject of a knowledge graph. In the biomedical context, nodes usually represent different kinds of biological entities like genes, proteins or diseases. |
| PKT Human Disease KG | PheKnowLator Ecosystem benchmark knowledge graphs that represent the molecular mechanisms of human disease. |
| PKT-KG | Phenotype Knowledge TransLator knowledge graph construction algorithm. |
| Relation | The relationship that connects two nodes in a triple or edge. Relations are used to specify different types of relationships (e.g., interaction, substance that treats) that can exist between a pair of nodes. |
| Relation Strategy | Within the PheKnowLator Ecosystem, relations can be modeled in two ways when constructing a knowledge graph: (i) Standard Relations (i.e., a unidirectional edge is used to connect a pair of nodes) and (ii) Inverse Relations (i.e., bidirectional edges created by inferring the inverse of relations from ontologies and implicitly symmetric relations like gene-gene interactions). |
| Semantic Abstraction | Within the PheKnowLator Ecosystem, the OWL-NETS algorithm is used to decode semantically complex OWL-based KGs into KGs that contain biologically meaningful information. Additionally, the Semantic Abstraction parameter, when used with OWL-NETS, includes functionality that can harmonize a KG to a specific kind of Knowledge Model. See the following link for more information: https://github.com/callahantiff/PheKnowLator/wiki/OWL-NETS-2.0. |

**Supplementary Table 2. Acronyms used in the Manuscript.**

| Concept | Definition |
|---|---|
| API | Application Programming Interfaces |
| ChEBI | Chemical Entities of Biological Interest Ontology |
| CL | Cell Ontology |
| CLO | Cell Line Ontology |
| FAIR | Findable, Accessible, Interoperable, and Reproducible |
| GB | Gigabyte |
| GO | Gene Ontology |
| HPO | Human Phenotype Ontology |
| KG | Knowledge Graph |
| Mondo | Mondo Disease Ontology |
| OBO | Open Biological and Biomedical Ontology |
| OWL | Web Ontology Language |
| PheKnowLator | Phenotype Knowledge TransLator |
| PKT | PheKnowLator |
| PRO | Protein Ontology |
| PW | Pathway Ontology |
| RDF | Resource Description Framework |
| RDFS | Resource Description Framework Schema |
| RO | Relation Ontology |
| SO | Sequence Ontology |
| SPARQL | SPARQL Protocol and RDF Query Language |
| Uberon | Uber-Anatomy Ontology |
| VO | Vaccine Ontology |
| XML | Extensible Markup Language |

**Supplementary Table 3. PheKnowLator Ecosystem Resources.**

| Resource | URL |
|---|---|
| Ecosystem Component 1: Knowledge Graph Construction Resources | |
| GitHub | https://github.com/callahantiff/PheKnowLator |
| PyPI | https://pypi.org/project/pkt-kg/ |
| Docker Container | https://github.com/callahantiff/PheKnowLator/blob/master/Dockerfile |
| DockerHub | https://hub.docker.com/repository/docker/callahantiff/pheknowlator |
| GitHub Actions | https://github.com/callahantiff/PheKnowLator/blob/master/.github/workflows/build-qa.yml |
| Algorithm Dependencies | https://github.com/callahantiff/PheKnowLator/wiki/Dependencies |
| Dependency Automation Script | https://github.com/callahantiff/PheKnowLator/blob/master/generates_dependency_documents.py |
| Testing Suite | https://github.com/callahantiff/PheKnowLator/tree/master/tests |
| Data Processing Jupyter Notebooks | https://github.com/callahantiff/PheKnowLator/blob/master/notebooks/Data_Preparation.ipynb<br>https://github.com/callahantiff/PheKnowLator/blob/master/notebooks/Ontology_Cleaning.ipynb |
| Ecosystem Component 2: Knowledge Graph Benchmarks | |
| Human Disease KG Benchmark Details | https://github.com/callahantiff/PheKnowLator/wiki/Benchmarks-and-Builds |
| Human Disease KG Benchmark Builds Archive | Zenodo<br>- https://zenodo.org/communities/pheknowlator-benchmark-human-disease-kg<br>- Monthly Build Archive: https://zenodo.org/doi/10.5281/zenodo.7030039<br>GitHub<br>- https://github.com/callahantiff/PheKnowLator/wiki/Archived-Builds |
| Human Disease KG Benchmark Builds File Descriptions | https://zenodo.org/records/10065431/files/PheKnowLator_HumanDiseaseKG_Output_FileInformation.xlsx |
| *Knowledge Graph Build Workflow Scripts* | |
| Build Documentation[a] | https://github.com/callahantiff/PheKnowLator/blob/master/builds |
| Docker Containers | https://github.com/callahantiff/PheKnowLator/blob/master/builds/Dockerfile.phases12<br>https://github.com/callahantiff/PheKnowLator/blob/master/builds/Dockerfile.phase3 |
| GitHub Actions | https://github.com/callahantiff/PheKnowLator/blob/master/.github/workflows/kg-build-part1.yml<br>https://github.com/callahantiff/PheKnowLator/blob/master/.github/workflows/kg-build-part2.yml |
| Build Requirements | https://github.com/callahantiff/PheKnowLator/blob/master/builds/build_requirements.txt |
| Build Utilities | https://github.com/callahantiff/PheKnowLator/blob/master/builds/build_utilities.py |
| Build Logging | https://github.com/callahantiff/PheKnowLator/blob/master/builds/job_monitoring.py<br>https://github.com/callahantiff/PheKnowLator/blob/master/builds/logging.ini |
| Phase 1 Build Scripts | https://github.com/callahantiff/PheKnowLator/blob/master/builds/phases1_2_entrypoint.py<br>https://github.com/callahantiff/PheKnowLator/blob/master/builds/build_phase_1.py<br>https://github.com/callahantiff/PheKnowLator/blob/master/builds/data_to_download.txt |
| Phase 2 Build Scripts | https://github.com/callahantiff/PheKnowLator/blob/master/builds/phases1_2_entrypoint.py<br>https://github.com/callahantiff/PheKnowLator/blob/master/builds/build_phase_2.py<br>https://github.com/callahantiff/PheKnowLator/blob/master/builds/data_preprocessing.py<br>https://github.com/callahantiff/PheKnowLator/blob/master/builds/ontology_cleaning.py |
| Phase 3 Build Scripts | https://github.com/callahantiff/PheKnowLator/blob/master/builds/build_phase_3.py |

| Resource | URL |
| --- | --- |
| | https://github.com/callahantiff/PheKnowLator/blob/master/builds/phase3_log_daemon.py |
| Ecosystem Component 3: Knowledge Graphs Tools | |
| Zenodo Community | https://zenodo.org/communities/pheknowlator-ecosystem |
| Jupyter Notebooks | https://github.com/callahantiff/PheKnowLator/blob/master/main.ipynb<br>https://github.com/callahantiff/PheKnowLator/blob/master/notebooks/OWLNETS_Example_Application.ipynb<br>https://github.com/callahantiff/PheKnowLator/blob/master/notebooks/RDF_Graph_Processing_Example.ipynb<br>https://github.com/callahantiff/PheKnowLator/blob/master/notebooks/Tutorials/entity_search/Entity_Search.ipynb |
| SPARQL Endpoint | http://sparql.pheknowlator.com/ (deprecated 2024)<br>Code to facilitate the hosting of a PheKnowLator KG via a SPARQL Endpoint with custom front end web application and serve data from a Blazegraph triple store are available from GitHub: https://github.com/callahantiff/PheKnowLator/tree/36fd2d1dad805a58f4a7fb801b2a7c26dfb130a8/builds/deploy/triple-store#readme |

[a]The build documentation provides a detailed description of all of the processes and code needed to generate the knowledge graph builds using Google Cloud Platform resources..

Acronyms: KG (knowledge graph); PKT (PheKnowlator); PKT-KG (PheKnowLator knowledge graph construction software).

**Supplementary Table 4. PheKnowLator Ecosystem Evaluation Resources.**

| Resource | URL |
|---|---|
| Survey of Open Source KG Construction Software | |
| GitHub Scraper | https://zenodo.org/doi/10.5281/zenodo.10052113 |
| Survey | https://zenodo.org/doi/10.5281/zenodo.5790040 |
| PKT Human Disease KGs | |
| PKT-KG Zenodo Release | https://doi.org/10.5281/zenodo.4685943 |
| PyPI Release | https://pypi.org/project/pkt-kg/2.1.0/ |
| Data Source Descriptions | https://github.com/callahantiff/PheKnowLator/wiki/May-01%2C-2021 |
| GitHub Actions Workflows | https://github.com/callahantiff/PheKnowLator/.github/workflows/kg-build-part1.yml<br>https://github.com/callahantiff/PheKnowLator/.github/workflows/kg-build-part2.yml |
| Docker | Data Preparation: https://github.com/callahantiff/PheKnowLator/builds/Dockerfile.phases12<br>KG Construction: https://github.com/callahantiff/PheKnowLator/builds/Dockerfile.phase3 |
| Data Sources[a] | data_to_download.txt |
| PKT-KG Input Dependencies[a] | edge_source_list.txt<br>ontology_source_list.txt<br>resource_info.txt |
| Build Metadata[a] | downloaded_build_metadata.txt<br>edge_source_metadata.txt<br>ontology_source_metadata.txt<br>preprocessed_build_metadata.txt |
| Ontology Quality Report[a] | ontology_cleaning_report.txt |
| Build Logs[a] | pkt_builder_phases12_log.log<br>pkt_build_log.log |
| Human Disease KG Benchmark Builds Zenodo Archive | *Class-based Knowledge Model*<br>Standard Relations without Semantic Abstraction: https://zenodo.org/doi/10.5281/zenodo.8180774<br>Standard Relations with Semantic Abstraction: https://zenodo.org/doi/10.5281/zenodo.8180772<br>Inverse Relations without Semantic Abstraction: https://zenodo.org/doi/10.5281/zenodo.8180766<br>Inverse Relations with Semantic Abstraction: https://zenodo.org/doi/10.5281/zenodo.8180768<br>*Instance-based Knowledge Model*<br>Standard Relations without Semantic Abstraction: https://zenodo.org/doi/10.5281/zenodo.8180764<br>Standard Relations with Semantic Abstraction: https://zenodo.org/doi/10.5281/zenodo.8180762<br>Inverse Relations without Semantic Abstraction: https://zenodo.org/doi/10.5281/zenodo.8180758<br>Inverse Relations with Semantic Abstraction: https://zenodo.org/doi/10.5281/zenodo.8180756<br><br>A table describing all files for each build KG type can be found on the Zenodo Community archive: https://zenodo.org/records/10065431/files/PheKnowLator_HumanDiseaseKG_Output_FileInformation.xlsx |

[a]Each referenced file can be found within each knowledge graph type for every build on Zenodo.

Acronyms: KG (knowledge graph); PKT-KG (PheKnowLator knowledge graph construction algorithm).

**Supplementary Table 5. Open-Source Knowledge Graph Construction Methods Survey Criteria.**

| Criteria | Description | Example Questions |
|---|---|---|
| Construction Functionality | An assessment of how well the method covers the steps needed to construct a knowledge graph from downloading and processing data and building edge lists to generating and outputting a KG | Is there functionality to download data?<br>Can multiple types of KGs be constructed?<br>Is preprocessing or filtering performed as part of the construction process? |
| Maturity | An assessment of the level, stage or development phase of a method | Is a versioning system in place?<br>Have many releases been made?<br>Are procedures in place to enable collaboration? |
| Availability | An assessment of the openness of a method and the ease of obtaining a copy of the method | Is the method licensed?<br>What type of license is used? |
| Usability | An assessment of the efforts put in place to ensure that a user, with reasonable technical skills, could use the method | Is there a Wiki, Read the Docs, or GitPage associated with the method?<br>Are there examples of how to use the method? |
| Reproducibility | An assessment of whether or not the method provides tools or resources to help reproduce the KG construction process and maintain the code base | What tools are provided to help enable reproducibility (e.g., Docker container, Jupyter Notebook, R Markdown)?<br>Does the repository include any form of testing? |

Note. The survey questions were adapted from http://dx.doi.org/10.1109/aswec.2004.1290484.

**Supplementary Table 6. Open-Source Knowledge Graph Construction Methods.**

| Method | GitHub Repository[a] | Publication DOI | Primary Goal or Objective (from GitHub)[b] | Method Validation | Most Recent Repository Interaction[c] |
|---|---|---|---|---|---|
| Bio2BEL | bio2bel/ | 10.1101/631812v1 | Bio2BEL uses the Biological Expression Language as a common schema for integrating a wide variety of biomedical databases including causal, correlative, and associative relationships between entities on the molecular, process, cellular, systems, and population levels | N/A | Within the last month |
| Bio2RDF | bio2rdf | 10.1016/j.jbi.2008.03.004 | Bio2RDF is an open-source project that uses Semantic Web technologies to build and provide the largest network of Linked Data for the Life Sciences | Examined impact of four transcription factors in Parkinson's disease | Within the last week |
| Bio4J | bio4j/bio4j | 10.1101/016758 | Bio4j aims to offer a platform for the integration of semantically rich biological data using typed graph models | Tool use demonstration; no formal biological validation | > 1 year |
| BioGrakn | graknlabs/biograkn | 10.1007/978-3-319-61566-0_28 | BioGrakn is based on GRAKN.AI, which is a deductive database in the form of a knowledge graph, allowing complex data modelling, verification, scaling, querying and analysis | Illustrative queries spanning precision medicine, text mining, and disease | Within the last month |
| Clinical Knowledge Graph (CKG) | MannLabs/CKG | 10.1101/2020.05.09.084897 | Clinical Knowledge Graph is a platform with twofold objectives: 1) build a graph database with experimental data and data imported from diverse biomedical databases and 2) automate knowledge discovery making use of all the information contained in the graph | Biomarker studies to demonstrate CKG use for clinical decision-making | Within the last month |
| COVID-19-Community | covid-19-net/covid-19-community | NA | The COVID-19-Community is a community effort to build a Neo4j knowledge graph that links heterogenous data about COVID-19 | Tool use demonstration | Within the last week |
| Dipper | monarch-initiative/dipper | NA | Dipper is a Python package to generate RDF triples from common scientific resources | Tool use demonstration | Within the last week |
| Hetionet | hetio/hetionet | 10.7554/eLife.26726 | Hetionet is a hetnet — network with multiple node and edge (relationship) types — which encodes biology. Hetnet was designed for Project Rephetio | Predicted the probability of treatment for 209,168 compound–disease pairs | Within the last year |
| iASiS Open Data Graph | tasosnent/Biomedical-Knowledge-Integration | arXiv:1912.08633 | iASiS is a framework to automatically retrieve and integrate disease-specific knowledge into an up-to-date semantic graph | Examined use with lung cancer, dementia, and Duchenne Muscular Dystrophy | Within the last 6 months |

| Method | GitHub Repository[a] | Publication DOI | Primary Goal or Objective (from GitHub)[b] | Method Validation | Most Recent Repository Interaction[c] |
|---|---|---|---|---|---|
| KG-COVID-19 | Knowledge-Graph-Hub/kg-covid-19 | NA | KG-COVID-19 is a flexible framework to ingest, integrate, and remix biomedical data to produce KGs for COVID-19 response. The framework can be applied to other problems in which siloed biomedical data must be quickly integrated for different biomedical research applications, including for future pandemics | Tool use demonstration | Within the last week |
| Knowledge Base Of Biomedicine (KaBOB) | UCDenver-ccp/kabob | 10.1186/s12859-015-0559-3 | KaBOB is a knowledge base of semantically integrated data. The system introduces five processes for semantic data integration including making explicit the differences between biomedical concepts and database records, aggregating sets of identifiers denoting the same biomedical concepts across data sources, and using declaratively represented forward-chaining rules to take information that is variably represented in source databases and integrating it into a consistent biomedical representation | Constructed a multi-species KG | Within the last year |
| Knowledge Graph Exchange (KGX) | NCATS-Tangerine/kgx | NA | KGX is a library and set of command line utilities for exchanging Knowledge Graphs that conform to or are aligned to the Biolink Model | Tool use demonstration | Within the last month |
| Knowledge Graph Toolkit (KGTK) | usc-isi-i2/kgtk/ | arXiv:2006.00088 | KGTK is a data science-centric toolkit to represent, create, transform, enhance and analyze KGs. KGTK represents graphs in tables and leverages popular libraries developed for data science applications, enabling a wide audience of developers to easily construct KG pipelines for their applications | Demonstrated functionality using Wikidata, DBpedia, and ConceptNet | Within the last week |
| ProNet | cran/ProNet | NA | ProNet provides functions for biological network construction, visualization and analyses, including topological statistics, functional module clustering, and GO-profiling | Examined H1N1 IAV-human protein-protein interactions | > 1 year |
| SEmantic Modeling machIne (SEMi) | giuseppefutia/semi | 10.1016/j.softx.2020.100516 | SeMi (SEmantic Modeling machIne) is a tool to semi-automatically build large-scale Knowledge Graphs from structured sources such as CSV, JSON, and XML files | Validated using advertising data | Within the last 6 months |
| PheKnowLator | PheKnowLator | 10.1101/2020.04.30.071407 | PheKnowLator (Phenotype Knowledge Translator) is a novel framework and fully automated Python 3 library explicitly designed for optimized construction of semantically-rich, large-scale biomedical KGs | Built and compared 12 benchmark KGs including construction performance | Within the last week |

[a]All GitHub URLs begin with the following prefix: https://github.com/.
[b]Whenever possible, descriptions of methods and tools were copied verbatim from the associated GitHub site, documentation, and/or manuscript.
[c]The most recent repository interaction was documented at the time of completing the survey, which was May 2020 (updated in June 2021).

Acronyms: KG (Knowledge Graph).

**Supplementary Table 7. Open-Source Knowledge Graph Construction Survey - Functionality.**

| Method | Download Functionality | Edge list Functionality | Construction Functionality | Multiple KG Types | Other KG Construction Functionality | Process Ontology Data | Process Linked Open Data | Process Experimental Data | Process Clinical Data | Data Processing Limits |
|---|---|---|---|---|---|---|---|---|---|---|
| Bio2BEL | Yes | Yes | Yes | Yes | The entire PyBEL ecosystem tools are all available for all graphs generated by Bio2BEL | Yes | Yes | Yes | Yes | No |
| Bio2RDF | Yes | Yes | Yes | Yes | Talend RESTful API; community ontology mappings; SPARQL query repository | Yes | Yes | Yes | Yes | No |
| Bio4J | Yes | Yes | Yes | Yes | Titan, Angulillos API | Yes | Yes | No | No | No |
| BioGrakn | No | No | Yes | Yes | Provides different types of API clients (Java, Python, Node.js) and a Grakn Workbase | No | Yes | Yes | Yes | No |
| Clinical Knowledge Graph (CKG) | Yes | No | Yes | No | Data preparation (filtering, imputation, formattin); data analysis (dimensionality reduction, visualization, hypothesis testing) | Yes | Yes | Yes | No | No |
| COVID-19-Community | Yes | Yes | Yes | No | Neo4J Browser | No | Yes | No | Yes | No |
| Dipper | Yes | Yes | Yes | Yes | SciGraph RESTful API<br>Build KGs with evidence and provenance | Yes | Yes | Yes | No | No |
| Hetionet | Yes | No | Yes | No | Neo4J Browser<br>Creates permuted KGs | Yes | Yes | Yes | Yes | No |
| iASiS Open Data Graph | Yes | Yes | Yes | No | Biomedical Harvesters; MedKnow | Yes | Yes | No | Yes | No |
| KG-COVID-19 | Yes | Yes | Yes | No | Leverages BioLink | Yes | Yes | No | No | No |
| Knowledge Base Of Biomedicine (KaBOB) | Yes | Yes | Yes | Yes | Blazegraph | Yes | Yes | No | No | No |
| Knowledge Graph Exchange (KGX) | Yes | Yes | Yes | Yes | KG verified to confirm to the Biolink model, summary statistics | Yes | Yes | Yes | No | No |
| Knowledge Graph Toolkit (KGTK) | Yes | Yes | Yes | Yes | Data cleaning module, processes other KGs, KG querying modules, summary statistics, node embeddings | No | Yes | No | No | No |
| ProNet | No | Yes | Yes | No | KG visualization; enables topological analyses | No | Yes | Yes | No | No |
| SEmantic Modeling machIne (SEMi) | No | Yes | Yes | Yes | Semantic type detector; weighted graph generator; semantic model builder and refiner; link predictor | Yes | Yes | No | No | No |
| PheKnowLator | Yes | Yes | Yes | Yes | Data download and preprocessing tools; ontology quality control tools; export node metadata; property graphs; SPARQL Endpoint | Yes | Yes | Yes | No | No |

Note. For scoring, 1 point was awarded for an answer of "Yes" and for the presence of other KG construction functionality.

Acronyms: KG (Knowledge Graph).

**Supplementary Table 8. Open-Source Knowledge Graph Construction Survey - Availability.**

| Method | Open Source | License | Operating Systems | Coding Languages | External Dependencies |
|---|---|---|---|---|---|
| Bio2BEL | Yes | MIT | Linux, Windows, Mac OSX, Cloud-based systems and/or architectures | Python, SQL | Bioregistry, PyOBO, Bioversions, various other standard Python packages |
| Bio2RDF | Yes | MIT<br>Apache 2.0<br>CC0-1.0 | Linux, Windows, Mac OSX | Java, JavaScript, Shell, OWL | Virtuoso, GIT |
| Bio4J | No | AGPL-3.0 | Linux, Windows, Mac OSX, Cloud-based systems and/or architectures | Java, Scala | Angulillos, AWS EC2/S3, Titan |
| BioGrakn | No | None | Linux, Windows, Mac OSX, Cloud-based architectures | Python, Java, Node.js | GraknLabs, Maven |
| Clinical Knowledge Graph (CKG) | Yes | MIT | Linux, Windows, Mac OSX | Python | Java SE Runtime, Neo4j, R, Python 3.6 |
| COVID-19-Community | No | MIT | Linux, Windows, Mac OSX | Python, Shell | Neo4J, Anaconda |
| Dipper | No | BSD-3 | Linux, Windows, Mac OSX | Python, TSQL | |
| Hetionet | Yes | CC0 | Linux, Windows, Mac OSX | Python, Shell | Docker, Neo4J |
| iASiS Open Data Graph | No | Apache 2.0 | Linux, Windows, Mac OSX | Python, Java | MongoDB, UMLS, ReVerb, MetaMap, SemRep, YAJL, Neo4J |
| KG-COVID-19 | Yes | BSD-3 | Linux, Windows, Mac OSX | Python | KGX, BioLink |
| Knowledge Base Of Biomedicine (KaBOB) | Yes | GPL | Linux, Windows, Mac OSX | Groovy, Clojure, Shell | Docker, Maven |
| Knowledge Graph Exchange (KGX) | Yes | BSD-3 | Linux, Windows, Mac OSX | Python | Docker, BioLink |
| Knowledge Graph Toolkit (KGTK) | Yes | MIT | Linux, Windows, Mac OSX | Python | Anaconda, mlr |
| ProNet | Yes | GPL (>=2) | Linux, Windows, Mac OSX | R | BioGrid, GO |
| SEmantic Modeling machIne (SEMi) | Yes | GPL | Linux, Windows, Mac OSX | Python, JavaScript, Shell | Anaconda, Node.js (11.15.0), Java, Maven, Elasticsearch |
| PheKnowLator | Yes | Apache 2.0 | Linux, Windows, Mac OSX, Cloud-based systems and/or architectures | Python, Java, Shell | OWL Tools |

Note. For scoring, 1 point was awarded for an answer of “Yes” and for the presence of a license.

**Supplementary Table 9. Open-Source Knowledge Graph Construction Survey - Usability.**

| Method | README | Wiki, Docs, or GitPage | Example Use | Tutorials | Install Tools | Method Use Resources | Sample Data | Handles Different Sized Data | Output Types | Adoption Indicators |
|---|---|---|---|---|---|---|---|---|---|---|
| Bio2BEL | Yes | Yes | Yes | No | PyPI<br>Maven | None | Yes | Yes | NetworkX, Cytoscape, text files, n-triples, Biological Expression Language, several IO formats | Yes |
| Bio2RDF | Yes | Yes | Yes | No | None | None | Yes | Yes | Virtuoso dump, OWL, nq | Yes |
| Bio4J | Yes | Yes | Yes | Yes | AWS S3 | Angulillos API<br>Titan | Yes | Yes | Titan | Yes |
| BioGrakn | Yes | Yes | Yes | Yes | None | Grakn Clients | Yes | Yes | Grakn KG output types | Yes |
| Clinical Knowledge Graph (CKG) | Yes | Yes | Yes | Yes | Docker | Jupyter Notebook<br>Docker | Yes | Yes | Neo4j | Yes |
| COVID-19-Community | Yes | No | Yes | Yes | Jupyter Notebook | Jupyter Notebooks | Yes | Yes | Neo4J, CSV | Yes |
| Dipper | Yes | Yes | Yes | Yes | PYPI | Jupyter Notebooks | Yes | Yes | TTL, Neo4J, TSV | Yes |
| Hetionet | Yes | Yes | Yes | Yes | Jupyter Notebook | Jupyter Notebook<br>Docker | Yes | Yes | JSON, Neo4J, TSV, and Matrix | Yes |
| iASiS Open Data Graph | Yes | Yes | Yes | No | None | None | No | Yes | JSON, CSV, Neo4J, MongoDB | Yes |
| KG-COVID-19 | Yes | Yes | Yes | Yes | None | None | Yes | Yes | RDF, TSV | Yes |
| Knowledge Base Of Biomedicine (KaBOB) | Yes | Yes | Yes | Yes | Docker | Docker | Yes | Yes | RDF/XML | Yes |
| Knowledge Graph Exchange (KGX) | Yes | Yes | Yes | Yes | PyPI<br>Docker | None | Yes | Yes | OWL or RDF/XML, NetworkX, text files, n-triples files, tar, csv, graphML, TTL, JSON, RQ, RSA | Yes |
| Knowledge Graph Toolkit (KGTK) | Yes | Yes | Yes | Yes | Jupyter Notebook<br>Docker | Jupyter Notebook<br>Docker | Yes | Yes | n-triples files, JSON, Neo4J, GML | Yes |
| ProNet | No | Yes | Yes | Yes | CRAN | R Markdown | Yes | Yes | R data frame object (rda) | No |
| SEmantic Modeling machIne (SEMi) | Yes | Yes | Yes | No | PyPI | None | Yes | Yes | OWL or RDF/XML files, graph, json, TTL | No |
| PheKnowLator | Yes | Yes | Yes | Yes | PyPI<br>Jupyter Notebook<br>Docker | Jupyter Notebook<br>Docker | Yes | Yes | RDF/XML, NetworkX, a text files, n-triples, JSON | Yes |

Note. For scoring, 1 point was awarded for an answer of “Yes” and for the presence of tools to run and install the method.

**Supplementary Table 10. Open-Source Knowledge Graph Construction Survey - Maturity.**

| Method | Multiple Releases | Release Count | Method Published | Collaboration Encouraged | Collaboration Procedures |
|---|---|---|---|---|---|
| Bio2BEL | Yes | 1 | Yes | Yes | Yes |
| Bio2RDF | Yes | 2 | Yes | Yes | No |
| Bio4J | Yes | 100 | Yes | No | Yes |
| BioGrakn | Yes | 1 | Yes | No | No |
| Clinical Knowledge Graph (CKG) | No | 0 | Yes | Yes | Yes |
| COVID-19-Community | No | 0 | No | Yes | Yes |
| Dipper | Yes | 4 | No | No | No |
| Hetionet | No | 1 | Yes | Yes | No |
| iASiS Open Data Graph | No | 0 | Yes | No | No |
| KG-COVID-19 | No | 0 | No | Yes | Yes |
| Knowledge Base Of Biomedicine (KaBOB) | No | 1 | Yes | No | No |
| Knowledge Graph Exchange (KGX) | No | 0 | No | No | No |
| Knowledge Graph Toolkit (KGTK) | Yes | 3 | Yes | Yes | Yes |
| ProNet | Yes | 1 | Unclear | No | No |
| SEmantic Modeling machIne (SEMi) | No | 0 | Yes | No | No |
| PheKnowLator | Yes | 1 | Yes | Yes | Yes |

Note. For scoring, 1 point was awarded for an answer of “Yes” and for the presence of at least one release.

**Supplementary Table 11. Open-Source Knowledge Graph Construction Survey - Reproducibility.**

| Method | Reproducibility Tools | Install Services | Deployment Services | Maintainability Measures | Well-Documented Codebase | Actively Used Issue Tracker |
|---|---|---|---|---|---|---|
| Bio2BEL | CLI Tool | Yes | No | Yes | Yes | Yes |
| Bio2RDF | None | No | No | No | Yes | Yes |
| Bio4J | AWS S3 Titan distribution | No | Yes | No | Yes | Yes |
| BioGrakn | Grakn Tools | Yes | Yes | No | Yes | Yes |
| Clinical Knowledge Graph (CKG) | Jupyter Notebook<br>Docker | No | No | No | Yes | Yes |
| COVID-19-Community | Jupyter Notebooks | No | No | No | Yes | Yes |
| Dipper | Jupyter Notebook | Yes | Yes | No | Yes | Yes |
| Hetionet | Jupyter Notebook<br>Docker | No | No | No | Yes | Yes |
| iASiS Open Data Graph | None | Partial | No | No | Yes | Yes |
| KG-COVID-19 | None | Yes | Yes | Yes | Yes | Yes |
| Knowledge Base Of Biomedicine (KaBOB) | Docker | Yes | Yes | No | Yes | Yes |
| Knowledge Graph Exchange (KGX) | Jupyter Notebook<br>Docker | Yes | Yes | No | Yes | Yes |
| Knowledge Graph Toolkit (KGTK) | Jupyter Notebook<br>Docker | Yes | Yes | Yes | Yes | Yes |
| ProNet | R Markdown | No | No | No | Yes | No |
| SEmantic Modeling machIne (SEMi) | None | Yes | Yes | No | Yes | Yes |
| PheKnowLator | PyPI<br>Docker<br>Jupyter Notebook | Yes | Yes | Yes | Yes | Yes |

Note. For scoring, 1 point was awarded for an answer of “Yes” and for the presence of at least one reproducibility tool.

## Supplementary Table 12. PKT Human Disease Knowledge Graph Resources - Ontologies.

| Provider | Filename | URLs and Citations | License | Node or Edge Type and Usage |
|---|---|---|---|---|
| *ONTOLOGY RESOURCES* | | | | |
| Chemical Entities of Biological Interest (ChEBI) | http://purl.obolibrary.org/obo/chebi.owl | **URL:** https://www.ebi.ac.uk/chebi/ **Citation:** PMID:26467479 | CC BY 4.0 | Utilized to connect chemicals to complexes, diseases, genes, GO biological processes, GO cellular components, GO molecular functions, pathways, phenotypes, reactions, and transcripts. |
| Cell Ontology (CL)[a] | http://purl.obolibrary.org/obo/uberon/ext.owl | **URL:** https://github.com/obophenotype/cell-ontology **Citation:** PMID:27377652 | CC BY 4.0 | Utilized to connect transcripts and proteins to cells. Additionally, this ontology imports the following ontologies: ChEBI, GO, PATO, PRO, RO, Uberon. |
| Cell Line Ontology (CLO) | http://purl.obolibrary.org/obo/clo.owl | **URL:** https://obofoundry.org/ontology/clo.html **Citation:** PMID:25852852 | CC BY 3.0 | Utilized to map cell lines to transcripts and proteins. Additionally, this ontology imports the following ontologies: CL, DOID, NCBITaxon, Uberon. |
| Gene Ontology (GO) | http://purl.obolibrary.org/obo/go.owl | **URL:** http://geneontology.org/ **Citations:** PMID:10802651; PMID:36866529 | CC BY 4.0 | Utilized to connect biological processes, cellular components, and molecular functions to chemicals, pathways, and proteins. Additionally, this ontology imports the following ontologies: CL, NCBITaxon, RO, and Uberon. |
| Human Phenotype Ontology (HPO) | http://purl.obolibrary.org/obo/hp.owl | **URL:** https://ontology.jax.org/api/hp/docs/ **Citation:** PMID:33264411 | MIT License | Utilized to connect phenotypes to chemicals, diseases, genes, and variants. Additionally, this ontology imports the following ontologies: CL, ChEBI, GO, and Uberon. |
| Mondo Disease Ontology (Mondo) | http://purl.obolibrary.org/obo/mondo.owl | **URL:** https://mondo.monarchinitiative.org/ **Citation:** DOI:10.1101/2022.04.13.22273750 | CC BY 4.0 | Utilized to connect diseases to chemicals, phenotypes, genes, and variants. Additionally, this ontology imports the following ontologies: CL, NCBITaxon, GO, HPO, and Uberon. |
| Pathway Ontology (PW) | http://purl.obolibrary.org/obo/pw.owl | **URL:** https://rgd.mcw.edu/wg/home/pathway2/ **Citation:** PMID:24499703 | CC BY 4.0 | Utilized to connect pathways to GO biological processes, GO cellular components, GO molecular functions,and Reactome pathways. |
| Protein Ontology (PRO) | http://purl.obolibrary.org/obo/pr.owl | **URL:** https://proconsortium.org/ **Citation:** PMID:20935045 | CC BY 4.0 | Utilized to connect proteins to chemicals, genes, anatomy, catalysts, cell lines, cofactors, complexes, GO biological processes, GO cellular components, GO molecular functions, pathways, proteins, reactions, and transcripts. Additionally, this ontology imports the following ontologies: ChEBI, DOID, and GO. The ontology was subset to only include homo sapien proteins. |
| Relations Ontology (RO) | http://purl.obolibrary.org/obo/ro.owl | **URL:** https://github.com/oborel/obo-relations/ **Citation:** PMID:15892874 | CC 1.0 Universal | Utilized this to connect all data sources added to the core set of merged ontologies. |
| Sequence Ontology (SO) | http://purl.obolibrary.org/obo/so.owl | **URL:** https://github.com/The-Sequence-Ontology/SO-Ontologies **Citation:** PMID:15892872 | CC BY 4.0 | Utilized to connect transcripts and other genomic material like genes and variants. |
| Uber-Anatomy Ontology (Uberon) | http://purl.obolibrary.org/obo/uberon/ext.owl | **URL:** http://obophenotype.github.io/uberon/ **Citation:** PMID:22293552; PMID: 25009735 | CC BY 3.0 | Utilized to connect tissues, fluids, and cells to proteins and transcripts. Additionally, this ontology imports the following ontologies: ChEBI, CL, GO, PRO. |
| Vaccine Ontology (VO) | http://purl.obolibrary.org/obo/vo.owl | **URL:** https://github.com/vaccineontology/VO | CC BY 3.0 | Utilized the edges between this ontology and those it |

| Provider | Filename | URLs and Citations | License | Node or Edge Type and Usage |
|---|---|---|---|---|
| | | **Citation:** PMID:23256535; PMID:21624163 | | imports: ChEBI, DOID, GO, PRO, and Uberon. |
| *EDGE SET RESOURCES* | | | | |
| ClinVar | variant_summary.txt | **URL:** https://ftp.ncbi.nlm.nih.gov/pub/clinvar<br>**Citation:** PMID:29165669 | MIT License | variant-gene; variant-disease; variant-phenotype |
| Comparative Toxicogenomics Database (CTD) | CTD_chemicals_diseases.tsv<br>CTD_chem_gene_ixns.tsv<br>CTD_chem_go_enriched.tsv<br>CTD_genes_pathways.tsv | **URL:** http://ctdbase.org/<br>**Citation:** PMID:36169237 | CC BY 4.0 | chemical-disease; chemical-phenotype<br>chemical-gene; chemical-protein<br>chemical-biological process; chemical-cellular component; chemical-molecular function<br>gene-pathway |
| DisGeNET | Curated_gene_disease_associations.tsv | **URL:** https://www.disgenet.org/<br>**Citation:** PMID:31680165 | CC BY-NC-SA 4.0 | gene-disease; gene-phenotype |
| Ensembl | Homo_sapiens.GRCh38.102.gtf<br>Homo_sapiens.GRCh38.102.uniprot.tsv.gz<br>Homo_sapiens.GRCh38.102.entrez.tsv.gz | **URL:** https://useast.ensembl.org/<br>**Citation:** PMID:36318249 | Apache 2.0 | gene-protein; gene-transcript; transcript-protein |
| Gene MANIA | COMBINED.DEFAULT_NETWORKS.BP_COMBINING.txt | **URL:** https://genemania.org/<br>**Citation:** PMID:20576703 | CC BY 4.0 | gene-gene |
| Gene Ontology | goa_human.gaf | **URL:** http://geneontology.org/<br>**Citation:** PMID:10802651; PMID:36866529 | CC BY 4.0 | protein-biological process; protein-cellular component; protein-molecular function |
| The Genotype-Tissue Expression (GTEx) Project | https://storage.googleapis.com/gtex_analysis_v8/rna_seq_data/GTEx_Analysis_2017-06-05_v8_RNASeQCv1.1.9_gene_median_tpm.gct.gz | **URL:** https://gtexportal.org/home/<br>**Citation:** PMID:23715323 | CC BY 4.0 | protein-anatomy; protein-cell; transcript-anatomy; transcript-cell |
| HUGO Gene Nomenclature Committee (HGNC) | hgnc_complete_set.txt | **URL:** https://www.genenames.org/<br>**Citation:** PMID:36243972 | CC0 | gene-protein; gene-transcript; transcript-protein |
| The Human Phenotype Ontology (HPO) | phenotype.hpoa | **URL:** https://hpo.jax.org/<br>**Citation:** PMID:33264411 | MIT License | disease-phenotype |
| The Human Protein Atlas | API Query | **URL:** https://www.proteinatlas.org/<br>**Citation:** PubMed:25613900 | CC BY 3.0 | protein-anatomy; protein-cell; transcript-anatomy; transcript-cell |
| The National Center for Biotechnology Information (NCBI) | Homo_sapiens.gene_info.gz | **URL:** https://www.ncbi.nlm.nih.gov/gene/<br>**Citation:** PMID:21115458 | Terms and Conditions | gene-protein; gene-transcript; transcript-protein |
| Reactome Pathway Database | ChEBI2Reactome_All_Levels.txt<br>gene_association.reactome<br>UniProt2Reactome_All_Levels.txt | **URL:** https://reactome.org/<br>**Citation:** PMID:34788843 | CC0 | chemical-pathway<br>biological processes-pathway; pathway-cellular component; pathway-molecular function<br>protein-pathway |
| The Search Tool for Recurring Instances of | 9606.protein.links.v11.0.txt | **URL:** https://string-db.org/<br>**Citation:** PMID:36370105 | CC BY 4.0 | protein-protein |

| Provider | Filename | URLs and Citations | License | Node or Edge Type and Usage |
|---|---|---|---|---|
| Neighbouring Genes (STRING) Database | | | | |
| Universal Protein Resource (UniProt) | API Query | **URL:** https://www.uniprot.org/<br>**Citation:** PMID:36408920 | CC BY 4.0 | gene-protein; gene-transcript; transcript-protein<br>protein-catalyst; protein-cofactor |
| *MAPPING AND FILTERING RESOURCES* | | | | |
| Chemical Entities of Biological Interest (ChEBI) | names.tsv | **URL:** https://www.ebi.ac.uk/chebi/<br>**Citation:** PMID:26467479 | CC BY 4.0 | Used in combination with MeSH to to obtain mappings between MeSH identifiers and ChEBI identifiers for chemicals-diseases, chemicals-genes, chemical-GO biological processes, chemicals-GO cellular components, chemicals-GO molecular functions, chemicals-phenotypes, chemicals-proteins, and chemicals-transcripts. |
| Compath | curated_mappings.txt<br>kegg_reactome.csv | **URLs:**<br>- https://compath.scai.fraunhofer.de/<br>- https://github.com/ComPath/compath-resources/tree/master/mappings<br>**Citation:** PMID:30564458 | Imprint<br>MIT License | To align Reactome pathway concept identifiers to PW. |
| DisGeNET | disease_mappings.tsv | **URL:** https://www.disgenet.org/<br>**Citation:** PMID:31680165 | CC BY-NC-SA 4.0 | Obtain mappings between different disease terminologies and vocabularies including: DOID, OMIM, Orphanet, ICD9, ICD10, the UMLS and MeSH to Mondo and the HPO. |
| The Medical Subject Headings (MeSH) | mesh2021.nt | **URL:** https://www.nlm.nih.gov/mesh/meshhome.html<br>**Citation:** PMID:13982385 | Terms and Conditions | Used in combination with ChEBI to obtain mappings between MeSH identifiers and ChEBI identifiers for chemicals-diseases, chemicals-genes, chemical-GO biological processes, chemicals-GO cellular components, chemicals-GO molecular functions, chemicals-phenotypes, chemicals-proteins, and chemicals-transcripts. |
| Protein Ontology (PRO) | pro_mapping.txt | **URL:** https://proconsortium.org/<br>**Citation:** PMID:20935045 | CC BY 4.0 | To obtain mappings between PRO ontology concepts to other protein, gene, and transcript identifiers form UniProt, Entrez Gene, HGNC, |
| Universal Protein Resource (UniProt) | API Query | **URL:** https://www.uniprot.org/<br>**Citation:** PMID:36408920 | CC BY 4.0 | To obtain mappings between PRO ontology concepts to other protein, gene, and transcript identifiers from the PRO, Entrez Gene, HGNC, |

Note. Sources are reported for the v.2.1.0 knowledge graphs (built May 2021). The full URLs are provided here: https://github.com/callahantiff/PheKnowLator/blob/549e6e1e882e9ea579508ae24a90e64d962deb8c/builds/data_to_download.txt.
[a]The Cell Ontology is included with the extended version of Uberon.

Acronyms: CL (Cell ontology); CLO (Cell Line Ontology); ChEBI (Chemical Entities of Biological Interest); CTD (Comparative Toxicogenomics Database); DOID (Human Disease Ontology); GOOGLE_CLOUD_STORAGE (Google Cloud Storage); GO (Gene Ontology); HGNC (Human Gene Nomenclature Committee); HPO (Human Phenotype Ontology); HPA (Human Protein Atlas); ICD (International Classification of Diseases); MeSH (Medical Subject Headings); Mondo (Mondo Disease Ontology); OMIM (Online Mendelian Inheritance in Man); PRO (Protein Ontology); PRO (Protein Ontology); PW (Pathway Ontology); SO (Sequence Ontology); VO (Vaccine Ontology); Uberon (Uber-Anatomy Ontology); UMLS (Unified Medical Language System).

**Supplementary Table 13. Application of Data Quality Checks to OBO Foundry Ontologies.**

| Statistics[a] | CLO | ChEBI | GO | HPO | Mondo | PRO[b] | PW | RO | SO | Uberon | VO | Merged[c] |
|---|---|---|---|---|---|---|---|---|---|---|---|---|
| Pre-Processed Statistics | | | | | | | | | | | | |
| Edges | 1,387,096 | 5,264,571 | 1,425,434 | 884,999 | 2,313,343 | 2,079,356 | 35,291 | 7,970 | 44,655 | 752,291 | 86,454 | 13,746,883 |
| Classes | 111,712 | 156,098 | 62,237 | 38,843 | 55,478 | 148,243 | 2,642 | 116 | 2,910 | 28,738 | 7,089 | 548,947 |
| Individuals | 41 | 0 | 0 | 0 | 18 | 0 | 0 | 5 | 0 | 0 | 165 | 195 |
| Object Properties | 116 | 10 | 9 | 231 | 331 | 12 | 1 | 604 | 50 | 242 | 232 | 847 |
| Annotation Properties | 192 | 37 | 53 | 257 | 119 | 11 | 19 | 106 | 41 | 284 | 97 | 656 |
| Connected Components | 7 | 1 | 2 | 1 | 1 | 3 | 1 | 3 | 1 | 2 | 5 | 8 |
| Data Quality Check Errors | | | | | | | | | | | | |
| Value Errors | 1 | 0 | 0 | 0 | 0 | 0 | 0 | 0 | 0 | 0 | 0 | 0 |
| Identifier Errors | 0 | 0 | 0 | 0 | 0 | 0 | 0 | 0 | 0 | 0 | 2 | 2 |
| Deprecated Entities | 2 | 18,506 | 6,430 | 304 | 2,305 | 0 | 42 | 11 | 341 | 1,570 | 0 | 0 |
| Obsolete Entities | 13 | 0 | 0 | 0 | 0 | 0 | 0 | 1 | 0 | 0 | 0 | 0 |
| Punning | 16 | 0 | 0 | 0 | 0 | 0 | 0 | 0 | 0 | 0 | 0 | 8 |
| Consistency[d] | Yes | Yes | Yes | Yes | Yes | Yes | Yes | Yes | Yes | Yes | Yes | --- |
| Semantic Heterogeneity | --- | --- | --- | --- | --- | --- | --- | --- | --- | --- | --- | 7 |
| Identifier Alignment | --- | --- | --- | --- | --- | --- | --- | --- | --- | --- | --- | 23,624 |
| Post-Processed Statistics | | | | | | | | | | | | |
| Edges | 1,422,153 | 5,190,485 | 1,343,218 | 885,379 | 2,277,425 | 2,079,356 | 34,901 | 7,873 | 41,980 | 734,768 | 89,764 | 13,748,009 |
| Classes | 111,696 | 137,592 | 55,807 | 38,530 | 52,937 | 148,243 | 2,600 | 115 | 2,569 | 27,170 | 7,085 | 545,259 |
| Individuals | 33 | 0 | 0 | 0 | 17 | 0 | 0 | 5 | 0 | 0 | 165 | 188 |
| Object Properties | 112 | 10 | 9 | 231 | 330 | 12 | 1 | 594 | 50 | 238 | 232 | 846 |
| Annotation Properties | 187 | 37 | 53 | 257 | 119 | 11 | 19 | 106 | 41 | 284 | 97 | 656 |
| Connected Components | 7 | 1 | 2 | 1 | 1 | 3 | 1 | 3 | 1 | 2 | 5 | 8 |

Note. The OBO Foundry ontologies reported above apply to the PKT Human Disease KG v2.1.0. The extended version of Uberon used in this graph imports the full version of the Cell Ontology.

[a]The numbers for the ontologies are calculated using the versions of the ontologies which include all imported ontologies referenced by the primary ontology. This means that the counts of classes include all Web Ontology classes used for logical definitions, not only those that are explicitly part of the primary ontology's namespace.

[b]The PRO version references the human (NCBITaxon_9606) subset created for the PheKnowLator ecosystem.

[c]Merged represents all of the OBO Foundry ontologies merged into a single ontology.

[d]Consistency was evaluated using the ELK reasoner. The reasoner was only applied to individual OBO Foundry ontologies.

Acronyms: OBO (Open Biological and Biomedical Ontologies); CLO (Cell Line Ontology); ChEBI (Chemical Entities of Biological Interest); GO (Gene Ontology); HPO (Human Phenotype Ontology); Mondo (Mondo Disease Ontology); PRO (Protein Ontology); PW (Pathway Ontology); RO (Relation Ontology); SO (Sequence Ontology); Uberon (Uber-Anatomy Ontology); VO (Vaccine Ontology).

**Supplementary Table 14. PheKnowLator Knowledge Modeling Approaches.**

**Example:** Add <<EDNRB, Causes, ABCD syndrome>> to an ontologically-grounded knowledge graph.

**Challenge:** EDNRB is not currently represented in an ontology. ABCD syndrome is a class in the Human Phenotype Ontology, and is included in the knowledge graph.

**Solution:** Gene is a class in the Sequence Ontology and can be used to add EDNRB to the knowledge graph using two different strategies.

| Instance-based Knowledge Model (ABox) | Class-based Knowledge Model (TBox) |
| --- | --- |
| EDNRB, rdfs:subClassOf, Gene<br>EDNRB, rdf:type, owl:Class<br><br>UUID1, rdf:type, EDNRB<br>UUID1, rdf:type, owl:NamedIndividual<br><br>UUID2, rdf:type,ABCD syndrome<br>UUID2, rdf:type, owl:NamedIndividual<br><br>UUID1, Causes, UUID2 | EDNRB, rdfs:subClassOf, Gene<br>EDNRB, rdf:type, owl:Class<br><br>UUID1, rdfs:subClassOf, EDNRB<br>UUID1, rdfs:subClassOf, UUID2<br>UUID2, rdf:type, owl:Restriction<br>UUID2, owl:someValuesFrom, ABCD syndrome<br>UUID2, owl:onProperty, Causes |

Note. UUID1 and UUID2 are blank nodes or existential variables.[162] Pink highlighting is used for the EDNRB gene instance, yellow highlighting is used for the gene class, and green is used for the ABCD syndrome class.

Acronyms: EDNRB (endothelin receptor type B); OWL (Web Ontology Language); RDF (Resource Description Framework); RDFS (Resource Description Framework Syntax).

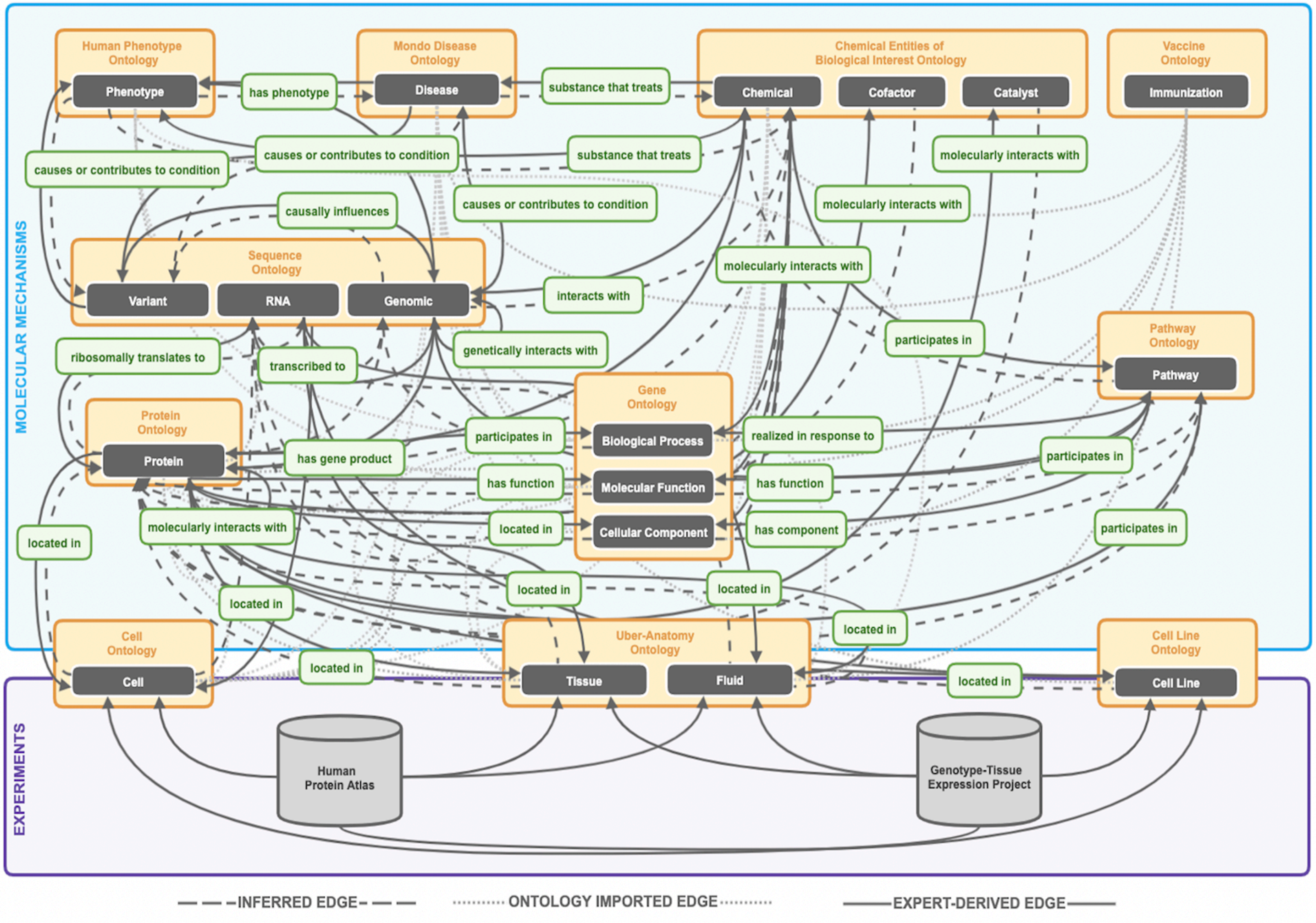

**Supplementary Figure 1. Human Disease Mechanism Graph Knowledge Representation.**

This figure illustrates the knowledge representation used to construct the human disease mechanisms knowledge graphs. The purple box represents experimental data and the blue box contains the molecular mechanisms created by integrating Open Biological and Biomedical (OBO) Foundry ontologies (gold and green). Edges between the ontologies are created by integrating other data sources that are not part of an OBO Foundry ontology (solid black lines). Dashed lines represent relationships that are inferred from the Relation Ontology and dotted lines represent relationships that exist between imported ontologies.

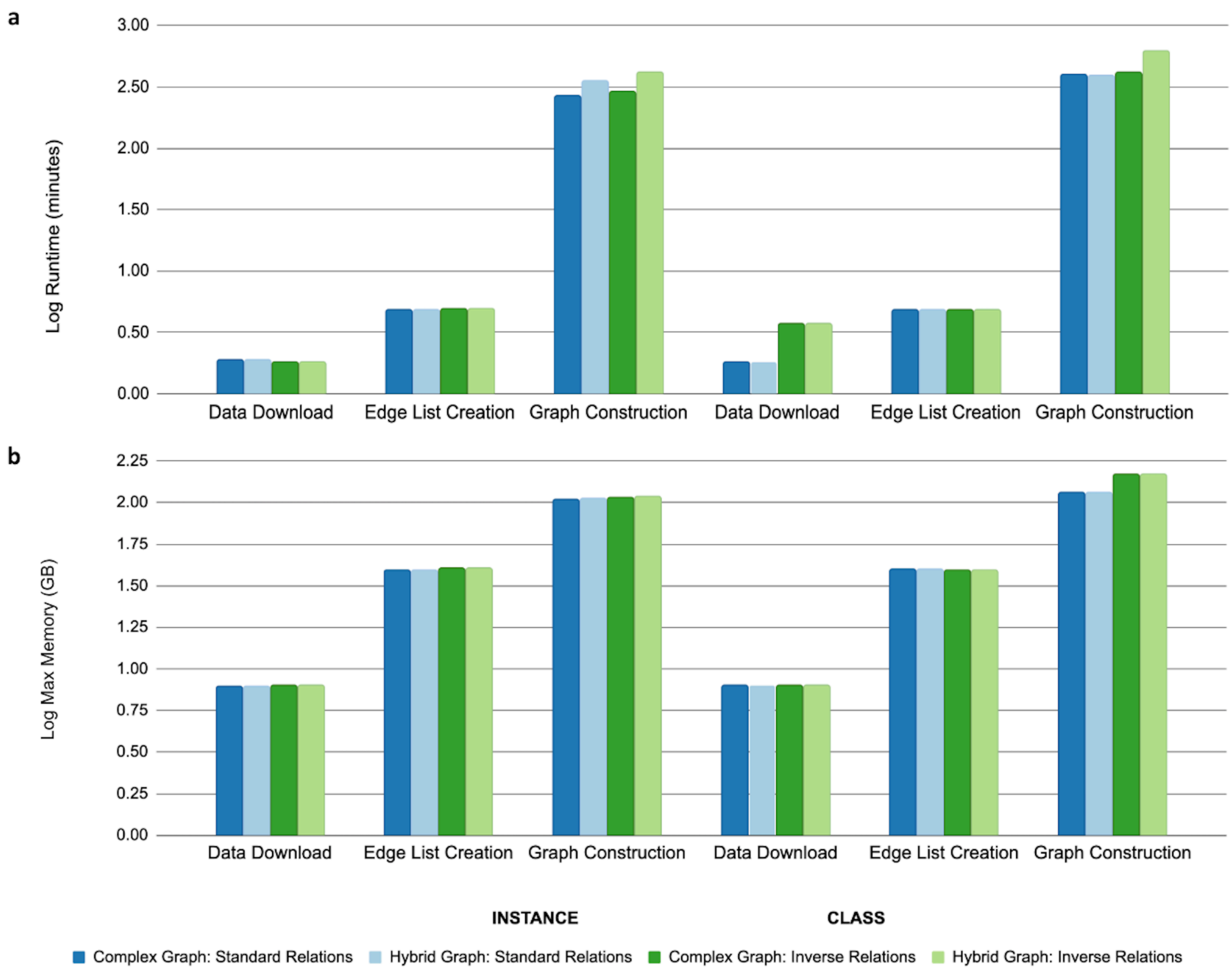

**Supplementary Figure 2. PKT Human Disease Knowledge Graph Construction - Computational Performance.**

This figure illustrates the (**A**) log runtime and (**B**) log max memory use (GB) performance for each build step with respect to the different build parameterizations or benchmarks provided by the PheKnowLator ecosystem. The ecosystem enables users to fully customize KGs generated by the Graph Construction build step through the following parameters: knowledge model (i.e., complex graphs constructed using class- or instance-based knowledge models), relation strategy (i.e., standard directed relations or inverse bidirectional relations), and semantic abstraction (i.e., transformation of complex graphs into hybrid graphs). The Data Download and Edge List Creation steps are the same regardless of how the Graph Construction step is parameterized. Computational performance was determined using an unreleased build (April 11, 2021) while testing the v.2.1 release

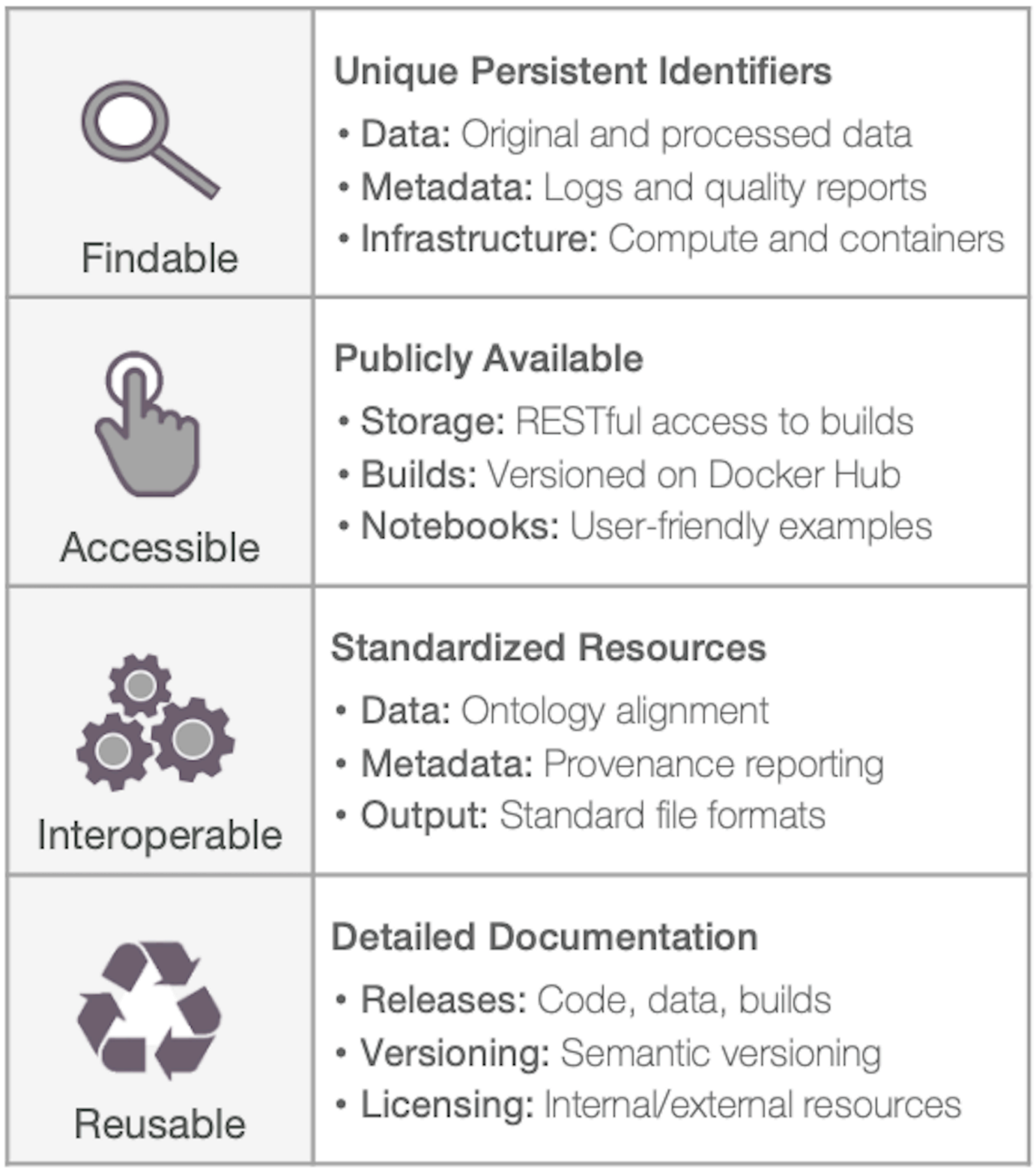

**Supplementary Figure 3. The PheKnowLator Ecosystem on FAIR Principles.**

The PheKnowLator Ecosystem is built on the FAIR principles of Findability, Accessibility, Interoperability, and Reusability.[13] Findability. Use of unique persistent identifiers for all downloaded and processed data, Docker containers, and compute instances and generation of metadata, reports, and logs. Accessibility. All resources are accessible via RESTful API access to a dedicated Google Cloud Storage Bucket (pre-2024) or Zenodo Archive (2024), all builds are versioned, and Jupyter Notebooks are used to improve the usability of the Ecosystem resources. Interoperability. Built on Semantic Web standards, grounded in Open Biological and Biomedical Foundry ontologies, and adoption of standard identifiers for all resources. Reusability. Builds are automated, containerized, and deployed through GitHub Actions workflows, resources, scripts, and workflows are versioned using Semantic Versioning, the Ecosystem is licensed, and licensing constraints are enforced for all ingested data.

# Supplementary Documents

## Knowledge Graph Build Output Metadata

This section provides examples of the metadata files that are output by the ecosystem for each constructed knowledge graph. All of the example files were pulled from the most recent PKT Human Disease KG build, generated on November 1, 2021. For additional details on this build, please see the associated GitHub wiki: https://github.com/callahantiff/PheKnowLator/wiki/November-01%2C-2021. The logs included in this section are for the class-based + standard relations + OWL-NETS knowledge graph type from this build.

### Table of Contents

**Supplementary Document 1. downloaded_build_metadata.txt.**

```
==================================
Mon Nov 01 01:45:15 UTC 2021
==================================

DATA INFO
  - DOWNLOAD_URL = http://purl.obolibrary.org/obo/hp.owl
  - DOWNLOAD_DATE = 11/01/2021
  - FILE_SIZE_IN_BYTES = 84212277
GOOGLE_CLOUD_STORAGE_URL= https://storage.googleapis.com/pheknowlator/archived_builds/release_v3.0.2/build_01NOV2021/data/original_data/hp_with_imports.owl

DATA INFO
  - DOWNLOAD_URL = http://purl.obolibrary.org/obo/go.owl
  - DOWNLOAD_DATE = 11/01/2021
  - FILE_SIZE_IN_BYTES = 132081899
  - GOOGLE_CLOUD_STORAGE_URL = https://storage.googleapis.com/pheknowlator/archived_builds/release_v3.0.2/build_01NOV2021/data/original_data/go_with_imports.owl

DATA INFO
  - DOWNLOAD_URL = http://purl.obolibrary.org/obo/mondo.owl
  - DOWNLOAD_DATE = 11/01/2021
  - FILE_SIZE_IN_BYTES = 235314739
  - GOOGLE_CLOUD_STORAGE_URL = https://storage.googleapis.com/pheknowlator/archived_builds/release_v3.0.2/build_01NOV2021/data/original_data/mondo_with_imports.owl

DATA INFO
  - DOWNLOAD_URL = http://purl.obolibrary.org/obo/vo.owl
  - DOWNLOAD_DATE = 11/01/2021
  - FILE_SIZE_IN_BYTES = 8110267
  - GOOGLE_CLOUD_STORAGE_URL = https://storage.googleapis.com/pheknowlator/archived_builds/release_v3.0.2/build_01NOV2021/data/original_data/vo_with_imports.owl

DATA INFO
  - DOWNLOAD_URL = http://purl.obolibrary.org/obo/chebi.owl
  - DOWNLOAD_DATE = 11/01/2021
  - FILE_SIZE_IN_BYTES = 651338784
  - GOOGLE_CLOUD_STORAGE_URL = https://storage.googleapis.com/pheknowlator/archived_builds/release_v3.0.2/build_01NOV2021/data/original_data/chebi_with_imports.owl

DATA INFO
  - DOWNLOAD_URL = http://purl.obolibrary.org/obo/uberon/ext.owl
  - DOWNLOAD_DATE = 11/01/2021
  - FILE_SIZE_IN_BYTES = 65910831
  - GOOGLE_CLOUD_STORAGE_URL = https://storage.googleapis.com/pheknowlator/archived_builds/release_v3.0.2/build_01NOV2021/data/original_data/ext_with_imports.owl

DATA INFO
  - DOWNLOAD_URL = http://purl.obolibrary.org/obo/clo.owl
  - DOWNLOAD_DATE = 11/01/2021
  - FILE_SIZE_IN_BYTES = 119273027
  - GOOGLE_CLOUD_STORAGE_URL = https://storage.googleapis.com/pheknowlator/archived_builds/release_v3.0.2/build_01NOV2021/data/original_data/clo_with_imports.owl

DATA INFO
  - DOWNLOAD_URL = http://purl.obolibrary.org/obo/pr.owl
  - DOWNLOAD_DATE = 11/01/2021
  - FILE_SIZE_IN_BYTES = 1223936557
  - GOOGLE_CLOUD_STORAGE_URL = https://storage.googleapis.com/pheknowlator/archived_builds/release_v3.0.2/build_01NOV2021/data/original_data/pr_with_imports.owl

DATA INFO
  - DOWNLOAD_URL = http://purl.obolibrary.org/obo/so.owl
  - DOWNLOAD_DATE = 11/01/2021
  - FILE_SIZE_IN_BYTES = 5225970
  - GOOGLE_CLOUD_STORAGE_URL = https://storage.googleapis.com/pheknowlator/archived_builds/release_v3.0.2/build_01NOV2021/data/original_data/so_with_imports.owl

DATA INFO
  - DOWNLOAD_URL = http://purl.obolibrary.org/obo/pw.owl
  - DOWNLOAD_DATE = 11/01/2021
  - FILE_SIZE_IN_BYTES = 4965358
```

- GOOGLE_CLOUD_STORAGE_URL = https://storage.googleapis.com/pheknowlator/archived_builds/release_v3.0.2/build_01NOV2021/data/original_data/pw_with_imports.owl

DATA INFO
- DOWNLOAD_URL = http://purl.obolibrary.org/obo/ro.owl
- DOWNLOAD_DATE = 11/01/2021
- FILE_SIZE_IN_BYTES = 867789
- GOOGLE_CLOUD_STORAGE_URL = https://storage.googleapis.com/pheknowlator/archived_builds/release_v3.0.2/build_01NOV2021/data/original_data/ro_with_imports.owl

DATA INFO
- DOWNLOAD_URL = http://ftp.ebi.ac.uk/pub/databases/genenames/hgnc/tsv/hgnc_complete_set.txt
- DOWNLOAD_DATE = 11/01/2021
- FILE_SIZE_IN_BYTES = 15972451
- GOOGLE_CLOUD_STORAGE_URL = https://storage.googleapis.com/pheknowlator/archived_builds/release_v3.0.2/build_01NOV2021/data/original_data/hgnc_complete_set.txt

DATA INFO
- DOWNLOAD_URL = ftp://ftp.ensembl.org/pub/release-102/gtf/homo_sapiens/Homo_sapiens.GRCh38.102.gtf.gz
- DOWNLOAD_DATE = 11/01/2021
- FILE_SIZE_IN_BYTES = 1280942256
- GOOGLE_CLOUD_STORAGE_URL = https://storage.googleapis.com/pheknowlator/archived_builds/release_v3.0.2/build_01NOV2021/data/original_data/Homo_sapiens.GRCh38.102.gtf

DATA INFO
- DOWNLOAD_URL = ftp://ftp.ensembl.org/pub/release-102/tsv/homo_sapiens/Homo_sapiens.GRCh38.102.uniprot.tsv.gz
- DOWNLOAD_DATE = 11/01/2021
- FILE_SIZE_IN_BYTES = 13485382

GOOGLE_CLOUD_STORAGE_URL= https://storage.googleapis.com/pheknowlator/archived_builds/release_v3.0.2/build_01NOV2021/data/original_data/Homo_sapiens.GRCh38.102.uniprot.tsv

DATA INFO
- DOWNLOAD_URL = ftp://ftp.ensembl.org/pub/release-102/tsv/homo_sapiens/Homo_sapiens.GRCh38.102.entrez.tsv.gz
- DOWNLOAD_DATE = 11/01/2021
- FILE_SIZE_IN_BYTES = 17055479

GOOGLE_CLOUD_STORAGE_URL= https://storage.googleapis.com/pheknowlator/archived_builds/release_v3.0.2/build_01NOV2021/data/original_data/Homo_sapiens.GRCh38.102.entrez.tsv

DATA INFO
DOWNLOAD_URL=
https://www.uniprot.org/uniprot/?query=&fil=organism%3A%22Homo%20sapiens%20(Human)%20%5B9606%5D%22&columns=id%2Creviewed%2Cdatabase(GeneID)%2Cdatabase(Ensembl)%2Cdatabase(HGNC)%2Cgenes(ALTERNATIVE)%2Cgenes(PREF
ERRED)&format=tab
- DOWNLOAD_DATE = 11/01/2021
- FILE_SIZE_IN_BYTES = 8759737
- GOOGLE_CLOUD_STORAGE_URL = https://storage.googleapis.com/pheknowlator/archived_builds/release_v3.0.2/build_01NOV2021/data/original_data/uniprot_identifier_mapping.tab

DATA INFO
- DOWNLOAD_URL = ftp://ftp.ncbi.nih.gov/gene/DATA/GENE_INFO/Mammalia/Homo_sapiens.gene_info.gz
- DOWNLOAD_DATE = 11/01/2021
- FILE_SIZE_IN_BYTES = 13909469
- GOOGLE_CLOUD_STORAGE_URL = https://storage.googleapis.com/pheknowlator/archived_builds/release_v3.0.2/build_01NOV2021/data/original_data/Homo_sapiens.gene_info

DATA INFO
- DOWNLOAD_URL = https://proconsortium.org/download/current/promapping.txt
- DOWNLOAD_DATE = 11/01/2021
- FILE_SIZE_IN_BYTES = 15271039
- GOOGLE_CLOUD_STORAGE_URL = https://storage.googleapis.com/pheknowlator/archived_builds/release_v3.0.2/build_01NOV2021/data/original_data/promapping.txt

DATA INFO
- DOWNLOAD_URL = ftp://nlmpubs.nlm.nih.gov/online/mesh/rdf/2021/mesh2021.nt
- DOWNLOAD_DATE = 11/01/2021
- FILE_SIZE_IN_BYTES = 1948657299
- GOOGLE_CLOUD_STORAGE_URL = https://storage.googleapis.com/pheknowlator/archived_builds/release_v3.0.2/build_01NOV2021/data/original_data/mesh2021.nt

DATA INFO
- DOWNLOAD_URL = ftp://ftp.ebi.ac.uk/pub/databases/chebi/Flat_file_tab_delimited/names.tsv.gz
- DOWNLOAD_DATE = 11/01/2021
- FILE_SIZE_IN_BYTES = 30365252
- GOOGLE_CLOUD_STORAGE_URL = https://storage.googleapis.com/pheknowlator/archived_builds/release_v3.0.2/build_01NOV2021/data/original_data/names.tsv

DATA INFO
- DOWNLOAD_URL = https://www.disgenet.org/static/disgenet_ap1/files/downloads/disease_mappings.tsv.gz
- DOWNLOAD_DATE = 11/01/2021

- FILE_SIZE_IN_BYTES = 18367848
- GOOGLE_CLOUD_STORAGE_URL = https://storage.googleapis.com/pheknowlator/archived_builds/release_v3.0.2/build_01NOV2021/data/original_data/disease_mappings.tsv

DATA INFO
DOWNLOAD_URL=
https://www.proteinatlas.org/api/search_download.php?search=&columns=g,eg,up,pe,rnatsm,rnaclsm,rnacasm,rnabrsm,rnabcsm,rnablsm,scl,t_RNA_adipose_tissue,t_RNA_adrenal_gland,t_RNA_amygdala,t_RNA_appendix,t_RNA_basal_ganglia,t_RNA_bone_marrow,t_RNA_breast,t_RNA_cerebellum,t_RNA_cerebral_cortex,t_RNA_cervix,_uterine,t_RNA_colon,t_RNA_corpus_callosum,t_RNA_ductus_deferens,t_RNA_duodenum,t_RNA_endometrium_1,t_RNA_epididymis,t_RNA_esophagus,t_RNA_fallopian_tube,t_RNA_gallbladder,t_RNA_heart_muscle,t_RNA_hippocampal_formation,t_RNA_hypothalamus,t_RNA_kidney,t_RNA_liver,t_RNA_lung,t_RNA_lymph_node,t_RNA_midbrain,t_RNA_olfactory_region,t_RNA_ovary,t_RNA_pancreas,t_RNA_parathyroid_gland,t_RNA_pituitary_gland,t_RNA_placenta,t_RNA_pons_and_medulla,t_RNA_prostate,t_RNA_rectum,t_RNA_retina,t_RNA_salivary_gland,t_RNA_seminal_vesicle,t_RNA_skeletal_muscle,t_RNA_skin_1,t_RNA_small_intestine,t_RNA_smooth_muscle,t_RNA_spinal_cord,t_RNA_spleen,t_RNA_stomach_1,t_RNA_testis,t_RNA_thalamus,t_RNA_thymus,t_RNA_thyroid_gland,t_RNA_tongue,t_RNA_tonsil,t_RNA_urinary_bladder,t_RNA_vagina,t_RNA_B-cells,t_RNA_dendritic_cells,t_RNA_granulocytes,t_RNA_monocytes,t_RNA_NK-cells,t_RNA_T-cells,t_RNA_total_PBMC,cell_RNA_A-431,cell_RNA_A549,cell_RNA_AF22,cell_RNA_AN3-CA,cell_RNA_ASC_diff,cell_RNA_ASC_TERT1,cell_RNA_BEWO,cell_RNA_BJ,cell_RNA_BJ_hTERT+,cell_RNA_BJ_hTERT+_SV40_Large_T+,cell_RNA_BJ_hTERT+_SV40_Large_T+_RasG12V,cell_RNA_CACO-2,cell_RNA_CAPAN-2,cell_RNA_Daudi,cell_RNA_EFO-21,cell_RNA_fHDF/TERT166,cell_RNA_HaCaT,cell_RNA_HAP1,cell_RNA_HBEC3-KT,cell_RNA_HBF_TERT88,cell_RNA_HDLM-2,cell_RNA_HEK_293,cell_RNA_HEL,cell_RNA_HeLa,cell_RNA_Hep_G2,cell_RNA_HHSteC,cell_RNA_HL-60,cell_RNA_HMC-1,cell_RNA_HSkMC,cell_RNA_hTCEpi,cell_RNA_hTEC/SVTERT24-B,cell_RNA_hTERT-HME1,cell_RNA_HUVEC_TERT2,cell_RNA_K-562,cell_RNA_Karpas-707,cell_RNA_LHCN-M2,cell_RNA_MCF7,cell_RNA_MOLT-4,cell_RNA_NB-4,cell_RNA_NTERA-2,cell_RNA_PC-3,cell_RNA_REH,cell_RNA_RH-30,cell_RNA_RPMI-8226,cell_RNA_RPTEC_TERT1,cell_RNA_RT4,cell_RNA_SCLC-21H,cell_RNA_SH-SY5Y,cell_RNA_SiHa,cell_RNA_SK-BR-3,cell_RNA_SK-MEL-30,cell_RNA_T-47d,cell_RNA_THP-1,cell_RNA_TIME,cell_RNA_U-138_MG,cell_RNA_U-2_OS,cell_RNA_U-2197,cell_RNA_U-251_MG,cell_RNA_U-266/70,cell_RNA_U-266/84,cell_RNA_U-698,cell_RNA_U-87_MG,cell_RNA_U-937,cell_RNA_WM-115,blood_RNA_basophil,blood_RNA_classical_monocyte,blood_RNA_eosinophil,blood_RNA_gdT-cell,blood_RNA_intermediate_monocyte,blood_RNA_MAIT_T-cell,blood_RNA_memory_B-cell,blood_RNA_memory_CD4_T-cell,blood_RNA_memory_CD8_T-cell,blood_RNA_myeloid_DC,blood_RNA_naive_B-cell,blood_RNA_naive_CD4_T-cell,blood_RNA_naive_CD8_T-cell,blood_RNA_neutrophil,blood_RNA_NK-cell,blood_RNA_non-classical_monocyte,blood_RNA_plasmacytoid_DC,blood_RNA_T-reg,blood_RNA_total_PBMC,brain_RNA_amygdala,brain_RNA_basal_ganglia,brain_RNA_cerebellum,brain_RNA_cerebral_cortex,brain_RNA_hippocampal_formation,brain_RNA_hypothalamus,brain_RNA_midbrain,brain_RNA_olfactory_region,brain_RNA_pons_and_medulla,brain_RNA_thalamus&format=tsv
- DOWNLOAD_DATE = 11/01/2021
- FILE_SIZE_IN_BYTES = 15328159
- GOOGLE_CLOUD_STORAGE_URL = https://storage.googleapis.com/pheknowlator/archived_builds/release_v3.0.2/build_01NOV2021/data/original_data/proteinatlas_search.tsv

DATA INFO
- DOWNLOAD_URL = https://storage.googleapis.com/gtex_analysis_v8/rna_seq_data/GTEx_Analysis_2017-06-05_v8_RNASeQCv1.1.9_gene_median_tpm.gct.gz
- DOWNLOAD_DATE = 11/01/2021
- FILE_SIZE_IN_BYTES = 17780477

GOOGLE_CLOUD_STORAGE_URL= https://storage.googleapis.com/pheknowlator/archived_builds/release_v3.0.2/build_01NOV2021/data/original_data/GTEx_Analysis_2017-06-05_v8_RNASeQCv1.1.9_gene_median_tpm.gct

DATA INFO
- DOWNLOAD_URL = https://reactome.org/download/current/ReactomePathways.txt
- DOWNLOAD_DATE = 11/01/2021
- FILE_SIZE_IN_BYTES = 1423494
- GOOGLE_CLOUD_STORAGE_URL = https://storage.googleapis.com/pheknowlator/archived_builds/release_v3.0.2/build_01NOV2021/data/original_data/ReactomePathways.txt

DATA INFO
- DOWNLOAD_URL = https://reactome.org/download/current/gene_association.reactome.gz
- DOWNLOAD_DATE = 11/01/2021
- FILE_SIZE_IN_BYTES = 11955743
- GOOGLE_CLOUD_STORAGE_URL = https://storage.googleapis.com/pheknowlator/archived_builds/release_v3.0.2/build_01NOV2021/data/original_data/gene_association.reactome

DATA INFO
- DOWNLOAD_URL = https://reactome.org/download/current/ChEBI2Reactome_All_Levels.txt
- DOWNLOAD_DATE = 11/01/2021
- FILE_SIZE_IN_BYTES = 30485396
- GOOGLE_CLOUD_STORAGE_URL = https://storage.googleapis.com/pheknowlator/archived_builds/release_v3.0.2/build_01NOV2021/data/original_data/ChEBI2Reactome_All_Levels.txt

DATA INFO
- DOWNLOAD_URL = http://compath.scai.fraunhofer.de/export_mappings
- DOWNLOAD_DATE = 11/01/2021
- FILE_SIZE_IN_BYTES = 196388

GOOGLE_CLOUD_STORAGE_URL= https://storage.googleapis.com/pheknowlator/archived_builds/release_v3.0.2/build_01NOV2021/data/original_data/compath_canonical_pathway_mappings.txt

DATA INFO
- DOWNLOAD_URL = https://raw.githubusercontent.com/ComPath/resources/master/mappings/kegg_reactome.csv
- DOWNLOAD_DATE = 11/01/2021
- FILE_SIZE_IN_BYTES = 92309
- GOOGLE_CLOUD_STORAGE_URL = https://storage.googleapis.com/pheknowlator/archived_builds/release_v3.0.2/build_01NOV2021/data/original_data/kegg_reactome.csv

DATA INFO
- DOWNLOAD_URL = https://storage.googleapis.com/pheknowlator/curated_data/genomic_sequence_ontology_mappings.xlsx
- DOWNLOAD_DATE = 11/01/2021
- FILE_SIZE_IN_BYTES = 20641

GOOGLE_CLOUD_STORAGE_URL= https://storage.googleapis.com/pheknowlator/archived_builds/release_v3.0.2/build_01NOV2021/data/original_data/genomic_sequence_ontology_mappings.xlsx

DATA INFO

DOWNLOAD_URL=
https://sparql.proconsortium.org/virtuoso/sparql?query=PREFIX+obo%3A+%3Chttp%3A%2F%2Fpurl.obolibrary.org%2Fobo%2F%3E%0D%0A%0D%0ASELECT+%3FPRO_term%0D%0AFROM+%3Chttp%3A%2F%2Fpurl.obolibrary.org%2Fobo%2Fpr%3E%0D%0AWHERE+%7B%0D%0A+++++++%3FPRO_term+rdf%3Atype+owl%3AClass+.%0D%0A+++++++%3FPRO_term+rdfs%3AsubClassOf+%3Frestriction+.%0D%0A+++++++%3Frestriction+owl%3AonProperty+obo%3ARO_0002160+.%0D%0A+++++++%3Frestriction+owl%3AsomeValuesFrom+obo%3ANCBITaxon_9606+.%0D%0A%0D%0A+++++++%23+use+this+to+filter-out+things+like+hgnc+ids%0D%0A+++++++FILTER+%28regex%28%3FPRO_term%2C%22http%3A%2F%2Fpurl.obolibrary.org%2Fobo%2F*%22%29%29+.%0D%0A%7D&format=text%2Fhtml&debug=
- DOWNLOAD_DATE = 11/01/2021
- FILE_SIZE_IN_BYTES = 1301002
- GOOGLE_CLOUD_STORAGE_URL = https://storage.googleapis.com/pheknowlator/archived_builds/release_v3.0.2/build_01NOV2021/data/original_data/human_pro_classes.html

DATA INFO
- DOWNLOAD_URL = ftp://ftp.ncbi.nlm.nih.gov/pub/clinvar/tab_delimited/variant_summary.txt.gz
- DOWNLOAD_DATE = 11/01/2021
- FILE_SIZE_IN_BYTES = 954216609
- GOOGLE_CLOUD_STORAGE_URL = https://storage.googleapis.com/pheknowlator/archived_builds/release_v3.0.2/build_01NOV2021/data/original_data/variant_summary.txt

DATA INFO
DOWNLOAD_URL=
https://www.uniprot.org/uniprot/?query=&fil=organism%3A%22Homo%20sapiens%20(Human)%20%5B9606%5D%22&columns=id%2Creviewed%2Centry%20name%2Cdatabase(PRO)%2Cchebi(Cofactor)%2Cchebi(Catalytic%20activity)&format=tab
- DOWNLOAD_DATE = 11/01/2021
- FILE_SIZE_IN_BYTES = 10491909
- GOOGLE_CLOUD_STORAGE_URL = https://storage.googleapis.com/pheknowlator/archived_builds/release_v3.0.2/build_01NOV2021/data/original_data/uniprot-cofactor-catalyst.tab

DATA INFO
- DOWNLOAD_URL = http://ctdbase.org/reports/CTD_chem_gene_ixns.tsv.gz
- DOWNLOAD_DATE = 11/01/2021
- FILE_SIZE_IN_BYTES = 441097103
- GOOGLE_CLOUD_STORAGE_URL = https://storage.googleapis.com/pheknowlator/archived_builds/release_v3.0.2/build_01NOV2021/data/original_data/CTD_chem_gene_ixns.tsv

DATA INFO
- DOWNLOAD_URL = http://ctdbase.org/reports/CTD_chem_go_enriched.tsv.gz
- DOWNLOAD_DATE = 11/01/2021
- FILE_SIZE_IN_BYTES = 817733569
- GOOGLE_CLOUD_STORAGE_URL = https://storage.googleapis.com/pheknowlator/archived_builds/release_v3.0.2/build_01NOV2021/data/original_data/CTD_chem_go_enriched.tsv

DATA INFO
- DOWNLOAD_URL = https://reactome.org/download/current/ChEBI2Reactome_All_Levels.txt
- DOWNLOAD_DATE = 11/01/2021
- FILE_SIZE_IN_BYTES = 30485396
- GOOGLE_CLOUD_STORAGE_URL = https://storage.googleapis.com/pheknowlator/archived_builds/release_v3.0.2/build_01NOV2021/data/original_data/ChEBI2Reactome_All_Levels.txt

DATA INFO
- DOWNLOAD_URL = http://ctdbase.org/reports/CTD_chemicals_diseases.tsv.gz
- DOWNLOAD_DATE = 11/01/2021
- FILE_SIZE_IN_BYTES = 701201197
- GOOGLE_CLOUD_STORAGE_URL = https://storage.googleapis.com/pheknowlator/archived_builds/release_v3.0.2/build_01NOV2021/data/original_data/CTD_chemicals_diseases.tsv

DATA INFO
- DOWNLOAD_URL = http://purl.obolibrary.org/obo/hp/hpoa/phenotype.hpoa
- DOWNLOAD_DATE = 11/01/2021
- FILE_SIZE_IN_BYTES = 27315334
- GOOGLE_CLOUD_STORAGE_URL = https://storage.googleapis.com/pheknowlator/archived_builds/release_v3.0.2/build_01NOV2021/data/original_data/phenotype.hpoa

DATA INFO
- DOWNLOAD_URL = https://www.disgenet.org/static/disgenet_ap1/files/downloads/curated_gene_disease_associations.tsv.gz
- DOWNLOAD_DATE = 11/01/2021
- FILE_SIZE_IN_BYTES = 11542996

GOOGLE_CLOUD_STORAGE_URL= https://storage.googleapis.com/pheknowlator/archived_builds/release_v3.0.2/build_01NOV2021/data/original_data/curated_gene_disease_associations.tsv

DATA INFO
- DOWNLOAD_URL = http://genemania.org/data/current/Homo_sapiens.COMBINED/COMBINED.DEFAULT_NETWORKS.BP_COMBINING.txt
- DOWNLOAD_DATE = 11/01/2021
- FILE_SIZE_IN_BYTES = 246190113

GOOGLE_CLOUD_STORAGE_URL= https://storage.googleapis.com/pheknowlator/archived_builds/release_v3.0.2/build_01NOV2021/data/original_data/COMBINED.DEFAULT_NETWORKS.BP_COMBINING.txt

DATA INFO
- DOWNLOAD_URL = http://ctdbase.org/reports/CTD_genes_pathways.tsv.gz

- DOWNLOAD_DATE = 11/01/2021
- FILE_SIZE_IN_BYTES = 8192661
- GOOGLE_CLOUD_STORAGE_URL = https://storage.googleapis.com/pheknowlator/archived_builds/release_v3.0.2/build_01NOV2021/data/original_data/CTD_genes_pathways.tsv

DATA INFO
- DOWNLOAD_URL = https://reactome.org/download/current/gene_association.reactome.gz
- DOWNLOAD_DATE = 11/01/2021
- FILE_SIZE_IN_BYTES = 11955743
- GOOGLE_CLOUD_STORAGE_URL = https://storage.googleapis.com/pheknowlator/archived_builds/release_v3.0.2/build_01NOV2021/data/original_data/gene_association.reactome

DATA INFO
- DOWNLOAD_URL = http://current.geneontology.org/annotations/goa_human.gaf.gz
- DOWNLOAD_DATE = 11/01/2021
- FILE_SIZE_IN_BYTES = 109325311
- GOOGLE_CLOUD_STORAGE_URL = https://storage.googleapis.com/pheknowlator/archived_builds/release_v3.0.2/build_01NOV2021/data/original_data/goa_human.gaf

DATA INFO
- DOWNLOAD_URL = https://reactome.org/download/current/UniProt2Reactome_All_Levels.txt
- DOWNLOAD_DATE = 11/01/2021
- FILE_SIZE_IN_BYTES = 102191227
- GOOGLE_CLOUD_STORAGE_URL = https://storage.googleapis.com/pheknowlator/archived_builds/release_v3.0.2/build_01NOV2021/data/original_data/UniProt2Reactome_All_Levels.txt

DATA INFO
- DOWNLOAD_URL = https://stringdb-static.org/download/protein.links.v11.0/9606.protein.links.v11.0.txt.gz
- DOWNLOAD_DATE = 11/01/2021
- FILE_SIZE_IN_BYTES = 540934917
- GOOGLE_CLOUD_STORAGE_URL = https://storage.googleapis.com/pheknowlator/archived_builds/release_v3.0.2/build_01NOV2021/data/original_data/9606.protein.links.v11.0.txt

DATA INFO
- DOWNLOAD_URL = https://storage.googleapis.com/pheknowlator/curated_data/genomic_typing_dict.pkl
- DOWNLOAD_DATE = 11/01/2021
- FILE_SIZE_IN_BYTES = 2268
- GOOGLE_CLOUD_STORAGE_URL = https://storage.googleapis.com/pheknowlator/archived_builds/release_v3.0.2/build_01NOV2021/data/original_data/genomic_typing_dict.pkl

DATA INFO
- DOWNLOAD_URL = https://storage.googleapis.com/pheknowlator/curated_data/zooma_tissue_cell_mapping_04JAN2020.xlsx
- DOWNLOAD_DATE = 11/01/2021
- FILE_SIZE_IN_BYTES = 18225305

GOOGLE_CLOUD_STORAGE_URL= https://storage.googleapis.com/pheknowlator/archived_builds/release_v3.0.2/build_01NOV2021/data/original_data/zooma_tissue_cell_mapping_04JAN2020.xlsx

**Supplementary Document 2. preprocessed_build_metadata.txt**

```
==================================
Mon Nov 01 10:00:03 UTC 2021
==================================

DATA INFO
DOWNLOAD_URL= https://storage.googleapis.com/pheknowlator/archived_builds/release_v3.0.2/build_01NOV2021/data/original_data/CLINVAR_VARIANT_GENE_DISEASE_PHENOTYPE_EDGES.txt
  - DOWNLOAD_DATE = 11/01/2021
  - FILE_SIZE_IN_BYTES = 3662353455
GOOGLE_CLOUD_STORAGE_URL = https://storage.googleapis.com/pheknowlator/archived_builds/release_v3.0.2/build_01NOV2021/data/processed_data/CLINVAR_VARIANT_GENE_DISEASE_PHENOTYPE_EDGES.txt

DATA INFO
  - DOWNLOAD_URL = https://storage.googleapis.com/pheknowlator/archived_builds/release_v3.0.2/build_01NOV2021/data/original_data/DISEASE_MONDO_MAP.txt
  - DOWNLOAD_DATE = 11/01/2021
  - FILE_SIZE_IN_BYTES = 3871799
  - GOOGLE_CLOUD_STORAGE_URL = https://storage.googleapis.com/pheknowlator/archived_builds/release_v3.0.2/build_01NOV2021/data/processed_data/DISEASE_MONDO_MAP.txt

DATA INFO
  - DOWNLOAD_URL = https://storage.googleapis.com/pheknowlator/archived_builds/release_v3.0.2/build_01NOV2021/data/original_data/ENSEMBL_GENE_ENTREZ_GENE_MAP.txt
  - DOWNLOAD_DATE = 11/01/2021
  - FILE_SIZE_IN_BYTES = 3564948
  - GOOGLE_CLOUD_STORAGE_URL = https://storage.googleapis.com/pheknowlator/archived_builds/release_v3.0.2/build_01NOV2021/data/processed_data/ENSEMBL_GENE_ENTREZ_GENE_MAP.txt

DATA INFO
  - DOWNLOAD_URL = https://storage.googleapis.com/pheknowlator/archived_builds/release_v3.0.2/build_01NOV2021/data/original_data/ENSEMBL_TRANSCRIPT_PROTEIN_ONTOLOGY_MAP.txt
  - DOWNLOAD_DATE = 11/01/2021
  - FILE_SIZE_IN_BYTES = 2931638
  - GOOGLE_CLOUD_STORAGE_URL = https://storage.googleapis.com/pheknowlator/archived_builds/release_v3.0.2/build_01NOV2021/data/processed_data/ENSEMBL_TRANSCRIPT_PROTEIN_ONTOLOGY_MAP.txt

DATA INFO
  - DOWNLOAD_URL = https://storage.googleapis.com/pheknowlator/archived_builds/release_v3.0.2/build_01NOV2021/data/original_data/ENTREZ_GENE_ENSEMBL_TRANSCRIPT_MAP.txt
  - DOWNLOAD_DATE = 11/01/2021
  - FILE_SIZE_IN_BYTES = 15146432
  - GOOGLE_CLOUD_STORAGE_URL = https://storage.googleapis.com/pheknowlator/archived_builds/release_v3.0.2/build_01NOV2021/data/processed_data/ENTREZ_GENE_ENSEMBL_TRANSCRIPT_MAP.txt

DATA INFO
  - DOWNLOAD_URL = https://storage.googleapis.com/pheknowlator/archived_builds/release_v3.0.2/build_01NOV2021/data/original_data/ENTREZ_GENE_PRO_ONTOLOGY_MAP.txt
  - DOWNLOAD_DATE = 11/01/2021
  - FILE_SIZE_IN_BYTES = 1124944
  - GOOGLE_CLOUD_STORAGE_URL = https://storage.googleapis.com/pheknowlator/archived_builds/release_v3.0.2/build_01NOV2021/data/processed_data/ENTREZ_GENE_PRO_ONTOLOGY_MAP.txt

DATA INFO
  - DOWNLOAD_URL = https://storage.googleapis.com/pheknowlator/archived_builds/release_v3.0.2/build_01NOV2021/data/original_data/GENE_SYMBOL_ENSEMBL_TRANSCRIPT_MAP.txt
  - DOWNLOAD_DATE = 11/01/2021
  - FILE_SIZE_IN_BYTES = 19365937
  - GOOGLE_CLOUD_STORAGE_URL = https://storage.googleapis.com/pheknowlator/archived_builds/release_v3.0.2/build_01NOV2021/data/processed_data/GENE_SYMBOL_ENSEMBL_TRANSCRIPT_MAP.txt

DATA INFO
  - DOWNLOAD_URL = https://storage.googleapis.com/pheknowlator/archived_builds/release_v3.0.2/build_01NOV2021/data/original_data/HPA_GTEX_RNA_GENE_PROTEIN_EDGES.txt
  - DOWNLOAD_DATE = 11/01/2021
  - FILE_SIZE_IN_BYTES = 20611429
  - GOOGLE_CLOUD_STORAGE_URL = https://storage.googleapis.com/pheknowlator/archived_builds/release_v3.0.2/build_01NOV2021/data/processed_data/HPA_GTEX_RNA_GENE_PROTEIN_EDGES.txt

DATA INFO
  - DOWNLOAD_URL = https://storage.googleapis.com/pheknowlator/archived_builds/release_v3.0.2/build_01NOV2021/data/original_data/HPA_GTEx_TISSUE_CELL_MAP.txt
  - DOWNLOAD_DATE = 11/01/2021
  - FILE_SIZE_IN_BYTES = 8222
  - GOOGLE_CLOUD_STORAGE_URL = https://storage.googleapis.com/pheknowlator/archived_builds/release_v3.0.2/build_01NOV2021/data/processed_data/HPA_GTEx_TISSUE_CELL_MAP.txt

DATA INFO
  - DOWNLOAD_URL = https://storage.googleapis.com/pheknowlator/archived_builds/release_v3.0.2/build_01NOV2021/data/original_data/HPA_tissues.txt
  - DOWNLOAD_DATE = 11/01/2021
  - FILE_SIZE_IN_BYTES = 1530
  - GOOGLE_CLOUD_STORAGE_URL = https://storage.googleapis.com/pheknowlator/archived_builds/release_v3.0.2/build_01NOV2021/data/processed_data/HPA_tissues.txt
```

DATA INFO
- DOWNLOAD_URL = https://storage.googleapis.com/pheknowlator/archived_builds/release_v3.0.2/build_01NOV2021/data/original_data/INVERSE_RELATIONS.txt
- DOWNLOAD_DATE = 11/01/2021
- FILE_SIZE_IN_BYTES = 4602
- GOOGLE_CLOUD_STORAGE_URL = https://storage.googleapis.com/pheknowlator/archived_builds/release_v3.0.2/build_01NOV2021/data/processed_data/INVERSE_RELATIONS.txt

DATA INFO
- DOWNLOAD_URL = https://storage.googleapis.com/pheknowlator/archived_builds/release_v3.0.2/build_01NOV2021/data/original_data/MESH_CHEBI_MAP.txt
- DOWNLOAD_DATE = 11/01/2021
- FILE_SIZE_IN_BYTES = 358052
- GOOGLE_CLOUD_STORAGE_URL = https://storage.googleapis.com/pheknowlator/archived_builds/release_v3.0.2/build_01NOV2021/data/processed_data/MESH_CHEBI_MAP.txt

DATA INFO
- DOWNLOAD_URL = https://storage.googleapis.com/pheknowlator/archived_builds/release_v3.0.2/build_01NOV2021/data/original_data/Merged_gene_rna_protein_identifiers.pkl
- DOWNLOAD_DATE = 11/01/2021
- FILE_SIZE_IN_BYTES = 539048900
- GOOGLE_CLOUD_STORAGE_URL = https://storage.googleapis.com/pheknowlator/archived_builds/release_v3.0.2/build_01NOV2021/data/processed_data/Merged_gene_rna_protein_identifiers.pkl

DATA INFO
- DOWNLOAD_URL = https://storage.googleapis.com/pheknowlator/archived_builds/release_v3.0.2/build_01NOV2021/data/original_data/PHENOTYPE_HPO_MAP.txt
- DOWNLOAD_DATE = 11/01/2021
- FILE_SIZE_IN_BYTES = 719692
- GOOGLE_CLOUD_STORAGE_URL = https://storage.googleapis.com/pheknowlator/archived_builds/release_v3.0.2/build_01NOV2021/data/processed_data/PHENOTYPE_HPO_MAP.txt

DATA INFO
- DOWNLOAD_URL = https://storage.googleapis.com/pheknowlator/archived_builds/release_v3.0.2/build_01NOV2021/data/original_data/PheKnowLator_MergedOntologies.owl
- DOWNLOAD_DATE = 11/01/2021
- FILE_SIZE_IN_BYTES = 1450356910
- GOOGLE_CLOUD_STORAGE_URL = https://storage.googleapis.com/pheknowlator/archived_builds/release_v3.0.2/build_01NOV2021/data/processed_data/PheKnowLator_MergedOntologies.owl

DATA INFO
- DOWNLOAD_URL = https://storage.googleapis.com/pheknowlator/archived_builds/release_v3.0.2/build_01NOV2021/data/original_data/REACTOME_PW_GO_MAPPINGS.txt
- DOWNLOAD_DATE = 11/01/2021
- FILE_SIZE_IN_BYTES = 398373
- GOOGLE_CLOUD_STORAGE_URL = https://storage.googleapis.com/pheknowlator/archived_builds/release_v3.0.2/build_01NOV2021/data/processed_data/REACTOME_PW_GO_MAPPINGS.txt

DATA INFO
- DOWNLOAD_URL = https://storage.googleapis.com/pheknowlator/archived_builds/release_v3.0.2/build_01NOV2021/data/original_data/RELATIONS_LABELS.txt
- DOWNLOAD_DATE = 11/01/2021
- FILE_SIZE_IN_BYTES = 45000
- GOOGLE_CLOUD_STORAGE_URL = https://storage.googleapis.com/pheknowlator/archived_builds/release_v3.0.2/build_01NOV2021/data/processed_data/RELATIONS_LABELS.txt

DATA INFO
- DOWNLOAD_URL = https://storage.googleapis.com/pheknowlator/archived_builds/release_v3.0.2/build_01NOV2021/data/original_data/SO_GENE_TRANSCRIPT_VARIANT_TYPE_MAPPING.txt
- DOWNLOAD_DATE = 11/01/2021
- FILE_SIZE_IN_BYTES = 28069897
- GOOGLE_CLOUD_STORAGE_URL = https://storage.googleapis.com/pheknowlator/archived_builds/release_v3.0.2/build_01NOV2021/data/processed_data/SO_GENE_TRANSCRIPT_VARIANT_TYPE_MAPPING.txt

DATA INFO
- DOWNLOAD_URL = https://storage.googleapis.com/pheknowlator/archived_builds/release_v3.0.2/build_01NOV2021/data/original_data/STRING_PRO_ONTOLOGY_MAP.txt
- DOWNLOAD_DATE = 11/01/2021
- FILE_SIZE_IN_BYTES = 1326141
- GOOGLE_CLOUD_STORAGE_URL = https://storage.googleapis.com/pheknowlator/archived_builds/release_v3.0.2/build_01NOV2021/data/processed_data/STRING_PRO_ONTOLOGY_MAP.txt

DATA INFO
- DOWNLOAD_URL = https://storage.googleapis.com/pheknowlator/archived_builds/release_v3.0.2/build_01NOV2021/data/original_data/UNIPROT_ACCESSION_PRO_ONTOLOGY_MAP.txt
- DOWNLOAD_DATE = 11/01/2021
- FILE_SIZE_IN_BYTES = 3736130
- GOOGLE_CLOUD_STORAGE_URL = https://storage.googleapis.com/pheknowlator/archived_builds/release_v3.0.2/build_01NOV2021/data/processed_data/UNIPROT_ACCESSION_PRO_ONTOLOGY_MAP.txt

DATA INFO
- DOWNLOAD_URL = https://storage.googleapis.com/pheknowlator/archived_builds/release_v3.0.2/build_01NOV2021/data/original_data/UNIPROT_PROTEIN_CATALYST.txt
- DOWNLOAD_DATE = 11/01/2021
- FILE_SIZE_IN_BYTES = 1639479
- GOOGLE_CLOUD_STORAGE_URL = https://storage.googleapis.com/pheknowlator/archived_builds/release_v3.0.2/build_01NOV2021/data/processed_data/UNIPROT_PROTEIN_CATALYST.txt

DATA INFO

- DOWNLOAD_URL = https://storage.googleapis.com/pheknowlator/archived_builds/release_v3.0.2/build_01NOV2021/data/original_data/UNIPROT_PROTEIN_COFACTOR.txt
- DOWNLOAD_DATE = 11/01/2021
- FILE_SIZE_IN_BYTES = 179314
- GOOGLE_CLOUD_STORAGE_URL = https://storage.googleapis.com/pheknowlator/archived_builds/release_v3.0.2/build_01NOV2021/data/processed_data/UNIPROT_PROTEIN_COFACTOR.txt

DATA INFO
- DOWNLOAD_URL = https://storage.googleapis.com/pheknowlator/archived_builds/release_v3.0.2/build_01NOV2021/data/original_data/chebi_with_imports.owl
- DOWNLOAD_DATE = 11/01/2021
- FILE_SIZE_IN_BYTES = 643473711
- GOOGLE_CLOUD_STORAGE_URL = https://storage.googleapis.com/pheknowlator/archived_builds/release_v3.0.2/build_01NOV2021/data/processed_data/chebi_with_imports.owl

DATA INFO
- DOWNLOAD_URL = https://storage.googleapis.com/pheknowlator/archived_builds/release_v3.0.2/build_01NOV2021/data/original_data/clo_with_imports.owl
- DOWNLOAD_DATE = 11/01/2021
- FILE_SIZE_IN_BYTES = 122349569
- GOOGLE_CLOUD_STORAGE_URL = https://storage.googleapis.com/pheknowlator/archived_builds/release_v3.0.2/build_01NOV2021/data/processed_data/clo_with_imports.owl

DATA INFO
- DOWNLOAD_URL = https://storage.googleapis.com/pheknowlator/archived_builds/release_v3.0.2/build_01NOV2021/data/original_data/ensembl_identifier_data_cleaned.txt
- DOWNLOAD_DATE = 11/01/2021
- FILE_SIZE_IN_BYTES = 31849231
- GOOGLE_CLOUD_STORAGE_URL = https://storage.googleapis.com/pheknowlator/archived_builds/release_v3.0.2/build_01NOV2021/data/processed_data/ensembl_identifier_data_cleaned.txt

DATA INFO
- DOWNLOAD_URL = https://storage.googleapis.com/pheknowlator/archived_builds/release_v3.0.2/build_01NOV2021/data/original_data/ext_with_imports.owl
- DOWNLOAD_DATE = 11/01/2021
- FILE_SIZE_IN_BYTES = 64227360
- GOOGLE_CLOUD_STORAGE_URL = https://storage.googleapis.com/pheknowlator/archived_builds/release_v3.0.2/build_01NOV2021/data/processed_data/ext_with_imports.owl

DATA INFO
- DOWNLOAD_URL = https://storage.googleapis.com/pheknowlator/archived_builds/release_v3.0.2/build_01NOV2021/data/original_data/go_with_imports.owl
- DOWNLOAD_DATE = 11/01/2021
- FILE_SIZE_IN_BYTES = 122488692
- GOOGLE_CLOUD_STORAGE_URL = https://storage.googleapis.com/pheknowlator/archived_builds/release_v3.0.2/build_01NOV2021/data/processed_data/go_with_imports.owl

DATA INFO
- DOWNLOAD_URL = https://storage.googleapis.com/pheknowlator/archived_builds/release_v3.0.2/build_01NOV2021/data/original_data/hp_with_imports.owl
- DOWNLOAD_DATE = 11/01/2021
- FILE_SIZE_IN_BYTES = 84205700
- GOOGLE_CLOUD_STORAGE_URL = https://storage.googleapis.com/pheknowlator/archived_builds/release_v3.0.2/build_01NOV2021/data/processed_data/hp_with_imports.owl

DATA INFO
- DOWNLOAD_URL = https://storage.googleapis.com/pheknowlator/archived_builds/release_v3.0.2/build_01NOV2021/data/original_data/human_pro.owl
- DOWNLOAD_DATE = 11/01/2021
- FILE_SIZE_IN_BYTES = 217854250
- GOOGLE_CLOUD_STORAGE_URL = https://storage.googleapis.com/pheknowlator/archived_builds/release_v3.0.2/build_01NOV2021/data/processed_data/human_pro.owl

DATA INFO
- DOWNLOAD_URL = https://storage.googleapis.com/pheknowlator/archived_builds/release_v3.0.2/build_01NOV2021/data/original_data/mondo_with_imports.owl
- DOWNLOAD_DATE = 11/01/2021
- FILE_SIZE_IN_BYTES = 231020989
- GOOGLE_CLOUD_STORAGE_URL = https://storage.googleapis.com/pheknowlator/archived_builds/release_v3.0.2/build_01NOV2021/data/processed_data/mondo_with_imports.owl

DATA INFO
- DOWNLOAD_URL = https://storage.googleapis.com/pheknowlator/archived_builds/release_v3.0.2/build_01NOV2021/data/original_data/node_metadata_dict.pkl
- DOWNLOAD_DATE = 11/01/2021
- FILE_SIZE_IN_BYTES = 325174359
- GOOGLE_CLOUD_STORAGE_URL = https://storage.googleapis.com/pheknowlator/archived_builds/release_v3.0.2/build_01NOV2021/data/processed_data/node_metadata_dict.pkl

DATA INFO
- DOWNLOAD_URL = https://storage.googleapis.com/pheknowlator/archived_builds/release_v3.0.2/build_01NOV2021/data/original_data/ontology_cleaning_report.txt
- DOWNLOAD_DATE = 11/01/2021
- FILE_SIZE_IN_BYTES = 1436530
- GOOGLE_CLOUD_STORAGE_URL = https://storage.googleapis.com/pheknowlator/archived_builds/release_v3.0.2/build_01NOV2021/data/processed_data/ontology_cleaning_report.txt

DATA INFO
- DOWNLOAD_URL = https://storage.googleapis.com/pheknowlator/archived_builds/release_v3.0.2/build_01NOV2021/data/original_data/pr_with_imports.owl

- DOWNLOAD_DATE = 11/01/2021
- FILE_SIZE_IN_BYTES = 217870791
- GOOGLE_CLOUD_STORAGE_URL = https://storage.googleapis.com/pheknowlator/archived_builds/release_v3.0.2/build_01NOV2021/data/processed_data/pr_with_imports.owl

DATA INFO
- DOWNLOAD_URL = https://storage.googleapis.com/pheknowlator/archived_builds/release_v3.0.2/build_01NOV2021/data/original_data/pw_with_imports.owl
- DOWNLOAD_DATE = 11/01/2021
- FILE_SIZE_IN_BYTES = 4915785
- GOOGLE_CLOUD_STORAGE_URL = https://storage.googleapis.com/pheknowlator/archived_builds/release_v3.0.2/build_01NOV2021/data/processed_data/pw_with_imports.owl

DATA INFO
- DOWNLOAD_URL = https://storage.googleapis.com/pheknowlator/archived_builds/release_v3.0.2/build_01NOV2021/data/original_data/ro_with_imports.owl
- DOWNLOAD_DATE = 11/01/2021
- FILE_SIZE_IN_BYTES = 857709
- GOOGLE_CLOUD_STORAGE_URL = https://storage.googleapis.com/pheknowlator/archived_builds/release_v3.0.2/build_01NOV2021/data/processed_data/ro_with_imports.owl

DATA INFO
- DOWNLOAD_URL = https://storage.googleapis.com/pheknowlator/archived_builds/release_v3.0.2/build_01NOV2021/data/original_data/so_with_imports.owl
- DOWNLOAD_DATE = 11/01/2021
- FILE_SIZE_IN_BYTES = 4890370
- GOOGLE_CLOUD_STORAGE_URL = https://storage.googleapis.com/pheknowlator/archived_builds/release_v3.0.2/build_01NOV2021/data/processed_data/so_with_imports.owl

DATA INFO
- DOWNLOAD_URL = https://storage.googleapis.com/pheknowlator/archived_builds/release_v3.0.2/build_01NOV2021/data/original_data/subclass_construction_map.pkl
- DOWNLOAD_DATE = 11/01/2021
- FILE_SIZE_IN_BYTES = 21922553
- GOOGLE_CLOUD_STORAGE_URL = https://storage.googleapis.com/pheknowlator/archived_builds/release_v3.0.2/build_01NOV2021/data/processed_data/subclass_construction_map.pkl

DATA INFO
- DOWNLOAD_URL = https://storage.googleapis.com/pheknowlator/archived_builds/release_v3.0.2/build_01NOV2021/data/original_data/vo_with_imports.owl
- DOWNLOAD_DATE = 11/01/2021
- FILE_SIZE_IN_BYTES = 8388461
- GOOGLE_CLOUD_STORAGE_URL = https://storage.googleapis.com/pheknowlator/archived_builds/release_v3.0.2/build_01NOV2021/data/processed_data/vo_with_imports.owl

**Supplementary Document 3. edge_source_metadata.txt.**

```
===================================
#Tue Nov 02 01:06:42 UTC 2021
===================================
EDGE: chemical-disease
DATA PROCESSING INFO
  - IDENTIFIER MAPPING = chemical (./resources/processed_data/MESH_CHEBI_MAP.txt) | disease (./resources/processed_data/DISEASE_MONDO_MAP.txt)
  - FILTERING CRITERIA = None
  - EVIDENCE CRITERIA = data[5]!=''
DATA INFO
  - DOWNLOAD_URL = https://storage.googleapis.com/pheknowlator/archived_builds/release_v3.0.2/build_01NOV2021/data/original_data/CTD_chemicals_diseases.tsv
  - DOWNLOAD_DATE = 11/02/2021
  - FILE_SIZE_IN_BYTES = 701201197
  - DOWNLOADED_FILE_LOCATION = resources/edge_data/chemical-disease_CTD_chemicals_diseases.tsv

EDGE: chemical-gene
DATA PROCESSING INFO
  - IDENTIFIER MAPPING = chemical (./resources/processed_data/MESH_CHEBI_MAP.txt)
  - FILTERING CRITERIA = data[6]==Homo sapiens | data[5].startswith('gene')
  - EVIDENCE CRITERIA = data[9]affectsnot in x
DATA INFO
  - DOWNLOAD_URL = https://storage.googleapis.com/pheknowlator/archived_builds/release_v3.0.2/build_01NOV2021/data/original_data/CTD_chem_gene_ixns.tsv
  - DOWNLOAD_DATE = 11/02/2021
  - FILE_SIZE_IN_BYTES = 441097103
  - DOWNLOADED_FILE_LOCATION = resources/edge_data/chemical-gene_CTD_chem_gene_ixns.tsv

EDGE: chemical-gobp
DATA PROCESSING INFO
  - IDENTIFIER MAPPING = chemical (./resources/processed_data/MESH_CHEBI_MAP.txt)
  - FILTERING CRITERIA = data[3]==Biological Process
  - EVIDENCE CRITERIA = data[8]<=1.04e-47
DATA INFO
  - DOWNLOAD_URL = https://storage.googleapis.com/pheknowlator/archived_builds/release_v3.0.2/build_01NOV2021/data/original_data/CTD_chem_go_enriched.tsv
  - DOWNLOAD_DATE = 11/02/2021
  - FILE_SIZE_IN_BYTES = 817733569
  - DOWNLOADED_FILE_LOCATION = resources/edge_data/chemical-gobp_CTD_chem_go_enriched.tsv

EDGE: chemical-gocc
DATA PROCESSING INFO
  - IDENTIFIER MAPPING = chemical (./resources/processed_data/MESH_CHEBI_MAP.txt)
  - FILTERING CRITERIA = data[3]==Cellular Component
  - EVIDENCE CRITERIA = data[8]<=1.04e-47
DATA INFO
  - DOWNLOAD_URL = https://storage.googleapis.com/pheknowlator/archived_builds/release_v3.0.2/build_01NOV2021/data/original_data/CTD_chem_go_enriched.tsv
  - DOWNLOAD_DATE = 11/02/2021
  - FILE_SIZE_IN_BYTES = 817733569
  - DOWNLOADED_FILE_LOCATION = resources/edge_data/chemical-gocc_CTD_chem_go_enriched.tsv

EDGE: chemical-gomf
DATA PROCESSING INFO
  - IDENTIFIER MAPPING = chemical (./resources/processed_data/MESH_CHEBI_MAP.txt)
  - FILTERING CRITERIA = data[3]==Molecular Function
```

- EVIDENCE CRITERIA = data[8]<=1.04e-47

DATA INFO
- DOWNLOAD_URL = https://storage.googleapis.com/pheknowlator/archived_builds/release_v3.0.2/build_01NOV2021/data/original_data/CTD_chem_go_enriched.tsv
- DOWNLOAD_DATE = 11/02/2021
- FILE_SIZE_IN_BYTES = 817733569
- DOWNLOADED_FILE_LOCATION = resources/edge_data/chemical-gomf_CTD_chem_go_enriched.tsv

EDGE: chemical-pathway

DATA PROCESSING INFO
- IDENTIFIER MAPPING = None
- FILTERING CRITERIA = data[5]==Homo sapiens
- EVIDENCE CRITERIA = None

DATA INFO
- DOWNLOAD_URL = https://storage.googleapis.com/pheknowlator/archived_builds/release_v3.0.2/build_01NOV2021/data/original_data/ChEBI2Reactome_All_Levels.txt
- DOWNLOAD_DATE = 11/02/2021
- FILE_SIZE_IN_BYTES = 30485396
- DOWNLOADED_FILE_LOCATION = resources/edge_data/chemical-pathway_ChEBI2Reactome_All_Levels.txt

EDGE: chemical-phenotype

DATA PROCESSING INFO
- IDENTIFIER MAPPING = chemical (./resources/processed_data/MESH_CHEBI_MAP.txt) | phenotype (./resources/processed_data/PHENOTYPE_HPO_MAP.txt)
- FILTERING CRITERIA = None
- EVIDENCE CRITERIA = data[5]!=''

DATA INFO
- DOWNLOAD_URL = https://storage.googleapis.com/pheknowlator/archived_builds/release_v3.0.2/build_01NOV2021/data/original_data/CTD_chemicals_diseases.tsv
- DOWNLOAD_DATE = 11/02/2021
- FILE_SIZE_IN_BYTES = 701201197
- DOWNLOADED_FILE_LOCATION = resources/edge_data/chemical-phenotype_CTD_chemicals_diseases.tsv

EDGE: chemical-protein

DATA PROCESSING INFO
- IDENTIFIER MAPPING = chemical (./resources/processed_data/MESH_CHEBI_MAP.txt) | protein (./resources/processed_data/ENTREZ_GENE_PRO_ONTOLOGY_MAP.txt)
- FILTERING CRITERIA = data[6]==Homo sapiens | data[5] .startswith('protein')
- EVIDENCE CRITERIA = data[9]affectsnot in x

DATA INFO
- DOWNLOAD_URL = https://storage.googleapis.com/pheknowlator/archived_builds/release_v3.0.2/build_01NOV2021/data/original_data/CTD_chem_gene_ixns.tsv
- DOWNLOAD_DATE = 11/02/2021
- FILE_SIZE_IN_BYTES = 441097103
- DOWNLOADED_FILE_LOCATION = resources/edge_data/chemical-protein_CTD_chem_gene_ixns.tsv

EDGE: disease-phenotype

DATA PROCESSING INFO
- IDENTIFIER MAPPING = None
- FILTERING CRITERIA = None
- EVIDENCE CRITERIA = None

DATA INFO
- DOWNLOAD_URL = https://storage.googleapis.com/pheknowlator/archived_builds/release_v3.0.2/build_01NOV2021/data/original_data/phenotype.hpoa
- DOWNLOAD_DATE = 11/02/2021
- FILE_SIZE_IN_BYTES = 27315334
- DOWNLOADED_FILE_LOCATION = resources/edge_data/disease-phenotype_phenotype.hpoa

EDGE: gene-disease

DATA PROCESSING INFO

- IDENTIFIER MAPPING = disease (./resources/processed_data/DISEASE_MONDO_MAP.txt)
- FILTERING CRITERIA = None
- EVIDENCE CRITERIA = data[10]>=1.0

DATA INFO
- DOWNLOAD_URL = https://storage.googleapis.com/pheknowlator/archived_builds/release_v3.0.2/build_01NOV2021/data/original_data/curated_gene_disease_associations.tsv
- DOWNLOAD_DATE = 11/02/2021
- FILE_SIZE_IN_BYTES = 11542996
- DOWNLOADED_FILE_LOCATION = resources/edge_data/gene-disease_curated_gene_disease_associations.tsv

EDGE: gene-gene

DATA PROCESSING INFO
- IDENTIFIER MAPPING = gene (./resources/processed_data/ENSEMBL_GENE_ENTREZ_GENE_MAP.txt) | gene (./resources/processed_data/ENSEMBL_GENE_ENTREZ_GENE_MAP.txt)
- FILTERING CRITERIA = None
- EVIDENCE CRITERIA = None

DATA INFO
- DOWNLOAD_URL = https://storage.googleapis.com/pheknowlator/archived_builds/release_v3.0.2/build_01NOV2021/data/original_data/COMBINED.DEFAULT_NETWORKS.BP_COMBINING.txt
- DOWNLOAD_DATE = 11/02/2021
- FILE_SIZE_IN_BYTES = 246190113
- DOWNLOADED_FILE_LOCATION = resources/edge_data/gene-gene_COMBINED.DEFAULT_NETWORKS.BP_COMBINING.txt

EDGE: gene-pathway

DATA PROCESSING INFO
- IDENTIFIER MAPPING = None
- FILTERING CRITERIA = data[3].startswith('REACT:R-HSA-')
- EVIDENCE CRITERIA = None

DATA INFO
- DOWNLOAD_URL = https://storage.googleapis.com/pheknowlator/archived_builds/release_v3.0.2/build_01NOV2021/data/original_data/CTD_genes_pathways.tsv
- DOWNLOAD_DATE = 11/02/2021
- FILE_SIZE_IN_BYTES = 8192661
- DOWNLOADED_FILE_LOCATION = resources/edge_data/gene-pathway_CTD_genes_pathways.tsv

EDGE: gene-phenotype

DATA PROCESSING INFO
- IDENTIFIER MAPPING = phenotype (./resources/processed_data/PHENOTYPE_HPO_MAP.txt)
- FILTERING CRITERIA = None
- EVIDENCE CRITERIA = data[10]>=1.0

DATA INFO
- DOWNLOAD_URL = https://storage.googleapis.com/pheknowlator/archived_builds/release_v3.0.2/build_01NOV2021/data/original_data/curated_gene_disease_associations.tsv
- DOWNLOAD_DATE = 11/02/2021
- FILE_SIZE_IN_BYTES = 11542996
- DOWNLOADED_FILE_LOCATION = resources/edge_data/gene-phenotype_curated_gene_disease_associations.tsv

EDGE: gene-protein

DATA PROCESSING INFO
- IDENTIFIER MAPPING = None
- FILTERING CRITERIA = data[4]==protein-coding
- EVIDENCE CRITERIA = None

DATA INFO
- DOWNLOAD_URL = https://storage.googleapis.com/pheknowlator/archived_builds/release_v3.0.2/build_01NOV2021/data/processed_data/ENTREZ_GENE_PRO_ONTOLOGY_MAP.txt
- DOWNLOAD_DATE = 11/02/2021
- FILE_SIZE_IN_BYTES = 1124944
- DOWNLOADED_FILE_LOCATION = resources/edge_data/gene-protein_ENTREZ_GENE_PRO_ONTOLOGY_MAP.txt

EDGE: gene-rna
DATA PROCESSING INFO
- IDENTIFIER MAPPING = None
- FILTERING CRITERIA = None
- EVIDENCE CRITERIA = None
DATA INFO
- DOWNLOAD_URL = https://storage.googleapis.com/pheknowlator/archived_builds/release_v3.0.2/build_01NOV2021/data/processed_data/ENTREZ_GENE_ENSEMBL_TRANSCRIPT_MAP.txt
- DOWNLOAD_DATE = 11/02/2021
- FILE_SIZE_IN_BYTES = 15146432
- DOWNLOADED_FILE_LOCATION = resources/edge_data/gene-rna_ENTREZ_GENE_ENSEMBL_TRANSCRIPT_MAP.txt

EDGE: gobp-pathway
DATA PROCESSING INFO
- IDENTIFIER MAPPING = None
- FILTERING CRITERIA = data[8]==P | data[12]==taxon:9606 | data[5].startswith('REACTOME')
- EVIDENCE CRITERIA = None
DATA INFO
- DOWNLOAD_URL = https://storage.googleapis.com/pheknowlator/archived_builds/release_v3.0.2/build_01NOV2021/data/original_data/gene_association.reactome
- DOWNLOAD_DATE = 11/02/2021
- FILE_SIZE_IN_BYTES = 11955743
- DOWNLOADED_FILE_LOCATION = resources/edge_data/gobp-pathway_gene_association.reactome

EDGE: pathway-gocc
DATA PROCESSING INFO
- IDENTIFIER MAPPING = None
- FILTERING CRITERIA = data[8]==C | data[12]==taxon:9606 | data[5].startswith('REACTOME')
- EVIDENCE CRITERIA = None
DATA INFO
- DOWNLOAD_URL = https://storage.googleapis.com/pheknowlator/archived_builds/release_v3.0.2/build_01NOV2021/data/original_data/gene_association.reactome
- DOWNLOAD_DATE = 11/02/2021
- FILE_SIZE_IN_BYTES = 11955743
- DOWNLOADED_FILE_LOCATION = resources/edge_data/pathway-gocc_gene_association.reactome

EDGE: pathway-gomf
DATA PROCESSING INFO
- IDENTIFIER MAPPING = None
- FILTERING CRITERIA = data[8]==F | data[12]==taxon:9606 | data[5].startswith('REACTOME')
- EVIDENCE CRITERIA = None
DATA INFO
- DOWNLOAD_URL = https://storage.googleapis.com/pheknowlator/archived_builds/release_v3.0.2/build_01NOV2021/data/original_data/gene_association.reactome
- DOWNLOAD_DATE = 11/02/2021
- FILE_SIZE_IN_BYTES = 11955743
- DOWNLOADED_FILE_LOCATION = resources/edge_data/pathway-gomf_gene_association.reactome

EDGE: protein-anatomy
DATA PROCESSING INFO
- IDENTIFIER MAPPING = protein (./resources/processed_data/UNIPROT_ACCESSION_PRO_ONTOLOGY_MAP.txt) | anatomy (./resources/processed_data/HPA_GTEx_TISSUE_CELL_MAP.txt)
- FILTERING CRITERIA = data[3]==Evidence at protein level | data[4]==anatomy
- EVIDENCE CRITERIA = None
DATA INFO
- DOWNLOAD_URL = https://storage.googleapis.com/pheknowlator/archived_builds/release_v3.0.2/build_01NOV2021/data/processed_data/HPA_GTEX_RNA_GENE_PROTEIN_EDGES.txt
- DOWNLOAD_DATE = 11/02/2021
- FILE_SIZE_IN_BYTES = 20611429
- DOWNLOADED_FILE_LOCATION = resources/edge_data/protein-anatomy_HPA_GTEX_RNA_GENE_PROTEIN_EDGES.txt

EDGE: protein-catalyst
DATA PROCESSING INFO
 - IDENTIFIER MAPPING = None
 - FILTERING CRITERIA = None
 - EVIDENCE CRITERIA = None
DATA INFO
 - DOWNLOAD_URL = https://storage.googleapis.com/pheknowlator/archived_builds/release_v3.0.2/build_01NOV2021/data/processed_data/UNIPROT_PROTEIN_CATALYST.txt
 - DOWNLOAD_DATE = 11/02/2021
 - FILE_SIZE_IN_BYTES = 1639479
 - DOWNLOADED_FILE_LOCATION = resources/edge_data/protein-catalyst_UNIPROT_PROTEIN_CATALYST.txt

EDGE: protein-cell
DATA PROCESSING INFO
 - IDENTIFIER MAPPING = protein (./resources/processed_data/UNIPROT_ACCESSION_PRO_ONTOLOGY_MAP.txt) | cell (./resources/processed_data/HPA_GTEx_TISSUE_CELL_MAP.txt)
 - FILTERING CRITERIA = data[3]==Evidence at protein level | data[4]==cell line
 - EVIDENCE CRITERIA = None
DATA INFO
 - DOWNLOAD_URL = https://storage.googleapis.com/pheknowlator/archived_builds/release_v3.0.2/build_01NOV2021/data/processed_data/HPA_GTEX_RNA_GENE_PROTEIN_EDGES.txt
 - DOWNLOAD_DATE = 11/02/2021
 - FILE_SIZE_IN_BYTES = 20611429
 - DOWNLOADED_FILE_LOCATION = resources/edge_data/protein-cell_HPA_GTEX_RNA_GENE_PROTEIN_EDGES.txt

EDGE: protein-cofactor
DATA PROCESSING INFO
 - IDENTIFIER MAPPING = None
 - FILTERING CRITERIA = None
 - EVIDENCE CRITERIA = None
DATA INFO
 - DOWNLOAD_URL = https://storage.googleapis.com/pheknowlator/archived_builds/release_v3.0.2/build_01NOV2021/data/processed_data/UNIPROT_PROTEIN_COFACTOR.txt
 - DOWNLOAD_DATE = 11/02/2021
 - FILE_SIZE_IN_BYTES = 179314
 - DOWNLOADED_FILE_LOCATION = resources/edge_data/protein-cofactor_UNIPROT_PROTEIN_COFACTOR.txt

EDGE: protein-gobp
DATA PROCESSING INFO
 - IDENTIFIER MAPPING = protein (./resources/processed_data/UNIPROT_ACCESSION_PRO_ONTOLOGY_MAP.txt)
 - FILTERING CRITERIA = data[8]==P | data[12]==taxon:9606
 - EVIDENCE CRITERIA = None
DATA INFO
 - DOWNLOAD_URL = https://storage.googleapis.com/pheknowlator/archived_builds/release_v3.0.2/build_01NOV2021/data/original_data/goa_human.gaf
 - DOWNLOAD_DATE = 11/02/2021
 - FILE_SIZE_IN_BYTES = 109325311
 - DOWNLOADED_FILE_LOCATION = resources/edge_data/protein-gobp_goa_human.gaf

EDGE: protein-gocc
DATA PROCESSING INFO
 - IDENTIFIER MAPPING = protein (./resources/processed_data/UNIPROT_ACCESSION_PRO_ONTOLOGY_MAP.txt)
 - FILTERING CRITERIA = data[8]==C | data[12]==taxon:9606
 - EVIDENCE CRITERIA = None
DATA INFO
 - DOWNLOAD_URL = https://storage.googleapis.com/pheknowlator/archived_builds/release_v3.0.2/build_01NOV2021/data/original_data/goa_human.gaf
 - DOWNLOAD_DATE = 11/02/2021

- FILE_SIZE_IN_BYTES = 109325311
- DOWNLOADED_FILE_LOCATION = resources/edge_data/protein-gocc_goa_human.gaf

EDGE: protein-gomf
DATA PROCESSING INFO
- IDENTIFIER MAPPING = protein (./resources/processed_data/UNIPROT_ACCESSION_PRO_ONTOLOGY_MAP.txt)
- FILTERING CRITERIA = data[8]==F | data[12]==taxon:9606
- EVIDENCE CRITERIA = None

DATA INFO
- DOWNLOAD_URL = https://storage.googleapis.com/pheknowlator/archived_builds/release_v3.0.2/build_01NOV2021/data/original_data/goa_human.gaf
- DOWNLOAD_DATE = 11/02/2021
- FILE_SIZE_IN_BYTES = 109325311
- DOWNLOADED_FILE_LOCATION = resources/edge_data/protein-gomf_goa_human.gaf

EDGE: protein-pathway
DATA PROCESSING INFO
- IDENTIFIER MAPPING = protein (./resources/processed_data/UNIPROT_ACCESSION_PRO_ONTOLOGY_MAP.txt)
- FILTERING CRITERIA = data[5]==Homo sapiens
- EVIDENCE CRITERIA = None

DATA INFO
- DOWNLOAD_URL = https://storage.googleapis.com/pheknowlator/archived_builds/release_v3.0.2/build_01NOV2021/data/original_data/UniProt2Reactome_All_Levels.txt
- DOWNLOAD_DATE = 11/02/2021
- FILE_SIZE_IN_BYTES = 102191227
- DOWNLOADED_FILE_LOCATION = resources/edge_data/protein-pathway_UniProt2Reactome_All_Levels.txt

EDGE: protein-protein
DATA PROCESSING INFO
- IDENTIFIER MAPPING = protein (./resources/processed_data/STRING_PRO_ONTOLOGY_MAP.txt) | protein (./resources/processed_data/STRING_PRO_ONTOLOGY_MAP.txt)
- FILTERING CRITERIA = None
- EVIDENCE CRITERIA = data[2]>=700

DATA INFO
- DOWNLOAD_URL = https://storage.googleapis.com/pheknowlator/archived_builds/release_v3.0.2/build_01NOV2021/data/original_data/9606.protein.links.v11.0.txt
- DOWNLOAD_DATE = 11/02/2021
- FILE_SIZE_IN_BYTES = 540934917
- DOWNLOADED_FILE_LOCATION = resources/edge_data/protein-protein_9606.protein.links.v11.0.txt

EDGE: rna-anatomy
DATA PROCESSING INFO
- IDENTIFIER MAPPING = rna (./resources/processed_data/GENE_SYMBOL_ENSEMBL_TRANSCRIPT_MAP.txt) | anatomy (./resources/processed_data/HPA_GTEx_TISSUE_CELL_MAP.txt)
- FILTERING CRITERIA = data[3]==Evidence at transcript level | data[4]==anatomy
- EVIDENCE CRITERIA = None

DATA INFO
- DOWNLOAD_URL = https://storage.googleapis.com/pheknowlator/archived_builds/release_v3.0.2/build_01NOV2021/data/processed_data/HPA_GTEX_RNA_GENE_PROTEIN_EDGES.txt
- DOWNLOAD_DATE = 11/02/2021
- FILE_SIZE_IN_BYTES = 20611429
- DOWNLOADED_FILE_LOCATION = resources/edge_data/rna-anatomy_HPA_GTEX_RNA_GENE_PROTEIN_EDGES.txt

EDGE: rna-cell
DATA PROCESSING INFO
- IDENTIFIER MAPPING = rna (./resources/processed_data/GENE_SYMBOL_ENSEMBL_TRANSCRIPT_MAP.txt) | cell (./resources/processed_data/HPA_GTEx_TISSUE_CELL_MAP.txt)
- FILTERING CRITERIA = data[3]==Evidence at transcript level | data[4]==cell line
- EVIDENCE CRITERIA = None

DATA INFO

- DOWNLOAD_URL = https://storage.googleapis.com/pheknowlator/archived_builds/release_v3.0.2/build_01NOV2021/data/processed_data/HPA_GTEX_RNA_GENE_PROTEIN_EDGES.txt
- DOWNLOAD_DATE = 11/02/2021
- FILE_SIZE_IN_BYTES = 20611429
- DOWNLOADED_FILE_LOCATION = resources/edge_data/rna-cell_HPA_GTEX_RNA_GENE_PROTEIN_EDGES.txt

EDGE: rna-protein
DATA PROCESSING INFO
- IDENTIFIER MAPPING = None
- FILTERING CRITERIA = data[4]==protein-coding
- EVIDENCE CRITERIA = None

DATA INFO
- DOWNLOAD_URL = https://storage.googleapis.com/pheknowlator/archived_builds/release_v3.0.2/build_01NOV2021/data/processed_data/ENSEMBL_TRANSCRIPT_PROTEIN_ONTOLOGY_MAP.txt
- DOWNLOAD_DATE = 11/02/2021
- FILE_SIZE_IN_BYTES = 2931638
- DOWNLOADED_FILE_LOCATION = resources/edge_data/rna-protein_ENSEMBL_TRANSCRIPT_PROTEIN_ONTOLOGY_MAP.txt

EDGE: variant-disease
DATA PROCESSING INFO
- IDENTIFIER MAPPING = disease (./resources/processed_data/DISEASE_MONDO_MAP.txt)
- FILTERING CRITERIA = data[9]!=-1 | data[16]==GRCh38 | data[8-9]dedupdesc
- EVIDENCE CRITERIA = data[24] in ["criteria provided, multiple submitters, no conflicts", "reviewed by expert panel", "practice guideline"] | data[7]==1

DATA INFO
- DOWNLOAD_URL = https://storage.googleapis.com/pheknowlator/archived_builds/release_v3.0.2/build_01NOV2021/data/processed_data/CLINVAR_VARIANT_GENE_DISEASE_PHENOTYPE_EDGES.txt
- DOWNLOAD_DATE = 11/02/2021
- FILE_SIZE_IN_BYTES = 3662353455
- DOWNLOADED_FILE_LOCATION = resources/edge_data/variant-disease_CLINVAR_VARIANT_GENE_DISEASE_PHENOTYPE_EDGES.txt

EDGE: variant-gene
DATA PROCESSING INFO
- IDENTIFIER MAPPING = None
- FILTERING CRITERIA = data[9]!=-1 | data[3]!=-1 | data[16]==GRCh38 | data[8-9]dedupdesc
- EVIDENCE CRITERIA = data[24] in ["criteria provided, multiple submitters, no conflicts", "reviewed by expert panel", "practice guideline"]

DATA INFO
- DOWNLOAD_URL = https://storage.googleapis.com/pheknowlator/archived_builds/release_v3.0.2/build_01NOV2021/data/processed_data/CLINVAR_VARIANT_GENE_DISEASE_PHENOTYPE_EDGES.txt
- DOWNLOAD_DATE = 11/02/2021
- FILE_SIZE_IN_BYTES = 3662353455
- DOWNLOADED_FILE_LOCATION = resources/edge_data/variant-gene_CLINVAR_VARIANT_GENE_DISEASE_PHENOTYPE_EDGES.txt

EDGE: variant-phenotype
DATA PROCESSING INFO
- IDENTIFIER MAPPING = phenotype (./resources/processed_data/PHENOTYPE_HPO_MAP.txt)
- FILTERING CRITERIA = data[9]!=-1 | data[16]==GRCh38 | data[8-9]dedupdesc
- EVIDENCE CRITERIA = data[24] in ["criteria provided, multiple submitters, no conflicts", "reviewed by expert panel", "practice guideline"] | data[7]==1

DATA INFO
- DOWNLOAD_URL = https://storage.googleapis.com/pheknowlator/archived_builds/release_v3.0.2/build_01NOV2021/data/processed_data/CLINVAR_VARIANT_GENE_DISEASE_PHENOTYPE_EDGES.txt
- DOWNLOAD_DATE = 11/02/2021
- FILE_SIZE_IN_BYTES = 3662353455
- DOWNLOADED_FILE_LOCATION = resources/edge_data/variant-phenotype_CLINVAR_VARIANT_GENE_DISEASE_PHENOTYPE_EDGES.txt

**Supplementary Document 4. ontology_source_metadata.txt.**

```
===================================
#Tue Nov 02 01:05:41 UTC 2021
===================================

EDGE: phenotype
DATA PROCESSING INFO
  - IDENTIFIER MAPPING = None
  - FILTERING CRITERIA = None
  - EVIDENCE CRITERIA = None
DATA INFO
  - DOWNLOAD_URL = https://storage.googleapis.com/pheknowlator/archived_builds/release_v3.0.2/build_01NOV2021/data/processed_data/hp_with_imports.owl
  - DOWNLOAD_DATE = 11/02/2021
  - FILE_SIZE_IN_BYTES = 84205700
  - DOWNLOADED_FILE_LOCATION = resources/ontologies/hp_with_imports_with_imports.owl

EDGE: go
DATA PROCESSING INFO
  - IDENTIFIER MAPPING = None
  - FILTERING CRITERIA = None
  - EVIDENCE CRITERIA = None
DATA INFO
  - DOWNLOAD_URL = https://storage.googleapis.com/pheknowlator/archived_builds/release_v3.0.2/build_01NOV2021/data/processed_data/go_with_imports.owl
  - DOWNLOAD_DATE = 11/02/2021
  - FILE_SIZE_IN_BYTES = 122488692
  - DOWNLOADED_FILE_LOCATION = resources/ontologies/go_with_imports_with_imports.owl

EDGE: disease
DATA PROCESSING INFO
  - IDENTIFIER MAPPING = None
  - FILTERING CRITERIA = None
  - EVIDENCE CRITERIA = None
DATA INFO
  - DOWNLOAD_URL = https://storage.googleapis.com/pheknowlator/archived_builds/release_v3.0.2/build_01NOV2021/data/processed_data/mondo_with_imports.owl
  - DOWNLOAD_DATE = 11/02/2021
  - FILE_SIZE_IN_BYTES = 231020989
  - DOWNLOADED_FILE_LOCATION = resources/ontologies/mondo_with_imports_with_imports.owl

EDGE: vaccine
DATA PROCESSING INFO
  - IDENTIFIER MAPPING = None
  - FILTERING CRITERIA = None
  - EVIDENCE CRITERIA = None
DATA INFO
  - DOWNLOAD_URL = https://storage.googleapis.com/pheknowlator/archived_builds/release_v3.0.2/build_01NOV2021/data/processed_data/vo_with_imports.owl
  - DOWNLOAD_DATE = 11/02/2021
  - FILE_SIZE_IN_BYTES = 8388461
  - DOWNLOADED_FILE_LOCATION = resources/ontologies/vo_with_imports_with_imports.owl

EDGE: chemical
DATA PROCESSING INFO
  - IDENTIFIER MAPPING = None
  - FILTERING CRITERIA = None
```

- EVIDENCE CRITERIA = None
DATA INFO
- DOWNLOAD_URL = https://storage.googleapis.com/pheknowlator/archived_builds/release_v3.0.2/build_01NOV2021/data/processed_data/chebi_with_imports.owl
- DOWNLOAD_DATE = 11/02/2021
- FILE_SIZE_IN_BYTES = 643473711
- DOWNLOADED_FILE_LOCATION = resources/ontologies/chebi_with_imports_with_imports.owl

EDGE: anatomy
DATA PROCESSING INFO
- IDENTIFIER MAPPING = None
- FILTERING CRITERIA = None
- EVIDENCE CRITERIA = None
DATA INFO
- DOWNLOAD_URL = https://storage.googleapis.com/pheknowlator/archived_builds/release_v3.0.2/build_01NOV2021/data/processed_data/ext_with_imports.owl
- DOWNLOAD_DATE = 11/02/2021
- FILE_SIZE_IN_BYTES = 64227360
- DOWNLOADED_FILE_LOCATION = resources/ontologies/ext_with_imports_with_imports.owl

EDGE: cell
DATA PROCESSING INFO
- IDENTIFIER MAPPING = None
- FILTERING CRITERIA = None
- EVIDENCE CRITERIA = None
DATA INFO
- DOWNLOAD_URL = https://storage.googleapis.com/pheknowlator/archived_builds/release_v3.0.2/build_01NOV2021/data/processed_data/clo_with_imports.owl
- DOWNLOAD_DATE = 11/02/2021
- FILE_SIZE_IN_BYTES = 122349569
- DOWNLOADED_FILE_LOCATION = resources/ontologies/clo_with_imports_with_imports.owl

EDGE: protein
DATA PROCESSING INFO
- IDENTIFIER MAPPING = None
- FILTERING CRITERIA = None
- EVIDENCE CRITERIA = None
DATA INFO
- DOWNLOAD_URL = https://storage.googleapis.com/pheknowlator/archived_builds/release_v3.0.2/build_01NOV2021/data/processed_data/pr_with_imports.owl
- DOWNLOAD_DATE = 11/02/2021
- FILE_SIZE_IN_BYTES = 217870791
- DOWNLOADED_FILE_LOCATION = resources/ontologies/pr_with_imports_with_imports.owl

EDGE: genomic
DATA PROCESSING INFO
- IDENTIFIER MAPPING = None
- FILTERING CRITERIA = None
- EVIDENCE CRITERIA = None
DATA INFO
- DOWNLOAD_URL = https://storage.googleapis.com/pheknowlator/archived_builds/release_v3.0.2/build_01NOV2021/data/processed_data/so_with_imports.owl
- DOWNLOAD_DATE = 11/02/2021
- FILE_SIZE_IN_BYTES = 4890370
- DOWNLOADED_FILE_LOCATION = resources/ontologies/so_with_imports_with_imports.owl

EDGE: pathway
DATA PROCESSING INFO

- IDENTIFIER MAPPING = None
- FILTERING CRITERIA = None
- EVIDENCE CRITERIA = None

DATA INFO
- DOWNLOAD_URL = https://storage.googleapis.com/pheknowlator/archived_builds/release_v3.0.2/build_01NOV2021/data/processed_data/pw_with_imports.owl
- DOWNLOAD_DATE = 11/02/2021
- FILE_SIZE_IN_BYTES = 4915785
- DOWNLOADED_FILE_LOCATION = resources/ontologies/pw_with_imports_with_imports.owl

EDGE: relation

DATA PROCESSING INFO
- IDENTIFIER MAPPING = None
- FILTERING CRITERIA = None
- EVIDENCE CRITERIA = None

DATA INFO
- DOWNLOAD_URL = https://storage.googleapis.com/pheknowlator/archived_builds/release_v3.0.2/build_01NOV2021/data/processed_data/ro_with_imports.owl
- DOWNLOAD_DATE = 11/02/2021
- FILE_SIZE_IN_BYTES = 857709
- DOWNLOADED_FILE_LOCATION = resources/ontologies/ro_with_imports_with_imports.owl

**Supplementary Document 5. ontology_cleaning_report.txt.**

==================================================
ONTOLOGY CLEANING REPORT
Mon Nov 01 09:57:17 UTC 2021
==================================================

ONTOLOGY: chebi_with_imports.owl
********************************

- Original GCS URL: https://storage.googleapis.com/pheknowlator/archived_builds/release_v3.0.2/build_01NOV2021/data/original_data/chebi_with_imports.owl
- Processed GCS URL: https://storage.googleapis.com/pheknowlator/archived_builds/release_v3.0.2/build_01NOV2021/data/processed_data/chebi_with_imports.owl
- Statistics Before Cleaning: 5644239 Triples; 168627 Classes; 0 Individuals; 10 Object Properties; 37 Annotation Properties; 1 Connected Components
- Statistics After Cleaning: 5569987 Triples; 150080 Classes; 0 Individuals; 10 Object Properties; 37 Annotation Properties; 1 Connected Components
- Value Errors: 0
- Identifier Errors: 0
- Deprecated Classes: 18547
- Obsolete Classes: 0
- Punning Errors: 0
    - Object Properties: 0

ONTOLOGY: clo_with_imports.owl
******************************

- Original GCS URL: https://storage.googleapis.com/pheknowlator/archived_builds/release_v3.0.2/build_01NOV2021/data/original_data/clo_with_imports.owl
- Processed GCS URL: https://storage.googleapis.com/pheknowlator/archived_builds/release_v3.0.2/build_01NOV2021/data/processed_data/clo_with_imports.owl
- Statistics Before Cleaning: 1387096 Triples; 111712 Classes; 41 Individuals; 116 Object Properties; 192 Annotation Properties; 7 Connected Components
- Statistics After Cleaning: 1422153 Triples; 111696 Classes; 33 Individuals; 112 Object Properties; 187 Annotation Properties; 7 Connected Components
- Value Errors (n=1):
    - RDF/XML parsing error in file builds/temp/clo_with_imports.owl, line 10971, column 99.
- Identifier Errors: 0
- Deprecated Classes: 2
- Obsolete Classes: 13
- Punning Errors:
    - Classes (n=10):
        - http://purl.obolibrary.org/obo/UBERON_0000062
        - http://purl.obolibrary.org/obo/UBERON_0001017
        - http://purl.obolibrary.org/obo/UBERON_0001004
        - http://purl.obolibrary.org/obo/UBERON_0000926
        - http://purl.obolibrary.org/obo/UBERON_0000029
        - http://purl.obolibrary.org/obo/UBERON_0001009
        - http://purl.obolibrary.org/obo/CLO_0054407
        - http://purl.obolibrary.org/obo/UBERON_0000383
        - http://purl.obolibrary.org/obo/CLO_0054409
        - http://purl.obolibrary.org/obo/UBERON_0001555
    - Object Properties (n=6):
        - http://purl.obolibrary.org/obo/RO_0002222
        - http://purl.obolibrary.org/obo/BFO_0000062
        - http://purl.obolibrary.org/obo/BFO_0000063
        - http://purl.obolibrary.org/obo/RO_0002161
        - http://purl.obolibrary.org/obo/RO_0002091
        - http://purl.obolibrary.org/obo/RO_0000087

ONTOLOGY: ext_with_imports.owl
******************************

- Original GCS URL: https://storage.googleapis.com/pheknowlator/archived_builds/release_v3.0.2/build_01NOV2021/data/original_data/ext_with_imports.owl

- Processed GCS URL: https://storage.googleapis.com/pheknowlator/archived_builds/release_v3.0.2/build_01NOV2021/data/processed_data/ext_with_imports.owl
- Statistics Before Cleaning: 769010 Triples; 28166 Classes; 0 Individuals; 239 Object Properties; 282 Annotation Properties; 19 Connected Components
- Statistics After Cleaning: 750999 Triples; 26577 Classes; 0 Individuals; 239 Object Properties; 282 Annotation Properties; 19 Connected Components
- Value Errors: 0
- Identifier Errors: 0
- Deprecated Classes: 1589
- Obsolete Classes: 0
- Punning Errors: 0

ONTOLOGY: go_with_imports.owl
*****************************
- Original GCS URL: https://storage.googleapis.com/pheknowlator/archived_builds/release_v3.0.2/build_01NOV2021/data/original_data/go_with_imports.owl
- Processed GCS URL: https://storage.googleapis.com/pheknowlator/archived_builds/release_v3.0.2/build_01NOV2021/data/processed_data/go_with_imports.owl
- Statistics Before Cleaning: 1425011 Triples; 62437 Classes; 0 Individuals; 9 Object Properties; 55 Annotation Properties; 2 Connected Components
- Statistics After Cleaning: 1334338 Triples; 55556 Classes; 0 Individuals; 9 Object Properties; 55 Annotation Properties; 2 Connected Components
- Value Errors: 0
- Identifier Errors: 0
- Deprecated Classes: 6881
- Obsolete Classes: 0
- Punning Errors: 0

ONTOLOGY: hp_with_imports.owl
*****************************
- Original GCS URL: https://storage.googleapis.com/pheknowlator/archived_builds/release_v3.0.2/build_01NOV2021/data/original_data/hp_with_imports.owl
- Processed GCS URL: https://storage.googleapis.com/pheknowlator/archived_builds/release_v3.0.2/build_01NOV2021/data/processed_data/hp_with_imports.owl
- Statistics Before Cleaning: 934712 Triples; 41442 Classes; 0 Individuals; 256 Object Properties; 226 Annotation Properties; 1 Connected Components
- Statistics After Cleaning: 934877 Triples; 41125 Classes; 0 Individuals; 256 Object Properties; 226 Annotation Properties; 1 Connected Components
- Value Errors: 0
- Identifier Errors: 0
- Deprecated Classes: 313
- Obsolete Classes: 0
- Punning Errors: 0

ONTOLOGY: mondo_with_imports.owl
********************************
- Original GCS URL: https://storage.googleapis.com/pheknowlator/archived_builds/release_v3.0.2/build_01NOV2021/data/original_data/mondo_with_imports.owl
- Processed GCS URL: https://storage.googleapis.com/pheknowlator/archived_builds/release_v3.0.2/build_01NOV2021/data/processed_data/mondo_with_imports.owl
- Statistics Before Cleaning: 2375692 Triples; 58553 Classes; 18 Individuals; 339 Object Properties; 153 Annotation Properties; 1 Connected Components
- Statistics After Cleaning: 2336440 Triples; 55929 Classes; 17 Individuals; 338 Object Properties; 153 Annotation Properties; 1 Connected Components
- Value Errors: 0
- Identifier Errors: 0
- Deprecated Classes: 2622
- Obsolete Classes: 0
- Punning Errors: 0

ONTOLOGY: pr_with_imports.owl
*****************************
- Original GCS URL: https://storage.googleapis.com/pheknowlator/archived_builds/release_v3.0.2/build_01NOV2021/data/original_data/pr_with_imports.owl
- Processed GCS URL: https://storage.googleapis.com/pheknowlator/archived_builds/release_v3.0.2/build_01NOV2021/data/processed_data/pr_with_imports.owl
- Statistics Before Cleaning: 2078128 Triples; 148164 Classes; 0 Individuals; 12 Object Properties; 11 Annotation Properties; 3 Connected Components
- Statistics After Cleaning: 2078128 Triples; 148164 Classes; 0 Individuals; 12 Object Properties; 11 Annotation Properties; 3 Connected Components
- Value Errors: 0
- Identifier Errors: 0

- Deprecated Classes: 0
- Obsolete Classes: 0
- Punning Errors: 0

ONTOLOGY: pw_with_imports.owl
*****************************
- Original GCS URL: https://storage.googleapis.com/pheknowlator/archived_builds/release_v3.0.2/build_01NOV2021/data/original_data/pw_with_imports.owl
- Processed GCS URL: https://storage.googleapis.com/pheknowlator/archived_builds/release_v3.0.2/build_01NOV2021/data/processed_data/pw_with_imports.owl
- Statistics Before Cleaning: 35291 Triples; 2642 Classes; 0 Individuals; 1 Object Properties; 19 Annotation Properties; 1 Connected Components
- Statistics After Cleaning: 34901 Triples; 2600 Classes; 0 Individuals; 1 Object Properties; 19 Annotation Properties; 1 Connected Components
- Value Errors: 0
- Identifier Errors: 0
- Deprecated Classes: 42
- Obsolete Classes: 0
- Punning Errors: 0

ONTOLOGY: ro_with_imports.owl
*****************************
- Original GCS URL: https://storage.googleapis.com/pheknowlator/archived_builds/release_v3.0.2/build_01NOV2021/data/original_data/ro_with_imports.owl
- Processed GCS URL: https://storage.googleapis.com/pheknowlator/archived_builds/release_v3.0.2/build_01NOV2021/data/processed_data/ro_with_imports.owl
- Statistics Before Cleaning: 8069 Triples; 99 Classes; 5 Individuals; 611 Object Properties; 113 Annotation Properties; 3 Connected Components
- Statistics After Cleaning: 7971 Triples; 99 Classes; 5 Individuals; 599 Object Properties; 113 Annotation Properties; 3 Connected Components
- Value Errors: 0
- Identifier Errors: 0
- Deprecated Classes: 12
- Obsolete Classes: 0
- Punning Errors: 0

ONTOLOGY: so_with_imports.owl
*****************************
- Original GCS URL: https://storage.googleapis.com/pheknowlator/archived_builds/release_v3.0.2/build_01NOV2021/data/original_data/so_with_imports.owl
- Processed GCS URL: https://storage.googleapis.com/pheknowlator/archived_builds/release_v3.0.2/build_01NOV2021/data/processed_data/so_with_imports.owl
- Statistics Before Cleaning: 44890 Triples; 2925 Classes; 0 Individuals; 50 Object Properties; 41 Annotation Properties; 1 Connected Components
- Statistics After Cleaning: 42198 Triples; 2583 Classes; 0 Individuals; 50 Object Properties; 41 Annotation Properties; 1 Connected Components
- Value Errors: 0
- Identifier Errors: 0
- Deprecated Classes: 342
- Obsolete Classes: 0
- Punning Errors: 0

ONTOLOGY: vo_with_imports.owl
*****************************
- Original GCS URL: https://storage.googleapis.com/pheknowlator/archived_builds/release_v3.0.2/build_01NOV2021/data/original_data/vo_with_imports.owl
- Processed GCS URL: https://storage.googleapis.com/pheknowlator/archived_builds/release_v3.0.2/build_01NOV2021/data/processed_data/vo_with_imports.owl
- Statistics Before Cleaning: 88373 Triples; 7198 Classes; 167 Individuals; 232 Object Properties; 99 Annotation Properties; 5 Connected Components
- Statistics After Cleaning: 91683 Triples; 7194 Classes; 167 Individuals; 232 Object Properties; 99 Annotation Properties; 5 Connected Components
- Value Errors: 0
- Identifier Errors (n=2):
    - http://purl.obolibrary.org/obo/PRO_000000001
    - http://purl.obolibrary.org/obo/PRO_000015399
- Deprecated Classes: 0
- Obsolete Classes: 0
- Punning Errors: 0

ONTOLOGY: PheKnowLator_MergedOntologies.owl
******************************************
- Original GCS URL: https://storage.googleapis.com/pheknowlator/archived_builds/release_v3.0.2/build_01NOV2021/data/original_data/PheKnowLator_MergedOntologies.owl
- Processed GCS URL: https://storage.googleapis.com/pheknowlator/archived_builds/release_v3.0.2/build_01NOV2021/data/processed_data/PheKnowLator_MergedOntologies.owl
- Statistics Before Cleaning: 13934582 Triples; 562412 Classes; 197 Individuals; 853 Object Properties; 635 Annotation Properties; 8 Connected Components
- Statistics After Cleaning: 13935691 Triples; 558715 Classes; 190 Individuals; 852 Object Properties; 635 Annotation Properties; 8 Connected Components
- Value Errors: 0
- Identifier Errors (n=2):
    - http://purl.obolibrary.org/obo/PRO_000000001
    - http://purl.obolibrary.org/obo/PRO_000015399
- Punning Errors:
    - Classes (n=8):
        - http://purl.obolibrary.org/obo/NCBITaxon_147099
        - http://purl.obolibrary.org/obo/NCBITaxon_6040
        - http://purl.obolibrary.org/obo/CLO_0054409
        - http://purl.obolibrary.org/obo/NCBITaxon_6157
        - http://purl.obolibrary.org/obo/NCBITaxon_6073
        - http://purl.obolibrary.org/obo/NCBITaxon_41324
        - http://purl.obolibrary.org/obo/NCBITaxon_8570
        - http://purl.obolibrary.org/obo/NCBITaxon_110815
    - Object Properties: 0
- Normalization:
    - Normalized Entities (n=7):
        - OGG_0000000002 rdfs:subClassOf SO_0000704
        - PR_000000001 rdfs:subClassOf SO_0000104
        - CHEBI_36080 rdfs:subClassOf SO_0000104
        - OGMS_0000045 rdfs:subClassOf MONDO_0000001
        - OBI_1110034 rdfs:subClassOf CHEBI_59132
        - VO_0003030 rdfs:subClassOf CHEBI_5291
        - FMA_12278 rdfs:subClassOf CHEBI_24621
    - Normalized HGNC IDs: 23631
    - Other Classes that May Need Normalizing: 404495
    - Deprecated Ontology HGNC Identifiers Needing Alignment: 0

## Supplementary Document 6. pkt_builder_phases12_log.log (Data Download and Preparation).

{"asctime": "2021-11-01 01:14:27,998", "levelname": "INFO", "name": "__main__", "module": "phases1_2_entrypoint", "funcName": "main", "lineno": 40, "message": "********** STARTING PHEKNOWLATOR KNOWLEDGE GRAPH BUILD **********"}
{"asctime": "2021-11-01 01:14:28,087", "levelname": "INFO", "name": "__main__", "module": "phases1_2_entrypoint", "funcName": "main", "lineno": 45, "message": "#####BUILD PHASE 1: DOWNLOADING BUILD DATA#####"}
{"asctime": "2021-11-01 01:14:28,148", "levelname": "INFO", "name": "builds.build_phase_1", "module": "build_phase_1", "funcName": "creates_build_directory_structure", "lineno": 46, "message": "Creating Build Directory Structure"}
{"asctime": "2021-11-01 01:14:29,081", "levelname": "INFO", "name": "builds.build_phase_1", "module": "build_phase_1", "funcName": "downloads_build_data", "lineno": 136, "message": "Downloading Build Data"}
{"asctime": "2021-11-01 01:14:29,111", "levelname": "INFO", "name": "builds.build_phase_1", "module": "build_phase_1", "funcName": "downloads_build_data", "lineno": 143, "message": "Downloading http://purl.obolibrary.org/obo/hp.owl"}
{"asctime": "2021-11-01 01:14:49,795", "levelname": "INFO", "name": "builds.build_phase_1", "module": "build_phase_1", "funcName": "downloads_build_data", "lineno": 143, "message": "Downloading http://purl.obolibrary.org/obo/go.owl"}
{"asctime": "2021-11-01 01:15:14,376", "levelname": "INFO", "name": "builds.build_phase_1", "module": "build_phase_1", "funcName": "downloads_build_data", "lineno": 143, "message": "Downloading http://purl.obolibrary.org/obo/mondo.owl"}
{"asctime": "2021-11-01 01:15:54,602", "levelname": "INFO", "name": "builds.build_phase_1", "module": "build_phase_1", "funcName": "downloads_build_data", "lineno": 143, "message": "Downloading http://purl.obolibrary.org/obo/vo.owl"}
{"asctime": "2021-11-01 01:15:59,539", "levelname": "INFO", "name": "builds.build_phase_1", "module": "build_phase_1", "funcName": "downloads_build_data", "lineno": 143, "message": "Downloading http://purl.obolibrary.org/obo/chebi.owl"}
{"asctime": "2021-11-01 01:35:06,926", "levelname": "INFO", "name": "builds.build_phase_1", "module": "build_phase_1", "funcName": "downloads_build_data", "lineno": 143, "message": "Downloading http://purl.obolibrary.org/obo/uberon/ext.owl"}
{"asctime": "2021-11-01 01:35:26,945", "levelname": "INFO", "name": "builds.build_phase_1", "module": "build_phase_1", "funcName": "downloads_build_data", "lineno": 143, "message": "Downloading http://purl.obolibrary.org/obo/clo.owl"}
{"asctime": "2021-11-01 01:35:56,608", "levelname": "INFO", "name": "builds.build_phase_1", "module": "build_phase_1", "funcName": "downloads_build_data", "lineno": 143, "message": "Downloading http://purl.obolibrary.org/obo/pr.owl"}
{"asctime": "2021-11-01 01:39:19,579", "levelname": "INFO", "name": "builds.build_phase_1", "module": "build_phase_1", "funcName": "downloads_build_data", "lineno": 143, "message": "Downloading http://purl.obolibrary.org/obo/so.owl"}
{"asctime": "2021-11-01 01:39:23,524", "levelname": "INFO", "name": "builds.build_phase_1", "module": "build_phase_1", "funcName": "downloads_build_data", "lineno": 143, "message": "Downloading http://purl.obolibrary.org/obo/pw.owl"}
{"asctime": "2021-11-01 01:39:26,893", "levelname": "INFO", "name": "builds.build_phase_1", "module": "build_phase_1", "funcName": "downloads_build_data", "lineno": 143, "message": "Downloading http://purl.obolibrary.org/obo/ro.owl"}
{"asctime": "2021-11-01 01:39:32,386", "levelname": "INFO", "name": "builds.build_phase_1", "module": "build_phase_1", "funcName": "downloads_build_data", "lineno": 143, "message": "Downloading hgnc_complete_set.txt, http://ftp.ebi.ac.uk/pub/databases/genenames/hgnc/tsv/hgnc_complete_set.txt"}
{"asctime": "2021-11-01 01:39:39,226", "levelname": "INFO", "name": "builds.build_phase_1", "module": "build_phase_1", "funcName": "downloads_build_data", "lineno": 143, "message": "Downloading Homo_sapiens.GRCh38.102.gtf, ftp://ftp.ensembl.org/pub/release-102/gtf/homo_sapiens/Homo_sapiens.GRCh38.102.gtf.gz"}
{"asctime": "2021-11-01 01:39:57,391", "levelname": "INFO", "name": "builds.build_phase_1", "module": "build_phase_1", "funcName": "downloads_build_data", "lineno": 143, "message": "Downloading Homo_sapiens.GRCh38.102.uniprot.tsv, ftp://ftp.ensembl.org/pub/release-102/tsv/homo_sapiens/Homo_sapiens.GRCh38.102.uniprot.tsv.gz"}
{"asctime": "2021-11-01 01:39:59,843", "levelname": "INFO", "name": "builds.build_phase_1", "module": "build_phase_1", "funcName": "downloads_build_data", "lineno": 143, "message": "Downloading Homo_sapiens.GRCh38.102.entrez.tsv, ftp://ftp.ensembl.org/pub/release-102/tsv/homo_sapiens/Homo_sapiens.GRCh38.102.entrez.tsv.gz"}
{"asctime": "2021-11-01 01:40:01,973", "levelname": "INFO", "name": "builds.build_phase_1", "module": "build_phase_1", "funcName": "downloads_build_data", "lineno": 143, "message": "Downloading uniprot_identifier_mapping.tab, https://www.uniprot.org/uniprot/?query=&fil=organism%3A%22Homo%20sapiens%20(Human)%20%5B9606%5D%22&columns=id%2Creviewed%2Cdatabase(GeneID)%2Cdatabase(Ensembl)%2Cdatabase(HGNC)%2Cgenes(ALTERNATIVE)%2Cgenes(PREFERRED)&format=tab"}
{"asctime": "2021-11-01 01:41:20,156", "levelname": "INFO", "name": "builds.build_phase_1", "module": "build_phase_1", "funcName": "downloads_build_data", "lineno": 143, "message": "Downloading Homo_sapiens.gene_info, ftp://ftp.ncbi.nih.gov/gene/DATA/GENE_INFO/Mammalia/Homo_sapiens.gene_info.gz"}
{"asctime": "2021-11-01 01:41:21,503", "levelname": "INFO", "name": "builds.build_phase_1", "module": "build_phase_1", "funcName": "downloads_build_data", "lineno": 143, "message": "Downloading promapping.txt, https://proconsortium.org/download/current/promapping.txt"}
{"asctime": "2021-11-01 01:41:22,649", "levelname": "INFO", "name": "builds.build_phase_1", "module": "build_phase_1", "funcName": "downloads_build_data", "lineno": 143, "message": "Downloading mesh2021.nt, ftp://nlmpubs.nlm.nih.gov/online/mesh/rdf/2021/mesh2021.nt"}
{"asctime": "2021-11-01 01:41:54,362", "levelname": "INFO", "name": "builds.build_phase_1", "module": "build_phase_1", "funcName": "downloads_build_data", "lineno": 143, "message": "Downloading names.tsv, ftp://ftp.ebi.ac.uk/pub/databases/chebi/Flat_file_tab_delimited/names.tsv.gz"}
{"asctime": "2021-11-01 01:41:57,241", "levelname": "INFO", "name": "builds.build_phase_1", "module": "build_phase_1", "funcName": "downloads_build_data", "lineno": 143, "message": "Downloading disease_mappings.tsv, https://www.disgenet.org/static/disgenet_ap1/files/downloads/disease_mappings.tsv.gz"}
{"asctime": "2021-11-01 01:41:59,514", "levelname": "INFO", "name": "builds.build_phase_1", "module": "build_phase_1", "funcName": "downloads_build_data", "lineno": 143, "message": "Downloading proteinatlas_search.tsv.gz, https://www.proteinatlas.org/api/search_download.php?search=&columns=g,eg,up,pe,rnatsm,rnaclsm,rnacasm,rnabrsm,rnabcsm,rnablsm,scl,t_RNA_adipose_tissue,t_RNA_adrenal_gland,t_RNA_amygdala,t_RNA_appendix,t_RNA_basal_ganglia,t_RNA_bone_marrow,t_RNA_breast,t_RNA_cerebellum,t_RNA_cerebral_cortex,t_RNA_cervix,_uterine,t_RNA_colon,t_RNA_corpus_callosum,t_RNA_ductus_deferens,t_RNA_duodenum,t_RNA_endometrium_1,t_RNA_epididymis,t_RNA_esophagus,t_RNA_fallopian_tube,t_RNA_gallbladder,t_RNA_heart_muscle,t_RNA_hippocampal_formation,t_RNA_hypothalamus,t_RNA_kidney,t_RNA_liver,t_RNA_lung,t_RNA_lymph_node,t_RNA_midbrain,t_RNA_olfactory_region,t_RNA_ovary,t_RNA_pancreas,t_RNA_parathyroid_gland,t_RNA_pituitary_gland,t_RNA_placenta,t_RNA_pons_and_medulla,t_RNA_prostate,t_RNA_rectum,t_RNA_retina,t_RNA_salivary_gland,t_RNA_seminal_vesicle,t_RNA_skeletal_muscle,t_RNA_skin_1,t_RNA_small_intestine,t_RNA_smooth_muscle,t_RNA_spinal_cord,t_RNA_spleen,t_RNA_stomach_1,t_RNA_testis,t_RNA_thalamus,t_RNA_thymus,t_RNA_thyroid_gland,t_RNA_tongue,t_RNA_tonsil,t_RNA_urinary_bladder,t_RNA_vagina,t_RNA_B-cells,t_RNA_dendritic_cells,t_RNA_granulocytes,t_RNA_monocytes,t_RNA_NK-cells,t_RNA_T-cells,t_RNA_total_PBMC,cell_RNA_A-431,cell_RNA_A549,cell_RNA_AF22,cell_RNA_AN3-CA,cell_RNA_ASC_diff,cell_RNA_ASC_TERT1,cell_RNA_BEWO,cell_RNA_BJ,cell_RNA_BJ_hTERT+,cell_RNA_BJ_hTERT+_SV40_Large_T+,cell_RNA_BJ_hTERT+_SV40_Large_T+_RasG12V,cell_RNA_CACO-2,cell_RNA_CAPAN-2,cell_RNA_Daudi,cell_RNA_EFO-21,cell_RNA_fHDF/TERT166,cell_RNA_HaCaT,cell_RNA_HAP1,cell_RNA_HBEC3-KT,cell_RNA_HBF_TERT88,cell_RNA_HDLM-2,cell_RNA_HEK_293,cell_RNA_HEL,cell_RNA_HeLa,cell_RNA_Hep_G2,cell_RNA_HHSteC,cell_RNA_HL-60,cell_RNA_HMC-1,cell_RNA_HSkMC,cell_RNA_hTCEpi,cell_RNA_hTEC/SVTERT24-B,cell_RNA_hTERT-HME1,cell_RNA_HUVEC_TERT2,cell_RNA_K-562,cell_RNA_Karpas-707,cell_RNA_LHCN-M2,cell_RNA_MCF7,cell_RNA_MOLT-4,cell_RNA_NB-4,cell_RNA_NTERA-2,cell_RNA_PC-3,cell_RNA_REH,cell_RNA_RH-30,cell_RNA_RPMI-8226,cell_RNA_RPTEC_TERT1,cell_RNA_RT4,cell_RNA_SCLC-21H,cell_RNA_SH-SY5Y,cell_RNA_SiHa,cell_RNA_SK-BR-3,cell_RNA_SK-MEL-30,cell_RNA_T-47d,cell_RNA_THP-1,cell_RNA_TIME,cell_RNA_U-138_MG,cell_RNA_U-2_OS,cell_RNA_U-2197,cell_RNA_U-251_MG,cell_RNA_U-266/70,cell_RNA_U-266/84,cell_RNA_U-698,cell_RNA_U-87_MG,cell_RNA_U-937,cell_RNA_WM-115,blood_RNA_basophil,blood_RNA_classical_monocyte,blood_RNA_eosinophil,blood_RNA_gdT-cell,blood_RNA_intermediate_monocyte,blood_RNA_MAIT_T-cell,blood_RNA_memory_B-cell,blood_RNA_memory_CD4_T-cell,blood_RNA_memory_CD8_T-cell,blood_RNA_myeloid_DC,blood_RNA_naive_B-cell,blood_RNA_naive_CD4_T-cell,blood_RNA_naive_CD8_T-cell,blood_RNA_neutrophil,blood_RNA_NK-cell,blood_RNA_non-classical_monocyte,blood_RNA_plasmacytoid_DC,blood_RNA_T-reg,blood_RNA_total_PBMC,brain_RNA_amygdala,brain_RNA_basal_ganglia,brain_RNA_cerebellum,brain_RNA_cerebral_cortex,brain_RNA_hippocampal_formation,brain_RNA_hypothalamus,brain_RNA_midbrain,brain_RNA_olfactory_region,brain_RNA_pons_and_medulla,brain_RNA_thalamus&format=tsv"}
{"asctime": "2021-11-01 01:42:11,936", "levelname": "INFO", "name": "builds.build_phase_1", "module": "build_phase_1", "funcName": "downloads_build_data", "lineno": 143, "message": "Downloading GTEx_Analysis_2017-06-05_v8_RNASeQCv1.1.9_gene_median_tpm.gct, https://storage.googleapis.com/gtex_analysis_v8/rna_seq_data/GTEx_Analysis_2017-06-05_v8_RNASeQCv1.1.9_gene_median_tpm.gct.gz"}
{"asctime": "2021-11-01 01:42:12,902", "levelname": "INFO", "name": "builds.build_phase_1", "module": "build_phase_1", "funcName": "downloads_build_data", "lineno": 143, "message": "Downloading ReactomePathways.txt, https://reactome.org/download/current/ReactomePathways.txt"}
{"asctime": "2021-11-01 01:42:13,467", "levelname": "INFO", "name": "builds.build_phase_1", "module": "build_phase_1", "funcName": "downloads_build_data", "lineno": 143, "message": "Downloading gene_association.reactome, https://reactome.org/download/current/gene_association.reactome.gz"}
{"asctime": "2021-11-01 01:42:14,221", "levelname": "INFO", "name": "builds.build_phase_1", "module": "build_phase_1", "funcName": "downloads_build_data", "lineno": 143, "message": "Downloading ChEBI2Reactome_All_Levels.txt, https://reactome.org/download/current/ChEBI2Reactome_All_Levels.txt"}
{"asctime": "2021-11-01 01:42:15,421", "levelname": "INFO", "name": "builds.build_phase_1", "module": "build_phase_1", "funcName": "downloads_build_data", "lineno": 143, "message": "Downloading compath_canonical_pathway_mappings.txt, http://compath.scai.fraunhofer.de/export_mappings"}
{"asctime": "2021-11-01 01:42:17,215", "levelname": "INFO", "name": "builds.build_phase_1", "module": "build_phase_1", "funcName": "downloads_build_data", "lineno": 143, "message": "Downloading kegg_reactome.csv, https://raw.githubusercontent.com/ComPath/resources/master/mappings/kegg_reactome.csv"}
{"asctime": "2021-11-01 01:42:17,653", "levelname": "INFO", "name": "builds.build_phase_1", "module": "build_phase_1", "funcName": "downloads_build_data", "lineno": 143, "message": "Downloading genomic_sequence_ontology_mappings.xlsx, https://storage.googleapis.com/pheknowlator/curated_data/genomic_sequence_ontology_mappings.xlsx"}
{"asctime": "2021-11-01 01:42:18,028", "levelname": "INFO", "name": "builds.build_phase_1", "module": "build_phase_1", "funcName": "downloads_build_data", "lineno": 143, "message": "Downloading human_pro_classes.html, https://sparql.proconsortium.org/virtuoso/sparql?query=PREFIX+obo%3A+%3Chttp%3A%2F%2Fpurl.obolibrary.org%2Fobo%2F%3E%0D%0A%0D%0ASELECT+%3FPRO_term%0D%0AFROM+%3Chttp%3A%2F%2Fpurl.obolibrary.org%2Fobo%2Fpr%3E%0D%0AWHERE+%7B%0D%0A+++++++%3FPRO_term+rdf%3Atype+owl%3AClass+.%0D%0A+++++++%3FPRO_term+rdfs%3AsubClassOf+%3Frestriction+.%0D%0A+++++++%3Frestriction+owl%3AonProperty+obo%3ARO_0002160+.%0D%0A+++++++%3Frestriction+owl%3AsomeValuesFrom+obo%3ANCBITaxon_9606+.%0D%0A%0D%0A+++++++%23+use+this+to+filter-out+things+like+hgnc+ids%0D%0A+++++++FILTER+%28regex%28%3FPRO_term%2C%22http%3A%2F%2Fpurl.obolibrary.org%2Fobo%2F*%22%29%29+.%0D%0A%7D&format=text%2Fhtml&debug="}
{"asctime": "2021-11-01 01:42:19,108", "levelname": "INFO", "name": "builds.build_phase_1", "module": "build_phase_1", "funcName": "downloads_build_data", "lineno": 143, "message": "Downloading variant_summary.txt, ftp://ftp.ncbi.nlm.nih.gov/pub/clinvar/tab_delimited/variant_summary.txt.gz"}
{"asctime": "2021-11-01 01:42:34,238", "levelname": "INFO", "name": "builds.build_phase_1", "module": "build_phase_1", "funcName": "downloads_build_data", "lineno": 143, "message": "Downloading uniprot-cofactor-catalyst.tab, https://www.uniprot.org/uniprot/?query=&fil=organism%3A%22Homo%20sapiens%20(Human)%20%5B9606%5D%22&columns=id%2Creviewed%2Centry%20name%2Cdatabase(PRO)%2Cchebi(Cofactor)%2Cchebi(Catalytic%20activity)&format=tab"}
{"asctime": "2021-11-01 01:43:24,015", "levelname": "INFO", "name": "builds.build_phase_1", "module": "build_phase_1", "funcName": "downloads_build_data", "lineno": 143, "message": "Downloading CTD_chem_gene_ixns.tsv, http://ctdbase.org/reports/CTD_chem_gene_ixns.tsv.gz"}
{"asctime": "2021-11-01 01:43:30,537", "levelname": "INFO", "name": "builds.build_phase_1", "module": "build_phase_1", "funcName": "downloads_build_data", "lineno": 143, "message": "Downloading CTD_chem_go_enriched.tsv, http://ctdbase.org/reports/CTD_chem_go_enriched.tsv.gz"}

{"asctime": "2021-11-01 01:43:55,405", "levelname": "INFO", "name": "builds.build_phase_1", "module": "build_phase_1", "funcName": "downloads_build_data", "lineno": 143, "message": "Downloading ChEBI2Reactome_All_Levels.txt, https://reactome.org/download/current/ChEBI2Reactome_All_Levels.txt"}
{"asctime": "2021-11-01 01:43:56,770", "levelname": "INFO", "name": "builds.build_phase_1", "module": "build_phase_1", "funcName": "downloads_build_data", "lineno": 143, "message": "Downloading CTD_chemicals_diseases.tsv, http://ctdbase.org/reports/CTD_chemicals_diseases.tsv.gz"}
{"asctime": "2021-11-01 01:44:18,538", "levelname": "INFO", "name": "builds.build_phase_1", "module": "build_phase_1", "funcName": "downloads_build_data", "lineno": 143, "message": "Downloading phenotype.hpoa, http://purl.obolibrary.org/obo/hp/hpoa/phenotype.hpoa"}
{"asctime": "2021-11-01 01:44:20,280", "levelname": "INFO", "name": "builds.build_phase_1", "module": "build_phase_1", "funcName": "downloads_build_data", "lineno": 143, "message": "Downloading curated_gene_disease_associations.tsv, https://www.disgenet.org/static/disgenet_ap1/files/downloads/curated_gene_disease_associations.tsv.gz"}
{"asctime": "2021-11-01 01:44:22,546", "levelname": "INFO", "name": "builds.build_phase_1", "module": "build_phase_1", "funcName": "downloads_build_data", "lineno": 143, "message": "Downloading COMBINED.DEFAULT_NETWORKS.BP_COMBINING.txt, http://genemania.org/data/current/Homo_sapiens.COMBINED/COMBINED.DEFAULT_NETWORKS.BP_COMBINING.txt"}
{"asctime": "2021-11-01 01:44:53,645", "levelname": "INFO", "name": "builds.build_phase_1", "module": "build_phase_1", "funcName": "downloads_build_data", "lineno": 143, "message": "Downloading CTD_genes_pathways.tsv, http://ctdbase.org/reports/CTD_genes_pathways.tsv.gz"}
{"asctime": "2021-11-01 01:44:54,405", "levelname": "INFO", "name": "builds.build_phase_1", "module": "build_phase_1", "funcName": "downloads_build_data", "lineno": 143, "message": "Downloading gene_association.reactome, https://reactome.org/download/current/gene_association.reactome.gz"}
{"asctime": "2021-11-01 01:44:55,295", "levelname": "INFO", "name": "builds.build_phase_1", "module": "build_phase_1", "funcName": "downloads_build_data", "lineno": 143, "message": "Downloading goa_human.gaf, http://current.geneontology.org/annotations/goa_human.gaf.gz"}
{"asctime": "2021-11-01 01:44:58,335", "levelname": "INFO", "name": "builds.build_phase_1", "module": "build_phase_1", "funcName": "downloads_build_data", "lineno": 143, "message": "Downloading UniProt2Reactome_All_Levels.txt, https://reactome.org/download/current/UniProt2Reactome_All_Levels.txt"}
{"asctime": "2021-11-01 01:45:01,370", "levelname": "INFO", "name": "builds.build_phase_1", "module": "build_phase_1", "funcName": "downloads_build_data", "lineno": 143, "message": "Downloading 9606.protein.links.v11.0.txt, https://stringdb-static.org/download/protein.links.v11.0/9606.protein.links.v11.0.txt.gz"}
{"asctime": "2021-11-01 01:45:14,415", "levelname": "INFO", "name": "builds.build_phase_1", "module": "build_phase_1", "funcName": "downloads_build_data", "lineno": 143, "message": "Downloading genomic_typing_dict.pkl, https://storage.googleapis.com/pheknowlator/curated_data/genomic_typing_dict.pkl"}
{"asctime": "2021-11-01 01:45:14,726", "levelname": "INFO", "name": "builds.build_phase_1", "module": "build_phase_1", "funcName": "downloads_build_data", "lineno": 143, "message": "Downloading zooma_tissue_cell_mapping_04JAN2020.xlsx, https://storage.googleapis.com/pheknowlator/curated_data/zooma_tissue_cell_mapping_04JAN2020.xlsx"}
{"asctime": "2021-11-01 01:45:15,900", "levelname": "INFO", "name": "__main__", "module": "phases1_2_entrypoint", "funcName": "main", "lineno": 52, "message": "#####BUILD PHASE 2: DATA PRE-PROCESSING#####"}
{"asctime": "2021-11-01 01:45:16,014", "levelname": "INFO", "name": "builds.data_preprocessing", "module": "data_preprocessing", "funcName": "preprocesses_build_data", "lineno": 1616, "message": "*** PROCESSING LINKED OPEN DATA SOURCES ***"}
{"asctime": "2021-11-01 01:45:16,015", "levelname": "INFO", "name": "builds.data_preprocessing", "module": "data_preprocessing", "funcName": "preprocesses_build_data", "lineno": 1619, "message": "STEP 1: HUMAN TRANSCRIPT, GENE, PROTEIN ID MAPPING"}
{"asctime": "2021-11-01 01:45:16,015", "levelname": "INFO", "name": "builds.data_preprocessing", "module": "data_preprocessing", "funcName": "generates_specific_genomic_identifier_maps", "lineno": 506, "message": "Generating Pairwise Genomic Cross-Map Sets"}
{"asctime": "2021-11-01 01:45:16,015", "levelname": "INFO", "name": "builds.data_preprocessing", "module": "data_preprocessing", "funcName": "_loads_genomic_typing_dictionary", "lineno": 114, "message": "Loading Genomic Typing Dictionary"}
{"asctime": "2021-11-01 01:45:16,075", "levelname": "INFO", "name": "builds.data_preprocessing", "module": "data_preprocessing", "funcName": "creates_master_genomic_identifier_map", "lineno": 469, "message": "Creating Genomic ID Cross-Map Dictionary"}
{"asctime": "2021-11-01 01:45:16,075", "levelname": "INFO", "name": "builds.data_preprocessing", "module": "data_preprocessing", "funcName": "_cross_maps_genomic_identifier_data", "lineno": 436, "message": "Cross-Mapping Genomic Identifier Data"}
{"asctime": "2021-11-01 01:45:16,076", "levelname": "INFO", "name": "builds.data_preprocessing", "module": "data_preprocessing", "funcName": "_fixes_genomic_symbols", "lineno": 406, "message": "Fixing genomic Symbols"}
{"asctime": "2021-11-01 01:45:16,076", "levelname": "INFO", "name": "builds.data_preprocessing", "module": "data_preprocessing", "funcName": "_merges_genomic_identifier_data", "lineno": 373, "message": "Merging Genomic ID Data"}
{"asctime": "2021-11-01 01:45:16,076", "levelname": "INFO", "name": "builds.data_preprocessing", "module": "data_preprocessing", "funcName": "_preprocess_hgnc_data", "lineno": 133, "message": "Preprocessing HGNC Data"}
{"asctime": "2021-11-01 01:45:40,677", "levelname": "INFO", "name": "builds.data_preprocessing", "module": "data_preprocessing", "funcName": "merges_ensembl_mapping_data", "lineno": 218, "message": "Merging Ensembl Annotation Data"}
{"asctime": "2021-11-01 01:45:43,385", "levelname": "INFO", "name": "builds.data_preprocessing", "module": "data_preprocessing", "funcName": "_preprocess_ensembl_data", "lineno": 176, "message": "Preprocessing Ensembl Data"}
{"asctime": "2021-11-01 01:47:36,552", "levelname": "INFO", "name": "builds.data_preprocessing", "module": "data_preprocessing", "funcName": "_preprocess_uniprot_data", "lineno": 268, "message": "Preprocessing UniProt Data"}
{"asctime": "2021-11-01 01:47:47,746", "levelname": "INFO", "name": "builds.data_preprocessing", "module": "data_preprocessing", "funcName": "_preprocess_ncbi_data", "lineno": 299, "message": "Preprocessing Entrez Data"}
{"asctime": "2021-11-01 01:48:45,188", "levelname": "INFO", "name": "builds.data_preprocessing", "module": "data_preprocessing", "funcName": "_preprocess_protein_ontology_mapping_data", "lineno": 352, "message": "Preprocessing Protein Ontology Data"}
{"asctime": "2021-11-01 02:29:55,647", "levelname": "INFO", "name": "builds.data_preprocessing", "module": "data_preprocessing", "funcName": "preprocesses_build_data", "lineno": 1624, "message": "STEP 2: MESH-CHEBI ID MAPPING"}
{"asctime": "2021-11-01 02:29:55,647", "levelname": "INFO", "name": "builds.data_preprocessing", "module": "data_preprocessing", "funcName": "creates_chebi_to_mesh_identifier_mappings", "lineno": 607, "message": "Creating MeSH-ChEBI ID Cross-Map Data"}
{"asctime": "2021-11-01 02:29:55,648", "levelname": "INFO", "name": "builds.data_preprocessing", "module": "data_preprocessing", "funcName": "_processes_mesh_data", "lineno": 543, "message": "Preprocessing MeSH Data"}
{"asctime": "2021-11-01 02:31:23,566", "levelname": "INFO", "name": "builds.data_preprocessing", "module": "data_preprocessing", "funcName": "_processes_chebi_data", "lineno": 585, "message": "Preprocessing ChEBI Data"}
{"asctime": "2021-11-01 02:31:30,646", "levelname": "INFO", "name": "builds.data_preprocessing", "module": "data_preprocessing", "funcName": "preprocesses_build_data", "lineno": 1629, "message": "STEP 3: DISEASE-PHENOTYPE ID MAPPING"}
{"asctime": "2021-11-01 02:31:30,646", "levelname": "INFO", "name": "builds.data_preprocessing", "module": "data_preprocessing", "funcName": "creates_disease_identifier_mappings", "lineno": 676, "message": "Creating Phenotype and Disease ID Cross-Map Data"}
{"asctime": "2021-11-01 02:31:30,646", "levelname": "INFO", "name": "builds.data_preprocessing", "module": "data_preprocessing", "funcName": "_preprocess_mondo_mapping_data", "lineno": 637, "message": "Loading MonDO Disease Ontology Data"}
{"asctime": "2021-11-01 02:38:35,522", "levelname": "INFO", "name": "builds.data_preprocessing", "module": "data_preprocessing", "funcName": "_preprocess_hpo_mapping_data", "lineno": 657, "message": "Loading Human Phenotype Ontology Data"}
{"asctime": "2021-11-01 02:42:04,165", "levelname": "INFO", "name": "builds.data_preprocessing", "module": "data_preprocessing", "funcName": "preprocesses_build_data", "lineno": 1634, "message": "STEP 4: CREATING HPA + GTEX ID EDGE DATA"}
{"asctime": "2021-11-01 02:42:04,165", "levelname": "INFO", "name": "builds.data_preprocessing", "module": "data_preprocessing", "funcName": "_hpa_gtex_ontology_alignment", "lineno": 730, "message": "Preprocessing HPA Data"}
{"asctime": "2021-11-01 02:42:06,905", "levelname": "INFO", "name": "builds.data_preprocessing", "module": "data_preprocessing", "funcName": "processes_hpa_gtex_data", "lineno": 781, "message": "Creating Human Protein Atlas and GTEx Cross-Map Data"}
{"asctime": "2021-11-01 02:42:06,906", "levelname": "INFO", "name": "builds.data_preprocessing", "module": "data_preprocessing", "funcName": "_extracts_hpa_tissue_information", "lineno": 754, "message": "Extracting HPA Tissue and Cell Information"}
{"asctime": "2021-11-01 02:43:51,974", "levelname": "INFO", "name": "builds.data_preprocessing", "module": "data_preprocessing", "funcName": "preprocesses_build_data", "lineno": 1640, "message": "STEP 5: SEQUENCE ONTOLOGY + PATHWAY ID MAP"}
{"asctime": "2021-11-01 02:43:51,974", "levelname": "INFO", "name": "builds.data_preprocessing", "module": "data_preprocessing", "funcName": "combines_pathway_and_sequence_ontology_dictionaries", "lineno": 1112, "message": "Creating Pathway and Sequence Ontology Mapping Dictionary"}
{"asctime": "2021-11-01 02:43:51,975", "levelname": "INFO", "name": "builds.data_preprocessing", "module": "data_preprocessing", "funcName": "_creates_sequence_identifier_mappings", "lineno": 1084, "message": "Creating Sequence Ontology ID Cross-Map Data"}
{"asctime": "2021-11-01 02:43:52,107", "levelname": "INFO", "name": "builds.data_preprocessing", "module": "data_preprocessing", "funcName": "_preprocesses_gene_types", "lineno": 993, "message": "Mapping Sequence Ontology Classes to Gene IDs"}
{"asctime": "2021-11-01 02:44:06,476", "levelname": "INFO", "name": "builds.data_preprocessing", "module": "data_preprocessing", "funcName": "_preprocesses_transcript_types", "lineno": 1028, "message": "Mapping Sequence Ontology Classes to Transcript IDs"}
{"asctime": "2021-11-01 02:44:58,316", "levelname": "INFO", "name": "builds.data_preprocessing", "module": "data_preprocessing", "funcName": "_preprocesses_variant_types", "lineno": 1059, "message": "Mapping Sequence Ontology Classes to Variant IDs"}
{"asctime": "2021-11-01 02:51:39,102", "levelname": "INFO", "name": "builds.data_preprocessing", "module": "data_preprocessing", "funcName": "_creates_pathway_identifier_mappings", "lineno": 964, "message": "Creating Pathway Ontology ID Cross-Map Data"}
{"asctime": "2021-11-01 02:51:39,102", "levelname": "INFO", "name": "builds.data_preprocessing", "module": "data_preprocessing", "funcName": "_preprocess_pathway_mapping_data", "lineno": 831, "message": "Loading Protein Ontology Data"}
{"asctime": "2021-11-01 02:51:46,164", "levelname": "INFO", "name": "builds.data_preprocessing", "module": "data_preprocessing", "funcName": "_processes_reactome_data", "lineno": 852, "message": "Loading Reactome Annotation Data"}
{"asctime": "2021-11-01 02:51:47,090", "levelname": "INFO", "name": "builds.data_preprocessing", "module": "data_preprocessing", "funcName": "_processes_compath_pathway_data", "lineno": 882, "message": "Loading ComPath Canonical Pathway Data"}
{"asctime": "2021-11-01 02:51:47,482", "levelname": "INFO", "name": "builds.data_preprocessing", "module": "data_preprocessing", "funcName": "_processes_kegg_pathway_data", "lineno": 911, "message": "Loading KEGG Data"}
{"asctime": "2021-11-01 02:51:47,671", "levelname": "INFO", "name": "builds.data_preprocessing", "module": "data_preprocessing", "funcName": "_queries_reactome_api", "lineno": 940, "message": "Querying Reactome API for Reactome-GO BP Mappings"}
{"asctime": "2021-11-01 02:56:50,953", "levelname": "INFO", "name": "builds.data_preprocessing", "module": "data_preprocessing", "funcName": "preprocesses_build_data", "lineno": 1645, "message": "STEP 6: CREATING A HUMAN PROTEIN ONTOLOGY"}
{"asctime": "2021-11-01 02:56:50,954", "levelname": "INFO", "name": "builds.data_preprocessing", "module": "data_preprocessing", "funcName": "constructs_human_protein_ontology", "lineno": 1202, "message": "Construct a Human PRotein Ontology"}
{"asctime": "2021-11-01 02:56:50,954", "levelname": "INFO", "name": "builds.data_preprocessing", "module": "data_preprocessing", "funcName": "_processes_protein_ontology_data", "lineno": 1134, "message": "Loading Protein Ontology Data"}
{"asctime": "2021-11-01 04:31:21,804", "levelname": "INFO", "name": "builds.data_preprocessing", "module": "data_preprocessing", "funcName": "_logically_verifies_human_protein_ontology", "lineno": 1172, "message": "Logically Verifying Human Protein Ontology Subset"}
{"asctime": "2021-11-01 04:34:16,112", "levelname": "INFO", "name": "builds.data_preprocessing", "module": "data_preprocessing", "funcName": "preprocesses_build_data", "lineno": 1650, "message": "STEP 7: EXTRACTING RELATION ONTOLOGY INFO"}
{"asctime": "2021-11-01 04:34:16,112", "levelname": "INFO", "name": "builds.data_preprocessing", "module": "data_preprocessing", "funcName": "processes_relation_ontology_data", "lineno": 1237, "message": "Creating Required Relations Ontology Data"}
{"asctime": "2021-11-01 04:34:17,955", "levelname": "INFO", "name": "builds.data_preprocessing", "module": "data_preprocessing", "funcName": "preprocesses_build_data", "lineno": 1655, "message": "STEP 8: CREATING CLINVAR VARIANT-DISEASE-PHENOTYPE DATA"}
{"asctime": "2021-11-01 04:34:17,955", "levelname": "INFO", "name": "builds.data_preprocessing", "module": "data_preprocessing", "funcName": "processes_clinvar_data", "lineno": 1270, "message": "Generating ClinVar Cross-Mapping Data"}
{"asctime": "2021-11-01 04:41:16,913", "levelname": "INFO", "name": "builds.data_preprocessing", "module": "data_preprocessing", "funcName": "preprocesses_build_data", "lineno": 1660, "message": "STEP 9: CREATING COFACTOR + CATALYST EDGE DATA"}
{"asctime": "2021-11-01 04:41:16,913", "levelname": "INFO", "name": "builds.data_preprocessing", "module": "data_preprocessing", "funcName": "processes_cofactor_catalyst_data", "lineno": 1294, "message": "Creating Protein-Cofactor and Protein-Catalyst Cross-Mappings"}
{"asctime": "2021-11-01 04:41:18,013", "levelname": "INFO", "name": "builds.data_preprocessing", "module": "data_preprocessing", "funcName": "preprocesses_build_data", "lineno": 1665, "message": "STEP 10: CREATING OBO-ONTOLOGY METADATA DICTIONARY"}

{"asctime": "2021-11-01 04:41:18,013", "levelname": "INFO", "name": "builds.data_preprocessing", "module": "data_preprocessing", "funcName": "creates_non_ontology_class_metadata_dict", "lineno": 1582, "message": "Creating Master Metadata Dictionary for Non-Ontology Entities"}
{"asctime": "2021-11-01 04:41:18,013", "levelname": "INFO", "name": "builds.data_preprocessing", "module": "data_preprocessing", "funcName": "_creates_gene_metadata_dict", "lineno": 1330, "message": "Generating Metadata for Gene Identifiers"}
{"asctime": "2021-11-01 04:41:39,478", "levelname": "INFO", "name": "builds.data_preprocessing", "module": "data_preprocessing", "funcName": "_creates_transcript_metadata_dict", "lineno": 1375, "message": "Generating Metadata for Transcript Identifiers"}
{"asctime": "2021-11-01 04:42:28,420", "levelname": "INFO", "name": "builds.data_preprocessing", "module": "data_preprocessing", "funcName": "_creates_variant_metadata_dict", "lineno": 1418, "message": "Generating Metadata for Variant IDs"}
{"asctime": "2021-11-01 04:46:58,371", "levelname": "INFO", "name": "builds.data_preprocessing", "module": "data_preprocessing", "funcName": "_creates_pathway_metadata_dict", "lineno": 1500, "message": "Generating Metadata for Pathway IDs"}
{"asctime": "2021-11-01 04:52:02,207", "levelname": "INFO", "name": "builds.data_preprocessing", "module": "data_preprocessing", "funcName": "_creates_relations_metadata_dict", "lineno": 1541, "message": "Generating Metadata for Relations IDs"}
{"asctime": "2021-11-01 04:52:37,865", "levelname": "INFO", "name": "builds.ontology_cleaning", "module": "ontology_cleaning", "funcName": "cleans_ontology_data", "lineno": 577, "message": "*** CLEANING INDIVIDUAL ONTOLOGY DATA SOURCES ***"}
{"asctime": "2021-11-01 04:52:37,866", "levelname": "INFO", "name": "builds.ontology_cleaning", "module": "ontology_cleaning", "funcName": "cleans_ontology_data", "lineno": 582, "message": "\nProcessing Ontology: CHEBI_WITH_IMPORTS.OWL"}
{"asctime": "2021-11-01 05:09:39,392", "levelname": "INFO", "name": "builds.ontology_cleaning", "module": "ontology_cleaning", "funcName": "updates_ontology_reporter", "lineno": 201, "message": "Obtaining Ontology Statistics"}
{"asctime": "2021-11-01 05:20:44,620", "levelname": "INFO", "name": "builds.ontology_cleaning", "module": "ontology_cleaning", "funcName": "fixes_ontology_parsing_errors", "lineno": 253, "message": "Finding Parsing Errors"}
{"asctime": "2021-11-01 05:22:09,099", "levelname": "INFO", "name": "builds.ontology_cleaning", "module": "ontology_cleaning", "funcName": "fixes_identifier_errors", "lineno": 278, "message": "Fixing Identifier Errors"}
{"asctime": "2021-11-01 05:23:00,468", "levelname": "INFO", "name": "builds.ontology_cleaning", "module": "ontology_cleaning", "funcName": "removes_deprecated_obsolete_entities", "lineno": 332, "message": "Removing Deprecated and Obsolete Classes"}
{"asctime": "2021-11-01 05:23:47,669", "levelname": "INFO", "name": "builds.ontology_cleaning", "module": "ontology_cleaning", "funcName": "fixes_punning_errors", "lineno": 362, "message": "Resolving Punning Errors"}
{"asctime": "2021-11-01 05:24:57,896", "levelname": "INFO", "name": "builds.ontology_cleaning", "module": "ontology_cleaning", "funcName": "_logically_verifies_cleaned_ontologies", "lineno": 178, "message": "PKT: Logically Verifying Ontology"}
{"asctime": "2021-11-01 05:30:47,755", "levelname": "INFO", "name": "builds.ontology_cleaning", "module": "ontology_cleaning", "funcName": "cleans_ontology_data", "lineno": 592, "message": "Reading in Cleaned Ontology -- Needed to Calculate Final Statistics"}
{"asctime": "2021-11-01 05:47:36,351", "levelname": "INFO", "name": "builds.ontology_cleaning", "module": "ontology_cleaning", "funcName": "updates_ontology_reporter", "lineno": 201, "message": "Obtaining Ontology Statistics"}
{"asctime": "2021-11-01 05:58:44,280", "levelname": "INFO", "name": "builds.ontology_cleaning", "module": "ontology_cleaning", "funcName": "cleans_ontology_data", "lineno": 582, "message": "\nProcessing Ontology: CLO_WITH_IMPORTS.OWL"}
{"asctime": "2021-11-01 06:02:04,699", "levelname": "INFO", "name": "builds.ontology_cleaning", "module": "ontology_cleaning", "funcName": "updates_ontology_reporter", "lineno": 201, "message": "Obtaining Ontology Statistics"}
{"asctime": "2021-11-01 06:04:33,690", "levelname": "INFO", "name": "builds.ontology_cleaning", "module": "ontology_cleaning", "funcName": "fixes_ontology_parsing_errors", "lineno": 253, "message": "Finding Parsing Errors"}
{"asctime": "2021-11-01 06:04:46,221", "levelname": "INFO", "name": "builds.ontology_cleaning", "module": "ontology_cleaning", "funcName": "fixes_identifier_errors", "lineno": 278, "message": "Fixing Identifier Errors"}
{"asctime": "2021-11-01 06:04:59,128", "levelname": "INFO", "name": "builds.ontology_cleaning", "module": "ontology_cleaning", "funcName": "removes_deprecated_obsolete_entities", "lineno": 332, "message": "Removing Deprecated and Obsolete Classes"}
{"asctime": "2021-11-01 06:05:09,062", "levelname": "INFO", "name": "builds.ontology_cleaning", "module": "ontology_cleaning", "funcName": "fixes_punning_errors", "lineno": 362, "message": "Resolving Punning Errors"}
{"asctime": "2021-11-01 06:05:32,673", "levelname": "INFO", "name": "builds.ontology_cleaning", "module": "ontology_cleaning", "funcName": "_logically_verifies_cleaned_ontologies", "lineno": 178, "message": "PKT: Logically Verifying Ontology"}
{"asctime": "2021-11-01 06:07:15,029", "levelname": "INFO", "name": "builds.ontology_cleaning", "module": "ontology_cleaning", "funcName": "cleans_ontology_data", "lineno": 592, "message": "Reading in Cleaned Ontology -- Needed to Calculate Final Statistics"}
{"asctime": "2021-11-01 06:10:43,709", "levelname": "INFO", "name": "builds.ontology_cleaning", "module": "ontology_cleaning", "funcName": "updates_ontology_reporter", "lineno": 201, "message": "Obtaining Ontology Statistics"}
{"asctime": "2021-11-01 06:13:10,116", "levelname": "INFO", "name": "builds.ontology_cleaning", "module": "ontology_cleaning", "funcName": "cleans_ontology_data", "lineno": 582, "message": "\nProcessing Ontology: EXT_WITH_IMPORTS.OWL"}
{"asctime": "2021-11-01 06:15:09,658", "levelname": "INFO", "name": "builds.ontology_cleaning", "module": "ontology_cleaning", "funcName": "updates_ontology_reporter", "lineno": 201, "message": "Obtaining Ontology Statistics"}
{"asctime": "2021-11-01 06:16:26,494", "levelname": "INFO", "name": "builds.ontology_cleaning", "module": "ontology_cleaning", "funcName": "fixes_ontology_parsing_errors", "lineno": 253, "message": "Finding Parsing Errors"}
{"asctime": "2021-11-01 06:16:39,857", "levelname": "INFO", "name": "builds.ontology_cleaning", "module": "ontology_cleaning", "funcName": "fixes_identifier_errors", "lineno": 278, "message": "Fixing Identifier Errors"}
{"asctime": "2021-11-01 06:16:47,098", "levelname": "INFO", "name": "builds.ontology_cleaning", "module": "ontology_cleaning", "funcName": "removes_deprecated_obsolete_entities", "lineno": 332, "message": "Removing Deprecated and Obsolete Classes"}
{"asctime": "2021-11-01 06:16:54,469", "levelname": "INFO", "name": "builds.ontology_cleaning", "module": "ontology_cleaning", "funcName": "fixes_punning_errors", "lineno": 362, "message": "Resolving Punning Errors"}
{"asctime": "2021-11-01 06:17:05,375", "levelname": "INFO", "name": "builds.ontology_cleaning", "module": "ontology_cleaning", "funcName": "_logically_verifies_cleaned_ontologies", "lineno": 178, "message": "PKT: Logically Verifying Ontology"}
{"asctime": "2021-11-01 06:17:59,074", "levelname": "INFO", "name": "builds.ontology_cleaning", "module": "ontology_cleaning", "funcName": "cleans_ontology_data", "lineno": 592, "message": "Reading in Cleaned Ontology -- Needed to Calculate Final Statistics"}
{"asctime": "2021-11-01 06:19:57,152", "levelname": "INFO", "name": "builds.ontology_cleaning", "module": "ontology_cleaning", "funcName": "updates_ontology_reporter", "lineno": 201, "message": "Obtaining Ontology Statistics"}
{"asctime": "2021-11-01 06:21:12,617", "levelname": "INFO", "name": "builds.ontology_cleaning", "module": "ontology_cleaning", "funcName": "cleans_ontology_data", "lineno": 582, "message": "\nProcessing Ontology: GO_WITH_IMPORTS.OWL"}
{"asctime": "2021-11-01 06:24:56,179", "levelname": "INFO", "name": "builds.ontology_cleaning", "module": "ontology_cleaning", "funcName": "updates_ontology_reporter", "lineno": 201, "message": "Obtaining Ontology Statistics"}
{"asctime": "2021-11-01 06:27:23,258", "levelname": "INFO", "name": "builds.ontology_cleaning", "module": "ontology_cleaning", "funcName": "fixes_ontology_parsing_errors", "lineno": 253, "message": "Finding Parsing Errors"}
{"asctime": "2021-11-01 06:27:46,878", "levelname": "INFO", "name": "builds.ontology_cleaning", "module": "ontology_cleaning", "funcName": "fixes_identifier_errors", "lineno": 278, "message": "Fixing Identifier Errors"}
{"asctime": "2021-11-01 06:28:00,501", "levelname": "INFO", "name": "builds.ontology_cleaning", "module": "ontology_cleaning", "funcName": "removes_deprecated_obsolete_entities", "lineno": 332, "message": "Removing Deprecated and Obsolete Classes"}
{"asctime": "2021-11-01 06:28:23,759", "levelname": "INFO", "name": "builds.ontology_cleaning", "module": "ontology_cleaning", "funcName": "fixes_punning_errors", "lineno": 362, "message": "Resolving Punning Errors"}
{"asctime": "2021-11-01 06:28:42,579", "levelname": "INFO", "name": "builds.ontology_cleaning", "module": "ontology_cleaning", "funcName": "_logically_verifies_cleaned_ontologies", "lineno": 178, "message": "PKT: Logically Verifying Ontology"}
{"asctime": "2021-11-01 06:30:12,230", "levelname": "INFO", "name": "builds.ontology_cleaning", "module": "ontology_cleaning", "funcName": "cleans_ontology_data", "lineno": 592, "message": "Reading in Cleaned Ontology -- Needed to Calculate Final Statistics"}
{"asctime": "2021-11-01 06:33:40,575", "levelname": "INFO", "name": "builds.ontology_cleaning", "module": "ontology_cleaning", "funcName": "updates_ontology_reporter", "lineno": 201, "message": "Obtaining Ontology Statistics"}
{"asctime": "2021-11-01 06:36:01,584", "levelname": "INFO", "name": "builds.ontology_cleaning", "module": "ontology_cleaning", "funcName": "cleans_ontology_data", "lineno": 582, "message": "\nProcessing Ontology: HP_WITH_IMPORTS.OWL"}
{"asctime": "2021-11-01 06:38:29,033", "levelname": "INFO", "name": "builds.ontology_cleaning", "module": "ontology_cleaning", "funcName": "updates_ontology_reporter", "lineno": 201, "message": "Obtaining Ontology Statistics"}
{"asctime": "2021-11-01 06:40:09,978", "levelname": "INFO", "name": "builds.ontology_cleaning", "module": "ontology_cleaning", "funcName": "fixes_ontology_parsing_errors", "lineno": 253, "message": "Finding Parsing Errors"}
{"asctime": "2021-11-01 06:40:25,302", "levelname": "INFO", "name": "builds.ontology_cleaning", "module": "ontology_cleaning", "funcName": "fixes_identifier_errors", "lineno": 278, "message": "Fixing Identifier Errors"}
{"asctime": "2021-11-01 06:40:34,105", "levelname": "INFO", "name": "builds.ontology_cleaning", "module": "ontology_cleaning", "funcName": "removes_deprecated_obsolete_entities", "lineno": 332, "message": "Removing Deprecated and Obsolete Classes"}
{"asctime": "2021-11-01 06:40:41,258", "levelname": "INFO", "name": "builds.ontology_cleaning", "module": "ontology_cleaning", "funcName": "fixes_punning_errors", "lineno": 362, "message": "Resolving Punning Errors"}
{"asctime": "2021-11-01 06:40:55,062", "levelname": "INFO", "name": "builds.ontology_cleaning", "module": "ontology_cleaning", "funcName": "_logically_verifies_cleaned_ontologies", "lineno": 178, "message": "PKT: Logically Verifying Ontology"}
{"asctime": "2021-11-01 06:42:02,123", "levelname": "INFO", "name": "builds.ontology_cleaning", "module": "ontology_cleaning", "funcName": "cleans_ontology_data", "lineno": 592, "message": "Reading in Cleaned Ontology -- Needed to Calculate Final Statistics"}
{"asctime": "2021-11-01 06:44:32,077", "levelname": "INFO", "name": "builds.ontology_cleaning", "module": "ontology_cleaning", "funcName": "updates_ontology_reporter", "lineno": 201, "message": "Obtaining Ontology Statistics"}
{"asctime": "2021-11-01 06:46:13,544", "levelname": "INFO", "name": "builds.ontology_cleaning", "module": "ontology_cleaning", "funcName": "cleans_ontology_data", "lineno": 582, "message": "\nProcessing Ontology: MONDO_WITH_IMPORTS.OWL"}
{"asctime": "2021-11-01 06:52:57,054", "levelname": "INFO", "name": "builds.ontology_cleaning", "module": "ontology_cleaning", "funcName": "updates_ontology_reporter", "lineno": 201, "message": "Obtaining Ontology Statistics"}
{"asctime": "2021-11-01 06:57:17,692", "levelname": "INFO", "name": "builds.ontology_cleaning", "module": "ontology_cleaning", "funcName": "fixes_ontology_parsing_errors", "lineno": 253, "message": "Finding Parsing Errors"}
{"asctime": "2021-11-01 06:57:59,960", "levelname": "INFO", "name": "builds.ontology_cleaning", "module": "ontology_cleaning", "funcName": "fixes_identifier_errors", "lineno": 278, "message": "Fixing Identifier Errors"}
{"asctime": "2021-11-01 06:58:22,654", "levelname": "INFO", "name": "builds.ontology_cleaning", "module": "ontology_cleaning", "funcName": "removes_deprecated_obsolete_entities", "lineno": 332, "message": "Removing Deprecated and Obsolete Classes"}
{"asctime": "2021-11-01 06:58:45,214", "levelname": "INFO", "name": "builds.ontology_cleaning", "module": "ontology_cleaning", "funcName": "fixes_punning_errors", "lineno": 362, "message": "Resolving Punning Errors"}
{"asctime": "2021-11-01 06:59:18,777", "levelname": "INFO", "name": "builds.ontology_cleaning", "module": "ontology_cleaning", "funcName": "_logically_verifies_cleaned_ontologies", "lineno": 178, "message": "PKT: Logically Verifying Ontology"}
{"asctime": "2021-11-01 07:01:56,990", "levelname": "INFO", "name": "builds.ontology_cleaning", "module": "ontology_cleaning", "funcName": "cleans_ontology_data", "lineno": 592, "message": "Reading in Cleaned Ontology -- Needed to Calculate Final Statistics"}
{"asctime": "2021-11-01 07:08:40,760", "levelname": "INFO", "name": "builds.ontology_cleaning", "module": "ontology_cleaning", "funcName": "updates_ontology_reporter", "lineno": 201, "message": "Obtaining Ontology Statistics"}
{"asctime": "2021-11-01 07:13:09,817", "levelname": "INFO", "name": "builds.ontology_cleaning", "module": "ontology_cleaning", "funcName": "cleans_ontology_data", "lineno": 582, "message": "\nProcessing Ontology: PR_WITH_IMPORTS.OWL"}
{"asctime": "2021-11-01 07:19:05,834", "levelname": "INFO", "name": "builds.ontology_cleaning", "module": "ontology_cleaning", "funcName": "updates_ontology_reporter", "lineno": 201, "message": "Obtaining Ontology Statistics"}
{"asctime": "2021-11-01 07:23:36,231", "levelname": "INFO", "name": "builds.ontology_cleaning", "module": "ontology_cleaning", "funcName": "fixes_ontology_parsing_errors", "lineno": 253, "message": "Finding Parsing Errors"}
{"asctime": "2021-11-01 07:24:09,599", "levelname": "INFO", "name": "builds.ontology_cleaning", "module": "ontology_cleaning", "funcName": "fixes_identifier_errors", "lineno": 278, "message": "Fixing Identifier Errors"}
{"asctime": "2021-11-01 07:24:31,041", "levelname": "INFO", "name": "builds.ontology_cleaning", "module": "ontology_cleaning", "funcName": "removes_deprecated_obsolete_entities", "lineno": 332, "message": "Removing Deprecated and Obsolete Classes"}
{"asctime": "2021-11-01 07:24:47,202", "levelname": "INFO", "name": "builds.ontology_cleaning", "module": "ontology_cleaning", "funcName": "fixes_punning_errors", "lineno": 362, "message": "Resolving Punning Errors"}
{"asctime": "2021-11-01 07:25:18,307", "levelname": "INFO", "name": "builds.ontology_cleaning", "module": "ontology_cleaning", "funcName": "_logically_verifies_cleaned_ontologies", "lineno": 178, "message": "PKT: Logically Verifying Ontology"}
{"asctime": "2021-11-01 07:27:46,337", "levelname": "INFO", "name": "builds.ontology_cleaning", "module": "ontology_cleaning", "funcName": "cleans_ontology_data", "lineno": 592, "message": "Reading in Cleaned Ontology -- Needed to Calculate Final Statistics"}
{"asctime": "2021-11-01 07:33:46,590", "levelname": "INFO", "name": "builds.ontology_cleaning", "module": "ontology_cleaning", "funcName": "updates_ontology_reporter", "lineno": 201, "message": "Obtaining Ontology Statistics"}
{"asctime": "2021-11-01 07:38:14,914", "levelname": "INFO", "name": "builds.ontology_cleaning", "module": "ontology_cleaning", "funcName": "cleans_ontology_data", "lineno": 582, "message": "\nProcessing Ontology: PW_WITH_IMPORTS.OWL"}

{"asctime": "2021-11-01 07:38:21,747", "levelname": "INFO", "name": "builds.ontology_cleaning", "module": "ontology_cleaning", "funcName": "updates_ontology_reporter", "lineno": 201, "message": "Obtaining Ontology Statistics"}
{"asctime": "2021-11-01 07:38:26,004", "levelname": "INFO", "name": "builds.ontology_cleaning", "module": "ontology_cleaning", "funcName": "fixes_ontology_parsing_errors", "lineno": 253, "message": "Finding Parsing Errors"}
{"asctime": "2021-11-01 07:38:28,882", "levelname": "INFO", "name": "builds.ontology_cleaning", "module": "ontology_cleaning", "funcName": "fixes_identifier_errors", "lineno": 278, "message": "Fixing Identifier Errors"}
{"asctime": "2021-11-01 07:38:29,205", "levelname": "INFO", "name": "builds.ontology_cleaning", "module": "ontology_cleaning", "funcName": "removes_deprecated_obsolete_entities", "lineno": 332, "message": "Removing Deprecated and Obsolete Classes"}
{"asctime": "2021-11-01 07:38:29,488", "levelname": "INFO", "name": "builds.ontology_cleaning", "module": "ontology_cleaning", "funcName": "fixes_punning_errors", "lineno": 362, "message": "Resolving Punning Errors"}
{"asctime": "2021-11-01 07:38:29,933", "levelname": "INFO", "name": "builds.ontology_cleaning", "module": "ontology_cleaning", "funcName": "_logically_verifies_cleaned_ontologies", "lineno": 178, "message": "PKT: Logically Verifying Ontology"}
{"asctime": "2021-11-01 07:38:35,443", "levelname": "INFO", "name": "builds.ontology_cleaning", "module": "ontology_cleaning", "funcName": "cleans_ontology_data", "lineno": 592, "message": "Reading in Cleaned Ontology -- Needed to Calculate Final Statistics"}
{"asctime": "2021-11-01 07:38:41,903", "levelname": "INFO", "name": "builds.ontology_cleaning", "module": "ontology_cleaning", "funcName": "updates_ontology_reporter", "lineno": 201, "message": "Obtaining Ontology Statistics"}
{"asctime": "2021-11-01 07:38:46,242", "levelname": "INFO", "name": "builds.ontology_cleaning", "module": "ontology_cleaning", "funcName": "cleans_ontology_data", "lineno": 582, "message": "\nProcessing Ontology: RO_WITH_IMPORTS.OWL"}
{"asctime": "2021-11-01 07:38:47,661", "levelname": "INFO", "name": "builds.ontology_cleaning", "module": "ontology_cleaning", "funcName": "updates_ontology_reporter", "lineno": 201, "message": "Obtaining Ontology Statistics"}
{"asctime": "2021-11-01 07:38:48,418", "levelname": "INFO", "name": "builds.ontology_cleaning", "module": "ontology_cleaning", "funcName": "fixes_ontology_parsing_errors", "lineno": 253, "message": "Finding Parsing Errors"}
{"asctime": "2021-11-01 07:38:50,807", "levelname": "INFO", "name": "builds.ontology_cleaning", "module": "ontology_cleaning", "funcName": "fixes_identifier_errors", "lineno": 278, "message": "Fixing Identifier Errors"}
{"asctime": "2021-11-01 07:38:50,872", "levelname": "INFO", "name": "builds.ontology_cleaning", "module": "ontology_cleaning", "funcName": "removes_deprecated_obsolete_entities", "lineno": 332, "message": "Removing Deprecated and Obsolete Classes"}
{"asctime": "2021-11-01 07:38:50,935", "levelname": "INFO", "name": "builds.ontology_cleaning", "module": "ontology_cleaning", "funcName": "fixes_punning_errors", "lineno": 362, "message": "Resolving Punning Errors"}
{"asctime": "2021-11-01 07:38:51,053", "levelname": "INFO", "name": "builds.ontology_cleaning", "module": "ontology_cleaning", "funcName": "_logically_verifies_cleaned_ontologies", "lineno": 178, "message": "PKT: Logically Verifying Ontology"}
{"asctime": "2021-11-01 07:38:54,246", "levelname": "INFO", "name": "builds.ontology_cleaning", "module": "ontology_cleaning", "funcName": "cleans_ontology_data", "lineno": 592, "message": "Reading in Cleaned Ontology -- Needed to Calculate Final Statistics"}
{"asctime": "2021-11-01 07:38:55,667", "levelname": "INFO", "name": "builds.ontology_cleaning", "module": "ontology_cleaning", "funcName": "updates_ontology_reporter", "lineno": 201, "message": "Obtaining Ontology Statistics"}
{"asctime": "2021-11-01 07:38:56,574", "levelname": "INFO", "name": "builds.ontology_cleaning", "module": "ontology_cleaning", "funcName": "cleans_ontology_data", "lineno": 582, "message": "\nProcessing Ontology: SO_WITH_IMPORTS.OWL"}
{"asctime": "2021-11-01 07:39:04,809", "levelname": "INFO", "name": "builds.ontology_cleaning", "module": "ontology_cleaning", "funcName": "updates_ontology_reporter", "lineno": 201, "message": "Obtaining Ontology Statistics"}
{"asctime": "2021-11-01 07:39:16,320", "levelname": "INFO", "name": "builds.ontology_cleaning", "module": "ontology_cleaning", "funcName": "fixes_ontology_parsing_errors", "lineno": 253, "message": "Finding Parsing Errors"}
{"asctime": "2021-11-01 07:39:19,403", "levelname": "INFO", "name": "builds.ontology_cleaning", "module": "ontology_cleaning", "funcName": "fixes_identifier_errors", "lineno": 278, "message": "Fixing Identifier Errors"}
{"asctime": "2021-11-01 07:39:19,809", "levelname": "INFO", "name": "builds.ontology_cleaning", "module": "ontology_cleaning", "funcName": "removes_deprecated_obsolete_entities", "lineno": 332, "message": "Removing Deprecated and Obsolete Classes"}
{"asctime": "2021-11-01 07:39:20,357", "levelname": "INFO", "name": "builds.ontology_cleaning", "module": "ontology_cleaning", "funcName": "fixes_punning_errors", "lineno": 362, "message": "Resolving Punning Errors"}
{"asctime": "2021-11-01 07:39:20,914", "levelname": "INFO", "name": "builds.ontology_cleaning", "module": "ontology_cleaning", "funcName": "_logically_verifies_cleaned_ontologies", "lineno": 178, "message": "PKT: Logically Verifying Ontology"}
{"asctime": "2021-11-01 07:39:27,169", "levelname": "INFO", "name": "builds.ontology_cleaning", "module": "ontology_cleaning", "funcName": "cleans_ontology_data", "lineno": 592, "message": "Reading in Cleaned Ontology -- Needed to Calculate Final Statistics"}
{"asctime": "2021-11-01 07:39:34,919", "levelname": "INFO", "name": "builds.ontology_cleaning", "module": "ontology_cleaning", "funcName": "updates_ontology_reporter", "lineno": 201, "message": "Obtaining Ontology Statistics"}
{"asctime": "2021-11-01 07:39:39,877", "levelname": "INFO", "name": "builds.ontology_cleaning", "module": "ontology_cleaning", "funcName": "cleans_ontology_data", "lineno": 582, "message": "\nProcessing Ontology: VO_WITH_IMPORTS.OWL"}
{"asctime": "2021-11-01 07:39:55,084", "levelname": "INFO", "name": "builds.ontology_cleaning", "module": "ontology_cleaning", "funcName": "updates_ontology_reporter", "lineno": 201, "message": "Obtaining Ontology Statistics"}
{"asctime": "2021-11-01 07:40:04,778", "levelname": "INFO", "name": "builds.ontology_cleaning", "module": "ontology_cleaning", "funcName": "fixes_ontology_parsing_errors", "lineno": 253, "message": "Finding Parsing Errors"}
{"asctime": "2021-11-01 07:40:08,518", "levelname": "INFO", "name": "builds.ontology_cleaning", "module": "ontology_cleaning", "funcName": "fixes_identifier_errors", "lineno": 278, "message": "Fixing Identifier Errors"}
{"asctime": "2021-11-01 07:40:09,438", "levelname": "INFO", "name": "builds.ontology_cleaning", "module": "ontology_cleaning", "funcName": "removes_deprecated_obsolete_entities", "lineno": 332, "message": "Removing Deprecated and Obsolete Classes"}
{"asctime": "2021-11-01 07:40:10,082", "levelname": "INFO", "name": "builds.ontology_cleaning", "module": "ontology_cleaning", "funcName": "fixes_punning_errors", "lineno": 362, "message": "Resolving Punning Errors"}
{"asctime": "2021-11-01 07:40:11,394", "levelname": "INFO", "name": "builds.ontology_cleaning", "module": "ontology_cleaning", "funcName": "_logically_verifies_cleaned_ontologies", "lineno": 178, "message": "PKT: Logically Verifying Ontology"}
{"asctime": "2021-11-01 07:40:22,176", "levelname": "INFO", "name": "builds.ontology_cleaning", "module": "ontology_cleaning", "funcName": "cleans_ontology_data", "lineno": 592, "message": "Reading in Cleaned Ontology -- Needed to Calculate Final Statistics"}
{"asctime": "2021-11-01 07:40:37,907", "levelname": "INFO", "name": "builds.ontology_cleaning", "module": "ontology_cleaning", "funcName": "updates_ontology_reporter", "lineno": 201, "message": "Obtaining Ontology Statistics"}
{"asctime": "2021-11-01 07:40:48,694", "levelname": "INFO", "name": "builds.ontology_cleaning", "module": "ontology_cleaning", "funcName": "cleans_ontology_data", "lineno": 598, "message": "*** CLEANING MERGED ONTOLOGY DATA ***"}
{"asctime": "2021-11-01 07:40:48,695", "levelname": "INFO", "name": "builds.ontology_cleaning", "module": "ontology_cleaning", "funcName": "merge_ontologies", "lineno": 161, "message": "Merging Ontologies: vo_with_imports.owl, so_with_imports.owl"}
{"asctime": "2021-11-01 07:40:55,252", "levelname": "INFO", "name": "builds.ontology_cleaning", "module": "ontology_cleaning", "funcName": "merge_ontologies", "lineno": 161, "message": "Merging Ontologies: ro_with_imports.owl, PheKnowLator_MergedOntologies.owl"}
{"asctime": "2021-11-01 07:41:01,303", "levelname": "INFO", "name": "builds.ontology_cleaning", "module": "ontology_cleaning", "funcName": "merge_ontologies", "lineno": 161, "message": "Merging Ontologies: pw_with_imports.owl, PheKnowLator_MergedOntologies.owl"}
{"asctime": "2021-11-01 07:41:09,946", "levelname": "INFO", "name": "builds.ontology_cleaning", "module": "ontology_cleaning", "funcName": "merge_ontologies", "lineno": 161, "message": "Merging Ontologies: pr_with_imports.owl, PheKnowLator_MergedOntologies.owl"}
{"asctime": "2021-11-01 07:41:55,436", "levelname": "INFO", "name": "builds.ontology_cleaning", "module": "ontology_cleaning", "funcName": "merge_ontologies", "lineno": 161, "message": "Merging Ontologies: mondo_with_imports.owl, PheKnowLator_MergedOntologies.owl"}
{"asctime": "2021-11-01 07:43:32,816", "levelname": "INFO", "name": "builds.ontology_cleaning", "module": "ontology_cleaning", "funcName": "merge_ontologies", "lineno": 161, "message": "Merging Ontologies: hp_with_imports.owl, PheKnowLator_MergedOntologies.owl"}
{"asctime": "2021-11-01 07:45:20,567", "levelname": "INFO", "name": "builds.ontology_cleaning", "module": "ontology_cleaning", "funcName": "merge_ontologies", "lineno": 161, "message": "Merging Ontologies: go_with_imports.owl, PheKnowLator_MergedOntologies.owl"}
{"asctime": "2021-11-01 07:47:28,284", "levelname": "INFO", "name": "builds.ontology_cleaning", "module": "ontology_cleaning", "funcName": "merge_ontologies", "lineno": 161, "message": "Merging Ontologies: ext_with_imports.owl, PheKnowLator_MergedOntologies.owl"}
{"asctime": "2021-11-01 07:49:47,810", "levelname": "INFO", "name": "builds.ontology_cleaning", "module": "ontology_cleaning", "funcName": "merge_ontologies", "lineno": 161, "message": "Merging Ontologies: clo_with_imports.owl, PheKnowLator_MergedOntologies.owl"}
{"asctime": "2021-11-01 07:52:32,133", "levelname": "INFO", "name": "builds.ontology_cleaning", "module": "ontology_cleaning", "funcName": "merge_ontologies", "lineno": 161, "message": "Merging Ontologies: chebi_with_imports.owl, PheKnowLator_MergedOntologies.owl"}
{"asctime": "2021-11-01 07:56:49,039", "levelname": "INFO", "name": "builds.ontology_cleaning", "module": "ontology_cleaning", "funcName": "cleans_ontology_data", "lineno": 604, "message": "Loading Merged Ontology"}
{"asctime": "2021-11-01 08:37:18,196", "levelname": "INFO", "name": "builds.ontology_cleaning", "module": "ontology_cleaning", "funcName": "updates_ontology_reporter", "lineno": 201, "message": "Obtaining Ontology Statistics"}
{"asctime": "2021-11-01 09:07:05,918", "levelname": "INFO", "name": "builds.ontology_cleaning", "module": "ontology_cleaning", "funcName": "fixes_identifier_errors", "lineno": 278, "message": "Fixing Identifier Errors"}
{"asctime": "2021-11-01 09:09:38,640", "levelname": "INFO", "name": "builds.ontology_cleaning", "module": "ontology_cleaning", "funcName": "normalizes_duplicate_classes", "lineno": 411, "message": "Normalizing Duplicate Concepts"}
{"asctime": "2021-11-01 09:09:38,641", "levelname": "INFO", "name": "builds.ontology_cleaning", "module": "ontology_cleaning", "funcName": "normalizes_existing_classes", "lineno": 444, "message": "Normalizing Existing Classes"}
{"asctime": "2021-11-01 09:10:10,035", "levelname": "INFO", "name": "builds.ontology_cleaning", "module": "ontology_cleaning", "funcName": "fixes_punning_errors", "lineno": 362, "message": "Resolving Punning Errors"}
{"asctime": "2021-11-01 09:13:36,293", "levelname": "INFO", "name": "builds.ontology_cleaning", "module": "ontology_cleaning", "funcName": "updates_ontology_reporter", "lineno": 201, "message": "Obtaining Ontology Statistics"}
{"asctime": "2021-11-01 09:57:17,472", "levelname": "INFO", "name": "builds.ontology_cleaning", "module": "ontology_cleaning", "funcName": "cleans_ontology_data", "lineno": 619, "message": "*** GENERATING ONTOLOGY CLEANING REPORT ***"}
{"asctime": "2021-11-01 09:57:17,931", "levelname": "INFO", "name": "builds.build_phase_2", "module": "build_phase_2", "funcName": "run_phase_2", "lineno": 165, "message": "Generating and Writing Preprocessed Data Metadata"}
{"asctime": "2021-11-01 10:00:03,648", "levelname": "INFO", "name": "builds.build_phase_2", "module": "build_phase_2", "funcName": "run_phase_2", "lineno": 181, "message": "Updating Input Dependency Documents"}
{"asctime": "2021-11-01 10:00:04,945", "levelname": "INFO", "name": "builds.build_phase_2", "module": "build_phase_2", "funcName": "run_phase_2", "lineno": 196, "message": "Uploading Processed Data to the temp_build_inprogress/data Directory"}
{"asctime": "2021-11-01 10:00:04,972", "levelname": "INFO", "name": "builds.build_phase_2", "module": "build_phase_2", "funcName": "run_phase_2", "lineno": 200, "message": "Copying Data FROM: archived_builds/release_v3.0.2/build_01NOV2021/data/ TO: temp_build_inprogress/data/"}
{"asctime": "2021-11-01 10:00:16,536", "levelname": "INFO", "name": "__main__", "module": "phases1_2_entrypoint", "funcName": "main", "lineno": 60, "message": " COMPLETED BUILD PHASES 1-2: 525.81 MINUTES "}
{"asctime": "2021-11-01 10:00:16,536", "levelname": "INFO", "name": "__main__", "module": "phases1_2_entrypoint", "funcName": "main", "lineno": 61, "message": "EXIT BUILD PHASES 1-2"}

## Supplementary Document 7. pkt_build_log.log (Knowledge Graph Construction).

{"asctime": "2021-11-02 01:02:08,692", "levelname": "INFO", "name": "__main__", "module": "build_phase_3", "funcName": "main", "lineno": 72, "message": "#####\nBUILD PHASE 3: DATA PRE-PROCESSING\n#####"}
{"asctime": "2021-11-02 01:02:08,693", "levelname": "INFO", "name": "__main__", "module": "build_phase_3", "funcName": "main", "lineno": 77, "message": "STEP 1: INITIALIZE GCS BUCKET AND REFORMAT INPUT ARGUMENTS"}
{"asctime": "2021-11-02 01:02:08,985", "levelname": "INFO", "name": "__main__", "module": "build_phase_3", "funcName": "main", "lineno": 105, "message": "STEP 2: CONSTRUCT KNOWLEDGE GRAPH"}
{"asctime": "2021-11-02 01:02:08,985", "levelname": "INFO", "name": "__main__", "module": "build_phase_3", "funcName": "main", "lineno": 105, "message": "KG Build: subclass + relations_only.txt"}
{"asctime": "2021-11-02 01:02:12,818", "levelname": "INFO", "name": "pkt_kg.downloads", "module": "downloads", "funcName": "__init__", "lineno": 59, "message": "**********PKT STEP: DOWNLOADING KNOWLEDGE GRAPH DATA**********"}
{"asctime": "2021-11-02 01:02:12,821", "levelname": "INFO", "name": "pkt_kg.downloads", "module": "downloads", "funcName": "downloads_data_from_url", "lineno": 279, "message": "***Downloading Data: ontology_source_list to \"resources/ontologies/\" ***"}
{"asctime": "2021-11-02 01:02:12,830", "levelname": "INFO", "name": "pkt_kg.downloads", "module": "downloads", "funcName": "downloads_data_from_url", "lineno": 284, "message": "Downloading: hp_with_imports"}
{"asctime": "2021-11-02 01:02:28,895", "levelname": "INFO", "name": "pkt_kg.downloads", "module": "downloads", "funcName": "downloads_data_from_url", "lineno": 300, "message": "The knowledge graph contains 27169 classes, 341782 axioms, 256 object properties, and 0 individuals"}
{"asctime": "2021-11-02 01:02:28,896", "levelname": "INFO", "name": "pkt_kg.downloads", "module": "downloads", "funcName": "downloads_data_from_url", "lineno": 284, "message": "Downloading: go_with_imports"}
{"asctime": "2021-11-02 01:02:50,145", "levelname": "INFO", "name": "pkt_kg.downloads", "module": "downloads", "funcName": "downloads_data_from_url", "lineno": 300, "message": "The knowledge graph contains 43832 classes, 509854 axioms, 9 object properties, and 0 individuals"}
{"asctime": "2021-11-02 01:02:50,146", "levelname": "INFO", "name": "pkt_kg.downloads", "module": "downloads", "funcName": "downloads_data_from_url", "lineno": 284, "message": "Downloading: mondo_with_imports"}
{"asctime": "2021-11-02 01:03:26,234", "levelname": "INFO", "name": "pkt_kg.downloads", "module": "downloads", "funcName": "downloads_data_from_url", "lineno": 300, "message": "The knowledge graph contains 40975 classes, 612772 axioms, 338 object properties, and 17 individuals"}
{"asctime": "2021-11-02 01:03:26,235", "levelname": "INFO", "name": "pkt_kg.downloads", "module": "downloads", "funcName": "downloads_data_from_url", "lineno": 284, "message": "Downloading: vo_with_imports"}
{"asctime": "2021-11-02 01:03:30,130", "levelname": "INFO", "name": "pkt_kg.downloads", "module": "downloads", "funcName": "downloads_data_from_url", "lineno": 300, "message": "The knowledge graph contains 6825 classes, 62120 axioms, 232 object properties, and 167 individuals"}
{"asctime": "2021-11-02 01:03:30,131", "levelname": "INFO", "name": "pkt_kg.downloads", "module": "downloads", "funcName": "downloads_data_from_url", "lineno": 284, "message": "Downloading: chebi_with_imports"}
{"asctime": "2021-11-02 01:04:33,057", "levelname": "INFO", "name": "pkt_kg.downloads", "module": "downloads", "funcName": "downloads_data_from_url", "lineno": 300, "message": "The knowledge graph contains 150080 classes, 2719571 axioms, 10 object properties, and 0 individuals"}
{"asctime": "2021-11-02 01:04:33,058", "levelname": "INFO", "name": "pkt_kg.downloads", "module": "downloads", "funcName": "downloads_data_from_url", "lineno": 284, "message": "Downloading: ext_with_imports"}
{"asctime": "2021-11-02 01:04:47,341", "levelname": "INFO", "name": "pkt_kg.downloads", "module": "downloads", "funcName": "downloads_data_from_url", "lineno": 300, "message": "The knowledge graph contains 19096 classes, 266646 axioms, 239 object properties, and 0 individuals"}
{"asctime": "2021-11-02 01:04:47,342", "levelname": "INFO", "name": "pkt_kg.downloads", "module": "downloads", "funcName": "downloads_data_from_url", "lineno": 284, "message": "Downloading: clo_with_imports"}
{"asctime": "2021-11-02 01:05:07,293", "levelname": "INFO", "name": "pkt_kg.downloads", "module": "downloads", "funcName": "downloads_data_from_url", "lineno": 300, "message": "The knowledge graph contains 44858 classes, 548206 axioms, 112 object properties, and 33 individuals"}
{"asctime": "2021-11-02 01:05:07,294", "levelname": "INFO", "name": "pkt_kg.downloads", "module": "downloads", "funcName": "downloads_data_from_url", "lineno": 284, "message": "Downloading: pr_with_imports"}
{"asctime": "2021-11-02 01:05:34,557", "levelname": "INFO", "name": "pkt_kg.downloads", "module": "downloads", "funcName": "downloads_data_from_url", "lineno": 300, "message": "The knowledge graph contains 117081 classes, 1385427 axioms, 12 object properties, and 0 individuals"}
{"asctime": "2021-11-02 01:05:34,558", "levelname": "INFO", "name": "pkt_kg.downloads", "module": "downloads", "funcName": "downloads_data_from_url", "lineno": 284, "message": "Downloading: so_with_imports"}
{"asctime": "2021-11-02 01:05:37,171", "levelname": "INFO", "name": "pkt_kg.downloads", "module": "downloads", "funcName": "downloads_data_from_url", "lineno": 300, "message": "The knowledge graph contains 2363 classes, 23204 axioms, 50 object properties, and 0 individuals"}
{"asctime": "2021-11-02 01:05:37,171", "levelname": "INFO", "name": "pkt_kg.downloads", "module": "downloads", "funcName": "downloads_data_from_url", "lineno": 284, "message": "Downloading: pw_with_imports"}
{"asctime": "2021-11-02 01:05:39,867", "levelname": "INFO", "name": "pkt_kg.downloads", "module": "downloads", "funcName": "downloads_data_from_url", "lineno": 300, "message": "The knowledge graph contains 2600 classes, 21868 axioms, 1 object properties, and 0 individuals"}
{"asctime": "2021-11-02 01:05:39,868", "levelname": "INFO", "name": "pkt_kg.downloads", "module": "downloads", "funcName": "downloads_data_from_url", "lineno": 284, "message": "Downloading: ro_with_imports"}
{"asctime": "2021-11-02 01:05:41,962", "levelname": "INFO", "name": "pkt_kg.downloads", "module": "downloads", "funcName": "downloads_data_from_url", "lineno": 300, "message": "The knowledge graph contains 69 classes, 5823 axioms, 600 object properties, and 5 individuals"}
{"asctime": "2021-11-02 01:05:41,964", "levelname": "INFO", "name": "pkt_kg.downloads", "module": "downloads", "funcName": "generates_source_metadata", "lineno": 198, "message": "*** Generating Metadata ***"}
{"asctime": "2021-11-02 01:05:42,124", "levelname": "INFO", "name": "pkt_kg.downloads", "module": "downloads", "funcName": "__init__", "lineno": 59, "message": "**********PKT STEP: DOWNLOADING KNOWLEDGE GRAPH DATA**********"}
{"asctime": "2021-11-02 01:05:42,128", "levelname": "INFO", "name": "pkt_kg.downloads", "module": "downloads", "funcName": "downloads_data_from_url", "lineno": 348, "message": "*** Downloading Data: edge_source_list to \"resources/edge_data/\" ***"}
{"asctime": "2021-11-02 01:05:42,129", "levelname": "INFO", "name": "pkt_kg.downloads", "module": "downloads", "funcName": "downloads_data_from_url", "lineno": 353, "message": "Edge: chemical-disease"}
{"asctime": "2021-11-02 01:05:46,847", "levelname": "INFO", "name": "pkt_kg.downloads", "module": "downloads", "funcName": "downloads_data_from_url", "lineno": 353, "message": "Edge: chemical-gene"}
{"asctime": "2021-11-02 01:05:50,002", "levelname": "INFO", "name": "pkt_kg.downloads", "module": "downloads", "funcName": "downloads_data_from_url", "lineno": 353, "message": "Edge: chemical-gobp"}
{"asctime": "2021-11-02 01:05:55,511", "levelname": "INFO", "name": "pkt_kg.downloads", "module": "downloads", "funcName": "downloads_data_from_url", "lineno": 353, "message": "Edge: chemical-gocc"}
{"asctime": "2021-11-02 01:05:56,544", "levelname": "INFO", "name": "pkt_kg.downloads", "module": "downloads", "funcName": "downloads_data_from_url", "lineno": 353, "message": "Edge: chemical-gomf"}
{"asctime": "2021-11-02 01:05:57,597", "levelname": "INFO", "name": "pkt_kg.downloads", "module": "downloads", "funcName": "downloads_data_from_url", "lineno": 353, "message": "Edge: chemical-pathway"}
{"asctime": "2021-11-02 01:05:57,852", "levelname": "INFO", "name": "pkt_kg.downloads", "module": "downloads", "funcName": "downloads_data_from_url", "lineno": 353, "message": "Edge: chemical-phenotype"}
{"asctime": "2021-11-02 01:05:58,730", "levelname": "INFO", "name": "pkt_kg.downloads", "module": "downloads", "funcName": "downloads_data_from_url", "lineno": 353, "message": "Edge: chemical-protein"}
{"asctime": "2021-11-02 01:05:59,290", "levelname": "INFO", "name": "pkt_kg.downloads", "module": "downloads", "funcName": "downloads_data_from_url", "lineno": 353, "message": "Edge: disease-phenotype"}
{"asctime": "2021-11-02 01:05:59,665", "levelname": "INFO", "name": "pkt_kg.downloads", "module": "downloads", "funcName": "downloads_data_from_url", "lineno": 353, "message": "Edge: gene-disease"}
{"asctime": "2021-11-02 01:05:59,810", "levelname": "INFO", "name": "pkt_kg.downloads", "module": "downloads", "funcName": "downloads_data_from_url", "lineno": 353, "message": "Edge: gene-gene"}
{"asctime": "2021-11-02 01:06:01,624", "levelname": "INFO", "name": "pkt_kg.downloads", "module": "downloads", "funcName": "downloads_data_from_url", "lineno": 353, "message": "Edge: gene-pathway"}
{"asctime": "2021-11-02 01:06:01,694", "levelname": "INFO", "name": "pkt_kg.downloads", "module": "downloads", "funcName": "downloads_data_from_url", "lineno": 353, "message": "Edge: gene-phenotype"}
{"asctime": "2021-11-02 01:06:01,709", "levelname": "INFO", "name": "pkt_kg.downloads", "module": "downloads", "funcName": "downloads_data_from_url", "lineno": 353, "message": "Edge: gene-protein"}
{"asctime": "2021-11-02 01:06:01,735", "levelname": "INFO", "name": "pkt_kg.downloads", "module": "downloads", "funcName": "downloads_data_from_url", "lineno": 353, "message": "Edge: gene-rna"}
{"asctime": "2021-11-02 01:06:01,910", "levelname": "INFO", "name": "pkt_kg.downloads", "module": "downloads", "funcName": "downloads_data_from_url", "lineno": 353, "message": "Edge: gobp-pathway"}
{"asctime": "2021-11-02 01:06:02,114", "levelname": "INFO", "name": "pkt_kg.downloads", "module": "downloads", "funcName": "downloads_data_from_url", "lineno": 353, "message": "Edge: pathway-gocc"}
{"asctime": "2021-11-02 01:06:02,130", "levelname": "INFO", "name": "pkt_kg.downloads", "module": "downloads", "funcName": "downloads_data_from_url", "lineno": 353, "message": "Edge: pathway-gomf"}
{"asctime": "2021-11-02 01:06:02,146", "levelname": "INFO", "name": "pkt_kg.downloads", "module": "downloads", "funcName": "downloads_data_from_url", "lineno": 353, "message": "Edge: protein-anatomy"}
{"asctime": "2021-11-02 01:06:02,313", "levelname": "INFO", "name": "pkt_kg.downloads", "module": "downloads", "funcName": "downloads_data_from_url", "lineno": 353, "message": "Edge: protein-catalyst"}
{"asctime": "2021-11-02 01:06:02,342", "levelname": "INFO", "name": "pkt_kg.downloads", "module": "downloads", "funcName": "downloads_data_from_url", "lineno": 353, "message": "Edge: protein-cell"}
{"asctime": "2021-11-02 01:06:02,369", "levelname": "INFO", "name": "pkt_kg.downloads", "module": "downloads", "funcName": "downloads_data_from_url", "lineno": 353, "message": "Edge: protein-cofactor"}
{"asctime": "2021-11-02 01:06:02,384", "levelname": "INFO", "name": "pkt_kg.downloads", "module": "downloads", "funcName": "downloads_data_from_url", "lineno": 353, "message": "Edge: protein-gobp"}
{"asctime": "2021-11-02 01:06:03,132", "levelname": "INFO", "name": "pkt_kg.downloads", "module": "downloads", "funcName": "downloads_data_from_url", "lineno": 353, "message": "Edge: protein-gocc"}
{"asctime": "2021-11-02 01:06:03,274", "levelname": "INFO", "name": "pkt_kg.downloads", "module": "downloads", "funcName": "downloads_data_from_url", "lineno": 353, "message": "Edge: protein-gomf"}
{"asctime": "2021-11-02 01:06:03,410", "levelname": "INFO", "name": "pkt_kg.downloads", "module": "downloads", "funcName": "downloads_data_from_url", "lineno": 353, "message": "Edge: protein-pathway"}
{"asctime": "2021-11-02 01:06:04,093", "levelname": "INFO", "name": "pkt_kg.downloads", "module": "downloads", "funcName": "downloads_data_from_url", "lineno": 353, "message": "Edge: protein-protein"}
{"asctime": "2021-11-02 01:06:07,765", "levelname": "INFO", "name": "pkt_kg.downloads", "module": "downloads", "funcName": "downloads_data_from_url", "lineno": 353, "message": "Edge: rna-anatomy"}
{"asctime": "2021-11-02 01:06:07,793", "levelname": "INFO", "name": "pkt_kg.downloads", "module": "downloads", "funcName": "downloads_data_from_url", "lineno": 353, "message": "Edge: rna-cell"}

{"asctime": "2021-11-02 01:06:07,820", "levelname": "INFO", "name": "pkt_kg.downloads", "module": "downloads", "funcName": "downloads_data_from_url", "lineno": 353, "message": "Edge: rna-protein"}
{"asctime": "2021-11-02 01:06:07,872", "levelname": "INFO", "name": "pkt_kg.downloads", "module": "downloads", "funcName": "downloads_data_from_url", "lineno": 353, "message": "Edge: variant-disease"}
{"asctime": "2021-11-02 01:06:32,380", "levelname": "INFO", "name": "pkt_kg.downloads", "module": "downloads", "funcName": "downloads_data_from_url", "lineno": 353, "message": "Edge: variant-gene"}
{"asctime": "2021-11-02 01:06:37,059", "levelname": "INFO", "name": "pkt_kg.downloads", "module": "downloads", "funcName": "downloads_data_from_url", "lineno": 353, "message": "Edge: variant-phenotype"}
{"asctime": "2021-11-02 01:06:42,265", "levelname": "INFO", "name": "pkt_kg.downloads", "module": "downloads", "funcName": "generates_source_metadata", "lineno": 198, "message": "*** Generating Metadata ***"}
{"asctime": "2021-11-02 01:06:46,061", "levelname": "INFO", "name": "pkt_kg.edge_list", "module": "edge_list", "funcName": "runs_creates_knowledge_graph_edges", "lineno": 401, "message": "**********PKT STEP: GENERATING KNOWLEDGE GRAPH MASTER EDGE LIST**********"}
{"asctime": "2021-11-02 01:08:34,019", "levelname": "INFO", "name": "pkt_kg.edge_list", "module": "edge_list", "funcName": "creates_knowledge_graph_edges", "lineno": 383, "message": "Finished Edge: chemical-gene (chemical = 466, gene = 11978); 16708 unique edges"}
{"asctime": "2021-11-02 01:08:41,757", "levelname": "INFO", "name": "pkt_kg.edge_list", "module": "edge_list", "funcName": "creates_knowledge_graph_edges", "lineno": 383, "message": "Finished Edge: chemical-pathway (chemical = 2247, pathway = 2243); 29988 unique edges"}
{"asctime": "2021-11-02 01:08:46,193", "levelname": "INFO", "name": "pkt_kg.edge_list", "module": "edge_list", "funcName": "creates_knowledge_graph_edges", "lineno": 383, "message": "Finished Edge: gene-disease (gene = 5060, disease = 4436); 12842 unique edges"}
{"asctime": "2021-11-02 01:08:46,943", "levelname": "INFO", "name": "pkt_kg.edge_list", "module": "edge_list", "funcName": "creates_knowledge_graph_edges", "lineno": 383, "message": "Finished Edge: gene-protein (gene = 19316, protein = 19134); 19521 unique edges"}
{"asctime": "2021-11-02 01:08:50,589", "levelname": "INFO", "name": "pkt_kg.edge_list", "module": "edge_list", "funcName": "creates_knowledge_graph_edges", "lineno": 383, "message": "Finished Edge: pathway-gomf (pathway = 2422, gomf = 728); 2426 unique edges"}
{"asctime": "2021-11-02 01:08:50,680", "levelname": "INFO", "name": "pkt_kg.edge_list", "module": "edge_list", "funcName": "creates_knowledge_graph_edges", "lineno": 383, "message": "Finished Edge: protein-cofactor (protein = 1584, cofactor = 44); 1998 unique edges"}
{"asctime": "2021-11-02 01:09:16,316", "levelname": "INFO", "name": "pkt_kg.edge_list", "module": "edge_list", "funcName": "creates_knowledge_graph_edges", "lineno": 383, "message": "Finished Edge: protein-pathway (protein = 10546, pathway = 2507); 117813 unique edges"}
{"asctime": "2021-11-02 01:09:18,343", "levelname": "INFO", "name": "pkt_kg.edge_list", "module": "edge_list", "funcName": "creates_knowledge_graph_edges", "lineno": 383, "message": "Finished Edge: rna-protein (rna = 44202, protein = 19200); 44205 unique edges"}
{"asctime": "2021-11-02 01:09:57,823", "levelname": "INFO", "name": "pkt_kg.edge_list", "module": "edge_list", "funcName": "creates_knowledge_graph_edges", "lineno": 383, "message": "Finished Edge: chemical-gocc (chemical = 1121, gocc = 262); 47716 unique edges"}
{"asctime": "2021-11-02 01:10:17,756", "levelname": "INFO", "name": "pkt_kg.edge_list", "module": "edge_list", "funcName": "creates_knowledge_graph_edges", "lineno": 383, "message": "Finished Edge: chemical-disease (chemical = 4341, disease = 4583); 172573 unique edges"}
{"asctime": "2021-11-02 01:11:24,875", "levelname": "INFO", "name": "pkt_kg.edge_list", "module": "edge_list", "funcName": "creates_knowledge_graph_edges", "lineno": 383, "message": "Finished Edge: chemical-protein (chemical = 4272, protein = 7946); 71679 unique edges"}
{"asctime": "2021-11-02 01:11:29,606", "levelname": "INFO", "name": "pkt_kg.edge_list", "module": "edge_list", "funcName": "creates_knowledge_graph_edges", "lineno": 383, "message": "Finished Edge: gene-pathway (gene = 10369, pathway = 1860); 107009 unique edges"}
{"asctime": "2021-11-02 01:11:32,172", "levelname": "INFO", "name": "pkt_kg.edge_list", "module": "edge_list", "funcName": "creates_knowledge_graph_edges", "lineno": 383, "message": "Finished Edge: gobp-pathway (gobp = 479, pathway = 672); 672 unique edges"}
{"asctime": "2021-11-02 01:11:32,499", "levelname": "INFO", "name": "pkt_kg.edge_list", "module": "edge_list", "funcName": "creates_knowledge_graph_edges", "lineno": 383, "message": "Finished Edge: protein-catalyst (protein = 3049, catalyst = 3758); 25136 unique edges"}
{"asctime": "2021-11-02 01:11:34,088", "levelname": "INFO", "name": "pkt_kg.edge_list", "module": "edge_list", "funcName": "creates_knowledge_graph_edges", "lineno": 383, "message": "Finished Edge: chemical-gobp (chemical = 1350, gobp = 1510); 288873 unique edges"}
{"asctime": "2021-11-02 01:11:53,756", "levelname": "INFO", "name": "pkt_kg.edge_list", "module": "edge_list", "funcName": "creates_knowledge_graph_edges", "lineno": 383, "message": "Finished Edge: protein-gocc (protein = 18451, gocc = 1752); 82526 unique edges"}
{"asctime": "2021-11-02 01:12:07,668", "levelname": "INFO", "name": "pkt_kg.edge_list", "module": "edge_list", "funcName": "creates_knowledge_graph_edges", "lineno": 383, "message": "Finished Edge: rna-anatomy (rna = 29121, anatomy = 103); 444974 unique edges"}
{"asctime": "2021-11-02 01:12:47,883", "levelname": "INFO", "name": "pkt_kg.edge_list", "module": "edge_list", "funcName": "creates_knowledge_graph_edges", "lineno": 383, "message": "Finished Edge: chemical-gomf (chemical = 1133, gomf = 214); 28077 unique edges"}
{"asctime": "2021-11-02 01:12:57,504", "levelname": "INFO", "name": "pkt_kg.edge_list", "module": "edge_list", "funcName": "creates_knowledge_graph_edges", "lineno": 383, "message": "Finished Edge: disease-phenotype (disease = 11930, phenotype = 10068); 435102 unique edges"}
{"asctime": "2021-11-02 01:13:01,176", "levelname": "INFO", "name": "pkt_kg.edge_list", "module": "edge_list", "funcName": "creates_knowledge_graph_edges", "lineno": 383, "message": "Finished Edge: gene-phenotype (gene = 6785, phenotype = 1597); 24760 unique edges"}
{"asctime": "2021-11-02 01:13:06,564", "levelname": "INFO", "name": "pkt_kg.edge_list", "module": "edge_list", "funcName": "creates_knowledge_graph_edges", "lineno": 383, "message": "Finished Edge: pathway-gocc (pathway = 11252, gocc = 99); 16014 unique edges"}
{"asctime": "2021-11-02 01:13:14,213", "levelname": "INFO", "name": "pkt_kg.edge_list", "module": "edge_list", "funcName": "creates_knowledge_graph_edges", "lineno": 383, "message": "Finished Edge: protein-cell (protein = 10044, cell = 128); 75313 unique edges"}
{"asctime": "2021-11-02 01:13:37,715", "levelname": "INFO", "name": "pkt_kg.edge_list", "module": "edge_list", "funcName": "creates_knowledge_graph_edges", "lineno": 383, "message": "Finished Edge: protein-gomf (protein = 17801, gomf = 4430); 69801 unique edges"}
{"asctime": "2021-11-02 01:13:47,777", "levelname": "INFO", "name": "pkt_kg.edge_list", "module": "edge_list", "funcName": "creates_knowledge_graph_edges", "lineno": 383, "message": "Finished Edge: rna-cell (rna = 14044, cell = 130); 65180 unique edges"}
{"asctime": "2021-11-02 01:14:26,632", "levelname": "INFO", "name": "pkt_kg.edge_list", "module": "edge_list", "funcName": "creates_knowledge_graph_edges", "lineno": 383, "message": "Finished Edge: chemical-phenotype (chemical = 4102, phenotype = 1742); 110898 unique edges"}
{"asctime": "2021-11-02 01:15:15,286", "levelname": "INFO", "name": "pkt_kg.edge_list", "module": "edge_list", "funcName": "creates_knowledge_graph_edges", "lineno": 383, "message": "Finished Edge: gene-gene (gene = 250, gene = 267); 1694 unique edges"}
{"asctime": "2021-11-02 01:15:18,594", "levelname": "INFO", "name": "pkt_kg.edge_list", "module": "edge_list", "funcName": "creates_knowledge_graph_edges", "lineno": 383, "message": "Finished Edge: gene-rna (gene = 25527, rna = 179872); 182692 unique edges"}
{"asctime": "2021-11-02 01:15:26,108", "levelname": "INFO", "name": "pkt_kg.edge_list", "module": "edge_list", "funcName": "creates_knowledge_graph_edges", "lineno": 383, "message": "Finished Edge: protein-anatomy (protein = 10746, anatomy = 68); 30681 unique edges"}
{"asctime": "2021-11-02 01:15:48,464", "levelname": "INFO", "name": "pkt_kg.edge_list", "module": "edge_list", "funcName": "creates_knowledge_graph_edges", "lineno": 383, "message": "Finished Edge: protein-gobp (protein = 17404, gobp = 12329); 129424 unique edges"}
{"asctime": "2021-11-02 01:20:08,039", "levelname": "INFO", "name": "pkt_kg.edge_list", "module": "edge_list", "funcName": "creates_knowledge_graph_edges", "lineno": 383, "message": "Finished Edge: protein-protein (protein = 14230, protein = 14230); 618069 unique edges"}
{"asctime": "2021-11-02 01:21:54,271", "levelname": "INFO", "name": "pkt_kg.edge_list", "module": "edge_list", "funcName": "creates_knowledge_graph_edges", "lineno": 383, "message": "Finished Edge: variant-gene (variant = 145129, gene = 3626); 145129 unique edges"}
{"asctime": "2021-11-02 01:21:59,856", "levelname": "INFO", "name": "pkt_kg.edge_list", "module": "edge_list", "funcName": "creates_knowledge_graph_edges", "lineno": 383, "message": "Finished Edge: variant-phenotype (variant = 2100, phenotype = 436); 3081 unique edges"}
{"asctime": "2021-11-02 01:26:14,303", "levelname": "INFO", "name": "pkt_kg.edge_list", "module": "edge_list", "funcName": "creates_knowledge_graph_edges", "lineno": 383, "message": "Finished Edge: variant-disease (variant = 14732, disease = 3753); 43439 unique edges"}
{"asctime": "2021-11-02 01:27:05,318", "levelname": "INFO", "name": "pkt_kg.knowledge_graph", "module": "knowledge_graph", "funcName": "construct_knowledge_graph", "lineno": 543, "message": "**********PKT STEP: CONSTRUCTING KNOWLEDGE GRAPH**********\n### Starting Knowledge Graph Build: FULL ###"}
{"asctime": "2021-11-02 01:27:05,318", "levelname": "INFO", "name": "pkt_kg.knowledge_graph", "module": "knowledge_graph", "funcName": "construct_knowledge_graph", "lineno": 546, "message": "*** Loading Relations Data ***"}
{"asctime": "2021-11-02 01:27:05,319", "levelname": "INFO", "name": "pkt_kg.knowledge_graph", "module": "knowledge_graph", "funcName": "construct_knowledge_graph", "lineno": 551, "message": "*** Loading Merged Ontologies ***"}
{"asctime": "2021-11-02 02:24:06,890", "levelname": "INFO", "name": "pkt_kg.knowledge_graph", "module": "knowledge_graph", "funcName": "construct_knowledge_graph", "lineno": 557, "message": "Merged Ontologies Graph Stats: 13933782 triples, 5716554 nodes, 355 predicates, 558715 classes, 190 individuals, 853 object props, 635 annotation props"}
{"asctime": "2021-11-02 02:24:06,890", "levelname": "INFO", "name": "pkt_kg.knowledge_graph", "module": "knowledge_graph", "funcName": "construct_knowledge_graph", "lineno": 560, "message": "*** Loading Node Metadata Data ***"}
{"asctime": "2021-11-02 02:24:06,891", "levelname": "INFO", "name": "pkt_kg.metadata", "module": "metadata", "funcName": "metadata_processor", "lineno": 80, "message": "Loading and Processing Node Metadata"}
{"asctime": "2021-11-02 02:24:09,201", "levelname": "INFO", "name": "pkt_kg.metadata", "module": "metadata", "funcName": "extract_metadata", "lineno": 116, "message": "Extracting Class and Relation Metadata"}
{"asctime": "2021-11-02 02:26:39,115", "levelname": "INFO", "name": "pkt_kg.knowledge_graph", "module": "knowledge_graph", "funcName": "construct_knowledge_graph", "lineno": 565, "message": "*** Splitting Graph ***"}
{"asctime": "2021-11-02 03:20:03,997", "levelname": "INFO", "name": "pkt_kg.knowledge_graph", "module": "knowledge_graph", "funcName": "construct_knowledge_graph", "lineno": 567, "message": "Merged Ontologies - Logic Subset Graph Stats: 4103821 triples, 1420834 nodes, 43 predicates, 558715 classes, 190 individuals, 853 object props, 635 annotation props"}
{"asctime": "2021-11-02 03:23:44,592", "levelname": "INFO", "name": "pkt_kg.knowledge_graph", "module": "knowledge_graph", "funcName": "construct_knowledge_graph", "lineno": 574, "message": "*** Building Knowledge Graph Edges ***"}
{"asctime": "2021-11-02 03:58:16,555", "levelname": "INFO", "name": "pkt_kg.knowledge_graph", "module": "knowledge_graph", "funcName": "creates_new_edges", "lineno": 347, "message": "Created CHEMICAL-GOBP (class-class) Edges: 1722549 OWL Edges, 286618 Original Edges; 576080 OWL Nodes, Original Nodes: 1350 chemical(s), 1490 gobp(s)"}
{"asctime": "2021-11-02 04:15:32,538", "levelname": "INFO", "name": "pkt_kg.knowledge_graph", "module": "knowledge_graph", "funcName": "creates_new_edges", "lineno": 347, "message": "Created DISEASE-PHENOTYPE (class-class) Edges: 2592087 OWL Edges, 428374 Original Edges; 878594 OWL Nodes, Original Nodes: 11780 disease(s), 10062 phenotype(s)"}
{"asctime": "2021-11-02 04:29:10,127", "levelname": "INFO", "name": "pkt_kg.knowledge_graph", "module": "knowledge_graph", "funcName": "creates_new_edges", "lineno": 347, "message": "Created GENE-RNA (entity-entity) Edges: 1686851 OWL Edges, 182692 Original Edges; 570815 OWL Nodes, Original Nodes: 25527 gene(s), 179872 rna(s)"}
{"asctime": "2021-11-02 04:34:08,778", "levelname": "INFO", "name": "pkt_kg.knowledge_graph", "module": "knowledge_graph", "funcName": "creates_new_edges", "lineno": 347, "message": "Created RNA-ANATOMY (entity-class) Edges: 2728768 OWL Edges, 440217 Original Edges; 909683 OWL Nodes, Original Nodes: 29110 rna(s), 102 anatomy(s)"}
{"asctime": "2021-11-02 04:34:47,245", "levelname": "INFO", "name": "pkt_kg.knowledge_graph", "module": "knowledge_graph", "funcName": "creates_new_edges", "lineno": 347, "message": "Created CHEMICAL-DISEASE (class-class) Edges: 1021872 OWL Edges, 168841 Original Edges; 346511 OWL Nodes, Original Nodes: 4339 chemical(s), 4486 disease(s)"}

{"asctime": "2021-11-02 04:40:18,626", "levelname": "INFO", "name": "pkt_kg.knowledge_graph", "module": "knowledge_graph", "funcName": "creates_new_edges", "lineno": 347, "message": "Created PROTEIN-PROTEIN (class-class) Edges: 3722645 OWL Edges, 618069 Original Edges; 1250372 OWL Nodes, Original Nodes: 14230 protein(s), 14230 protein(s)"}
{"asctime": "2021-11-02 04:44:04,526", "levelname": "INFO", "name": "pkt_kg.knowledge_graph", "module": "knowledge_graph", "funcName": "creates_new_edges", "lineno": 347, "message": "Created PROTEIN-GOBP (class-class) Edges: 806278 OWL Edges, 129424 Original Edges; 288585 OWL Nodes, Original Nodes: 17404 protein(s), 12329 gobp(s)"}
{"asctime": "2021-11-02 04:49:59,442", "levelname": "INFO", "name": "pkt_kg.knowledge_graph", "module": "knowledge_graph", "funcName": "creates_new_edges", "lineno": 347, "message": "Created PROTEIN-GOCC (class-class) Edges: 515360 OWL Edges, 82526 Original Edges; 185259 OWL Nodes, Original Nodes: 18451 protein(s), 1752 gocc(s)"}
{"asctime": "2021-11-02 04:55:23,205", "levelname": "INFO", "name": "pkt_kg.knowledge_graph", "module": "knowledge_graph", "funcName": "creates_new_edges", "lineno": 347, "message": "Created GENE-PATHWAY (entity-entity) Edges: 655865 OWL Edges, 104891 Original Edges; 222723 OWL Nodes, Original Nodes: 10369 gene(s), 1809 pathway(s)"}
{"asctime": "2021-11-02 04:56:44,758", "levelname": "INFO", "name": "pkt_kg.knowledge_graph", "module": "knowledge_graph", "funcName": "creates_new_edges", "lineno": 347, "message": "Created VARIANT-GENE (entity-entity) Edges: 1168894 OWL Edges, 145129 Original Edges; 439031 OWL Nodes, Original Nodes: 145129 variant(s), 3626 gene(s)"}
{"asctime": "2021-11-02 04:56:51,244", "levelname": "INFO", "name": "pkt_kg.knowledge_graph", "module": "knowledge_graph", "funcName": "creates_new_edges", "lineno": 347, "message": "Created CHEMICAL-PHENOTYPE (class-class) Edges: 667099 OWL Edges, 110211 Original Edges; 226258 OWL Nodes, Original Nodes: 4100 chemical(s), 1732 phenotype(s)"}
{"asctime": "2021-11-02 04:58:30,807", "levelname": "INFO", "name": "pkt_kg.knowledge_graph", "module": "knowledge_graph", "funcName": "creates_new_edges", "lineno": 347, "message": "Created PROTEIN-CELL (class-class) Edges: 451320 OWL Edges, 73525 Original Edges; 157223 OWL Nodes, Original Nodes: 10044 protein(s), 125 cell(s)"}
{"asctime": "2021-11-02 05:03:40,898", "levelname": "INFO", "name": "pkt_kg.knowledge_graph", "module": "knowledge_graph", "funcName": "creates_new_edges", "lineno": 347, "message": "Created CHEMICAL-GOCC (class-class) Edges: 264351 OWL Edges, 43832 Original Edges; 89026 OWL Nodes, Original Nodes: 1121 chemical(s), 237 gocc(s)"}
{"asctime": "2021-11-02 05:04:55,020", "levelname": "INFO", "name": "pkt_kg.knowledge_graph", "module": "knowledge_graph", "funcName": "creates_new_edges", "lineno": 347, "message": "Created PROTEIN-GOMF (class-class) Edges: 441038 OWL Edges, 69801 Original Edges; 161837 OWL Nodes, Original Nodes: 17801 protein(s), 4430 gomf(s)"}
{"asctime": "2021-11-02 05:05:12,368", "levelname": "INFO", "name": "pkt_kg.knowledge_graph", "module": "knowledge_graph", "funcName": "creates_new_edges", "lineno": 347, "message": "Created RNA-CELL (entity-class) Edges: 428998 OWL Edges, 64451 Original Edges; 143109 OWL Nodes, Original Nodes: 14044 rna(s), 127 cell(s)"}
{"asctime": "2021-11-02 05:06:21,474", "levelname": "INFO", "name": "pkt_kg.knowledge_graph", "module": "knowledge_graph", "funcName": "creates_new_edges", "lineno": 347, "message": "Created PROTEIN-CATALYST (class-class) Edges: 149562 OWL Edges, 23794 Original Edges; 54389 OWL Nodes, Original Nodes: 3048 protein(s), 3749 catalyst(s)"}
{"asctime": "2021-11-02 05:08:22,977", "levelname": "INFO", "name": "pkt_kg.knowledge_graph", "module": "knowledge_graph", "funcName": "creates_new_edges", "lineno": 347, "message": "Created PROTEIN-ANATOMY (class-class) Edges: 194901 OWL Edges, 30681 Original Edges; 72180 OWL Nodes, Original Nodes: 10746 protein(s), 68 anatomy(s)"}
{"asctime": "2021-11-02 05:08:45,554", "levelname": "INFO", "name": "pkt_kg.knowledge_graph", "module": "knowledge_graph", "funcName": "creates_new_edges", "lineno": 347, "message": "Created PATHWAY-GOCC (entity-class) Edges: 119256 OWL Edges, 16014 Original Edges; 43559 OWL Nodes, Original Nodes: 11252 pathway(s), 99 gocc(s)"}
{"asctime": "2021-11-02 05:09:15,503", "levelname": "INFO", "name": "pkt_kg.knowledge_graph", "module": "knowledge_graph", "funcName": "creates_new_edges", "lineno": 347, "message": "Created PROTEIN-COFACTOR (class-class) Edges: 13388 OWL Edges, 1960 Original Edges; 5551 OWL Nodes, Original Nodes: 1583 protein(s), 44 cofactor(s)"}
{"asctime": "2021-11-02 05:10:51,535", "levelname": "INFO", "name": "pkt_kg.knowledge_graph", "module": "knowledge_graph", "funcName": "creates_new_edges", "lineno": 347, "message": "Created VARIANT-DISEASE (entity-class) Edges: 291895 OWL Edges, 43088 Original Edges; 104581 OWL Nodes, Original Nodes: 14712 variant(s), 3683 disease(s)"}
{"asctime": "2021-11-02 05:13:24,844", "levelname": "INFO", "name": "pkt_kg.knowledge_graph", "module": "knowledge_graph", "funcName": "creates_new_edges", "lineno": 347, "message": "Created CHEMICAL-GOMF (class-class) Edges: 164592 OWL Edges, 27209 Original Edges; 55759 OWL Nodes, Original Nodes: 1133 chemical(s), 204 gomf(s)"}
{"asctime": "2021-11-02 05:14:08,016", "levelname": "INFO", "name": "pkt_kg.knowledge_graph", "module": "knowledge_graph", "funcName": "creates_new_edges", "lineno": 347, "message": "Created CHEMICAL-PATHWAY (class-entity) Edges: 188972 OWL Edges, 29973 Original Edges; 65285 OWL Nodes, Original Nodes: 2245 chemical(s), 2243 pathway(s)"}
{"asctime": "2021-11-02 05:14:57,582", "levelname": "INFO", "name": "pkt_kg.knowledge_graph", "module": "knowledge_graph", "funcName": "creates_new_edges", "lineno": 347, "message": "Created GENE-DISEASE (entity-class) Edges: 91269 OWL Edges, 12808 Original Edges; 35057 OWL Nodes, Original Nodes: 5056 gene(s), 4422 disease(s)"}
{"asctime": "2021-11-02 05:15:14,241", "levelname": "INFO", "name": "pkt_kg.knowledge_graph", "module": "knowledge_graph", "funcName": "creates_new_edges", "lineno": 347, "message": "Created PATHWAY-GOMF (entity-class) Edges: 20272 OWL Edges, 2426 Original Edges; 8065 OWL Nodes, Original Nodes: 2422 pathway(s), 728 gomf(s)"}
{"asctime": "2021-11-02 05:15:45,160", "levelname": "INFO", "name": "pkt_kg.knowledge_graph", "module": "knowledge_graph", "funcName": "creates_new_edges", "lineno": 347, "message": "Created PROTEIN-PATHWAY (class-entity) Edges: 722480 OWL Edges, 117410 Original Edges; 248711 OWL Nodes, Original Nodes: 10512 protein(s), 2495 pathway(s)"}
{"asctime": "2021-11-02 05:16:10,820", "levelname": "INFO", "name": "pkt_kg.knowledge_graph", "module": "knowledge_graph", "funcName": "creates_new_edges", "lineno": 347, "message": "Created GENE-PROTEIN (entity-class) Edges: 174902 OWL Edges, 19521 Original Edges; 77505 OWL Nodes, Original Nodes: 19316 gene(s), 19134 protein(s)"}
{"asctime": "2021-11-02 05:16:43,553", "levelname": "INFO", "name": "pkt_kg.knowledge_graph", "module": "knowledge_graph", "funcName": "creates_new_edges", "lineno": 347, "message": "Created VARIANT-PHENOTYPE (entity-class) Edges: 22689 OWL Edges, 3005 Original Edges; 8555 OWL Nodes, Original Nodes: 2100 variant(s), 435 phenotype(s)"}
{"asctime": "2021-11-02 05:16:53,519", "levelname": "INFO", "name": "pkt_kg.knowledge_graph", "module": "knowledge_graph", "funcName": "creates_new_edges", "lineno": 347, "message": "Created GENE-GENE (entity-entity) Edges: 10819 OWL Edges, 1694 Original Edges; 3721 OWL Nodes, Original Nodes: 250 gene(s), 267 gene(s)"}
{"asctime": "2021-11-02 05:20:24,790", "levelname": "INFO", "name": "pkt_kg.knowledge_graph", "module": "knowledge_graph", "funcName": "creates_new_edges", "lineno": 347, "message": "Created CHEMICAL-PROTEIN (class-class) Edges: 442293 OWL Edges, 71679 Original Edges; 155580 OWL Nodes, Original Nodes: 4272 chemical(s), 7946 protein(s)"}
{"asctime": "2021-11-02 05:23:49,688", "levelname": "INFO", "name": "pkt_kg.knowledge_graph", "module": "knowledge_graph", "funcName": "creates_new_edges", "lineno": 347, "message": "Created RNA-PROTEIN (entity-class) Edges: 417047 OWL Edges, 44205 Original Edges; 151826 OWL Nodes, Original Nodes: 44202 rna(s), 19200 protein(s)"}
{"asctime": "2021-11-02 05:25:28,141", "levelname": "INFO", "name": "pkt_kg.knowledge_graph", "module": "knowledge_graph", "funcName": "creates_new_edges", "lineno": 347, "message": "Created GENE-PHENOTYPE (entity-class) Edges: 163019 OWL Edges, 24664 Original Edges; 57673 OWL Nodes, Original Nodes: 6782 gene(s), 1594 phenotype(s)"}
{"asctime": "2021-11-02 05:26:43,080", "levelname": "INFO", "name": "pkt_kg.knowledge_graph", "module": "knowledge_graph", "funcName": "creates_new_edges", "lineno": 347, "message": "Created CHEMICAL-GENE (class-entity) Edges: 124162 OWL Edges, 16652 Original Edges; 45687 OWL Nodes, Original Nodes: 466 chemical(s), 11923 gene(s)"}
{"asctime": "2021-11-02 05:26:47,258", "levelname": "INFO", "name": "pkt_kg.knowledge_graph", "module": "knowledge_graph", "funcName": "creates_new_edges", "lineno": 347, "message": "Created GOBP-PATHWAY (class-entity) Edges: 6745 OWL Edges, 671 Original Edges; 2578 OWL Nodes, Original Nodes: 478 gobp(s), 671 pathway(s)"}
{"asctime": "2021-11-02 05:36:55,564", "levelname": "INFO", "name": "pkt_kg.knowledge_graph", "module": "knowledge_graph", "funcName": "construct_knowledge_graph", "lineno": 590, "message": "See log: /PheKnowLator/resources/construction_approach/subclass_map_log.json"}
{"asctime": "2021-11-02 06:08:53,745", "levelname": "INFO", "name": "pkt_kg.knowledge_graph", "module": "knowledge_graph", "funcName": "construct_knowledge_graph", "lineno": 594, "message": "Full Logic Graph Stats: 25611453 triples, 8644230 nodes, 43 predicates, 4348593 classes, 190 individuals, 853 object props, 635 annotation props"}
{"asctime": "2021-11-02 07:23:38,482", "levelname": "INFO", "name": "pkt_kg.knowledge_graph", "module": "knowledge_graph", "funcName": "construct_knowledge_graph", "lineno": 596, "message": "Full Logic Subset (OWL) Graph Stats: 8644230 nodes, 25611453 edges, 2 self-loops, 5 most most common edges: http://www.w3.org/2000/01/rdf-schema#subClassOf:8411703, http://www.w3.org/1999/02/22-rdf-syntax-ns#type:8364393, http://www.w3.org/2002/07/owl#onProperty:3957847, http://www.w3.org/2002/07/owl#someValuesFrom:3953628, http://www.w3.org/1999/02/22-rdf-syntax-ns#first:280278, http://www.w3.org/1999/02/22-rdf-syntax-ns#rest:280278, average degree 2.962837985569565, 5 highest degree nodes: http://www.w3.org/2002/07/owl#Class:4348593, http://www.w3.org/2002/07/owl#Restriction:3957847,

http://purl.obolibrary.org/obo/RO_0002436:1001470, http://purl.obolibrary.org/obo/RO_0001025:689237, http://purl.obolibrary.org/obo/RO_0002200:428381, http://purl.obolibrary.org/obo/RO_0000056:381743, density: 3.427532965137278e-07, 2 component(s): {0: 8644227, 1: '3 nodes: http://www.w3.org/2002/07/owl#Ontology | http://purl.obolibrary.org/obo/chebi/204/chebi.owl | http://purl.obolibrary.org/obo/chebi.owl'}"}
{"asctime": "2021-11-02 07:31:29,029", "levelname": "INFO", "name": "pkt_kg.owlnets", "module": "owlnets", "funcName": "runs_owlnets", "lineno": 766, "message": "*** Running OWL-NETS ***"}
{"asctime": "2021-11-02 07:31:29,030", "levelname": "INFO", "name": "pkt_kg.owlnets", "module": "owlnets", "funcName": "removes_disjoint_with_axioms", "lineno": 130, "message": "Removing owl:disjointWith Axioms"}
{"asctime": "2021-11-02 07:31:29,156", "levelname": "INFO", "name": "pkt_kg.owlnets", "module": "owlnets", "funcName": "removes_edges_with_owl_semantics", "lineno": 157, "message": "Filtering Triples"}
{"asctime": "2021-11-02 07:39:27,341", "levelname": "INFO", "name": "pkt_kg.owlnets", "module": "owlnets", "funcName": "cleans_owl_encoded_entities", "lineno": 610, "message": "Decoding 114668 OWL Classes and Axioms"}
{"asctime": "2021-11-02 07:39:31,150", "levelname": "INFO", "name": "pkt_kg.owlnets", "module": "owlnets", "funcName": "cleans_owl_encoded_entities", "lineno": 610, "message": "Decoding 114667 OWL Classes and Axioms"}
{"asctime": "2021-11-02 07:39:35,410", "levelname": "INFO", "name": "pkt_kg.owlnets", "module": "owlnets", "funcName": "cleans_owl_encoded_entities", "lineno": 610, "message": "Decoding 114667 OWL Classes and Axioms"}
{"asctime": "2021-11-02 07:40:22,156", "levelname": "INFO", "name": "pkt_kg.owlnets", "module": "owlnets", "funcName": "cleans_owl_encoded_entities", "lineno": 610, "message": "Decoding 114667 OWL Classes and Axioms"}
{"asctime": "2021-11-02 07:44:34,169", "levelname": "INFO", "name": "pkt_kg.owlnets", "module": "owlnets", "funcName": "cleans_decoded_graph", "lineno": 206, "message": "Filtering Triples"}
{"asctime": "2021-11-02 07:44:34,653", "levelname": "INFO", "name": "pkt_kg.owlnets", "module": "owlnets", "funcName": "cleans_decoded_graph", "lineno": 206, "message": "Filtering Triples"}
{"asctime": "2021-11-02 07:45:14,083", "levelname": "INFO", "name": "pkt_kg.owlnets", "module": "owlnets", "funcName": "cleans_decoded_graph", "lineno": 206, "message": "Filtering Triples"}
{"asctime": "2021-11-02 07:46:20,730", "levelname": "INFO", "name": "pkt_kg.owlnets", "module": "owlnets", "funcName": "cleans_decoded_graph", "lineno": 206, "message": "Filtering Triples"}
{"asctime": "2021-11-02 07:47:22,051", "levelname": "INFO", "name": "pkt_kg.owlnets", "module": "owlnets", "funcName": "removes_disjoint_with_axioms", "lineno": 130, "message": "Removing owl:disjointWith Axioms"}
{"asctime": "2021-11-02 07:47:22,052", "levelname": "INFO", "name": "pkt_kg.owlnets", "module": "owlnets", "funcName": "removes_edges_with_owl_semantics", "lineno": 157, "message": "Filtering Triples"}
{"asctime": "2021-11-02 07:50:18,265", "levelname": "INFO", "name": "pkt_kg.owlnets", "module": "owlnets", "funcName": "cleans_owl_encoded_entities", "lineno": 610, "message": "Decoding 9364 OWL Classes and Axioms"}
{"asctime": "2021-11-02 07:50:18,715", "levelname": "INFO", "name": "pkt_kg.owlnets", "module": "owlnets", "funcName": "cleans_owl_encoded_entities", "lineno": 610, "message": "Decoding 9363 OWL Classes and Axioms"}
{"asctime": "2021-11-02 07:50:26,971", "levelname": "INFO", "name": "pkt_kg.owlnets", "module": "owlnets", "funcName": "cleans_owl_encoded_entities", "lineno": 610, "message": "Decoding 9363 OWL Classes and Axioms"}
{"asctime": "2021-11-02 07:51:00,114", "levelname": "INFO", "name": "pkt_kg.owlnets", "module": "owlnets", "funcName": "cleans_owl_encoded_entities", "lineno": 610, "message": "Decoding 9363 OWL Classes and Axioms"}
{"asctime": "2021-11-02 07:58:59,179", "levelname": "INFO", "name": "pkt_kg.owlnets", "module": "owlnets", "funcName": "cleans_decoded_graph", "lineno": 206, "message": "Filtering Triples"}
{"asctime": "2021-11-02 07:59:05,319", "levelname": "INFO", "name": "pkt_kg.owlnets", "module": "owlnets", "funcName": "cleans_decoded_graph", "lineno": 206, "message": "Filtering Triples"}
{"asctime": "2021-11-02 07:59:30,560", "levelname": "INFO", "name": "pkt_kg.owlnets", "module": "owlnets", "funcName": "cleans_decoded_graph", "lineno": 206, "message": "Filtering Triples"}
{"asctime": "2021-11-02 08:00:08,372", "levelname": "INFO", "name": "pkt_kg.owlnets", "module": "owlnets", "funcName": "cleans_decoded_graph", "lineno": 206, "message": "Filtering Triples"}
{"asctime": "2021-11-02 08:01:33,118", "levelname": "INFO", "name": "pkt_kg.owlnets", "module": "owlnets", "funcName": "removes_disjoint_with_axioms", "lineno": 130, "message": "Removing owl:disjointWith Axioms"}
{"asctime": "2021-11-02 08:01:33,119", "levelname": "INFO", "name": "pkt_kg.owlnets", "module": "owlnets", "funcName": "removes_edges_with_owl_semantics", "lineno": 157, "message": "Filtering Triples"}
{"asctime": "2021-11-02 08:06:55,548", "levelname": "INFO", "name": "pkt_kg.owlnets", "module": "owlnets", "funcName": "cleans_owl_encoded_entities", "lineno": 610, "message": "Decoding 65038 OWL Classes and Axioms"}
{"asctime": "2021-11-02 08:06:58,339", "levelname": "INFO", "name": "pkt_kg.owlnets", "module": "owlnets", "funcName": "cleans_owl_encoded_entities", "lineno": 610, "message": "Decoding 65038 OWL Classes and Axioms"}
{"asctime": "2021-11-02 08:07:01,427", "levelname": "INFO", "name": "pkt_kg.owlnets", "module": "owlnets", "funcName": "cleans_owl_encoded_entities", "lineno": 610, "message": "Decoding 65038 OWL Classes and Axioms"}
{"asctime": "2021-11-02 08:07:43,352", "levelname": "INFO", "name": "pkt_kg.owlnets", "module": "owlnets", "funcName": "cleans_owl_encoded_entities", "lineno": 610, "message": "Decoding 65038 OWL Classes and Axioms"}
{"asctime": "2021-11-02 08:12:06,872", "levelname": "INFO", "name": "pkt_kg.owlnets", "module": "owlnets", "funcName": "cleans_decoded_graph", "lineno": 206, "message": "Filtering Triples"}
{"asctime": "2021-11-02 08:15:37,045", "levelname": "INFO", "name": "pkt_kg.owlnets", "module": "owlnets", "funcName": "cleans_decoded_graph", "lineno": 206, "message": "Filtering Triples"}
{"asctime": "2021-11-02 08:17:56,512", "levelname": "INFO", "name": "pkt_kg.owlnets", "module": "owlnets", "funcName": "cleans_decoded_graph", "lineno": 206, "message": "Filtering Triples"}
{"asctime": "2021-11-02 08:20:01,999", "levelname": "INFO", "name": "pkt_kg.owlnets", "module": "owlnets", "funcName": "cleans_decoded_graph", "lineno": 206, "message": "Filtering Triples"}
{"asctime": "2021-11-02 08:21:33,746", "levelname": "INFO", "name": "pkt_kg.owlnets", "module": "owlnets", "funcName": "removes_disjoint_with_axioms", "lineno": 130, "message": "Removing owl:disjointWith Axioms"}
{"asctime": "2021-11-02 08:21:33,747", "levelname": "INFO", "name": "pkt_kg.owlnets", "module": "owlnets", "funcName": "removes_edges_with_owl_semantics", "lineno": 157, "message": "Filtering Triples"}
{"asctime": "2021-11-02 08:24:47,308", "levelname": "INFO", "name": "pkt_kg.owlnets", "module": "owlnets", "funcName": "cleans_owl_encoded_entities", "lineno": 610, "message": "Decoding 18729 OWL Classes and Axioms"}
{"asctime": "2021-11-02 08:24:47,628", "levelname": "INFO", "name": "pkt_kg.owlnets", "module": "owlnets", "funcName": "cleans_owl_encoded_entities", "lineno": 610, "message": "Decoding 18729 OWL Classes and Axioms"}
{"asctime": "2021-11-02 08:24:52,252", "levelname": "INFO", "name": "pkt_kg.owlnets", "module": "owlnets", "funcName": "cleans_owl_encoded_entities", "lineno": 610, "message": "Decoding 18729 OWL Classes and Axioms"}
{"asctime": "2021-11-02 08:25:26,246", "levelname": "INFO", "name": "pkt_kg.owlnets", "module": "owlnets", "funcName": "cleans_owl_encoded_entities", "lineno": 610, "message": "Decoding 18729 OWL Classes and Axioms"}
{"asctime": "2021-11-02 08:32:14,445", "levelname": "INFO", "name": "pkt_kg.owlnets", "module": "owlnets", "funcName": "cleans_decoded_graph", "lineno": 206, "message": "Filtering Triples"}
{"asctime": "2021-11-02 08:32:54,439", "levelname": "INFO", "name": "pkt_kg.owlnets", "module": "owlnets", "funcName": "cleans_decoded_graph", "lineno": 206, "message": "Filtering Triples"}
{"asctime": "2021-11-02 08:33:02,489", "levelname": "INFO", "name": "pkt_kg.owlnets", "module": "owlnets", "funcName": "cleans_decoded_graph", "lineno": 206, "message": "Filtering Triples"}
{"asctime": "2021-11-02 08:33:26,285", "levelname": "INFO", "name": "pkt_kg.owlnets", "module": "owlnets", "funcName": "cleans_decoded_graph", "lineno": 206, "message": "Filtering Triples"}
{"asctime": "2021-11-02 08:34:53,681", "levelname": "INFO", "name": "pkt_kg.owlnets", "module": "owlnets", "funcName": "removes_disjoint_with_axioms", "lineno": 130, "message": "Removing owl:disjointWith Axioms"}
{"asctime": "2021-11-02 08:34:53,682", "levelname": "INFO", "name": "pkt_kg.owlnets", "module": "owlnets", "funcName": "removes_edges_with_owl_semantics", "lineno": 157, "message": "Filtering Triples"}
{"asctime": "2021-11-02 08:40:42,490", "levelname": "INFO", "name": "pkt_kg.owlnets", "module": "owlnets", "funcName": "cleans_owl_encoded_entities", "lineno": 610, "message": "Decoding 63702 OWL Classes and Axioms"}
{"asctime": "2021-11-02 08:40:45,645", "levelname": "INFO", "name": "pkt_kg.owlnets", "module": "owlnets", "funcName": "cleans_owl_encoded_entities", "lineno": 610, "message": "Decoding 63702 OWL Classes and Axioms"}
{"asctime": "2021-11-02 08:40:48,594", "levelname": "INFO", "name": "pkt_kg.owlnets", "module": "owlnets", "funcName": "cleans_owl_encoded_entities", "lineno": 610, "message": "Decoding 63701 OWL Classes and Axioms"}
{"asctime": "2021-11-02 08:41:31,818", "levelname": "INFO", "name": "pkt_kg.owlnets", "module": "owlnets", "funcName": "cleans_owl_encoded_entities", "lineno": 610, "message": "Decoding 63701 OWL Classes and Axioms"}
{"asctime": "2021-11-02 08:54:09,937", "levelname": "INFO", "name": "pkt_kg.owlnets", "module": "owlnets", "funcName": "cleans_decoded_graph", "lineno": 206, "message": "Filtering Triples"}
{"asctime": "2021-11-02 08:54:34,397", "levelname": "INFO", "name": "pkt_kg.owlnets", "module": "owlnets", "funcName": "cleans_decoded_graph", "lineno": 206, "message": "Filtering Triples"}
{"asctime": "2021-11-02 08:54:52,950", "levelname": "INFO", "name": "pkt_kg.owlnets", "module": "owlnets", "funcName": "cleans_decoded_graph", "lineno": 206, "message": "Filtering Triples"}
{"asctime": "2021-11-02 08:55:50,180", "levelname": "INFO", "name": "pkt_kg.owlnets", "module": "owlnets", "funcName": "cleans_decoded_graph", "lineno": 206, "message": "Filtering Triples"}
{"asctime": "2021-11-02 08:57:26,093", "levelname": "INFO", "name": "pkt_kg.owlnets", "module": "owlnets", "funcName": "makes_graph_connected", "lineno": 663, "message": "Ensuring OWL-NETS Graph Contains a Single Connected Component"}
{"asctime": "2021-11-02 08:57:26,093", "levelname": "INFO", "name": "pkt_kg.owlnets", "module": "owlnets", "funcName": "makes_graph_connected", "lineno": 667, "message": "Obtaining node list"}
{"asctime": "2021-11-02 09:19:25,234", "levelname": "INFO", "name": "pkt_kg.owlnets", "module": "owlnets", "funcName": "makes_graph_connected", "lineno": 683, "message": "Updating graph connectivity"}
{"asctime": "2021-11-02 09:19:25,287", "levelname": "INFO", "name": "pkt_kg.owlnets", "module": "owlnets", "funcName": "makes_graph_connected", "lineno": 688, "message": "1100 triples added to make connected"}
{"asctime": "2021-11-02 09:23:55,212", "levelname": "INFO", "name": "pkt_kg.owlnets", "module": "owlnets", "funcName": "write_out_results", "lineno": 737, "message": "Serializing OWL-NETS Graph"}
{"asctime": "2021-11-02 09:39:00,282", "levelname": "INFO", "name": "pkt_kg.owlnets", "module": "owlnets", "funcName": "write_out_results", "lineno": 751, "message": "OWL-NETS Graph Stats: 780753 nodes, 5072062 edges, 441 self-loops, 5 most most common edges: http://www.w3.org/2000/01/rdf-schema#subClassOf:1204474, http://purl.obolibrary.org/obo/RO_0002436:1001464, http://purl.obolibrary.org/obo/RO_0001025:689128, http://purl.obolibrary.org/obo/RO_0002200:428374, http://purl.obolibrary.org/obo/RO_0000056:381721, http://purl.obolibrary.org/obo/RO_0002606:279052, average degree 6.496372092070091, 5 highest degree nodes: http://purl.obolibrary.org/obo/SO_0000673:190850, http://purl.obolibrary.org/obo/SO_0001483:130579, http://purl.obolibrary.org/obo/NCBITaxon_9606:117091, http://purl.obolibrary.org/obo/SO_0001217:105056, http://purl.obolibrary.org/obo/SO_0002113:29340, http://purl.obolibrary.org/obo/SO_0001503:27620, density: 8.320660199487278e-06, 1 component(s): {0: 780753}"}

{"asctime": "2021-11-02 09:39:00,284", "levelname": "INFO", "name": "pkt_kg.owlnets", "module": "owlnets", "funcName": "purifies_graph_build", "lineno": 703, "message": "Purifying Graph Based on Construction Approach"}
{"asctime": "2021-11-02 09:39:00,284", "levelname": "INFO", "name": "pkt_kg.owlnets", "module": "owlnets", "funcName": "purifies_graph_build", "lineno": 708, "message": "Determining what triples need purification"}
{"asctime": "2021-11-02 09:39:00,287", "levelname": "INFO", "name": "pkt_kg.owlnets", "module": "owlnets", "funcName": "purifies_graph_build", "lineno": 711, "message": "Processing 161 http://www.w3.org/1999/02/22-rdf-syntax-ns#type triples"}
{"asctime": "2021-11-02 09:43:24,714", "levelname": "INFO", "name": "pkt_kg.owlnets", "module": "owlnets", "funcName": "write_out_results", "lineno": 737, "message": "Serializing Subclass-Purified OWL-NETS Graph"}
{"asctime": "2021-11-02 09:58:01,555", "levelname": "INFO", "name": "pkt_kg.owlnets", "module": "owlnets", "funcName": "write_out_results", "lineno": 751, "message": "Subclass-Purified OWL-NETS Graph Stats: 780753 nodes, 5072064 edges, 441 self-loops, 5 most most common edges: http://www.w3.org/2000/01/rdf-schema#subClassOf:1204637, http://purl.obolibrary.org/obo/RO_0002436:1001464, http://purl.obolibrary.org/obo/RO_0001025:689128, http://purl.obolibrary.org/obo/RO_0002200:428374, http://purl.obolibrary.org/obo/RO_0000056:381721, http://purl.obolibrary.org/obo/RO_0002606:279052, average degree 6.496374653699697, 5 highest degree nodes: http://purl.obolibrary.org/obo/SO_0000673:190850, http://purl.obolibrary.org/obo/SO_0001483:130579, http://purl.obolibrary.org/obo/NCBITaxon_9606:117091, http://purl.obolibrary.org/obo/SO_0001217:105056, http://purl.obolibrary.org/obo/SO_0002113:29340, http://purl.obolibrary.org/obo/SO_0001503:27620, density: 8.3206634804646e-06, 1 component(s): {0: 780753}"}
{"asctime": "2021-11-02 09:58:01,557", "levelname": "INFO", "name": "pkt_kg.owlnets", "module": "owlnets", "funcName": "runs_owlnets", "lineno": 796, "message": "\n\nOWL-NETS Graph Stats: 5072062 triples, 780753 nodes, 290 predicates, 0 classes, 0 individuals, 0 object props, 0 annotation props;\nPurified OWL-NETS Graph Stats: 5072064 triples, 780753 nodes, 289 predicates, 0 classes, 0 individuals, 0 object props, 0 annotation props"}
{"asctime": "2021-11-02 09:58:06,686", "levelname": "INFO", "name": "pkt_kg.owlnets", "module": "owlnets", "funcName": "runs_owlnets", "lineno": 812, "message": "Decoded 474111 owl-encoded classes and axioms. Note the following:\nPartially processed 364 cardinality elements\nRemoved 1645 owl:disjointWith axioms\nIgnored: 3 misc classes; 19 classes constructed with owl:complementOf; 4813 classes containing negation (e.g. pr#lacks_part, cl#has_not_completed)\nFiltering removed 17563627 semantic support triples"}
{"asctime": "2021-11-02 09:58:07,196", "levelname": "INFO", "name": "pkt_kg.knowledge_graph", "module": "knowledge_graph", "funcName": "construct_knowledge_graph", "lineno": 604, "message": "*** Writing Knowledge Graph Edge Lists ***"}
{"asctime": "2021-11-02 09:58:07,197", "levelname": "INFO", "name": "pkt_kg.knowledge_graph", "module": "knowledge_graph", "funcName": "construct_knowledge_graph", "lineno": 609, "message": "*** Processing OWL Graph ***"}
{"asctime": "2021-11-02 10:06:19,866", "levelname": "INFO", "name": "pkt_kg.metadata", "module": "metadata", "funcName": "output_metadata", "lineno": 258, "message": "Writing Class Metadata"}
{"asctime": "2021-11-02 10:13:05,566", "levelname": "INFO", "name": "pkt_kg.knowledge_graph", "module": "knowledge_graph", "funcName": "construct_knowledge_graph", "lineno": 609, "message": "*** Processing OWL-NETS Graph ***"}
{"asctime": "2021-11-02 10:14:37,082", "levelname": "INFO", "name": "pkt_kg.metadata", "module": "metadata", "funcName": "output_metadata", "lineno": 258, "message": "Writing Class Metadata"}
{"asctime": "2021-11-02 10:15:59,257", "levelname": "INFO", "name": "pkt_kg.knowledge_graph", "module": "knowledge_graph", "funcName": "construct_knowledge_graph", "lineno": 609, "message": "*** Processing Purified OWL-NETS Graph ***"}
{"asctime": "2021-11-02 10:17:29,232", "levelname": "INFO", "name": "pkt_kg.metadata", "module": "metadata", "funcName": "output_metadata", "lineno": 258, "message": "Writing Class Metadata"}
{"asctime": "2021-11-02 10:23:46,066", "levelname": "INFO", "name": "pkt_kg.knowledge_graph", "module": "knowledge_graph", "funcName": "construct_knowledge_graph", "lineno": 620, "message": "\nLoading Full (Logic + Annotation) Graph"}
{"asctime": "2021-11-02 11:44:38,694", "levelname": "INFO", "name": "pkt_kg.knowledge_graph", "module": "knowledge_graph", "funcName": "construct_knowledge_graph", "lineno": 621, "message": "Deriving Stats"}
{"asctime": "2021-11-02 12:05:30,840", "levelname": "INFO", "name": "pkt_kg.knowledge_graph", "module": "knowledge_graph", "funcName": "construct_knowledge_graph", "lineno": 622, "message": "Full (Logic + Annotation) Graph Stats: 36201468 triples, 13680807 nodes, 355 predicates, 4348593 classes, 190 individuals, 853 object props, 635 annotation props"}
{"asctime": "2021-11-02 12:10:31,817", "levelname": "INFO", "name": "__main__", "module": "build_phase_3", "funcName": "main", "lineno": 125, "message": "STEP 3: UPLOAD KNOWLEDGE GRAPH DATA TO GOOGLE CLOUD STORAGE"}
{"asctime": "2021-11-02 12:10:31,818", "levelname": "INFO", "name": "__main__", "module": "build_phase_3", "funcName": "main", "lineno": 128, "message": "Removing Existing Data from Current Build Directory on Google Cloud Storage"}
{"asctime": "2021-11-02 12:10:48,261", "levelname": "INFO", "name": "__main__", "module": "build_phase_3", "funcName": "main", "lineno": 138, "message": "Copying Data FROM: archived_builds/release_v3.0.2/build_01NOV2021/data/ TO: current_build/data/"}
{"asctime": "2021-11-02 12:10:58,297", "levelname": "INFO", "name": "__main__", "module": "build_phase_3", "funcName": "main", "lineno": 143, "message": "Uploading Knowledge Graph Data from Docker to the archived_builds Directory"}
{"asctime": "2021-11-02 12:16:11,232", "levelname": "INFO", "name": "__main__", "module": "build_phase_3", "funcName": "main", "lineno": 148, "message": "Copying Graph Data from archived_builds to current_builds"}
{"asctime": "2021-11-02 12:16:16,599", "levelname": "INFO", "name": "__main__", "module": "build_phase_3", "funcName": "main", "lineno": 160, "message": "STEP 4: BUILD CLEAN-UP"}
{"asctime": "2021-11-02 12:16:16,599", "levelname": "INFO", "name": "__main__", "module": "build_phase_3", "funcName": "main", "lineno": 163, "message": "COMPLETED BUILD PHASE 3: 674.132 MINUTES"}
{"asctime": "2021-11-02 12:16:16,599", "levelname": "INFO", "name": "__main__", "module": "build_phase_3", "funcName": "main", "lineno": 163, "message": "EXIT BUILD PHASE 3"}